\documentclass[acmsmall]{acmart} %
\usepackage{amsmath,bbm,bm}
\usepackage{gensymb}
\usepackage[utf8]{inputenc} % allow utf-8 input
\usepackage[T1]{fontenc}    % use 8-bit T1 fonts
\usepackage{booktabs}       % professional-quality tables
\usepackage{amsfonts}       % blackboard math symbols
\usepackage{nicefrac}       % compact symbols for 1/2, etc.
\usepackage{microtype}      % microtypography
\usepackage{graphicx}
\usepackage[font=footnotesize]{subcaption}
\usepackage{lipsum}
\usepackage{multirow}
\usepackage{comment}
\usepackage{color}
\usepackage{etoolbox}
\usepackage{caption}
\PassOptionsToPackage{table,xcdraw}{xcolor}

\newtoggle{comments}
\settoggle{comments}{true} %set true to show all comments, set false to hide all comments
\iftoggle{comments}{
  \newcommand {\alberto}[1]{{\color{orange}{~Alberto: #1}\normalfont}}
  \newcommand {\bjin}[1]{{\color{blue}{~Baihong: #1}\normalfont}}
  \newcommand {\yuxin}[1]{{\color{violet}{~Yuxin: #1}\normalfont}}
  \newcommand {\poolla}[1]{{\color{green}{~Kameshwar: #1}\normalfont}}
  \newcommand {\red}[1]{{\color{red}{#1}\normalfont}}
}{
  \newcommand {\alberto}[1]{{}}
  \newcommand {\bjin}[1]{{}}
  \newcommand {\yuxin}[1]{{}}
  \newcommand {\poolla}[1]{{}}
  \newcommand {\red}[1]{{}}
}

\usepackage{acronym}
\acrodef{o.o.d.}{out-of-distribution}
\acrodef{FDD}{anomaly detection}
\acrodef{LDA}{Linear Discriminant Analysis}
\acrodef{ML}{Machine Learning}
\acrodef{CFAR}{Constant False Alarm Rate}
\acrodef{RF}{Random Forest}
\acrodef{LR}{Logistic Regression}
\acrodef{NN}{Neural Network}
\acrodef{DT}{Decision Tree}
\acrodef{OC-SVM}{One-Class Support Vector Machine}
\acrodef{AE}{Autoencoder}
\acrodef{FN}{False Negative}

\acrodef{FPR}{False Positive Rate}
\acrodef{FNR}{False Negative Rate}
\acrodef{SL}{Severity Level}
\acrodef{ROC}{Receiver Operating Characteristic}
\acrodef{CI}{confidence interval}
\acrodef{incipient}{Incipient}
\acrodef{RBF}{Radial Basis Function}
\acrodef{PCA}{Principle Component Analysis}
\acrodef{LSTM}{Long Short-Term Memory}
\acrodef{RUL}{Remaining Useful Life}

\acrodef{FNP}{False Negative Precision}
\acrodef{AUC}{Area Under Curve}
\acrodef{FDD}{anomaly detection}
\acrodef{CNN}{Convolutional Neural Network}
\acrodef{KL}{Kullback–Leibler}
\acrodef{DR}{Diabetic Retinopathy}

\newcommand{\expct}[1]{\mathbb{E}\left[#1\right]}

\newcommand{\PrOf}[1]{\ensuremath{\mathbb{P}\left[#1\right]}}

\newcommand{\abs}[1]{\left\vert#1\right\vert}

\newcommand{\rvdiff}{Z}

\AtBeginDocument{%
  \providecommand\BibTeX{{%
    \normalfont B\kern-0.5em{\scshape i\kern-0.25em b}\kern-0.8em\TeX}}}

\begin{document}

\title{Using Ensemble Classifiers to Detect Incipient Anomalies}

\author{Baihong Jin}
\affiliation{%
  \institution{University of California, Berkeley}
}
\email{bjin@eecs.berkeley.edu}

\author{Yingshui Tan}
\affiliation{%
  \institution{University of California, Berkeley}
}
\email{tys@eecs.berkeley.edu}

\author{Albert Liu}
\affiliation{%
  \institution{University of California, Davis}
}
\email{adliu@ucdavis.edu}

\author{Xiangyu Yue}
\affiliation{%
  \institution{University of California, Berkeley}
}
\email{xyyue@eecs.berkeley.edu}

\author{Yuxin Chen}
\affiliation{%
  \institution{University of Chicago}
}
\email{chenyuxin@uchicago.edu}

\author{Alberto Sangiovanni~Vincentelli}
\affiliation{\institution{University of California, Berkeley}}
\email{alberto@eecs.berkeley.edu}

\renewcommand{\shortauthors}{B.~Jin et al.}

\begin{abstract}
Incipient anomalies present milder symptoms compared to severe ones, and are more difficult to detect and diagnose due to their close resemblance to normal operating conditions. The lack of incipient anomaly examples in the training data can pose severe risks to anomaly detection methods that are built upon \ac{ML} techniques, because these anomalies can be easily mistaken as normal operating conditions. To address this challenge, we propose to utilize the uncertainty information available from ensemble learning to identify potential misclassified incipient anomalies. We show in this paper that ensemble learning methods can give improved performance on incipient anomalies and identify common pitfalls in these models through extensive experiments on two real-world datasets. Then, we discuss how to design more effective ensemble models for detecting incipient anomalies.
\end{abstract}




\thanks{
  $^\ast$ Baihong Jin is the corresponding author.\\
  This work is supported by the National Research Foundation of Singapore through a grant to the Berkeley Education Alliance for Research in Singapore (BEARS) for the Singapore-Berkeley Building Efficiency and Sustainability in the Tropics (SinBerBEST) program, and by the Defence Science \& Technology Agency (DSTA) of Singapore. We would like to thank Prof. Jiantao Jiao for his valuable suggestions and comments.
%   Author's addresses and contact information: Baihong Jin, Yingshui Tan, Xiangyu Yue and Alberto Sangiovanni Vincentelli, Department of Electrical Engineering Computer Sciences, University of California, Berkeley, Berkeley, CA~94720. \{bjin,tys,xyyue,alberto\}@eecs.berkeley.edu; Berkeley, CA~94720; Albert Liu, University of California, Davis, albertdliu@gmail.com; Yuxin Chen, University of Chicago, chenyuxin@uchicago.edu.
}


\maketitle

\section{Introduction}\label{sec:intro}
\acresetall

\begin{figure}[tb]
    \centering
    \begin{subfigure}[t]{0.38\linewidth}
        \centering
        \includegraphics[height=3.5cm]{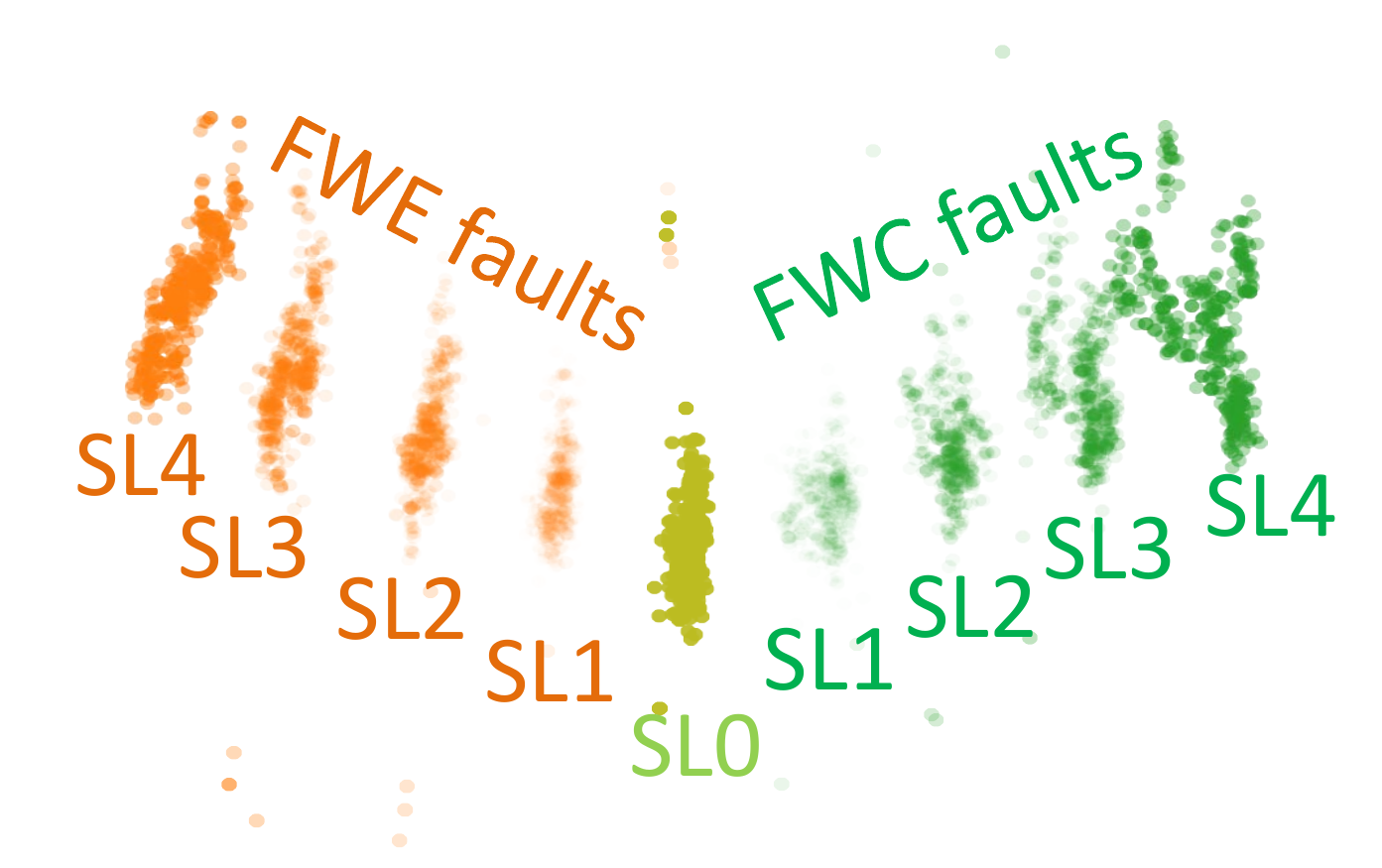}
        \caption{}
        \label{fig:visualization-chiller}
    \end{subfigure}
    \begin{subfigure}[t]{0.60\linewidth}
        \centering
        \includegraphics[height=5cm]{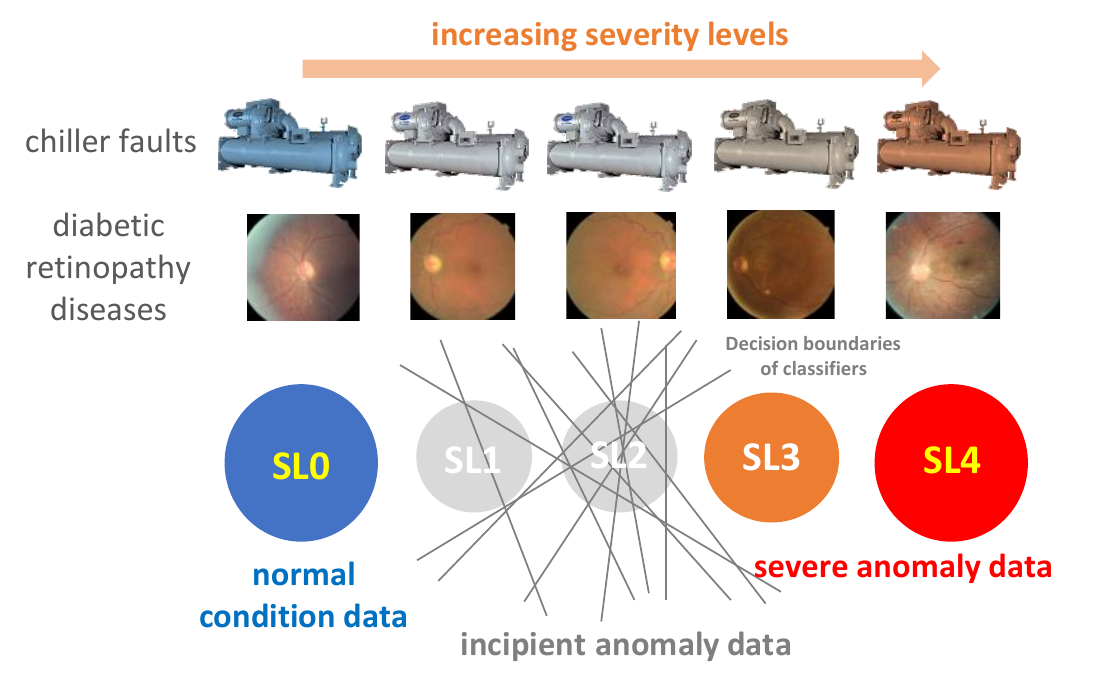}
        \caption{}
        \label{fig:ensemble-illustration}
    \end{subfigure}
    \caption{(a) Visualization of part of the RP-1043 chiller data~\cite{comstock1999development} where the ``severity spectrum'' for two fault conditions is visible. The normal condition and two fault conditions (each with four \acp{SL}) are shown. (b) Illustration showing how an ensemble classifier can conceptually help detect incipient anomalies. The gray lines represent the decision boundaries of base learners in the ensemble.}
\end{figure}

In anomaly detection applications\footnote{In this paper, an ``anomaly'' can mean either a machine fault in industrial applications or a human disease in health applications.}, it is common to encounter anomaly data examples whose symptoms correspond to different \acp{SL}. Fig.~\ref{fig:visualization-chiller} shows a real-world example where faults are categorized into four different \acp{SL}, from SL1 (slightest) to SL4 (most severe). The ability of accurately assessing the severity of faults/diseases is important for anomaly detection applications, yet very difficult on low-severity examples; SL1 data clusters are much closer to the normal cluster than to their corresponding SL4 clusters in Fig.~\ref{fig:visualization-chiller}. A anomaly detection system needs to be very sensitive to identify the low-severity faults; at the same time, it should keep the number of false positives low, which makes the design of such decision systems a challenging task.

If labeled data from different \acp{SL} are available, then regular regression or classification approaches are suitable, as already exemplified by previous research~\cite{krause2018grader,li2016fault}.
However, these fine-grained labeled datasets can take much effort to prepare and we may not always have \textit{a priori} access to a full spectrum of anomaly \acp{SL}. In an extreme case, as illustrated in Fig.~\ref{fig:ensemble-illustration}, suppose we only have access to the two ends (i.e. the normal condition SL0 and the most severe anomaly condition SL4) of the severity spectrum; incipient anomaly instances are not available to us. If we train a classification system only using the available SL0 and SL4 data, the resulting classifier may have great performance on in-distribution data (SL0 \& SL4). However, it may fail badly with identifying the incipient anomaly data. For example, most SL1 faults may be mistakenly recognized as normal by any of the decision boundaries shown in Fig.~\ref{fig:ensemble-illustration}. More generally, classical supervised learning approaches designed for achieving maximal separation between labeled classes (e.g. margin-based classifiers, discriminative neural networks, etc), are less effective in detecting such low-severity, incipient anomaly data examples.

In the absence of labeled data for certain categories of fault instances, common practices are to develop generative models, such as the Gaussian mixture model~\cite{ZongSMCLCC18}, \ac{PCA}~\cite{huang2007network,zhang2017improved}, \ac{LSTM} network~\cite{du2017deeplog} and autoencoder \cite{sakurada2014anomaly,tan2019encoder}. A potential problem for these models is that they may not always generalize well---that is, a single-trained model, when applied to an unseen incipient anomaly instance at test time, can be classified as normal~\cite{du2019lifelong}, i.e. becoming a false negative.

The solution we propose in this paper is based on ensemble learning~\cite{zhou2012ensemble}, i.e., on the process of training multiple classifiers and leveraging their joint decisions to recognize incipient anomalies. In literature, a variety of ensemble methods have been proposed on the estimation of decision uncertainties~\cite{leibig2017leveraging,gal2016uncertainty,lakshminarayanan2017simple}. Fig.~\ref{fig:ensemble-illustration} shows that the individual classifiers have much disagreement on the SL2 data. The amount of disagreement can be used to measure the decision uncertainties, and is therefore useful for indicating incipient anomalies. However, for SL1 data that are close to the normal cluster, the above approach will become less effective. We find this is a common phenomenon in our empirical studies. A remedy to this problem is to increase the \textit{statistical power} of the base learners by moving the decision boundaries towards the normal cluster. Another question is how to properly combine the anomaly scores from ensemble members into an \textit{uncertainty metric} to inform decision making. We will analyze two widely used uncertainty metrics and compare their performance on detecting incipient anomalies. 

We believe our proposed methods are a useful complement to the literature on multi-grade anomaly detection~\cite{krause2018grader,li2016fault,li2016data}, specifically under cases where the available anomaly data for training are insufficient to cover the entire \textit{severity spectrum}. In this paper, we give some caveats and an easy-to-use recipe for \ac{ML} practitioners to develop ensemble anomaly detection models that can more effectively recognize incipient anomaly examples, and will provide recommendations to address the aforementioned issues that can help produce more effective ensemble models for anomaly detection applications. We summarize our contributions in this paper as follows:
\begin{itemize}
    \item We show by experiments that incipient anomaly examples, when missing or underrepresented in the training distribution, can pose risks to popular supervised  \ac{ML}-based anomaly detection models such as \acp{DT} and \acp{NN}. Ensemble methods are in general helpful in improving the detection performance of both supervised and unsupervised models on such incipient anomalies.
    \item We compare and analyze two commonly used uncertainty metrics for ensemble learning, one based on ensemble mean (\textsc{mean}) and the other based on ensemble variance (\textsc{var}). Our theoretical analysis shows that \textsc{mean} is more preferable to \textsc{var} .
\end{itemize}
The rest of this paper is organized as follows. We formulate the anomaly detection problem and provide necessary background knowledge in Sec.~\ref{sec:problem-formulation}. Next in Sec.~\ref{sec:methodology}, we describe our methodology in details. The two datasets used in our empirical study are briefly described in Sec.~\ref{sec:dataset-descriptions}, and in Sec.~\ref{sec:experiment}  we present our experimental results. In Sec.~\ref{sec:related-work}, we review related research topics found in the literature. We summarize the findings in this paper and discuss future work in Sec.~\ref{sec:conclusion}.

\section{Background and Problem Formulation}\label{sec:problem-formulation}

\paragraph{Anomaly detection problem}
We formulate the anomaly detection problem in a \textit{binary classification} setting.
An anomaly detection system aims at differentiating the fault conditions from the normal condition by monitoring the system state. Let ${z}\in\{0,1\}$ be the ground-truth label of system state $\bm{x}\in\mathbb{R}^d$, where $z=0$ stands for the normal condition and $z=1$ the anomaly condition. An \textit{anomaly detector} is some rule, or function, that assigns (predicts) a class label $\hat{z}\in\{0,1\}$ to input $x$. 

Let $\mathcal{X}$ be the set of data points, and $\mathcal{M}$ be a model class of classification models. Suppose a classification model $M\in\mathcal{M}$ defines an \textit{anomaly score} function $s^M:\mathcal{X}\rightarrow\mathbb{R}$ that characterizes how likely a data point corresponds to an anomaly state; a larger $s^M(x)$ implies a higher chance of a data point $\bm{x}$ being an anomaly. The classifier's decision on whether or not $x$ corresponds to an anomaly can be made by introducing a \textit{decision threshold} $\tau^M$ to dichotomize the anomaly score $s^M(x)$. We can define the classifier's predicted label 
$\hat{z} = \mathbbm{1}\{s^M(x) > \tau^M\}$, 
i.e. $M$ predicts $x$ to be an anomaly if and only if the anomaly score $s^M(x)$ is above the threshold $\tau^M$. For evaluating the accuracy of anomaly detection, we can define the \acf{FNR} and \acf{FPR} of the model $M$ on the test data distribution as follows:
\begin{align}
    \text{FNR}(s^M,\tau^M) &= \PrOf{\hat{z}=0 \mid z=1},\\
    \text{FPR}(s^M,\tau^M) &= \PrOf{\hat{z}=1 \mid z=0}.
\end{align}
Let $\mathcal{X}^\text{train}$ be a subset of labeled data points for training. Ideally, our goal is to learn an anomaly score function $s^\ast$ by minimizing its classification error on $\mathcal{X}^\text{train}$, and then decide a corresponding threshold $\tau$, such that $(s^\ast,\tau)$ can optimally trade off the \ac{FNR} and the \ac{FPR} on unseen test data.

\paragraph{Setting the detection threshold $\tau$}
We leverage the prediction uncertainties given by ensemble learners to make uncertainty-informed decisions.
Consider an ensemble $\mathcal{E}$ comprising a diverse set of $K$ binary classifiers, $\mathcal{M}^{(1)},\mathcal{M}^{(2)},\ldots,\mathcal{M}^{(K)}$, that have been trained for the same detection task. Let $z_i\in\{0,1\}$ represent the ground-truth label of input $x_i$, and  $\hat{y}_i^{(k)}$  denote the output of the $k$th classifier where $k \in \{1,\dots, K\}$ and $\hat{y}_i^{(k)}\in[0,1]$. By using a threshold $\tau$ to dichotomize the continuous output $y_i^{(k)}$, each classifier $\mathcal{M}^{(k)}$ produces a predicted class label $\hat{z}_i^{(k)}$ for input $x_i$.

As mentioned above, one always has to make a trade-off between \ac{FNR} and \ac{FPR} by setting an appropriate decision threshold $\tau$ (a.k.a. operating point).
A simple approach is to directly set the decision threshold $\tau$ to a predefined value (e.g., $0.5$); this is often not a bad approach if most data points are well separated and receive an anomaly score close to $0$ or $1$. However, such approach usually does not returns us a high-sensitivity classifier that satisfies a given \ac{FPR} requirement. In real-practice, one often needs to decide a proper \textit{operating point} on the \ac{ROC} curve by taking \ac{FPR} and \ac{FNR} requirements into account. One way to do that is to set $\tau$ such that the \ac{FPR} on the development set reaches a predefined level $q$. The rationale behind such scheme is to fix the \ac{FPR} (type-1 errors) to a constant value on the development set while minimizing the number of false negatives (type-2 errors). Similar approaches are seen in other application domains. For example, in radar applications, this scheme is also known as \ac{CFAR} detection~\cite{richards2005fundamentals}. 

\begin{figure}[t]
    \centering
    \includegraphics[width=\linewidth]{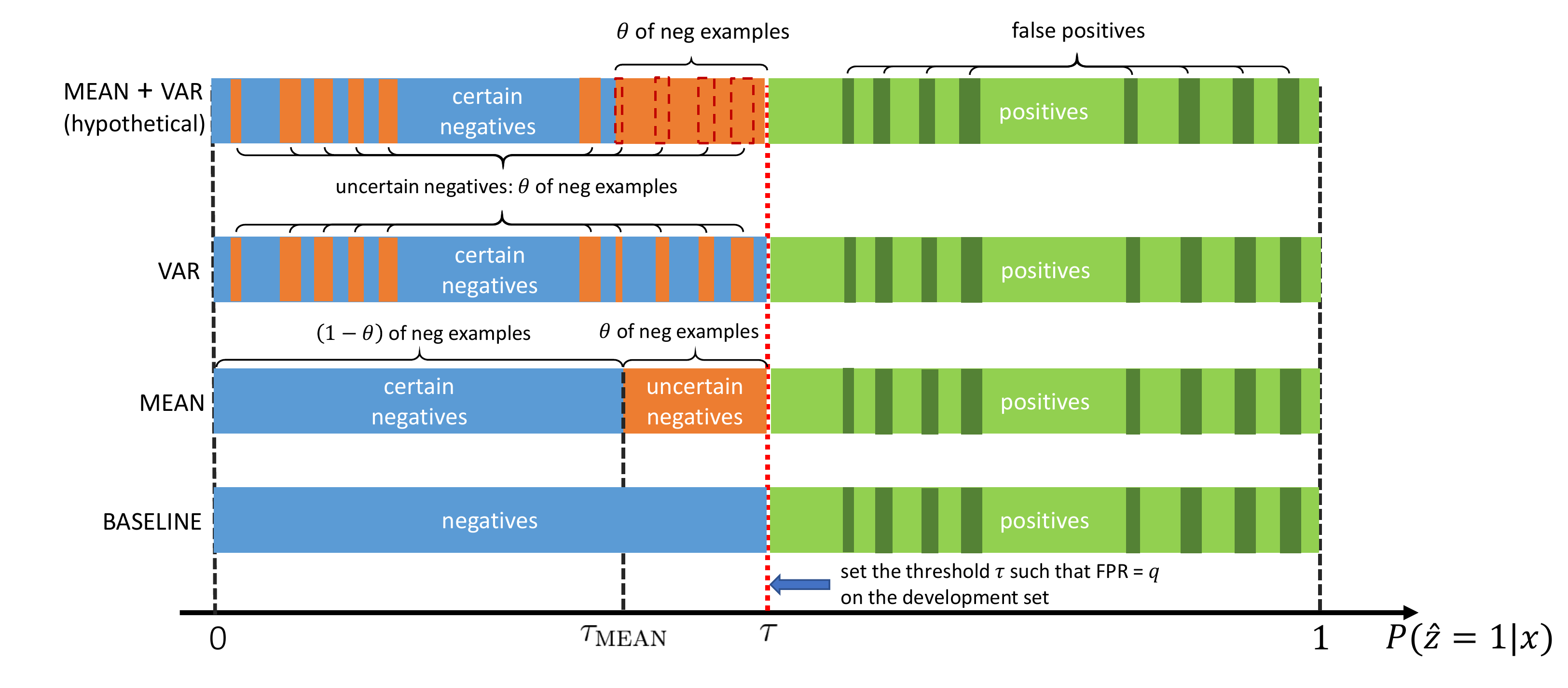}
    \caption{Illustration showing the concepts in an uncertainty-informed decision framework.}
    \label{fig:decision}
\end{figure}

\looseness -1 The decision scheme described above is illustrated in Fig.~\ref{fig:decision} as the \textsc{baseline} scheme. The goal is to come up with a proper $\tau$ that is used to identify positive examples. % (shown as yellow in diagram). 
Under most cases, there will be false positives among the examples predicted as positive; however, these false positives are not the utmost concern if the \ac{FPR} can be controlled to a low level. On the other hand, false negatives are anomalous instances mistaken as normal, which represents a more severe problem in anomaly detection. We propose utilizing decision uncertainty information from ensemble classifiers to identify potential false negatives in an uncertainty-informed decision scheme.   

\paragraph{Uncertainty-informed anomaly detection}
We consider an \textit{uncertainty-informed} diagnostic scheme as an application of prediction uncertainties that fosters the collaboration between human and AI systems. %\ac{ML}. 
In this scheme, an \ac{ML} model is first used to screen the cases (operational data for industrial machines, medical images for humans, etc.). Cases diagnosed as positive will be referred to a human reviewer for further inspection, who will confirm the case as positive if she agrees with the \ac{ML} model's decision. The \textsc{baseline} scheme suffers from the problem that false negatives from the \ac{ML} model's diagnoses would never be reviewed by human diagnosticians. In an \textit{uncertainty-informed} scheme, high-uncertainty negative examples will be identified and sent to human reviewers as well. The criterion used for picking out high-uncertainty examples does not have to be based on the classifier confidence $\hat{y}$; in fact, we can use a variety of \textit{uncertainty metrics} to be described below for ranking data examples by their associated uncertainties.

To identify false negatives in classification, we use an \textit{uncertainty metric} $U$ to rank the negative examples\footnote{Examples that are classified as negative by a classification model, i.e. $\{x_i\,\vert\,\hat{z}_i=0\}$.}. The uncertainty metric $U: \mathbb{R}^K\rightarrow \mathbb{R}$ takes as input the ensemble predictions $\{\hat{y}_i^{(k)}\}$ on $x_i$, and outputs an \textit{uncertainty score} $u(x_i)\doteq U(y_i^{(1)},y_i^{(2)},\ldots,y_i^{(K)})$. The interpretation of the uncertainty score $u(x_i)$ depends on the task. In our application, we seek to utilize prediction uncertainties to identify false negative decisions: a higher $u(x_i)$ indicates higher prediction uncertainty associated with $x_i$. In such situations, we may need human experts to join the decision process.

The uncertainty score $u(x_i)$ is a real-valued number, and to resolve a dichotomy between ``uncertain'' and ``certain'' we will need another threshold $\tilde{u}$. If $u(x_i)>\tilde{u}$ then $x_i$ is deemed an uncertain input example and otherwise a certain one. Once uncertain examples are identified, we will need external resources (e.g., human experts) to inspect them and determine their true labels; however, such external resources are often limited (e.g., due to budget constraints) so we need to determine a proper threshold $\tilde{u}$ so that the number of uncertain negatives is controlled. We determine $\tilde{u}$ by setting the \textit{uncertain negative ratio} on the development set to a fraction $\theta$, where the uncertain negative ratio is the fraction of uncertain examples among those examples predicted as negative. Only the predicted negative examples that receive the highest uncertainty scores are deemed \textit{uncertain negatives}. To evaluate how uncertain negatives overlap with the actual false negatives, we define the following performance measure.
\begin{definition}[False Negative Precision]
We define the false negative precision to be the fraction of false negative decisions among uncertain negative inputs, under given uncertainty metric $U$ and uncertain negative ratio $\theta$. Written in mathematical form,
\begin{align}
    \text{FN-precision}(U,q) \doteq \frac{\left\vert\left\{x_i\,\vert\,i\in\mathcal{I}^{-}_{q},z_i=1\right\}\right\vert}
    {\left\vert\mathcal{I}^{-}_{q}\right\vert}\in[0,1],
\end{align}
where $\mathcal{I}^{-}_{q}$ is the index set of uncertain negative examples.
\end{definition}
The FN-precision metric can be interpreted as the ratio of identified uncertain examples being actual false negatives. The higher the FN-precision value, the fewer false alarms are likely to be raised by an algorithm that detects uncertain negatives. We can similarly define a ``false negative recall'' metric that measures the fraction of false negatives identified by the algorithm. However, in this paper we choose to directly report the total number of false negatives and the number of false negatives that are deemed ``certain'' by the evaluated uncertainty estimation algorithms, as we think it a more straightforward way to make the comparison.
One of our goals in this paper is to rigorously analyze and compare two commonly used uncertainty metrics \textsc{mean} and \textsc{var}, which will be detailed in the upcoming section. Formally, we seek the uncertainty metric $U$ that maximizes $\text{FN-precision}(U,q)$.

% Similarly, we can define the false negative recall measure as below. 
% \begin{definition}[False Negative Recall]
% We define the false negative recall to be the fraction of false negative decisions recovered by the uncertain negative examples, under a given uncertainty metric $U$ and a given uncertain negative ratio $q$. As with the FN-precision, the FN-recall score also takes a value between 0 and 1. Written in mathematical form, 
% \begin{align}
%     \text{FN-recall}(U,q)\doteq\frac{\left\vert\left\{x_i\,\vert\, i\in\mathcal{I}^{-}_{q},z_i=1\right\}\right\vert}
%     {\left\vert\left\{x_i\,\vert\,\hat{z}_i=0, z_i=1\right\}\right\vert}\in[0,1].
% \end{align}
% \end{definition}

\section{Methodology}\label{sec:methodology}

In the sequel, we describe our methodology of using ensemble learning for improving incipient anomaly detection. We first discuss several design considerations when constructing ensemble classifiers.

\subsection{Creating Diversity among Base Learners in an Ensemble}
Diversity is recognized as one of the key factors that contribute to the success of ensemble approaches~\cite{brown2005diversity}. As illustrated in Fig.~\ref{fig:ensemble-illustration}, the diversity among ensemble members is crucial for improved detection performance on \ac{o.o.d.} data instances. In our empirical study to be described later, we will employ bagging~\cite{breiman1996bagging} to induce diversity among ensemble members.
Bagging~\cite{breiman1996bagging} (or bootstrap aggregation) is a classical approach for creating diversity among ensemble members. The core idea is to construct models from different training datasets using \textit{randomization}. In the original bagging approach~\cite{breiman1996bagging}, a random subset of the training samples is selected for training each member classifier. A later variant, the so-called ``feature bagging'' (a.k.a. random subspace method~\cite{ho1998random}) selects a random subset of the features for training. One famous application of bagging in \ac{ML} is the \acf{RF} model. In this study, we will use sample bagging in our experiments to induce diversity among ensemble classifiers.

\subsection{Combining Base Models into an Ensemble}
In ensemble analysis, one challenge is that the anomaly scores given by different models may not be directly comparable. This is known as the \textit{normalization issue}~\cite{aggarwal2013outlier}. This issue is more common with unsupervised or semi-supervised models, because the outputs from these models (e.g., the reconstruction errors from autoencoders) are often naturally unbounded. 
If the scores from different models are directly combined without normalization (e.g., by calculating the average or the maximum), models that give higher anomaly scores may be inadvertently favored~\cite{aggarwal2013outlier}. The normalization issue is less concerning for supervised classification models that use a softmax layer to produce probability vectors whose values are bounded within the $[0,1]$ interval; still, there are still concerns about whether or not these probability estimates are well \textit{calibrated} (known as the \textit{calibration issue}). For the ensemble supervised classifiers in this study, we assume minimal impacts from the calibration issues. 

On top of the normalization issue, how to properly aggregate the (normalized) anomaly scores from models in an ensemble, known as the \textit{combination issue}~\cite{aggarwal2013outlier}, constitutes another major challenge in ensemble analysis. Depending on how the base learners are combined into an ensemble detector, we can classify the combination scheme into a 1) hard voting or a 2) soft voting scheme. In hard voting schemes, each base learner predicts a binary label $\hat{y}\in\{0,1\}$ indicating whether an input example $x$ is normal or not, while in soft voting schemes base learners outputs real-valued anomaly scores.
In this work, we will mainly consider ensemble models made up of supervised classifiers, and focus on how to properly obtain uncertainty estimates from the score vectors in order to better detect incipient anomalies.

\subsection{Uncertainty Metrics for Ensemble Learners}
Suppose we have $m$ data points for testing, and they are organized into a design matrix $\bm{X}\in\mathbb{R}^{m\times d}$. The outputs from the ensemble of detection models can be accordingly written as an $m\times K$ matrix $\hat{\bm{Y}}$, where $K$ is the ensemble size. Note that entries in matrix $\hat{\bm{Y}}$ can either take discrete values from $\{0,1\}$ (in a hard voting scheme) or take continuous values from $[0,1]$ (in a soft voting scheme), depending on the nature of the underlying base learners. Using superscripts to differentiate ensemble members and subscripts to differentiate data points, we can denote the rows and columns of matrix $\hat{\bm{Y}}$ as follows
\begin{align}
    \hat{\bm{Y}} \doteq 
    \begin{bmatrix}
        \vert & \vert & \cdots & \vert\\
        \hat{\bm{Y}}^{(1)} & \hat{\bm{Y}}^{(2)} & \cdots & \hat{\bm{Y}}^{(K)}\\
        \vert & \vert & \cdots & \vert
    \end{bmatrix} = 
    \begin{bmatrix}
        \text{---} & \hat{\bm{Y}}_1 & \text{---} \\
        \text{---} & \hat{\bm{Y}}_2 & \text{---} \\
        \cdots & \cdots & \cdots \\ 
        \text{---} & \hat{\bm{Y}}_m & \text{---}
    \end{bmatrix}
\end{align}\label{eqn:prob-matrix}
where each $\hat{\bm{Y}}^{(k)}=\begin{bmatrix}
\hat{y}_{1}^{(k)}~\hat{y}_{2}^{(k)}\cdots\hat{y}_{m}^{(k)}
\end{bmatrix}^\intercal$ represents the predictions from the $k$th single learner ($k=1,2,\ldots,K$) on the $m$ data points, and each 
$\hat{\bm{Y}}_i = \begin{bmatrix}
\hat{y}_{i}^{(1)}~\hat{y}_{i}^{(2)}\cdots\hat{y}_{i}^{(K)}
\end{bmatrix}$ represents the $K$ predictions from the ensemble learner on $x_i$.

To come up with an uncertainty estimate for $x_i$, we calculate $U(\hat{y}_{i}^{(1)},\hat{y}_{i}^{(2)},\cdots, \hat{y}_{i}^{(K)})$ using $U$ as the uncertainty metric. A number of metrics have been proposed in literature for estimating the prediction uncertainties of ensemble learners. In %Lakshminarayanan~et~al.'s paper~
\citet{lakshminarayanan2017simple}, the metrics are broadly classified into two categories: \textit{confidence-based} and \textit{disagreement-based} metrics. The former category is designed to capture the consensus of the individual learners in an ensemble, while the latter aims to measure the degree of disagreement among their predictions; however, the two seemingly unrelated goals can have a significant overlap. In this paper, we propose a rigorous categorization for these uncertainty metrics depending on their mathematical forms to unveil their differences and to enable further analyses. Some metrics (hereinafter referred to as type-1 metrics) rely only on the ensemble output $\hat{y}_i^{e}$, while others (referred to as type-2 metrics) take all single learner's outputs into account. Type-1 metrics use the ensemble output $\hat{y}_i^{e}$ to compute the confidence level, without the need to know what the individual predictions are. A negative aspect of these metrics is that the disagreement among individual learners can be hidden beneath the ensemble output $\hat{y}_i^{e}$.

\paragraph{Confidence Gap Metric (\textsc{mean})}
An intuitive metric that measures the confidence of a classifier on input $x$ is to see how close the prediction $\hat{y}^{e}$ is to the decision threshold $\tau^{e}$. Here the superscripts in $\hat{y}^{e}$ and $\tau^{e}$ signify values associated with an ensemble classifier; in the special case where $K=1$, the ensemble classifier degenerates to a single learner classifier.
The smaller the gap $\left\vert\hat{y}_i^{e}-\tau^{e}\right\vert$ is, the higher the uncertainty with $x_i$. Since we prefer the convention that larger function values of $u^\textsc{mean}(x_i)$ corresponds to larger uncertainties, we define the uncertainty score under the margin metric can be formulated as 
\begin{align}
    u^\textsc{mean}(x_i) \doteq 1 - \left\vert\hat{y}_i^{e}-\tau^{e}\right\vert,
\end{align}\label{eqn:mean-metric}
where a constant $1$ is added to the definition so that the uncertainty value $u^\textsc{mean}(x)$ is always positive. Since the ensemble prediction $\hat{y}_i^{e}$ is obtained by taking the average of the individual outputs of classifiers in the ensemble, we will hereinafter refer to this metric as \textsc{mean}.

\paragraph{Binary Cross-Entropy Metric (\textsc{entropy})}
The binary cross-entropy $u^\textsc{entropy}$ as a function of $x_i$ takes the form
\begin{align}
    u^\textsc{entropy}(x_i) \doteq -\left[\hat{y}_i^{e}\log \hat{y}_i^{e} + \left(1-\hat{y}_i^{e}\right)\log\left(1-\hat{y}_i^{e}\right)\right]
\end{align}\label{eqn:entropy-metric}
\textsc{entropy} is equivalent to \textsc{mean} when the decision threshold $\tau^e=0.5$. It can be easily proved that when $\tau^e=0.5$,
\begin{align}
    u^\textsc{mean}(x_i) > u^\textsc{mean}(x_j) \Leftrightarrow u^\textsc{entropy}(x_i) > u^\textsc{entropy}(x_j).
\end{align}
In other words, when $\tau=0.5$ the rankings assigned by $u^\textsc{entropy}$ and by $u^\textsc{mean}(x)$ to the data points are the same. Since we identify uncertain examples by finding the top-ranked data points, $u^\textsc{entropy}$ and $u^\textsc{mean}(x)$ are equivalent. The $u^\textsc{entropy}$ metric can be useful for evaluating decision uncertainties when no decision threshold is a priori assigned.

% Let us use a simple example to illustrate the point. Consider a scenario where 10 models independently predict the disease likelihood. For the first data point, 10 individual models with half of them predicting 0.9 and the other half predicting 0.1. This will result in a ensemble prediction of 0.5 if simple averaging is used to combine all predictions. Consider another data point where the 10 models all predict 0.5, and the ensemble output is still 0.5, same as the previous data point. \yuxin{Move this to the next paragraph? We haven't introduced T-2 yet} Type-1 metrics fail to capture the disagreement among the 10 models in the former case. In contrast, type-2 metrics to be described next are designed to capture the disagreement among individual predictions.
% Type-2 metrics focus on quantifying the \textit{disagreement} among individual learners in an ensemble, which addresses the shortcomings of type-1 metrics. In the binary classification case, each model $j$ predicts $\hat{z}^{(j)}=\Pr\left(y=1\,\vert\,x\right)$. A type-2 metric compute a statistic that measures the ``distances'' from $\hat{z}^{(j)}$'s to their mean $\bar{\hat{z}}$. In literature, the variance (or standard deviation) and the \ac{KL} divergence (or Jensen-Shannon divergence, a symmetric variant of the \ac{KL} divergence) are most widely used as type-2 metrics.

Comparing to the type-1 metrics described above, type-2 metrics have the potential to give a more comprehensive characterization of the individual predictions (e.g., the disagreement among $\hat{y}_{i}^{(1)},\hat{y}_{i}^{(2)},\cdots, \hat{y}_{i}^{(K)}$). The following two existing type-2 metrics that are often used in literature, focus on quantifying the \textit{disagreement} among individual learners in an ensemble and for this reason, may be able to address the shortcomings of type-1 metrics.

\paragraph{Variance Metric (\textsc{var})}
The variance (or standard deviation) metric~\cite{leibig2017leveraging,jin2019detecting} measures how spread out the individual learners' predictions are from the ensemble prediction $\hat{y}_i^{e}$. The uncertainty score of input $x_i$ based on \textit{sample variance} can be written as
\begin{align}
    u^\textsc{var}(x_i)\doteq\frac{1}{K-1}\sum_{k=1}^K \left[\hat{y}_i^{(k)} - \hat{y}_i^{e}\right]
\end{align}\label{eqn:var-metric}

\paragraph{\ac{KL} Divergence Metric (\textsc{kl})}
Similar to the variance metric, the \ac{KL} divergence metric~\cite{goldberger2003efficient} measures the deviation of individual learner's predictions from the ensemble output $\hat{y}_i^{e}$. The uncertainty score $s^\textsc{kl}(x_i)$ of input $x_i$ under the \ac{KL} divergence metric can be written as
\begin{align}
    u^\textsc{kl}(x_i)\doteq \frac{1}{K}\sum_{k=1}^K D_\textsc{kl}\left(y_i^{(k)}\,\Big\Vert\, \hat{y}_i^{e}\right)=\sum_{k=1}^K \hat{y}_i^{(k)}\log\frac{\hat{y}_i^{(k)}}{\hat{y}_i^{e}}.
\end{align}\label{eqn:kl-metric}

A problem with \textsc{var} and \textsc{kl} is that they focus mainly on the disagreement among ensemble predictions but do not take in consideration the value of $\hat{y}_i^{e}$. Consider a scenario where the all ensemble members predict a probability of $0.5$. Both \textsc{var} and \textsc{kl} will produce an uncertainty score of $0$ and thus will not be able to capture any decision uncertainties; in fact, this case where all learners give an output of $0.5$ is highly uncertain. Next, we will compare two representative uncertainty metrics, \textsc{mean} and \textsc{var}, from a theoretical perspective. 

\subsection{Theoretical Analysis on Uncertainty Metrics \textsc{mean} and \textsc{var}}\label{sec:theory}
To model how different classifiers will respond to a given input $x_i$, we assume that the prediction $\hat{y}_i^{(k)}$ from classifier $\mathcal{M}^{(k)}$ is sampled from a beta distribution $\mathcal{B}(\alpha_i,\beta_i)$ that is characterized by two parameters by $\alpha_i$ and $\beta_i$. We further assume that $\alpha_i+\beta_i$ is fixed to the same constant value for all $i$'s.
Under this assumption, the \ac{SL} of the case represented by $x_i$ can be characterized by a single parameter $\alpha_i$, easing further analysis. The larger the value of $\alpha_i$, the more severe the case of $x_i$ is. When $\alpha_i$ and $\beta_i$ are close, the case is ambiguous as the distribution shifts towards being symmetric (i.e. signifying much disagreement) rather than being one-sided.

The main theoretical contribution of this paper is presented in the following theorem, which implies that if $x_i$ is more likely to be positive than $x_j$, then for ensemble learners of fixed size, the upper bound on the probability of $s_\textsc{mean}$ making a wrong decision is lower. In other words, $s_\textsc{mean}$ is likely to be a more robust measure than $s_\textsc{var}$. 
\begin{theorem}\label{thm:mean-vs-var-beta-finite}
Consider inputs $x_i$, $x_j$, with $y_i \sim \mathcal{B}(\alpha_i,\beta_i)$, $y_j \sim \mathcal{B}(\alpha_j,\beta_j)$, and $\alpha_i+\beta_i = \alpha_j+\beta_j$. Let $\Delta_{ij}(s) := \mathbb{E}[s(x_j)-s(x_i)]$ where $s(\cdot)$ denotes an uncertainty score estimated from $K$ \emph{i.i.d.} ensemble learners. If $\alpha_i < \alpha_j \leq \beta_j$, then $\Delta_{ij}(s_\textsc{mean}) > \Delta_{ij}(s_\textsc{var}) > 0$. Furthermore, it holds that $\Pr\left(s_\textsc{mean}(x_i) > s_\textsc{mean}(x_j)\right) = \mathcal{O}\left(\frac{1}{K \Delta_{ij}^2(s_\textsc{mean})}\right)$ and $\Pr\left(s_\textsc{var}(x_i) > s_\textsc{var}(x_j)\right) = \mathcal{O}\left(\frac{1}{K \Delta_{ij}^2(s_\textsc{var})}\right)$.  %\\\Delta_{ij}(s_\textsc{mean}) > \Delta_{ij}(s_\textsc{var})$. 
\end{theorem}

The choice of uncertainty metric $U$ determines how examples are ranked and therefore affects the detection performance of false negatives. We expect the final ranking negative examples due to the uncertainty metric $U$ matches the true severity ranking given by $\alpha_i$. Taking a microscopic view into the ranking process, we consider two negative examples $x_i$ and $x_j$, and assume $x_i$ represents a less severe case than $x_j$. Under the above beta distribution assumption, we will have $\alpha_i<\alpha_j\leq\beta_j$. Our theoretical analysis will focus on the chance that $x_i$ (the less ambiguous or more normal case) is considered more uncertain than $x_j$ (the more ambiguous case). If the following theorem holds, then those correctly ranked by \textsc{var} are also likely to be correctly ranked by \textsc{mean}, indicating that \textsc{mean} is a preferable uncertainty metric to \textsc{var}.

\begin{lemma}\label{lm:sample-uncertainty}
Consider two inputs $x_i, x_j$ with uncertainty score $s(x_i)$ and $s(x_j)$ estimated from $K$ \emph{i.i.d.} ensemble learners, and denote by $\Delta_{ij}(s) := \mathbb{E}[s(x_j)-s(x_i)]$ the difference of expected uncertainty score.
If $\Delta_{ij}(s) > 0$, then $\Pr\left({s(x_i)>s(x_j)}\right) = \mathcal{O}\left(\frac{\text{Var}(s(x_i)+\text{Var}(s(x_j))}{\Delta_{ij}^2(s)}\right)$.
\end{lemma}
The proof is provided below. Intuitively, Lemma~\ref{lm:sample-uncertainty} states that if input $x_j$ is more uncertain than $x_i$ w.r.t. the expected uncertainty $\mathbb{E}[s(\cdot)]$, then the probability of the sample uncertain measure $s$ making a wrong decision is bounded. Based on such result, we establish the following error bounds for uncertainty metrics \textsc{mean} and \textsc{var}.
\begin{proof}
Let $\rvdiff=s(x_j)-s(x_i)$ be a random variable, where $s(x_i)$ and $s(x_j)$ denotes the uncertainty score of $x_i$ and $x_j$ estimated from $K$ i.i.d. ensemble learners. Therefore $\Delta_{ij}(s) = \expct{\rvdiff} > 0$.
By Chebyshev's Inequality, we obtain
\begin{align*}
\Pr\left( \abs{\rvdiff - \expct{\rvdiff}} \geq \Delta_{ij}(s) \right) \leq \frac{\text{Var}(\rvdiff)}{\Delta_{ij}^2(s)}
\end{align*}
which implies that
\begin{align*}
\Pr\left( \rvdiff - \expct{\rvdiff} \leq -\Delta_{ij}(s) \right)
= \Pr\left( \rvdiff - \Delta_{ij}(s) \leq -\Delta_{ij}(s) \right)
= \Pr\left( s(x_j)-s(x_i) \leq 0 \right)
\leq \frac{\text{Var}(\rvdiff)}{\Delta_{ij}^2(s)} 
\end{align*}
% \begin{align*}
% \Pr\left( \rvdiff - \expct{\rvdiff} \leq -\Delta_{ij}(s) \right)
% = &\Pr\left( \rvdiff - \Delta_{ij}(s) \leq -\Delta_{ij}(s) \right) \\
% = &\Pr\left( s(x_j)-s(x_i) \leq 0 \right) \\
% = &\Pr\left( s(x_i) \geq s(x_j) \right) \\
% \leq &\frac{\text{Var}(\rvdiff)}{\Delta_{ij}^2(s)} 
% \end{align*}
Further noticing that $\text{Var}(\rvdiff) = \text{Var}(s(x_j)-s(x_i)) = \text{Var}(s(x_j)) + \text{Var}(s(x_i))$, we conclude that
\begin{align*}
\Pr\left(s(x_i) > s(x_j)\right) = \mathcal{O}\left(\frac{\text{Var}(s(x_i))+\text{Var}(s(x_j))}{\Delta_{ij}^2(s)}\right)  
\end{align*}
which completes the proof.
\end{proof}

Based on Lemma~\ref{lm:sample-uncertainty}, below we provide the proof of Theorem~\ref{thm:mean-vs-var-beta-finite}.
\begin{proof}
To prove the first statement, i.e. $\Delta_{ij}(s_\textsc{mean}) > \Delta_{ij}(s_\textsc{var}) > 0$, we consider the following properties of a beta distribution $\mathcal{B}(\alpha,\beta)$. 
\begin{align*}
\mu = \frac{\alpha}{\alpha + \beta},\quad
\sigma = \frac{\alpha \beta}{(\alpha+\beta)^2 (1+\alpha+\beta)} = \frac{\mu(1-\mu)}{1+\alpha+\beta}
\end{align*}
where $\mu_i$ and $\sigma_i$ respectively represent the mean and variance of the beta distribution $\mathcal{B}(\alpha_i, \beta_i)$.

Let $\alpha_i + \beta_i = \alpha_j + \beta_j = c$. Since $\alpha_i < \alpha_j \leq \beta_j$, we know 
\begin{align*}
    \mu_i = \frac{\alpha_i}{\alpha_i + \beta_i} < \frac{\alpha_j}{\alpha_j + \beta_j} = \mu_j \leq \frac{1}{2}, \quad \sigma_i = \frac{\mu_i(1-\mu_i)}{1+\alpha_i+\beta_i} < \frac{\mu_j(1-\mu_j)}{1+\alpha_j+\beta_j} = \sigma_j
\end{align*}
Therefore, we have 
\begin{align*}
    \Delta_{ij}(s_\textsc{mean}) &= \expct{s_\textsc{mean}(x_j) - s_\textsc{mean}(x_i)} = \mu_j - \mu_i> 0,\\
    \Delta_{ij}(s_\textsc{var}) &= \expct{s_\textsc{var}(x_j) - s_\textsc{var}(x_i)} = \sigma_j - \sigma_i > 0.
\end{align*}
Furthermore, notice that
\begin{align*}
    \Delta_{ij}(s_\textsc{var}) 
    =  \frac{\mu_j(1-\mu_j) - \mu_i(1-\mu_i)}{1+c}
    < \mu_j(1-\mu_j) - \mu_i(1-\mu_i)
    = \mu_j-\mu_i - (\mu_j^2 - \mu_i^2)
    < \Delta_{ij}(s_\textsc{mean}),
\end{align*}
% \begin{align*}
%     \Delta_{ij}(s_\textsc{var}) 
%     &=  \frac{\mu_j(1-\mu_j) - \mu_i(1-\mu_i)}{1+c} \\
%     &< \mu_j(1-\mu_j) - \mu_i(1-\mu_i) \\
%     &= \mu_j-\mu_i - (\mu^2(x_j) - \mu^2(x_i)) \\
%     &< \Delta_{ij}(s_\textsc{mean}),
% \end{align*}
which proves the first statement of Theorem~\ref{thm:mean-vs-var-beta-finite}. 

To prove the second statement, i.e., to provide an upper bound on the errors of $s_\textsc{mean}$ and $s_\textsc{var}$, we plug in the definition of $s_\textsc{mean}$ and $s_\textsc{var}$ to Lemma~\ref{thm:mean-vs-var-beta-finite}:
\begin{align*}
    \Pr\left(s_\textsc{mean}(x_i) > s_\textsc{mean}(x_j)\right)
    =\,&\mathcal{O}\left(\frac{\text{Var}(s_\textsc{mean}(x_i))+\text{Var}(s_\textsc{mean}(x_j))}{\Delta_{ij}^2(s_\textsc{mean})}\right)\\
    \stackrel{(a)}{=}\,&\mathcal{O}\left(\frac{\sigma_j+\sigma_i}{K \Delta_{ij}^2(s_\textsc{mean})}\right) \\
    =\,&\mathcal{O}\left(\frac{1}{K \Delta_{ij}^2(s_\textsc{mean})}\right)
\end{align*}
where step (a) is due to $\text{Var}(s_\textsc{mean}(x_i)) = \sigma_i/n$. Similarly,
\begin{align*}
    \Pr\left(s_\textsc{var}(x_i) > s_\textsc{var}(x_j)\right)
    =\mathcal{O}\left(\frac{\text{Var}(s_\textsc{var}(x_i))+\text{Var}(s_\textsc{var}(x_j))}{\Delta_{ij}^2(s_\textsc{var})}\right)
    \stackrel{(b)}{=}\mathcal{O}\left(\frac{1}{K \Delta_{ij}^2(s_\textsc{var})}\right)
\end{align*}
Here, step (b) is due to the variance of sample variance $\text{Var}(s_\textsc{var}(x_i)) = \frac{1}{K} (\mu_4 - \sigma^2(x_i)) + \mathcal{O}(n^{-2}) = \mathcal{O}\left(\frac{1}{K}\right)$~\cite{cho2008variance} where $\mu_4$ is the Kurtosis of the Beta distribution $\mathcal{B}(\alpha_i,\beta_i)$.
\end{proof}

A direct corollary of the above theorem states that under infinite ensemble size,  using either \textsc{mean} or \textsc{var} as the uncertainty metric does not make a difference. 
\begin{corollary}\label{thm:mean-vs-var-beta}
If the sample size is infinite, then under the conditions of Theorem~\ref{thm:mean-vs-var-beta-finite}, we have $s_\textsc{mean}(x_i)<s_\textsc{mean}(x_j) \Leftrightarrow s_\textsc{var}(x_i)<s_\textsc{var}(x_j)$.
\end{corollary}

\section{Dataset Descriptions}\label{sec:dataset-descriptions}

\subsection{ASHRAE RP-1043 Chiller Dataset}\label{sec:rp-1043}
We used the ASHRAE~RP-1043 Dataset~\cite{comstock1999development} (hereinafter referred to as the ``chiller dataset'') to examine the proposed ensemble approach for detecting incipient anomalies. In the chiller dataset, sensor measurements of a typical cooling system---a 90-ton centrifugal water-cooled chiller---were recorded under both anomaly-free and various fault conditions. In this study, we included the six faults (FWE, FWC, RO, RL, CF, NC) used in our previous study~\cite{jin2019detecting} as the anomaly (positive) class; each fault was introduced at four levels of severity (SL1--SL4, from the slightest to the severest). We consider SL3 and SL4 cases as severe faults, and SL1 and SL2 cases as incipient faults. For feature selection, we also followed our previous study~\cite{jin2019detecting} and used the sixteen key features therein for training our models. To give the readers an intuitive view about the distribution of the chiller data, we employed the \ac{LDA} algorithm to reduce the data into two dimensions, and visualized part of the reduced-dimension data in Fig.~\ref{fig:visualization-chiller} described earlier. We can observe a general trend in the visualization: data points will deviate further away from the normal cluster when the corresponding fault develops into a higher \ac{SL}.

\subsection{Kaggle Diabetic Retinopathy Dataset}\label{sec:kaggle-DR}
\begin{figure}[tb]
  \centering
  \begin{subfigure}[t]{0.19\linewidth}
    \centering
    \includegraphics[height=2cm]{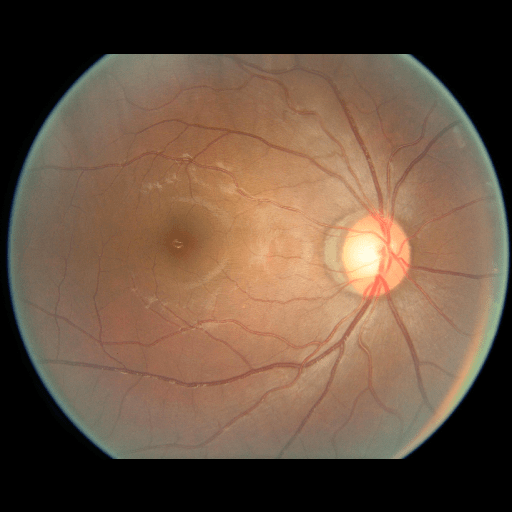}
    \includegraphics[trim=0.3cm 0.9cm 0cm 0.85cm,clip, height=2cm]{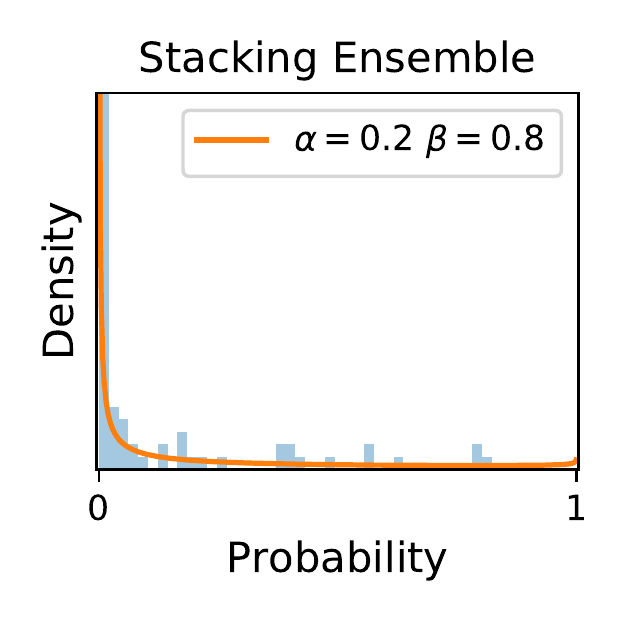}
    \caption{SL0 (No-DR)}
  \end{subfigure}
  \begin{subfigure}[t]{0.19\linewidth}
    \centering
    \includegraphics[height=2cm]{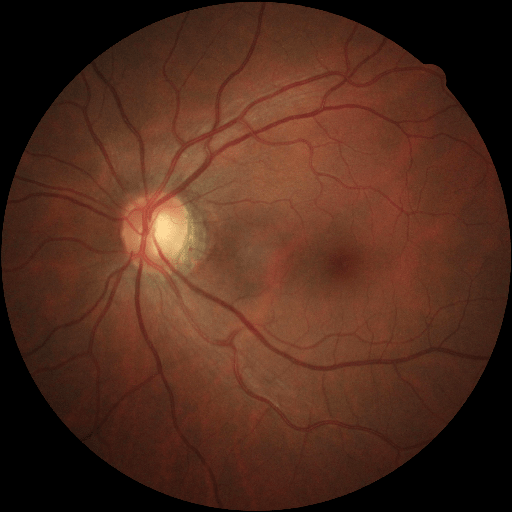}
    \includegraphics[trim=0.85cm 0.9cm 0cm 0.85cm,clip, height=2cm]{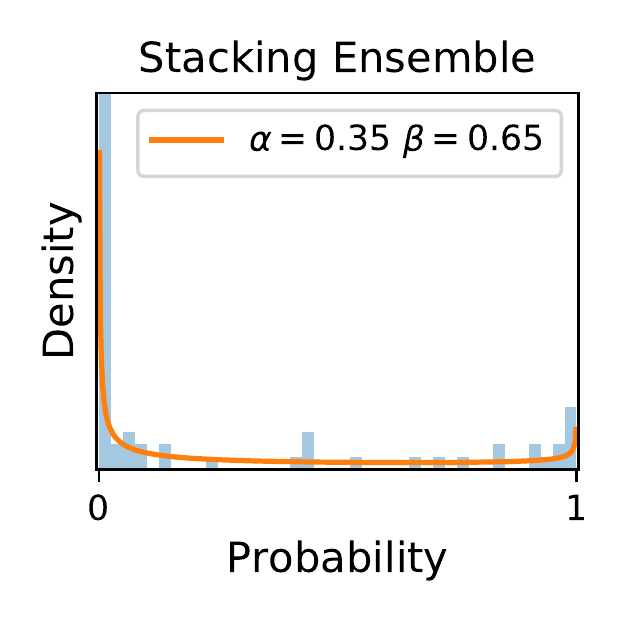}
    \caption{{SL1 (Mild)}}
  \end{subfigure}
  \begin{subfigure}[t]{0.19\linewidth}
    \centering
    \includegraphics[height=2cm]{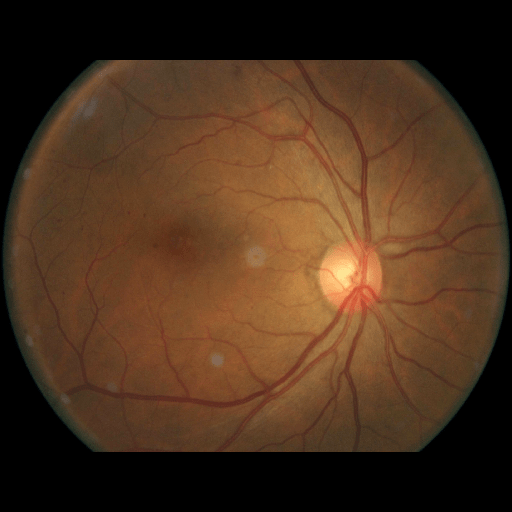}
    \includegraphics[trim=0.85cm 0.9cm 0cm 0.85cm,clip, height=2cm]{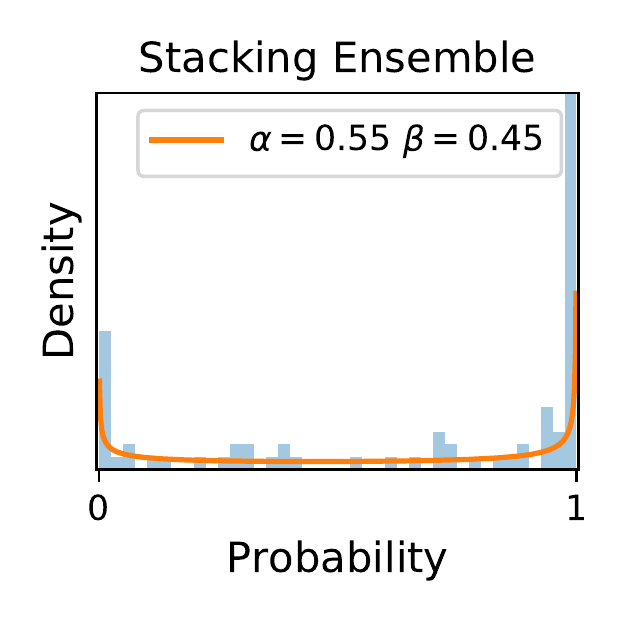}
    \caption{{SL2 (Moderate)}}
  \end{subfigure}
  \begin{subfigure}[t]{0.19\linewidth}
    \centering
    \includegraphics[height=2cm]{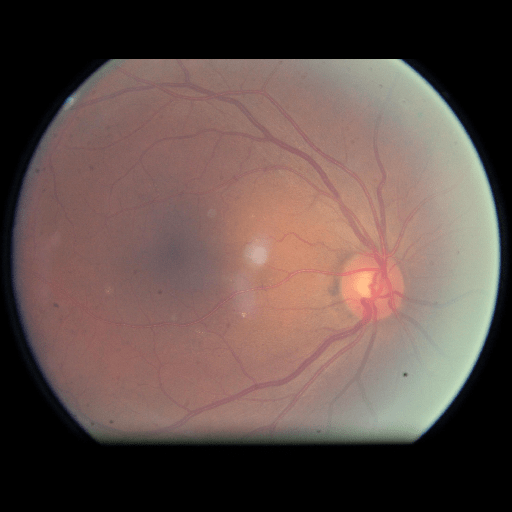}
    \includegraphics[trim=0.85cm 0.9cm 0cm 0.85cm,clip, height=2cm]{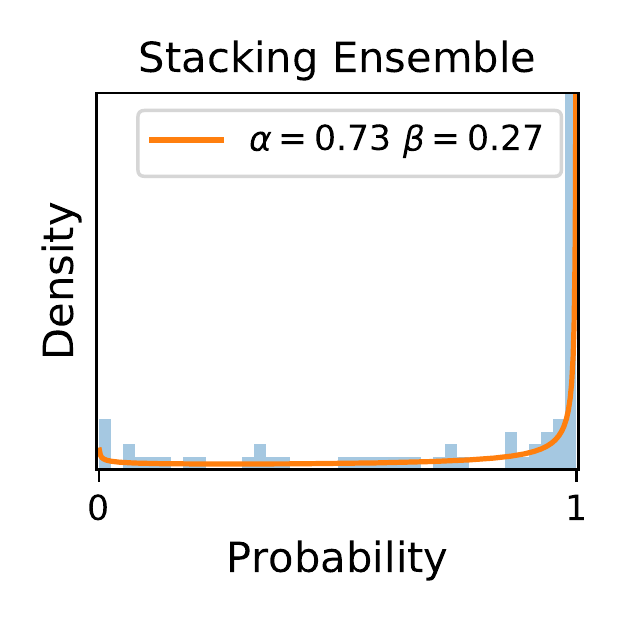}
    \caption{{SL3 (Severe)}}
  \end{subfigure}
  \begin{subfigure}[t]{0.19\linewidth}
    \centering
    \includegraphics[height=2cm]{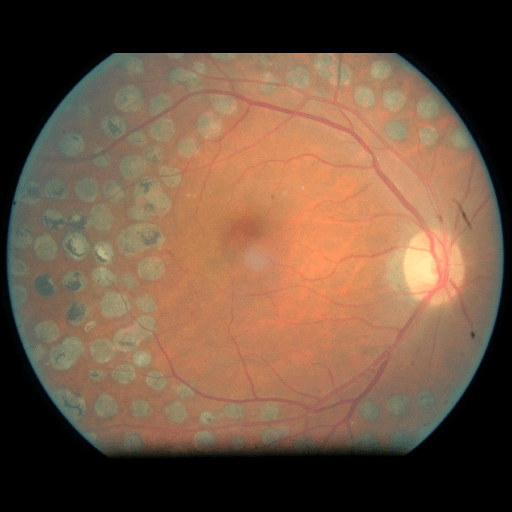}
    \includegraphics[trim=0.85cm 0.9cm 0cm 0.85cm,clip, height=2cm]{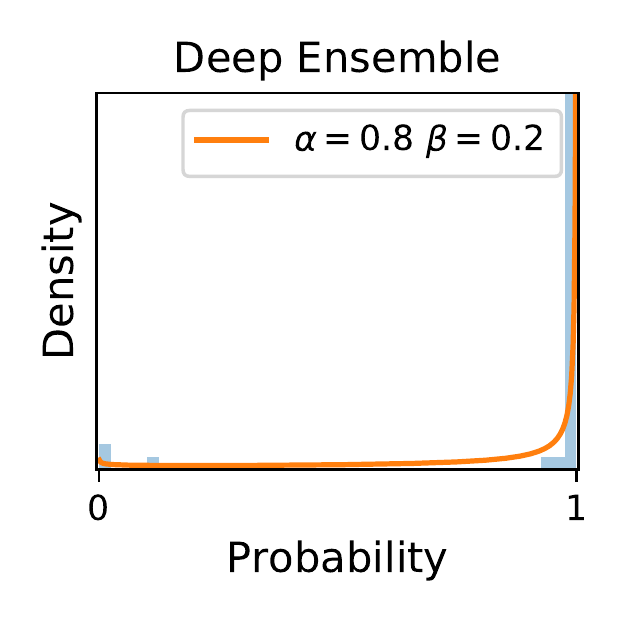}
    \caption{{SL4 (Proliferate)}}
  \end{subfigure}
\caption{Fundus images (top panel) of the five \acp{SL} of the \ac{DR} disease and the distributions (shown as histograms over probability bins) of their corresponding classifier predictions under ensembles (bottom panel). A beta distribution displayed by the orange curve is fitted to each distribution.}
\label{fig:beta-dist-five-SL}
\end{figure}

As another case study, we examined the efficacy of our proposed approach on a medical image dataset of \acp{DR} diseases. \ac{DR} is a common complication of the diabetic disease and the leading cause of blindness in the working-age population of the developed world~\cite{gulshan2016development}. The Kaggle-DR dataset~\cite{cuadros2009eyepacs} (hereinafter referred to as the ``\ac{DR} dataset'') comprises $88,702$ high resolution images. Similar to the above-mentioned chiller faults, the presence of \ac{DR} is also rated into five different \acp{SL}: \textit{no-DR} ({SL0}), \textit{mild} ({SL1}), \textit{moderate} ({SL2}), \textit{severe} ({SL3}), and \textit{proliferate} ({SL4}), as illustrated in Fig.~\ref{fig:beta-dist-five-SL}. 
Again, SL3 and SL4 are considered severe anomalies. It is worthy to note one key difference between the two datasets: the SL1 cases in the \ac{DR} dataset are considered non-referable disease type and thus belong to the negative class. 
% The data in the development set were used to train and validate individual deep learning models that will later be used to construct ensembles. 
% Similar to what we have done with the chiller dataset, we divided the \ac{DR} dataset into a development set and a test set, which respectively consisted of $58,125$ and $30,577$ images 

\subsection{Partitioning the Datasets}
\begin{figure}[tb]
    \centering
    \includegraphics[width=0.8\linewidth]{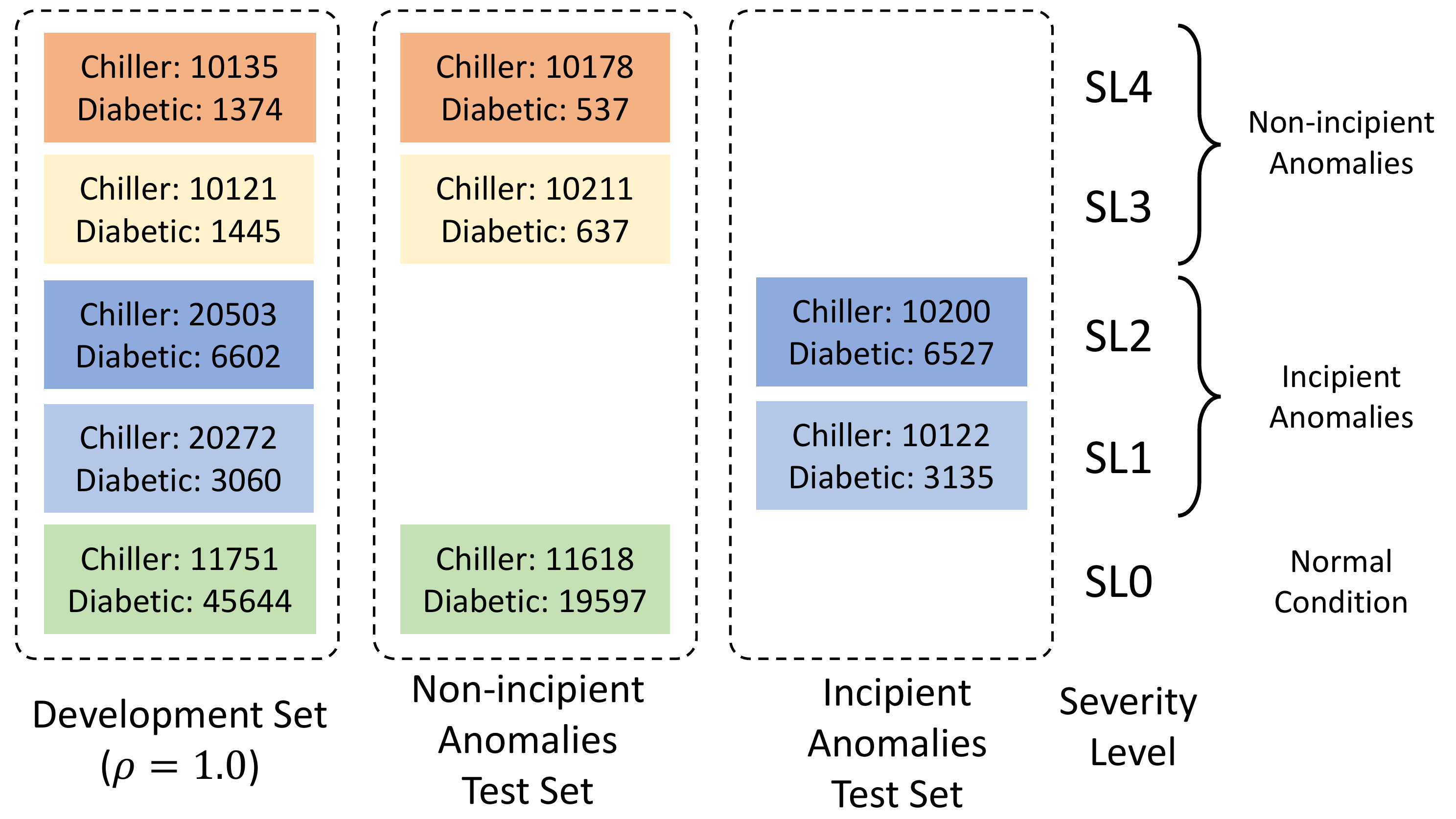}
    \caption{Layout of the development sets and test sets used in this study.}
    \label{fig:dataset-layout}
\end{figure}

We divided each dataset into a \textit{development set} and a \textit{test set}. The test set can be further divided into two parts; one contains only the normal data (SL0) and the non-incipient anomalies (SL3\,\&\,4), the other containing only the incipient anomalies; see Fig.~\ref{fig:dataset-layout} for an illustration. All five \acp{SL} were present in the development set data. To model how the availability of incipient anomaly data affected the detection performance, we introduce a parameter, the incipient anomaly ratio $\rho$, to control the proportion of incipient anomaly data that enters the development set. In our experiment, we tested $\rho=0, 0.2, 0.4, 0.6, 0.8, 1.0$. It is worthy to note that when $\rho=0$, no incipient anomaly data appeared in the development set; in other words, the incipient anomaly data became \ac{o.o.d.} because they were not present at training time. We specifically included this scenario to see if the models can learn useful knowledge from only non-incipient anomalies that is useful for identifying incipient anomalies. Further details about the two datasets as well as the preprocessing steps will be given in the appendix at the end of this paper.

\section{Experimental Study}\label{sec:experiment}

\subsection{Experimental Setup}
Real-world \ac{ML} practitioners perform extensive \textit{model selection} to search for models on the development set data, and pick those that are more likely to perform well on test sets. We employed a similar workflow in our empirical study. For each model class under study, we swept over a wide range of hyperparameter settings, picked out a set of best-performing hyperparameters, and assess whether or not our proposed ensemble method could deliver consistent performance improvement compared to the baseline scenarios. 

We experimented using \ac{DT} and \ac{NN} base learners to construct ensembles for the chiller dataset where sensor data assume a tabular form. For the \ac{DR} dataset, we trained multiple \ac{CNN} models of different architectures and data augmentation settings, and combined them into ensembles. Each ensemble model only consisted of base learners of the same type. In our empirical study, we evaluated ensembles of four different sizes: $5$, $10$, $15$ and $25$, and compared their performance to the single learner case. Further implementation details will be deferred to the appendix of this paper. Next, we will report the experimental results.

\subsection{Detection Performance of Ensemble Classifiers}

\subsubsection{\acf{FNR}}
\begin{figure}[tb]
  \centering
  \begin{subfigure}[t]{0.48\linewidth}
    \centering
    \includegraphics[height=3.5cm]{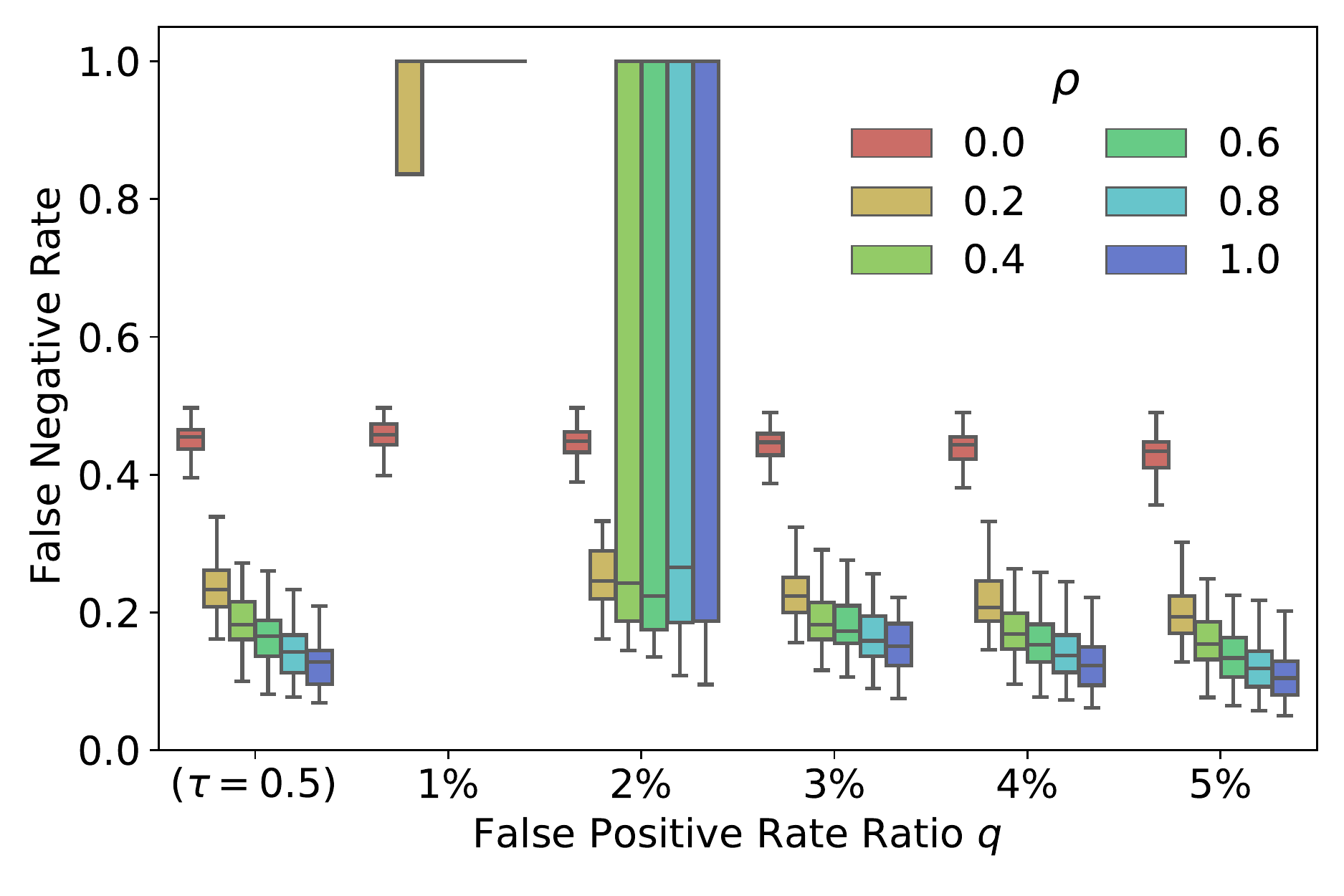}
    \caption{\acs{DT} ($K=1$) + chiller dataset}
  \end{subfigure}
  \begin{subfigure}[t]{0.48\linewidth}
    \centering
    \includegraphics[height=3.5cm]{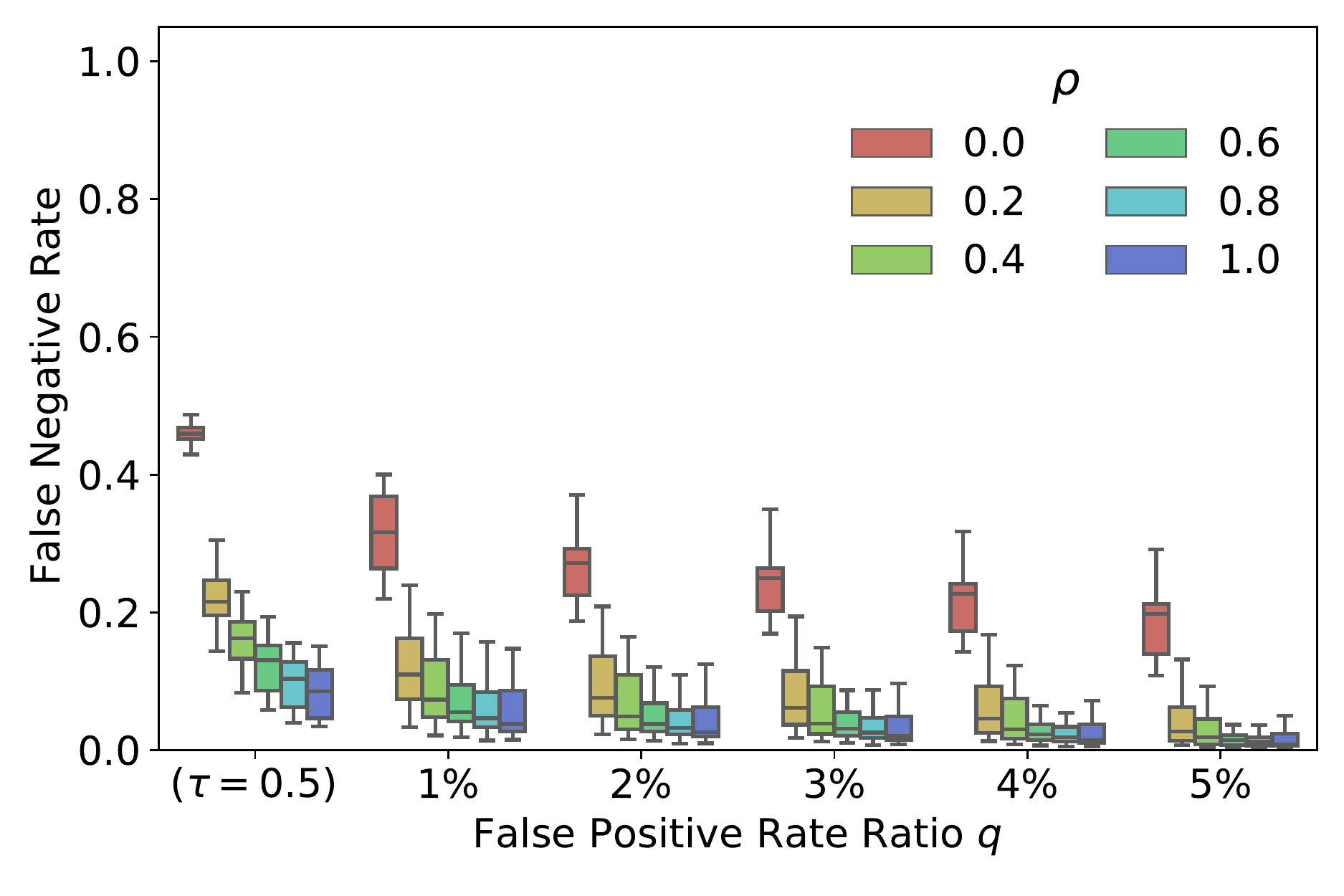}
    \caption{\acs{DT} ensemble ($K=25$) + chiller dataset}
  \end{subfigure}
  
  \begin{subfigure}[t]{0.48\linewidth}
    \centering
    \includegraphics[height=3.5cm]{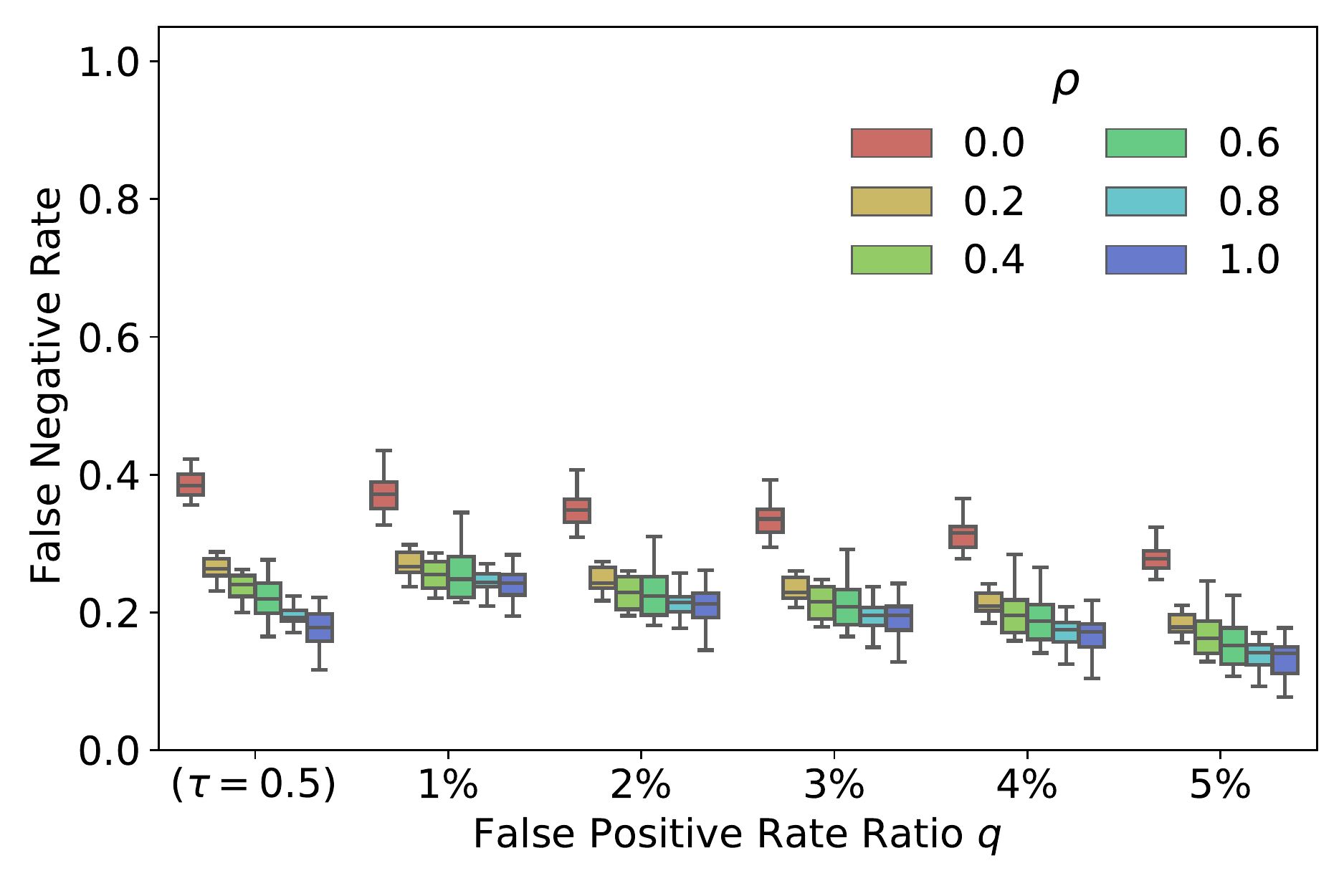}
    \caption{\acs{NN} ($K=1$) + chiller dataset}
  \end{subfigure}
  \begin{subfigure}[t]{0.48\linewidth}
    \centering
    \includegraphics[height=3.5cm]{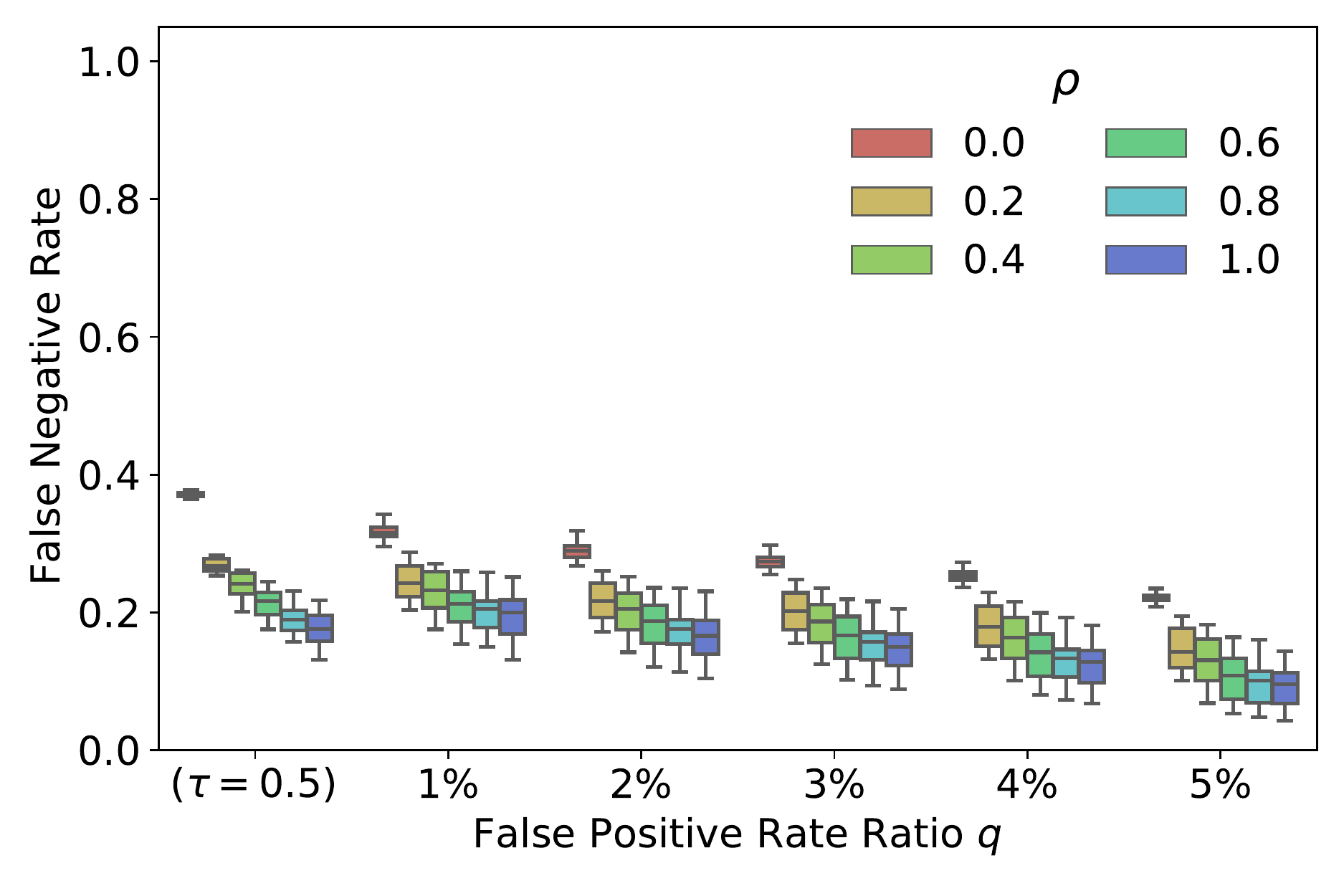}
    \caption{\acs{NN} ($K=25$) + chiller dataset}
  \end{subfigure}
  
  \begin{subfigure}[t]{0.48\linewidth}
    \centering
    \includegraphics[height=3.5cm]{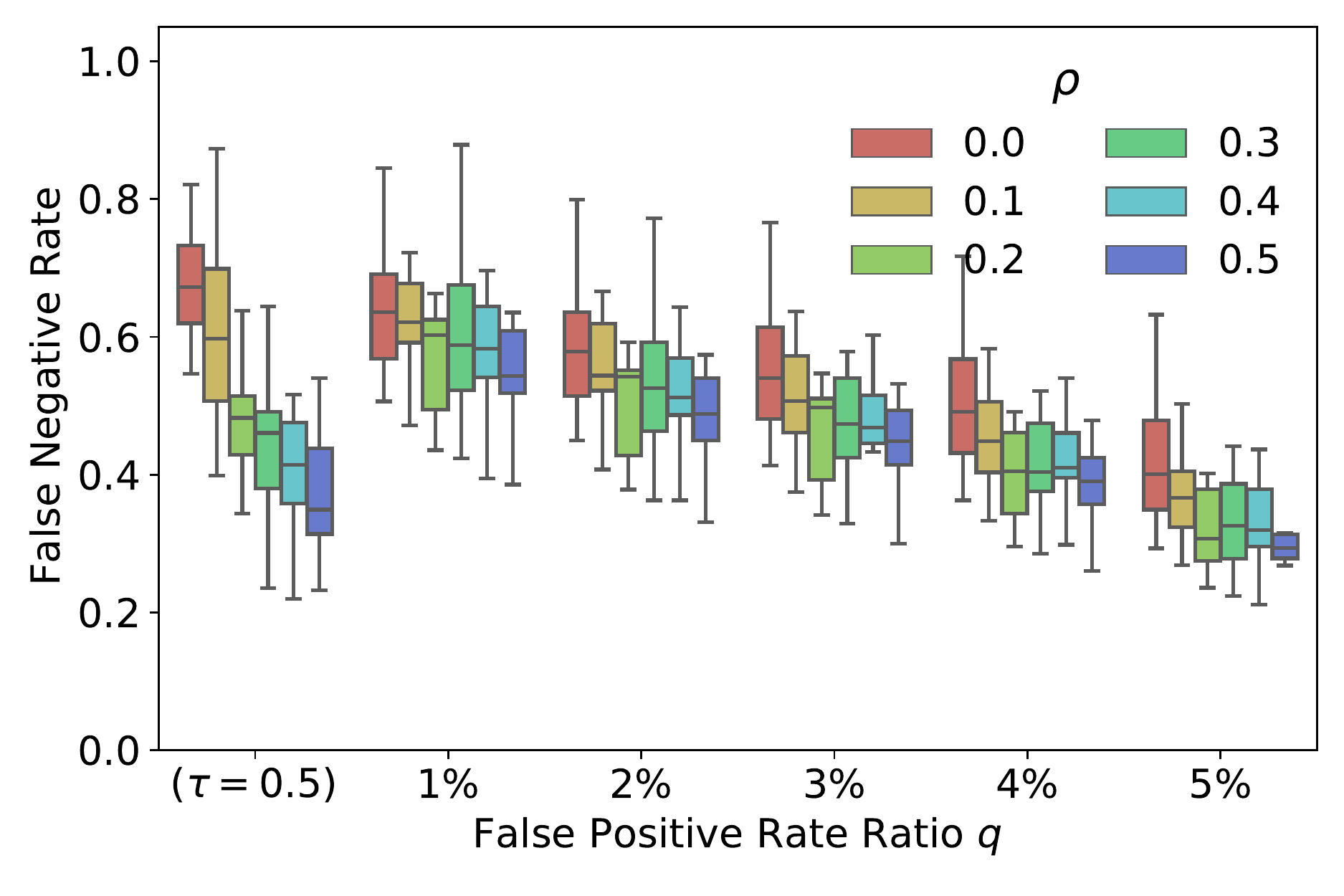}
    \caption{\acs{NN} ($K=1$) + \acs{DR} dataset}
  \end{subfigure}
  \begin{subfigure}[t]{0.48\linewidth}
    \centering
    \includegraphics[height=3.5cm]{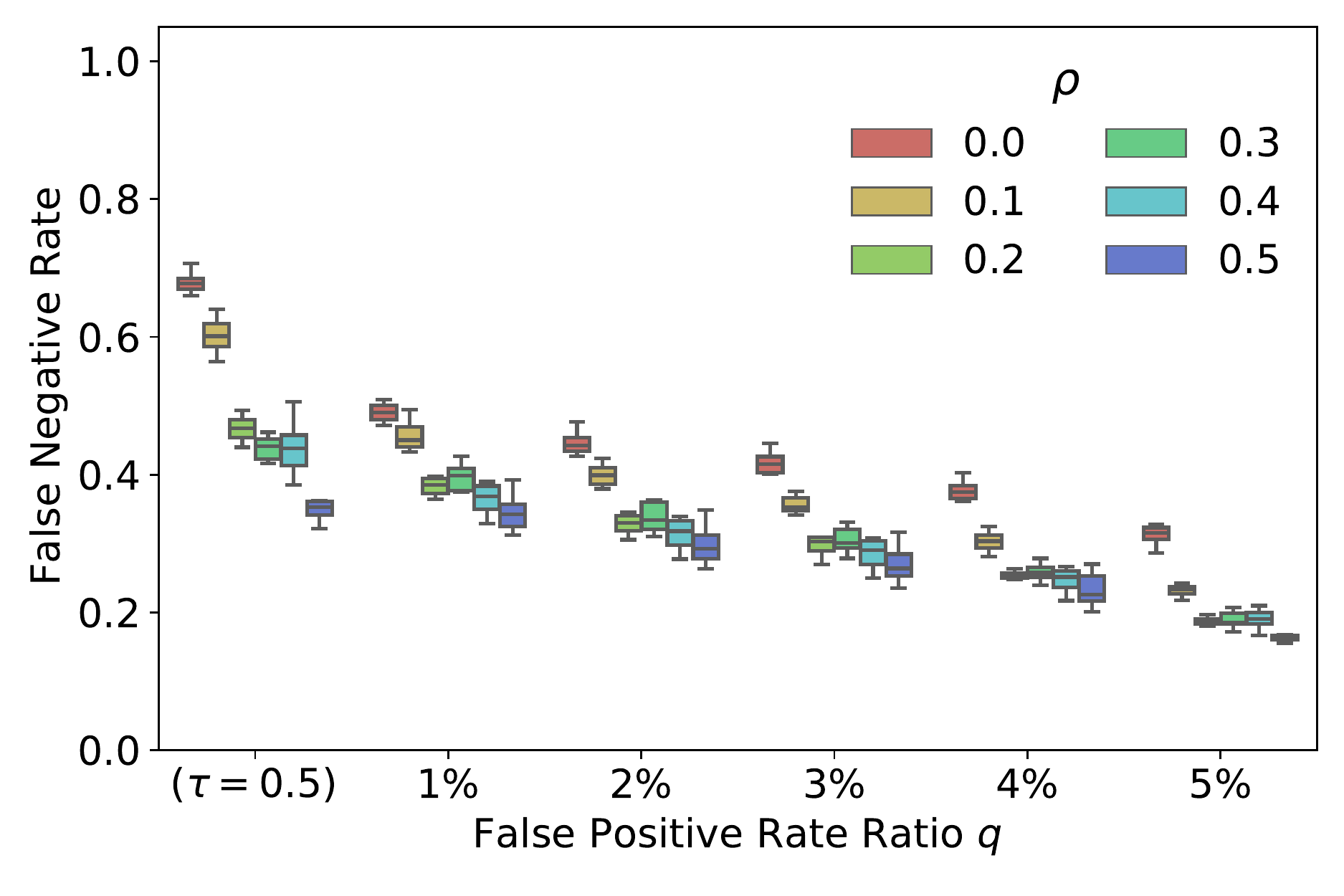}
    \caption{\acs{NN} ($K=25$) + \acs{DR} dataset}
  \end{subfigure}
  \caption{Detection performance in terms of \ac{FNR} on incipient anomalies for single learners ($K=1$) and for ensemble models ($K=25$). In plot (a) we see large \ac{FNR} with $q=1\%$ and $q=2\%$, which indicates that the resulting classifiers are too sensitive under small $q$ values so that a lot of false negatives decisions appear.}
  \label{fig:FNR-fpr-pct}
\end{figure}

% The \ac{FNR} is defined as the fraction of positive examples that are wrongly classified as negative, i.e.
% \begin{align}
%     \text{False Negative Rate} = \frac{FN}{TP + FN}.\notag
% \end{align}
We first report the detection performance of the trained ensembles on the two datasets in terms of \ac{FNR}. We examine the \ac{FNR} for both incipient and non-incipient anomalies under different settings of \ac{FPR} percentile $q$ and the incipient anomaly ratio $\rho$, and show the results as box plots in Fig.~\ref{fig:FNR-fpr-pct}. We only show the \ac{FNR} results on incipient anomaly cases for single learners (left panel) and for ensemble learners of size $25$ (right panel); the \ac{FNR} for non-incipient anomalies are all close to zero, which indicates near-perfect classification performance between SL0 (normal conditions) and SL3 \& SL4 (non-incipient anomalies). The results for non-incipient anomalies are not displayed here due to limited space. By comparing the two cases ($K=1$ vs. $K=25$), we can immediately see performance improvement for ensemble learners over single learners. In addition, we can observe a decreasing trend in \ac{FNR} with increasing $q$, which indicates that more incipient anomalies can be detected when we lower the detection threshold $\tau$; in other words, more incipient anomalies can be detected when the classifiers are working at more sensitive operating points. 

\subsubsection{Remaining False Negatives}
\begin{figure}[tb]
  \centering
  \captionsetup{justification=centering}
  \begin{subfigure}[t]{0.32\linewidth}
    \centering
    \includegraphics[height=3cm]{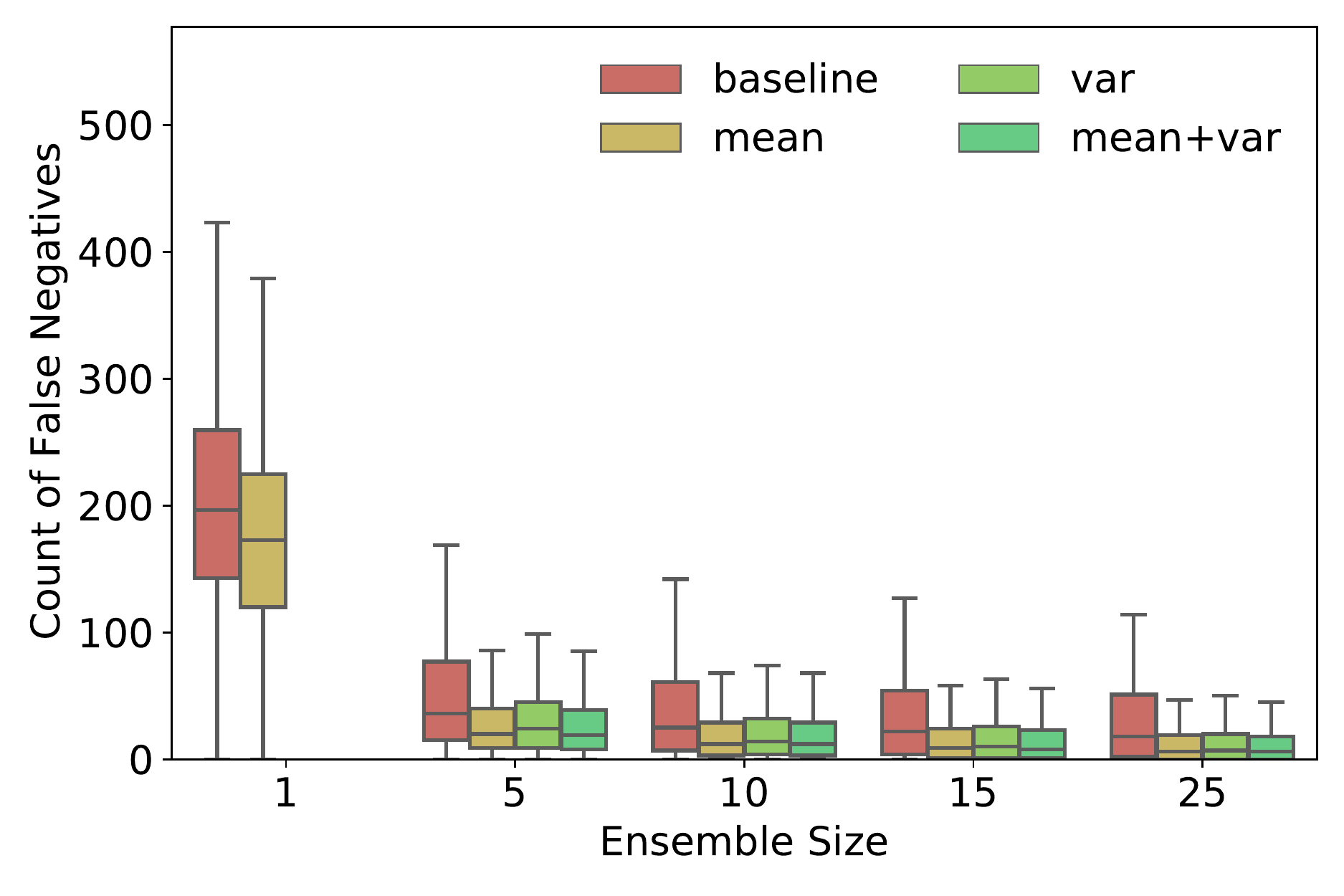}
    \caption{\acs{DT}  + non-incipient ($\rho=0$)}
  \end{subfigure}
  \begin{subfigure}[t]{0.32\linewidth}
    \centering
    \includegraphics[height=3cm]{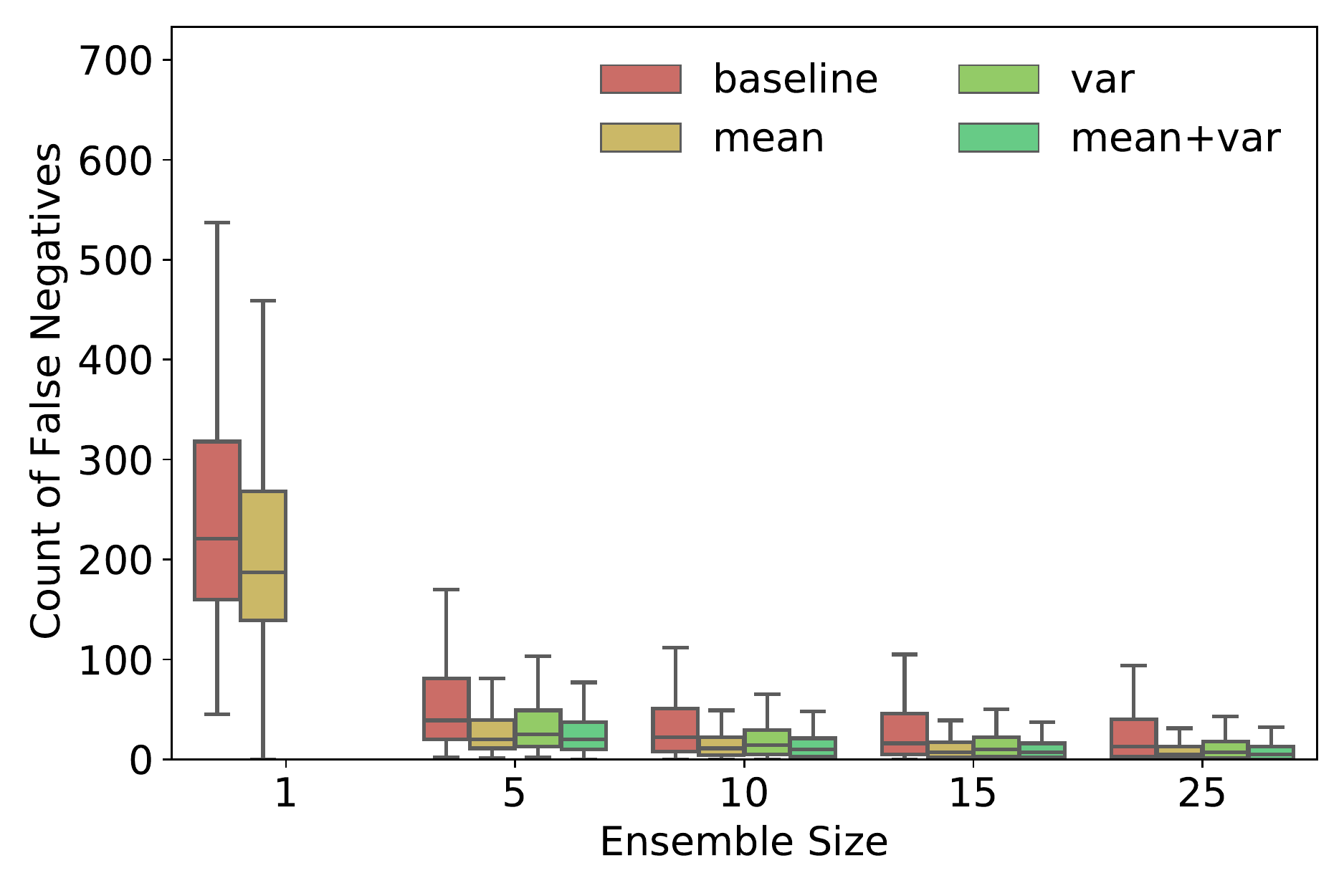}
    \caption{\acs{DT}  + non-incipient ($\rho=0.2$)}
  \end{subfigure}
  \begin{subfigure}[t]{0.32\linewidth}
    \centering
    \includegraphics[height=3cm]{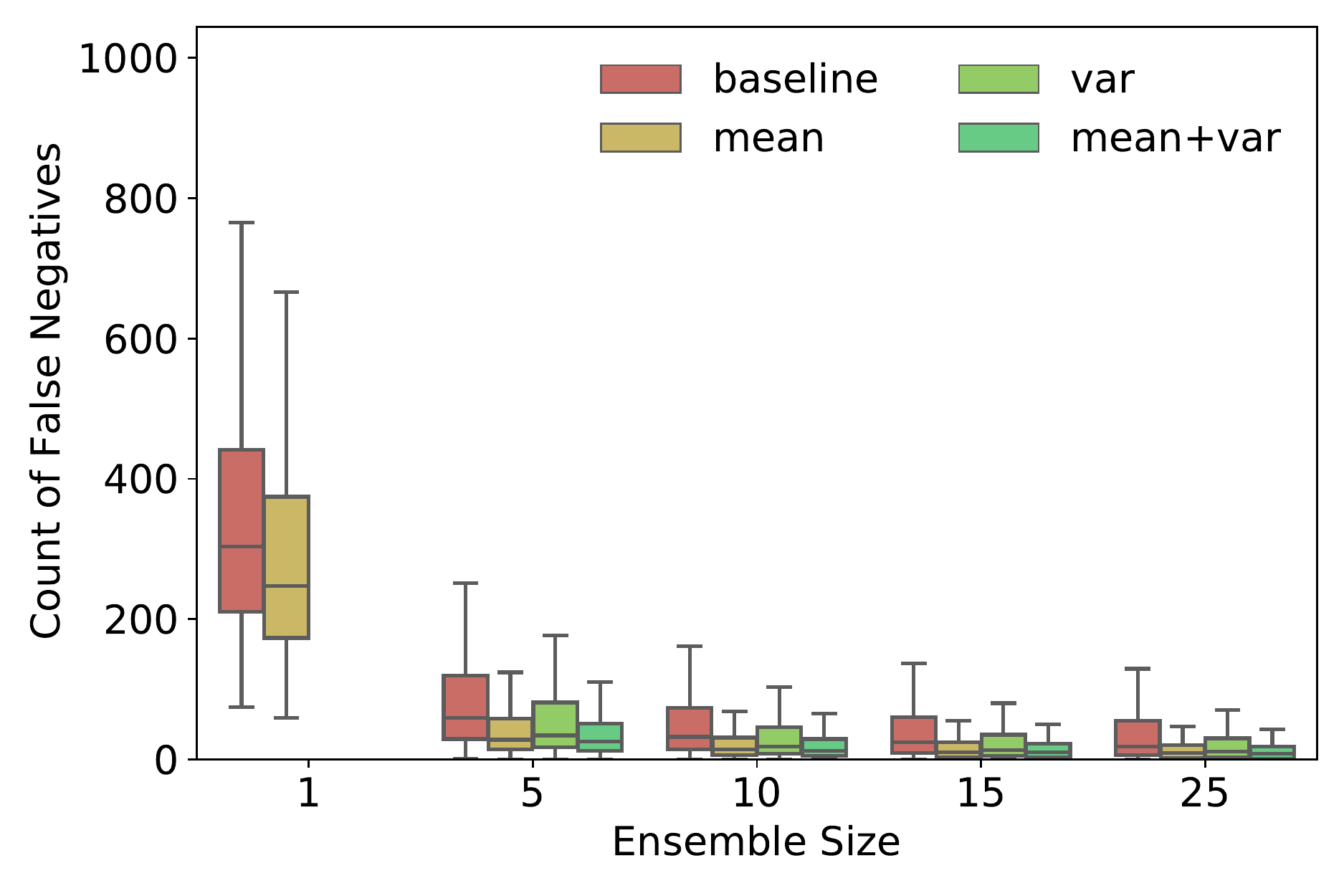}
    \caption{\acs{DT}  + non-incipient ($\rho=1.0$)}
  \end{subfigure}
  
  \begin{subfigure}[t]{0.32\linewidth}
    \centering
    \includegraphics[height=3cm]{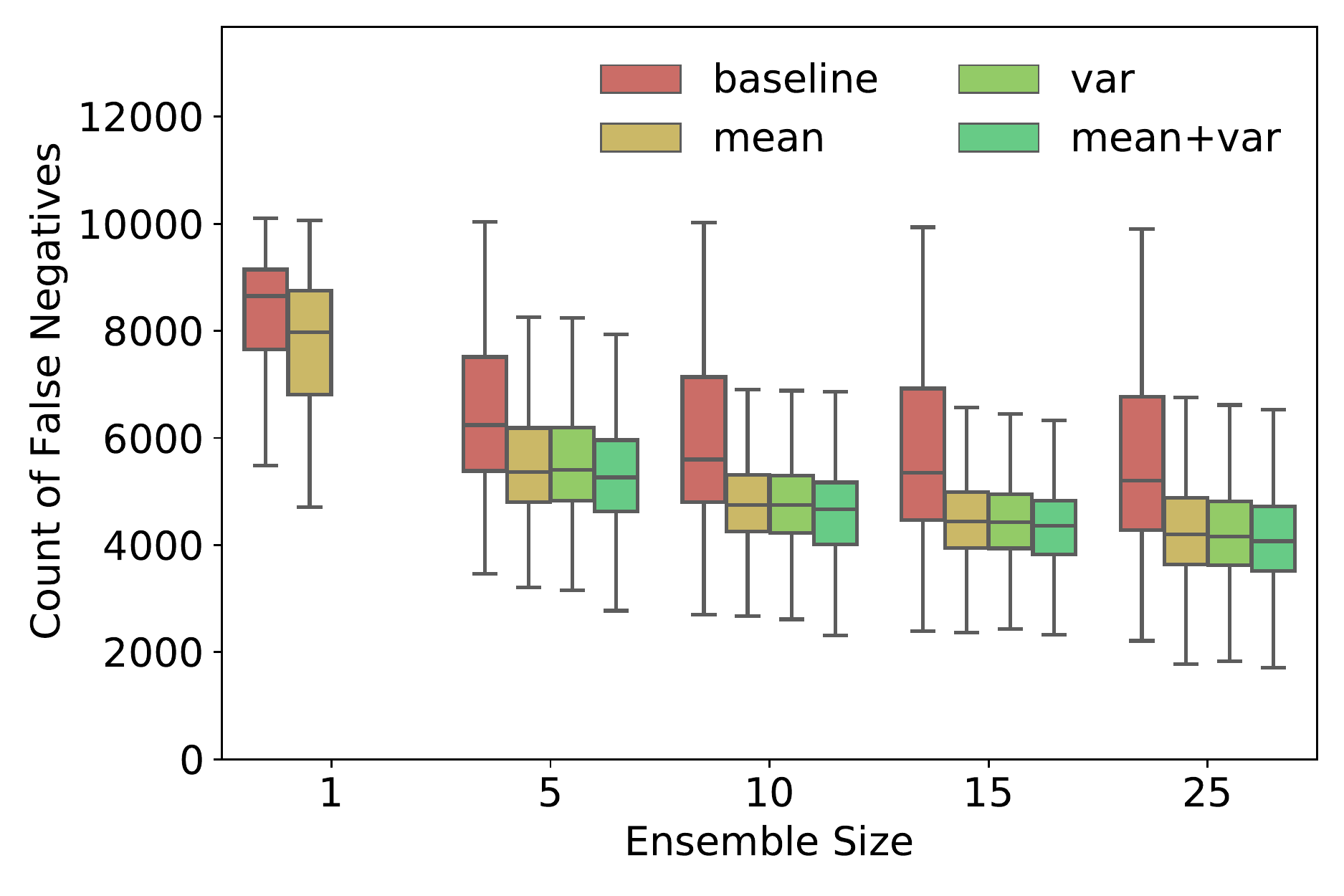}
    \caption{\acs{DT}  + incipient ($\rho=0$)}
  \end{subfigure}
  \begin{subfigure}[t]{0.32\linewidth}
    \centering
    \includegraphics[height=3cm]{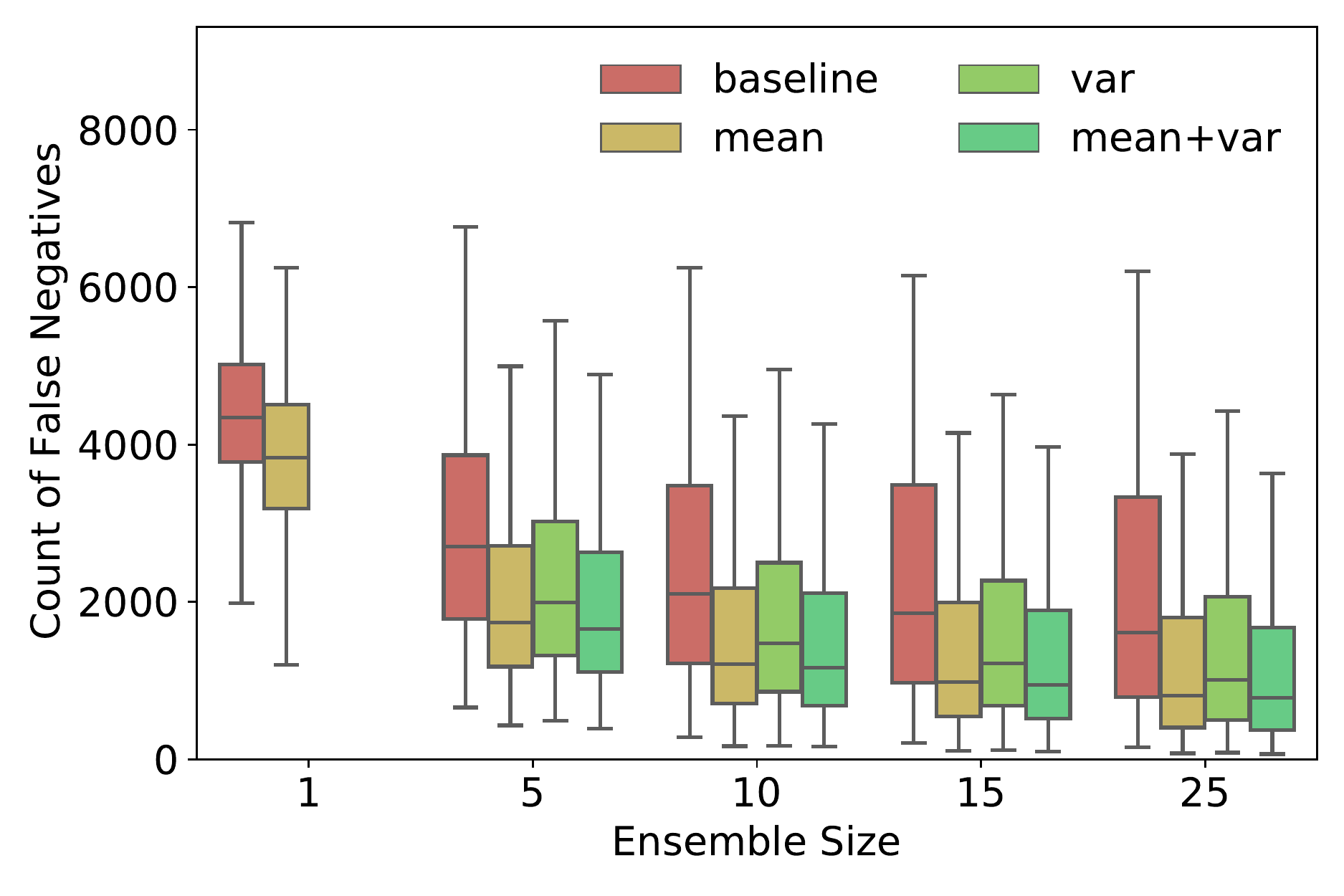}
    \caption{\acs{DT}  + incipient ($\rho=0.2$)}
  \end{subfigure}
  \begin{subfigure}[t]{0.32\linewidth}
    \centering
    \includegraphics[height=3cm]{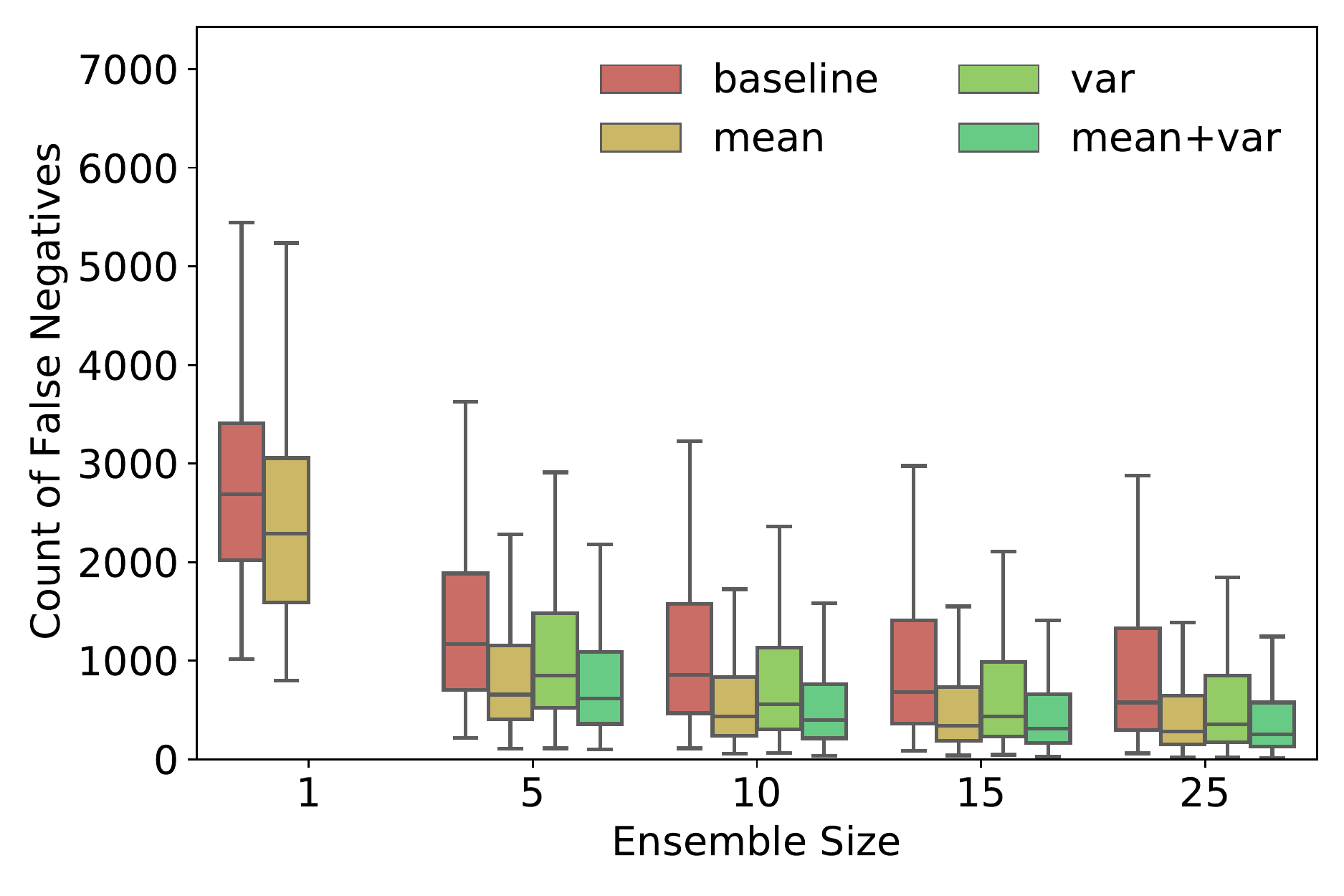}
    \caption{\acs{DT}  + incipient ($\rho=1.0$)}
  \end{subfigure}
  
  \begin{subfigure}[t]{0.32\linewidth}
    \centering
    \includegraphics[height=3cm]{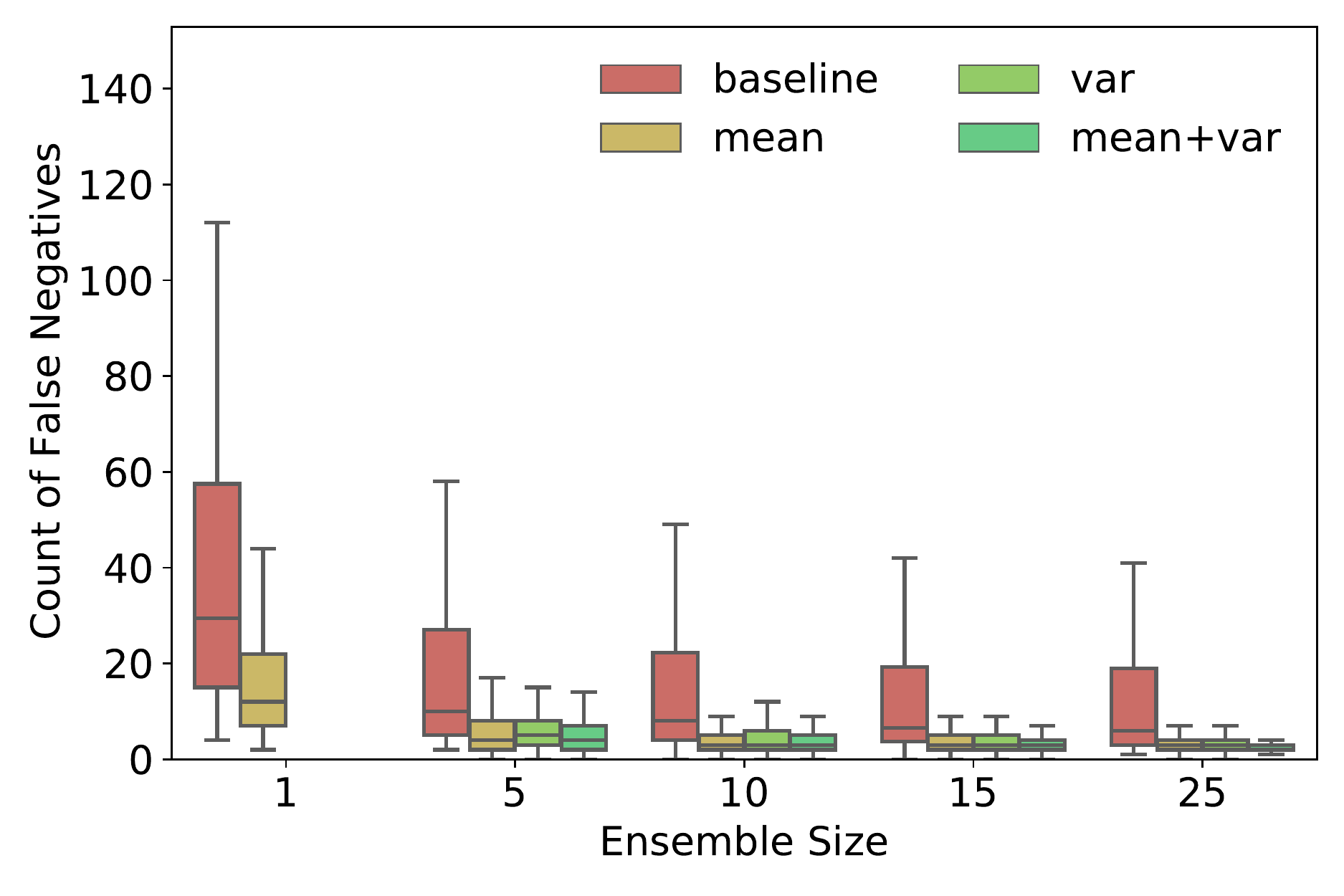}
    \caption{\acs{NN} + non-incipient ($\rho=0$)}
  \end{subfigure}
  \begin{subfigure}[t]{0.32\linewidth}
    \centering
    \includegraphics[height=3cm]{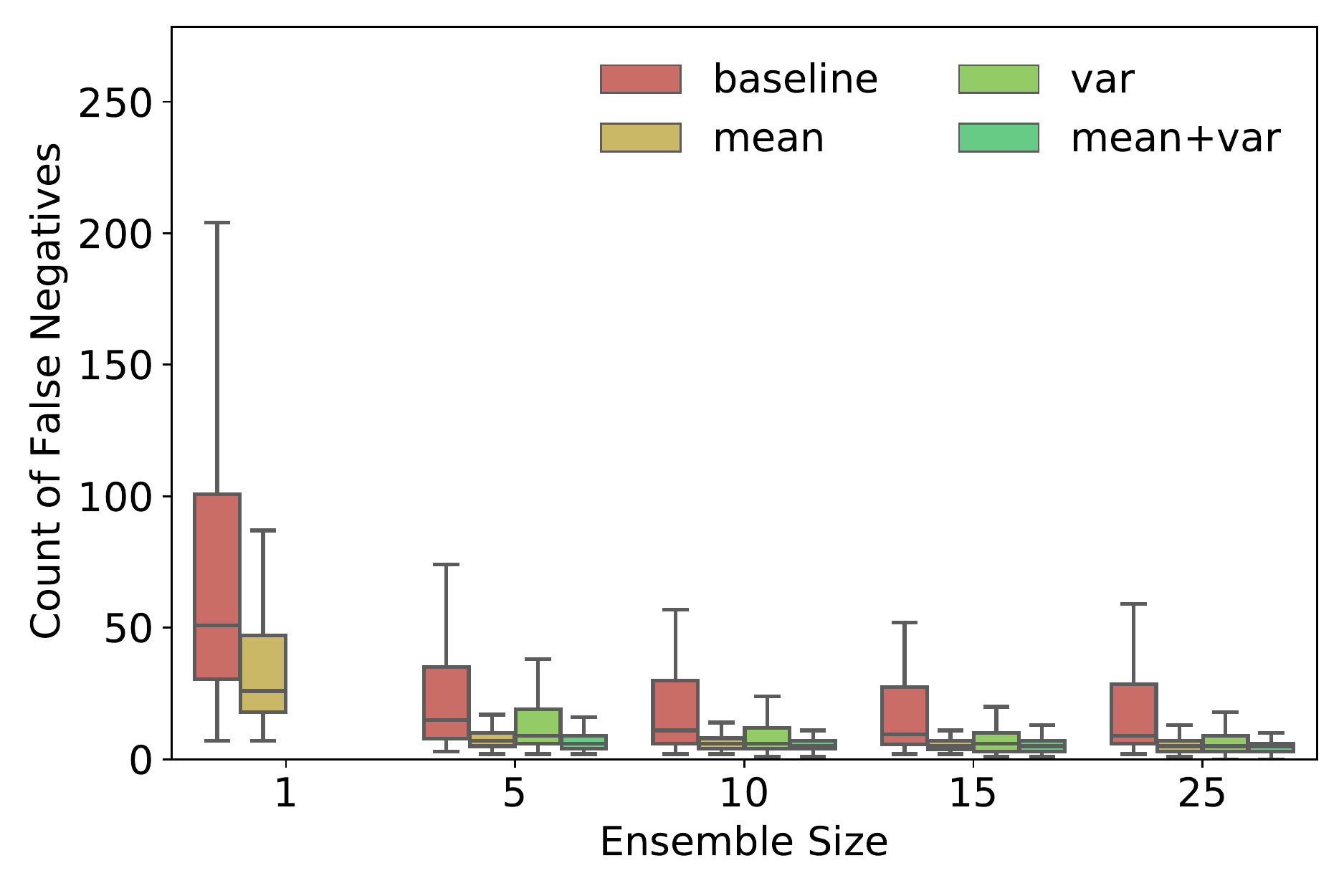}
    \caption{\acs{NN} + non-incipient ($\rho=0.2$)}
  \end{subfigure}
  \begin{subfigure}[t]{0.32\linewidth}
    \centering
    \includegraphics[height=3cm]{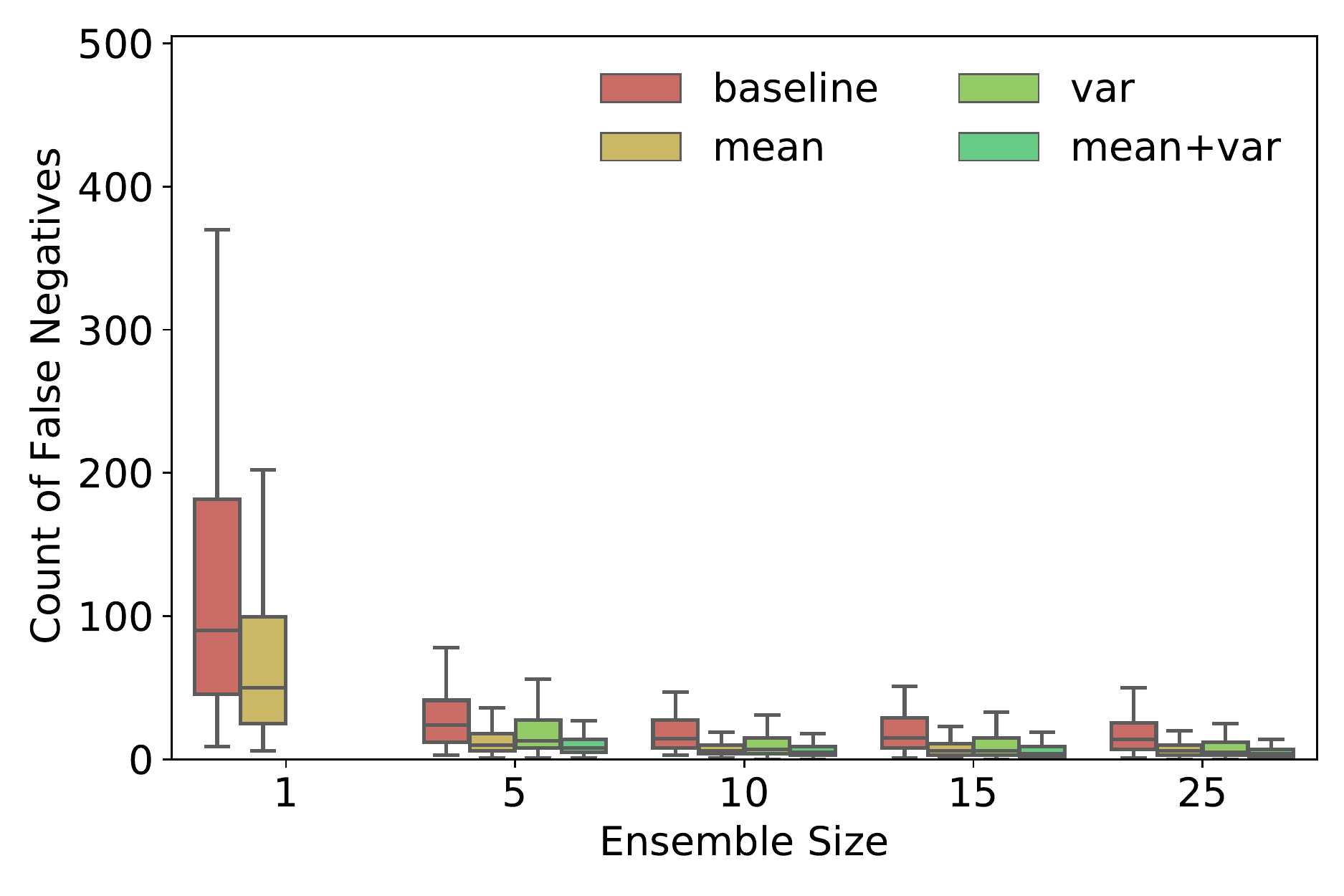}
    \caption{\acs{NN} + non-incipient ($\rho=1.0$)}
  \end{subfigure}
  
  \begin{subfigure}[t]{0.32\linewidth}
    \centering
    \includegraphics[height=3cm]{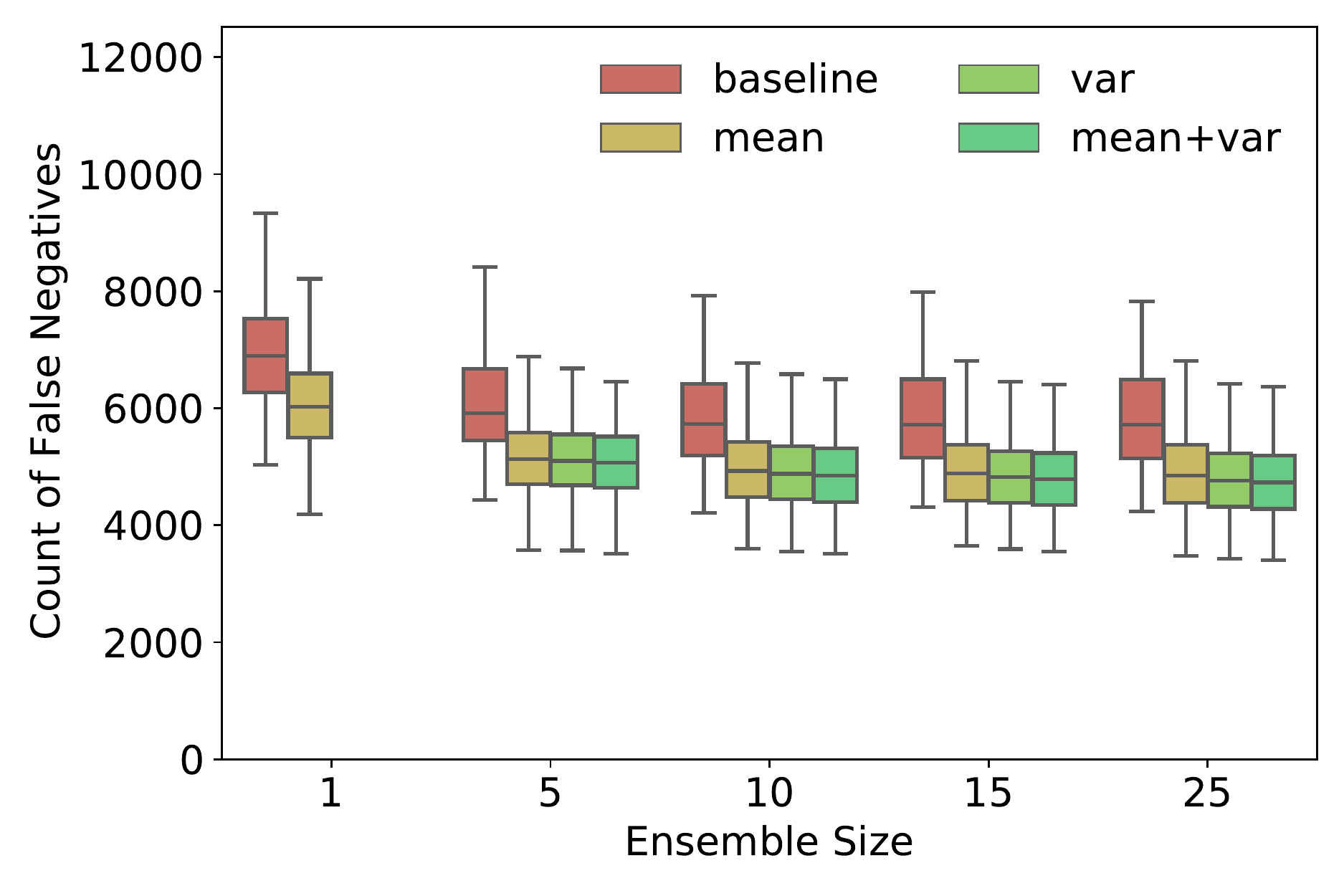}
    \caption{\acs{NN} + incipient ($\rho=0$)}
  \end{subfigure}
  \begin{subfigure}[t]{0.32\linewidth}
    \centering
    \includegraphics[height=3cm]{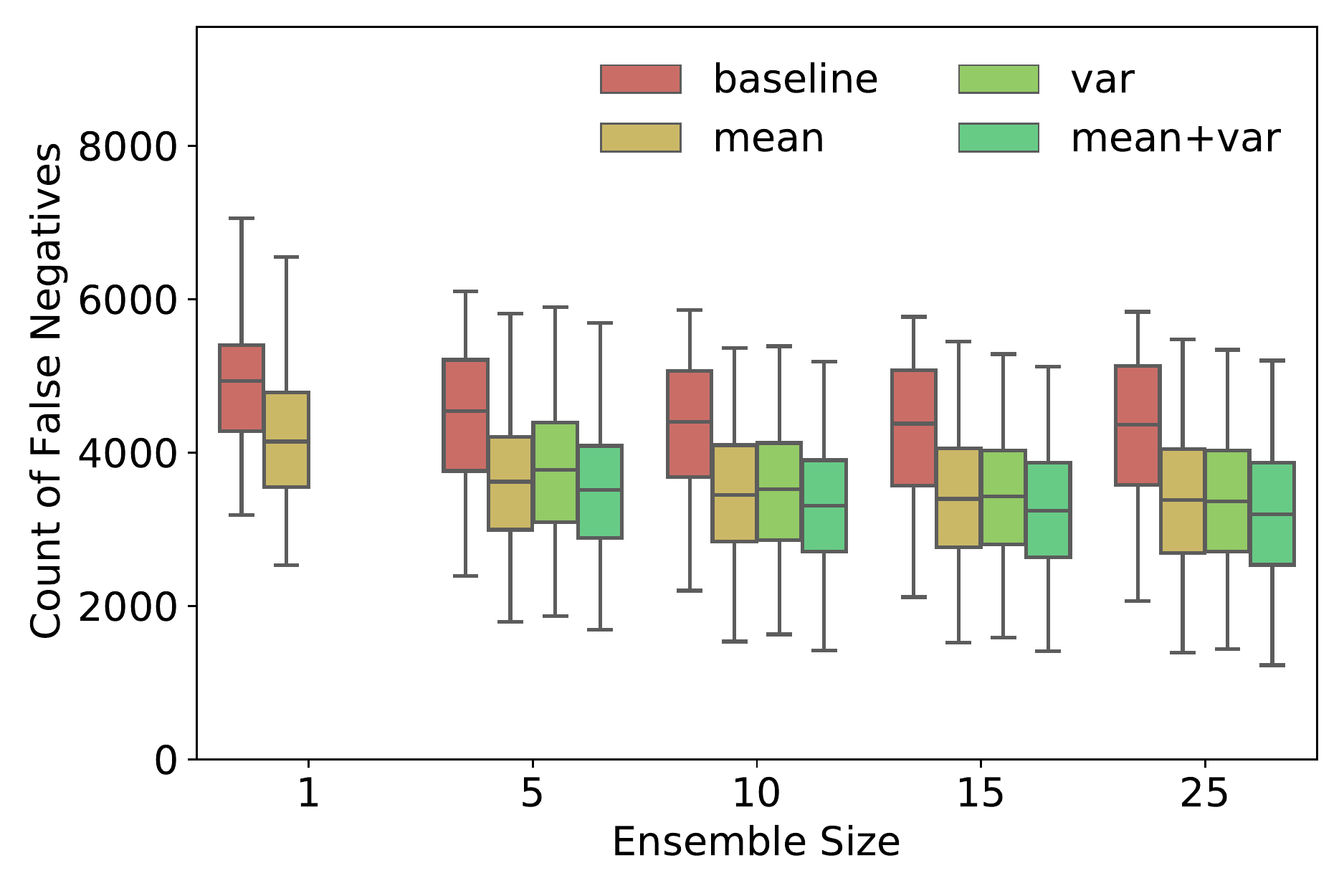}
    \caption{\acs{NN} + incipient ($\rho=0.2$)}
  \end{subfigure}
  \begin{subfigure}[t]{0.32\linewidth}
    \centering
    \includegraphics[height=3cm]{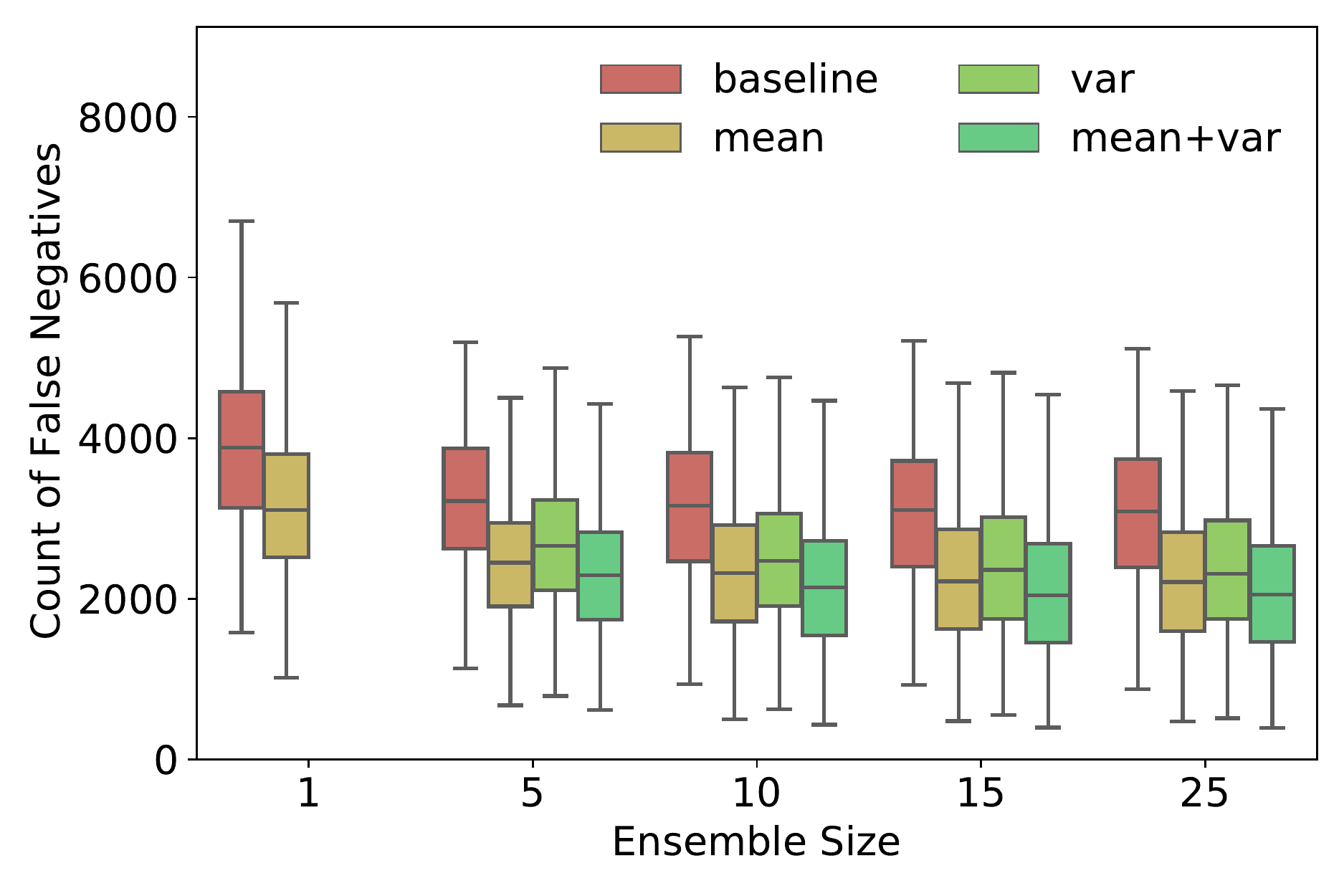}
    \caption{\acs{NN} + incipient ($\rho=1.0$)}
  \end{subfigure}
  \captionsetup{justification=raggedright}
  \caption{Box plots showing the number of certain false negatives (incipient anomalies wrongly classified as negative) after the rest are identified by uncertainty estimation for the chiller dataset. Results for $\rho=0,0.2,1.0$ are displayed respectively in the left, the middle and the right columns.}
  \label{fig:chiller-remaining-FN}
\end{figure}

\begin{figure}[tb]
  \centering
  \captionsetup{justification=centering}
  \begin{subfigure}[t]{0.32\linewidth}
    \centering
    \includegraphics[height=3cm]{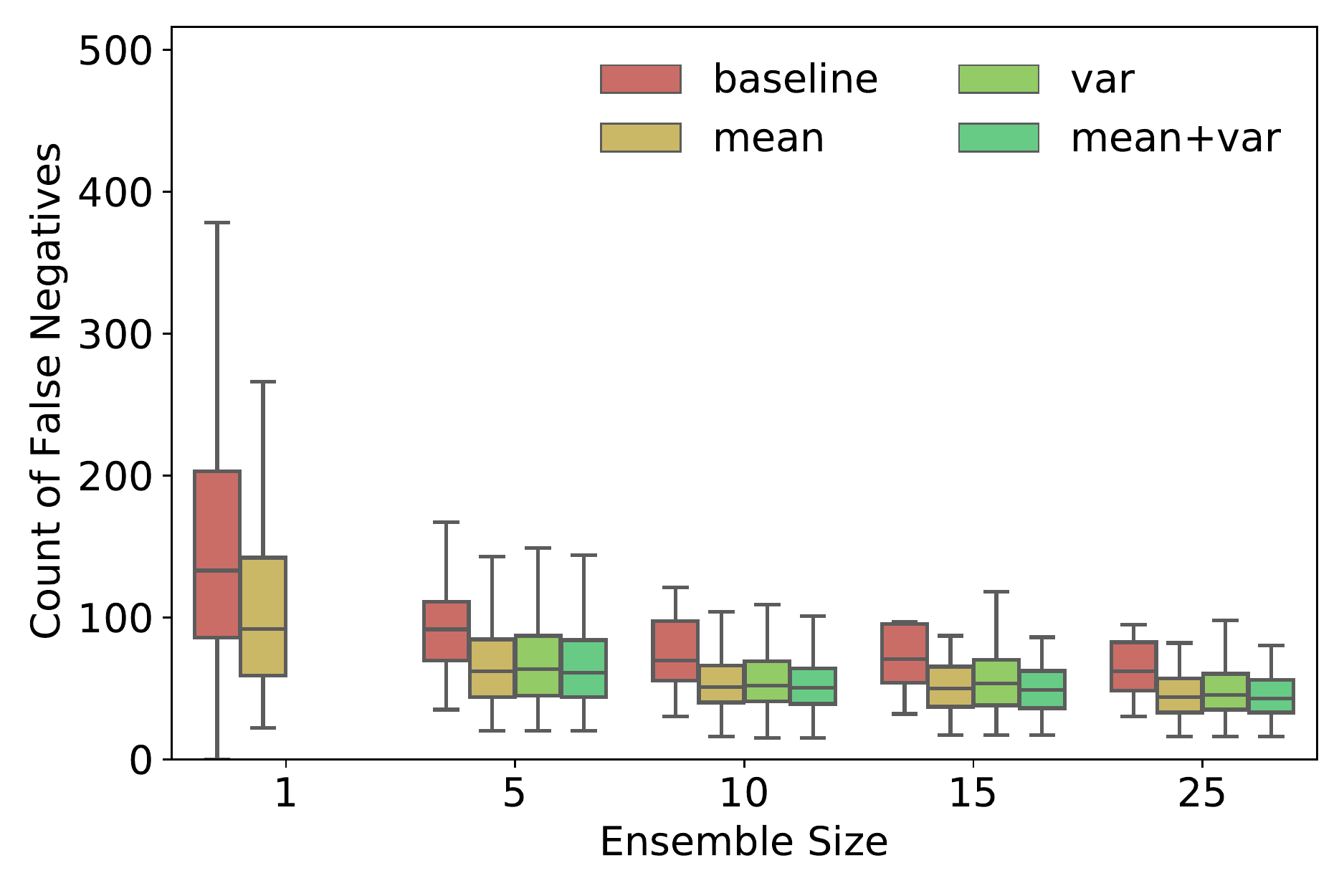}
    \caption{\acs{NN} + non-incipient ($\rho=0$)}
  \end{subfigure}
  \begin{subfigure}[t]{0.32\linewidth}
    \centering
    \includegraphics[height=3cm]{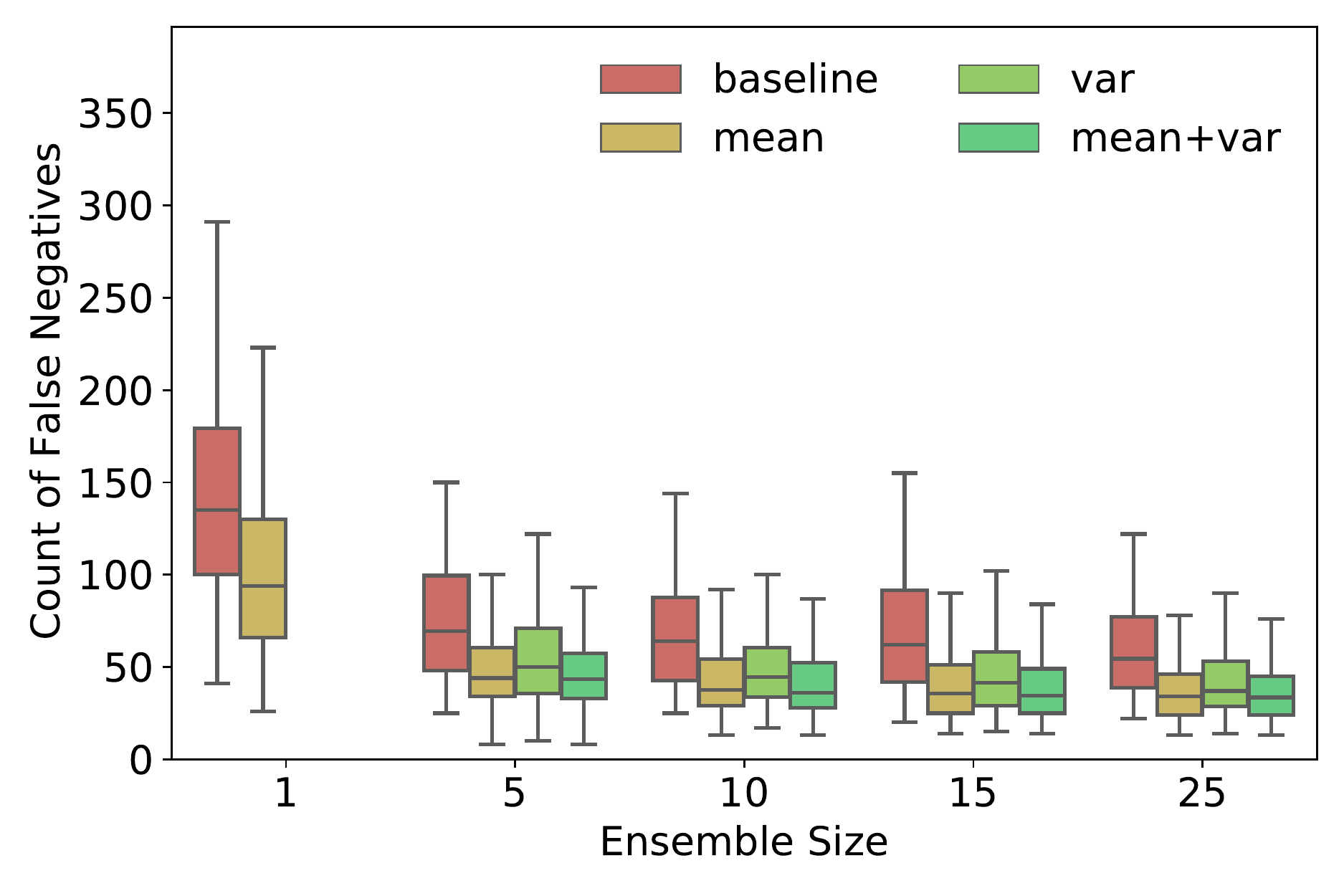}
    \caption{\acs{NN} + non-incipient ($\rho=0.2$)}
  \end{subfigure}
  \begin{subfigure}[t]{0.32\linewidth}
    \centering
    \includegraphics[height=3cm]{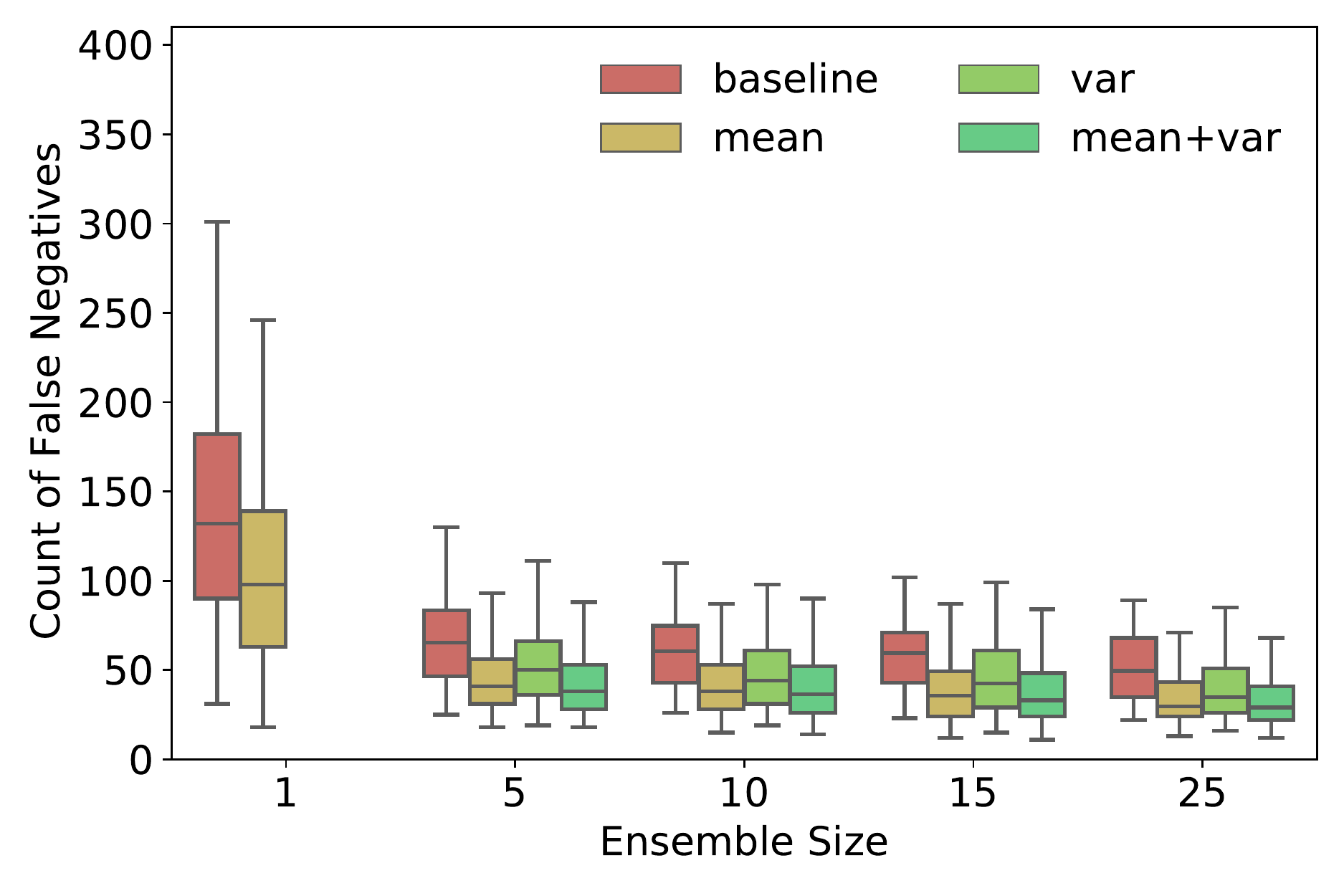}
    \caption{\acs{NN} + non-incipient ($\rho=1.0$)}
  \end{subfigure}
  
  \begin{subfigure}[t]{0.32\linewidth}
    \centering
    \includegraphics[height=3cm]{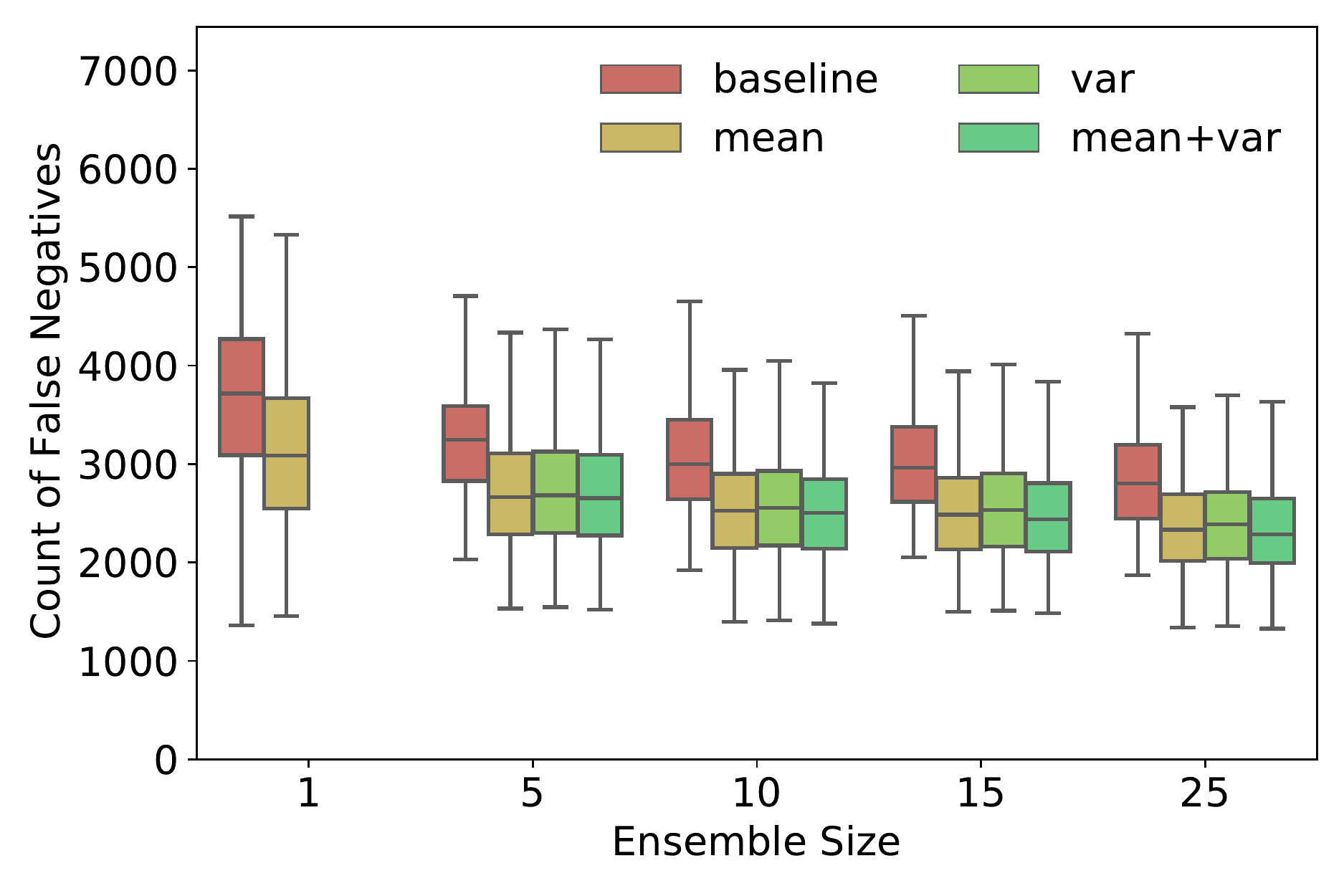}
    \caption{\acs{NN} + incipient ($\rho=0$)}
  \end{subfigure}
  \begin{subfigure}[t]{0.32\linewidth}
    \centering
    \includegraphics[height=3cm]{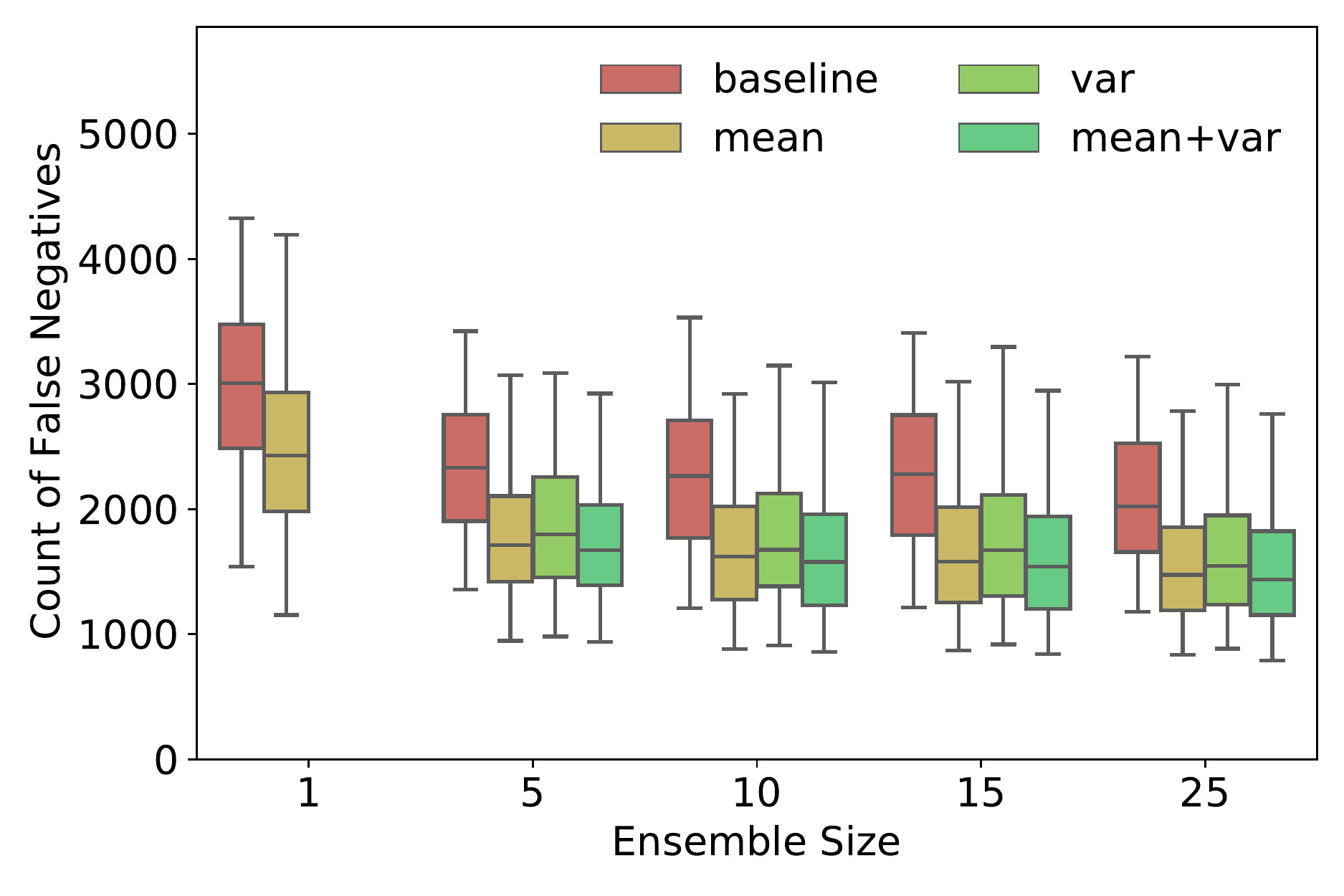}
    \caption{\acs{NN} + incipient ($\rho=0.2$)}
  \end{subfigure}
  \begin{subfigure}[t]{0.32\linewidth}
    \centering
    \includegraphics[height=3cm]{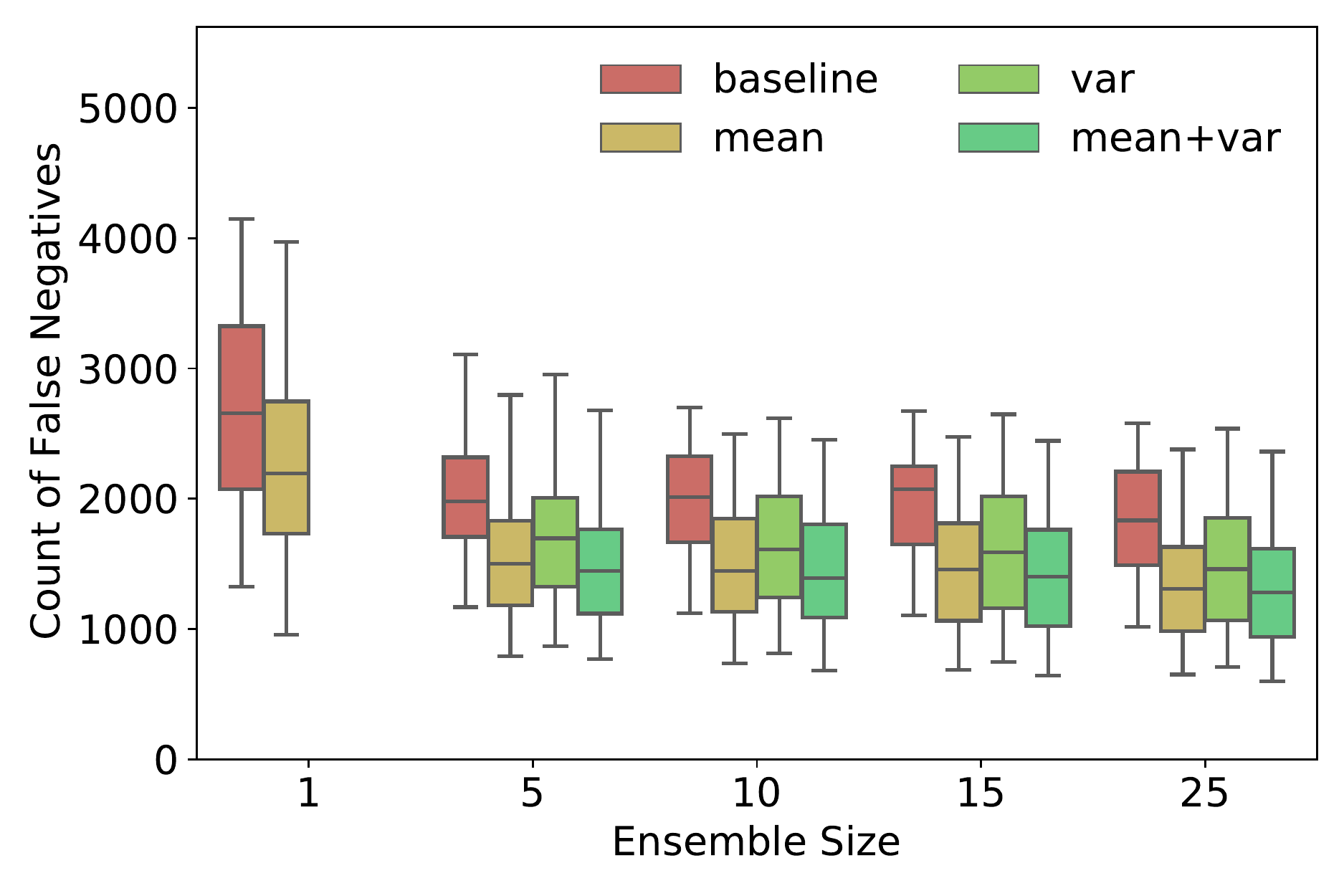}
    \caption{\acs{NN} + incipient ($\rho=1.0$)}
  \end{subfigure}
  \captionsetup{justification=raggedright}
  \caption{Box plots showing the number of certain false negatives (incipient anomalies wrongly classified as negative) after the rest are identified by uncertainty estimation for the \acs{DR} dataset. Results for $\rho=0,0.2,1.0$ are displayed respectively in the left, the middle and the right columns.}
  \label{fig:diabetic-remaining-FN}
\end{figure}

The next performance index we evaluate is the number of remaining false negatives after applying uncertainty estimation. The numbers of remaining false negatives are obtained by assuming that all identified uncertain false negatives will receive corrected labels. We are interested in knowing the number of remaining false negatives because these are mistakes that the uncertainty estimation techniques fail to identify. We visualize the performance variations of the trained models for the two datasets as box plots in Fig.~\ref{fig:chiller-remaining-FN} and in Fig.~\ref{fig:diabetic-remaining-FN}, respectively.

As displayed in the plots, besides \textsc{mean} and \textsc{var} we also included two other scenarios, \textsc{baseline} and \textsc{mean+var}, that respectively set the lower bound and the upper bound of performance of \textsc{mean} and \textsc{var}. Under \textsc{baseline}, no uncertainty information from output probabilities is utilized, i.e. $q=0$. \textsc{mean+var} is a hypothetical uncertainty metric where the uncertain examples identified by \textsc{mean+var} are the \textit{union} of the two sets of uncertain examples identified by \textsc{mean} and by \textsc{var}, not subject to the constraint imposed by $q$; see Fig.~\ref{fig:decision} for an illustration. Therefore, it is at least as good as \textsc{mean} or \textsc{var}. If \textsc{mean} and \textsc{var} do not have much overlapping, \textsc{mean+var} will identify many more false negatives than either of them alone; however, we can see from Fig.~\ref{fig:chiller-remaining-FN} and Fig.~\ref{fig:diabetic-remaining-FN} that this is not the case. The results given by \textsc{mean+var} do not have much improvement over those given by \textsc{mean}, indicating that many of the false negatives identified by \textsc{var} are also captured by \textsc{mean}, matching the expectation of Theorem~\ref{thm:mean-vs-var-beta-finite}.

An immediate observation from Figs.~\ref{fig:chiller-remaining-FN}\,\&\,\ref{fig:diabetic-remaining-FN} is that ensemble learning can achieve substantial performance improvement even for small ensemble sizes ($K=5$). For $K>5$, we can still see significant improvement when $K$ grows larger for tree ensembles; however, for \ac{NN} ensembles the marginal improvement from increasing ensemble sizes is smaller, which is probably due to the fact that individual \ac{DT} classifiers are relatively weak compared to individual \ac{NN} classifiers. By comparing the performance of \textsc{mean} and that of \textsc{var} in the plots, we can see that \textsc{mean} leads to fewer remaining false negatives in general; in other words, the \textsc{mean} uncertainty metric can identify more false negatives than the \textsc{var} metric can.

\subsubsection{False Negative Precision (FN-precision)}
\begin{figure}[tb]
  \centering
  \begin{subfigure}[t]{0.32\linewidth}
    \centering
    \includegraphics[height=3cm]{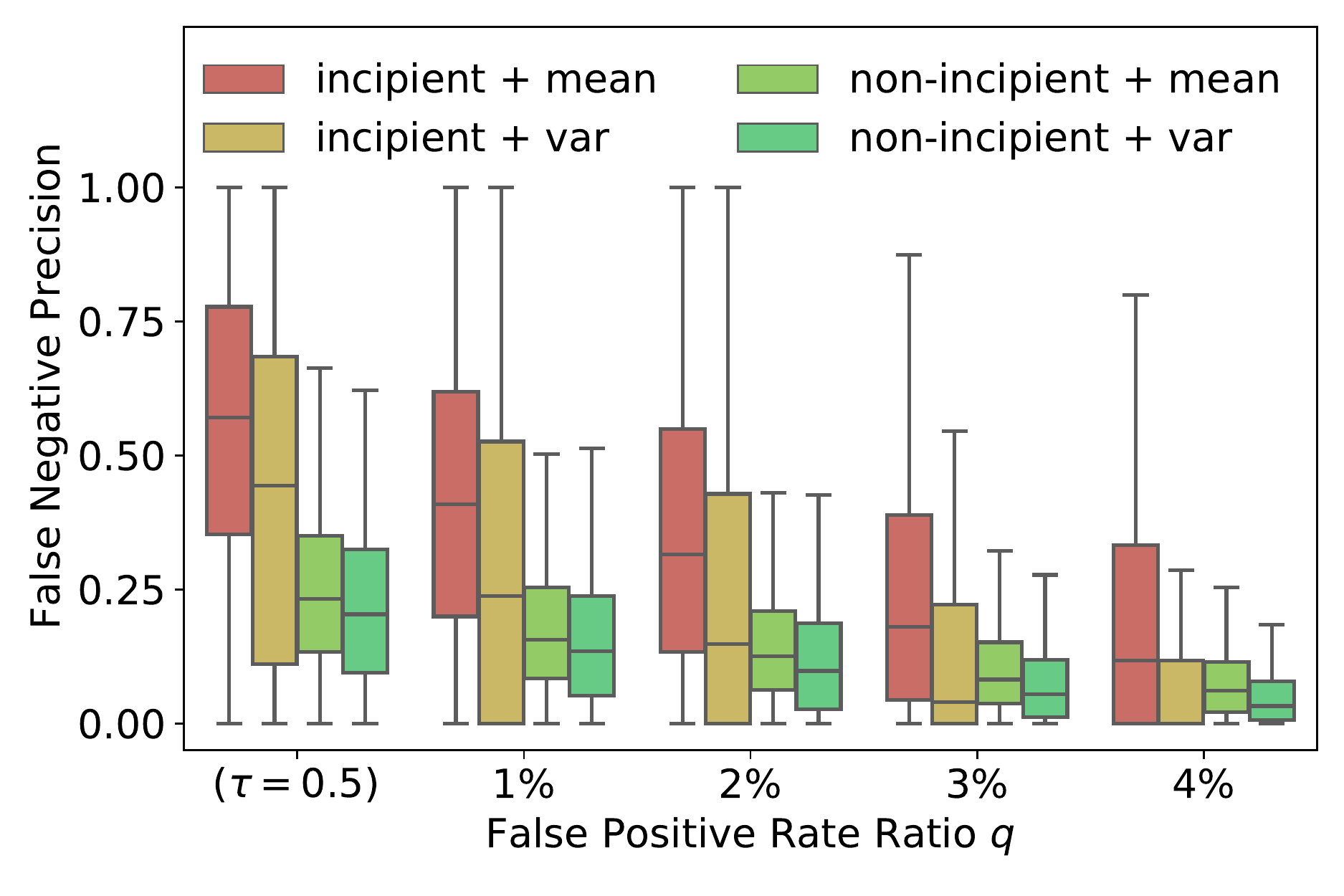}
    \caption{\acs{DT} + chiller ($\rho=0$)}
  \end{subfigure}
  \begin{subfigure}[t]{0.32\linewidth}
    \centering
    \includegraphics[height=3cm]{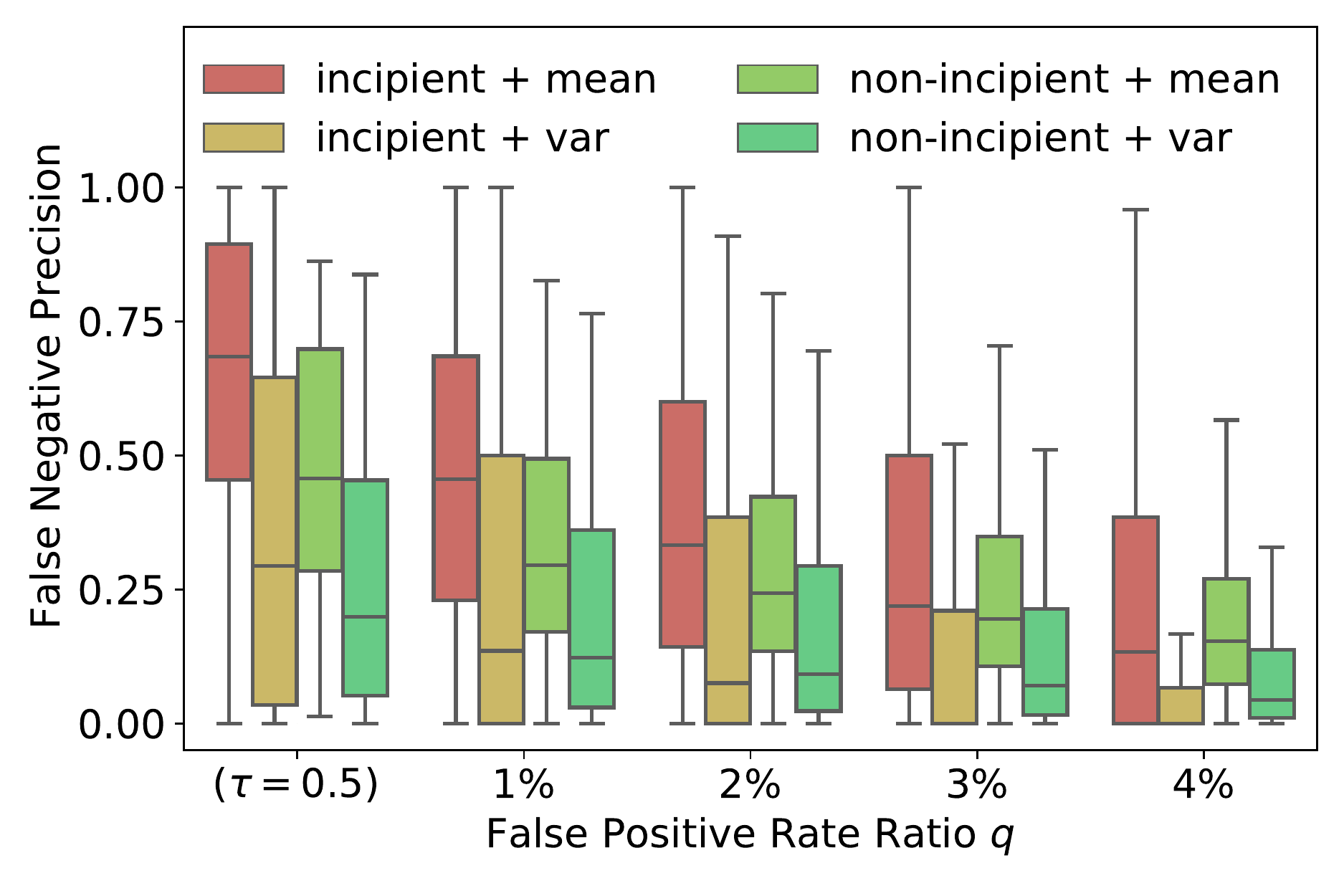}
    \caption{\acs{DT} + chiller ($\rho=0.2$)}
  \end{subfigure}
  \begin{subfigure}[t]{0.32\linewidth}
    \centering
    \includegraphics[height=3cm]{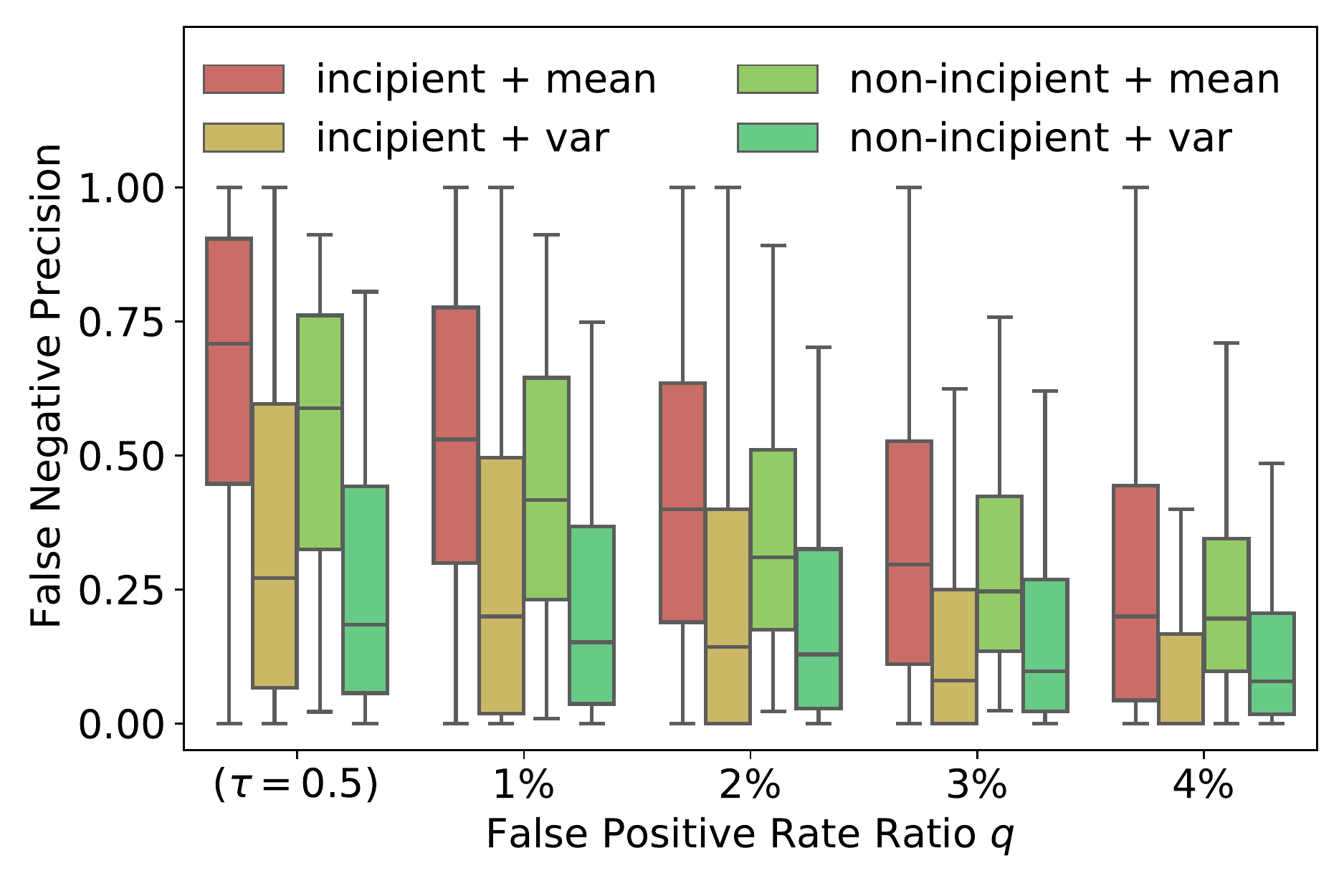}
    \caption{\acs{DT} + chiller ($\rho=1.0$)}
  \end{subfigure}
  
  \begin{subfigure}[t]{0.32\linewidth}
    \centering
    \includegraphics[height=3cm]{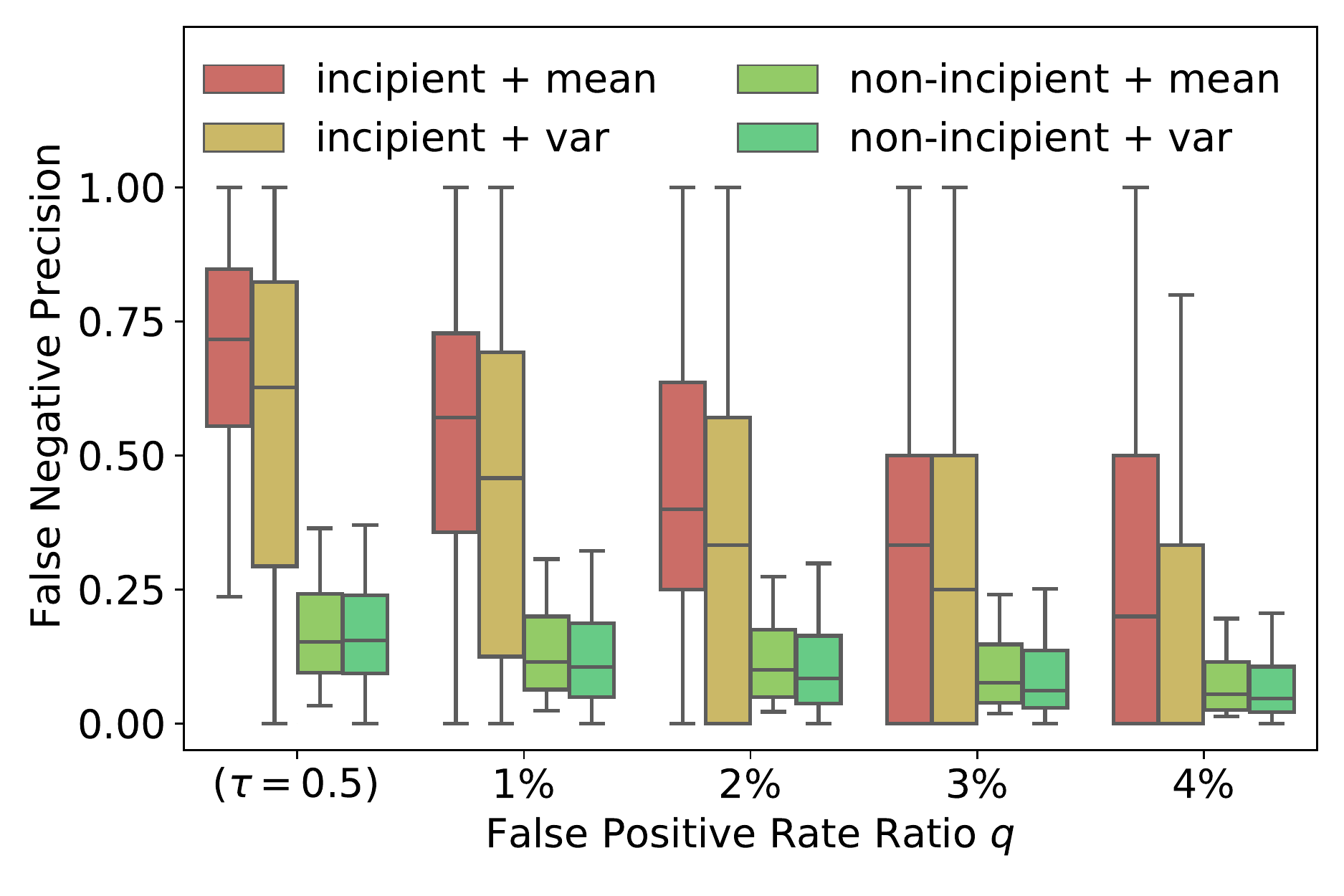}
    \caption{\acs{NN} + chiller ($\rho=0$)}
  \end{subfigure}
  \begin{subfigure}[t]{0.32\linewidth}
    \centering
    \includegraphics[height=3cm]{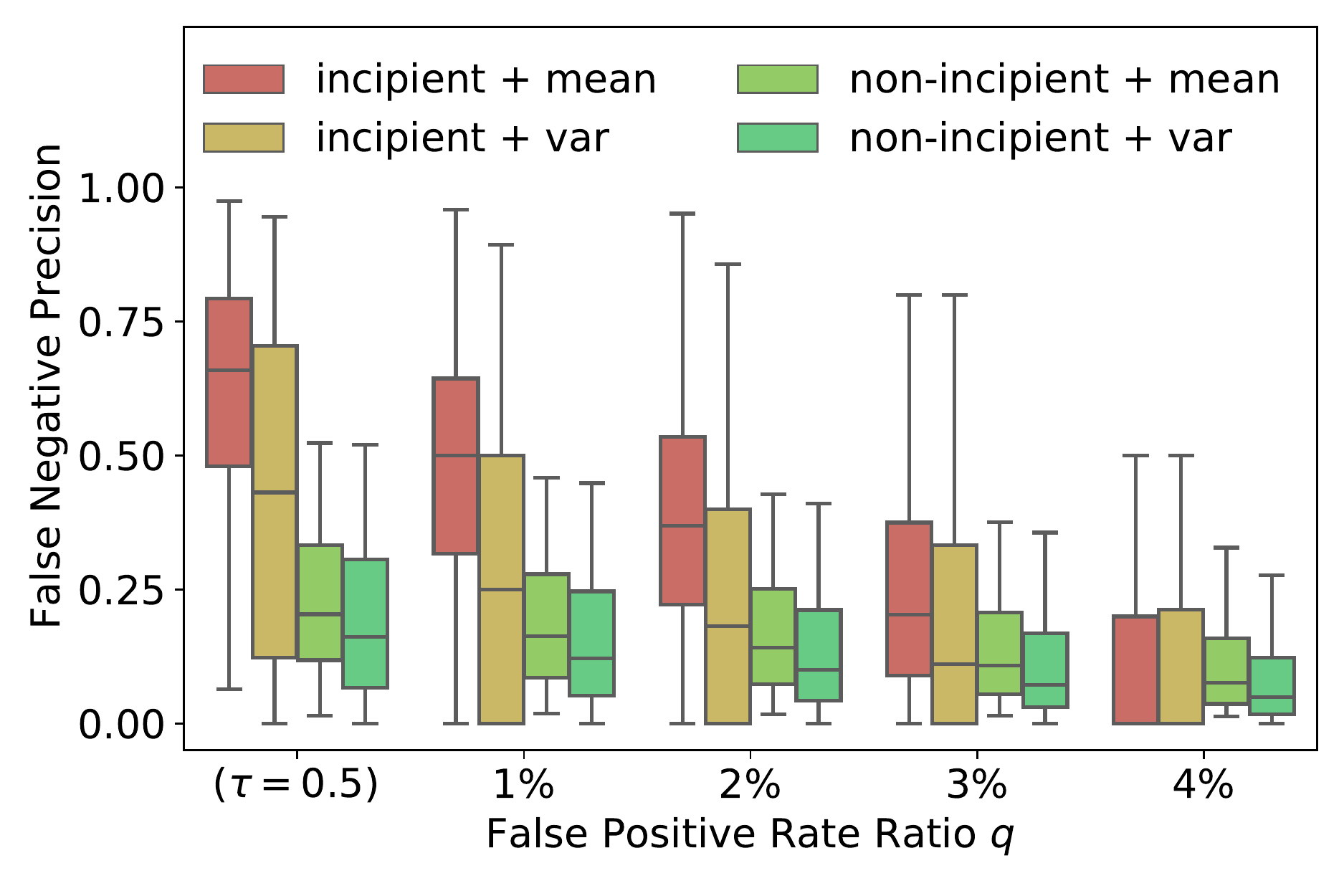}
    \caption{\acs{NN} + chiller ($\rho=0.2$)}
  \end{subfigure}
  \begin{subfigure}[t]{0.32\linewidth}
    \centering
    \includegraphics[height=3cm]{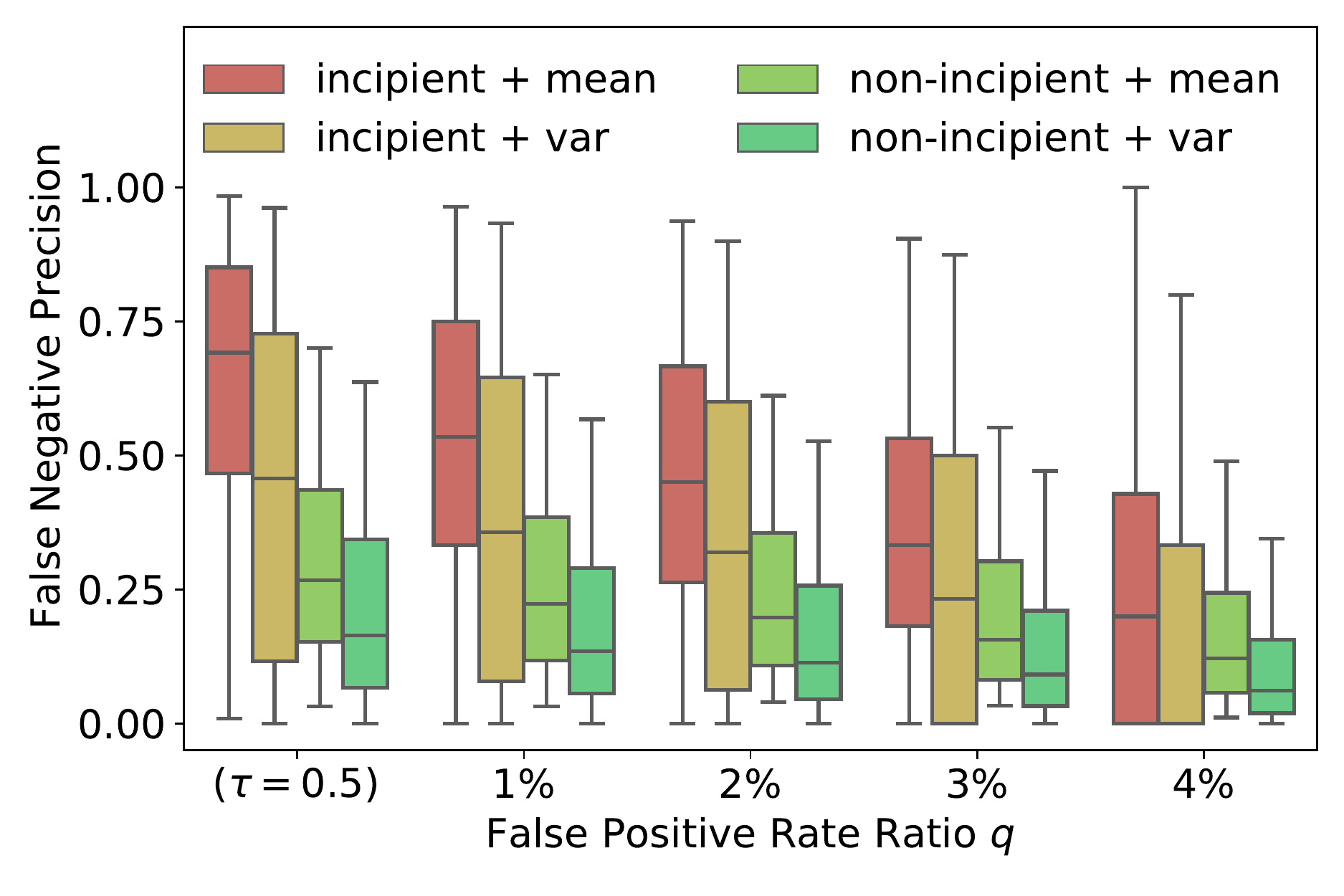}
    \caption{\acs{NN} + chiller ($\rho=1.0$)}
  \end{subfigure}

  \begin{subfigure}[t]{0.32\linewidth}
    \centering
    \includegraphics[height=3cm]{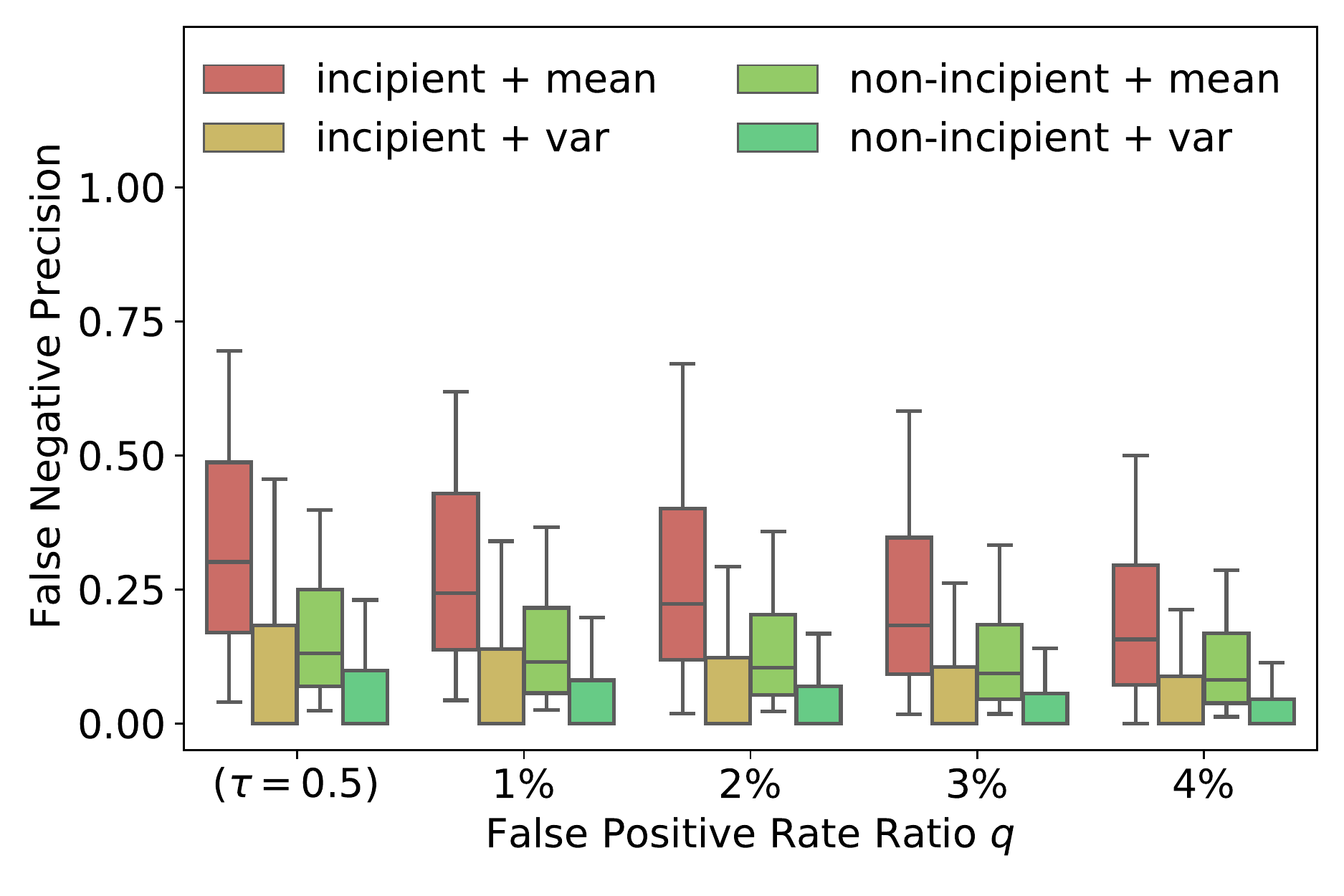}
    \caption{\acs{NN} + \acs{DR} ($\rho=0$)}
  \end{subfigure}
  \begin{subfigure}[t]{0.32\linewidth}
    \centering
    \includegraphics[height=3cm]{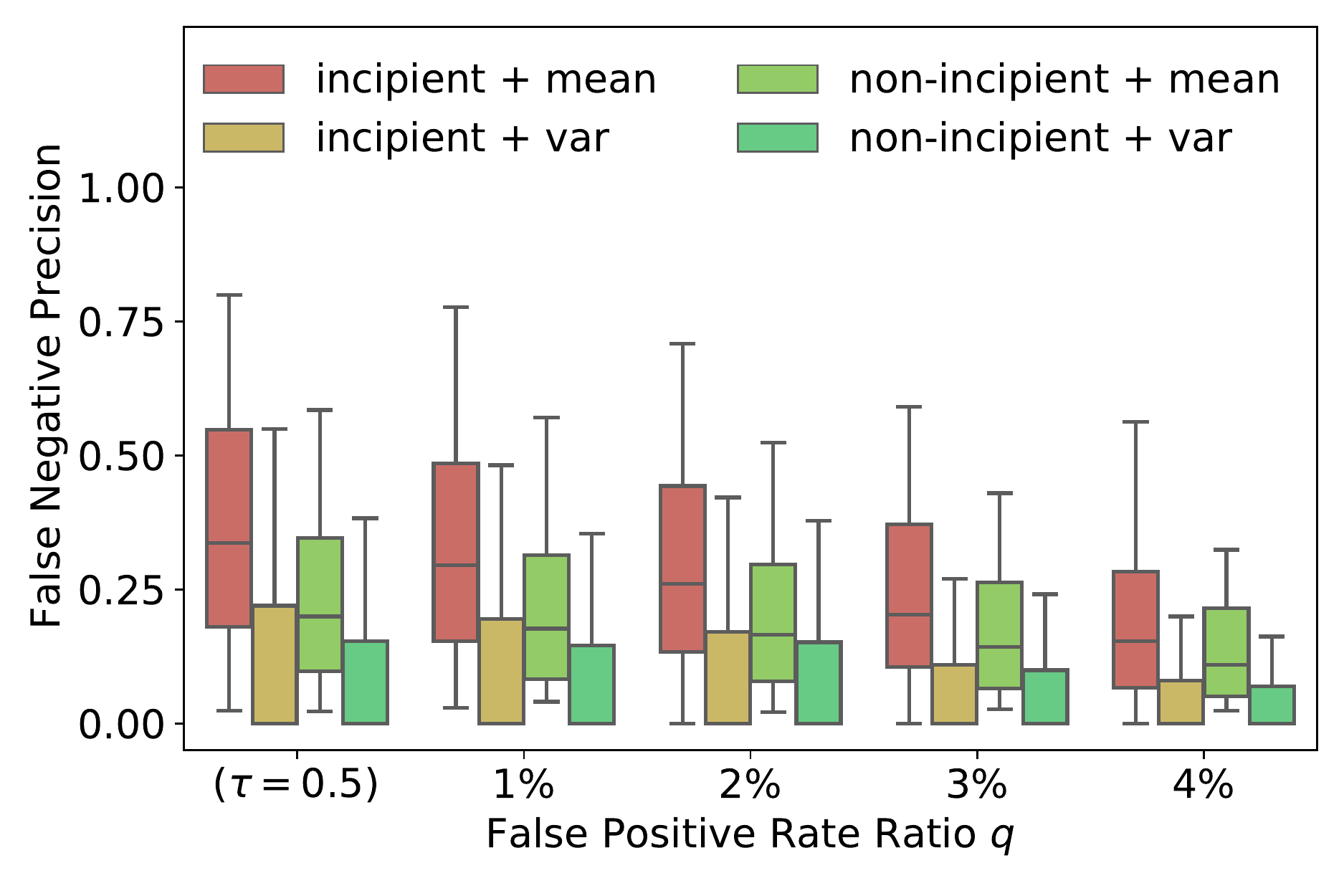}
    \caption{\acs{NN} + \acs{DR} ($\rho=0.2$)}
  \end{subfigure}
  \begin{subfigure}[t]{0.32\linewidth}
    \centering
    \includegraphics[height=3cm]{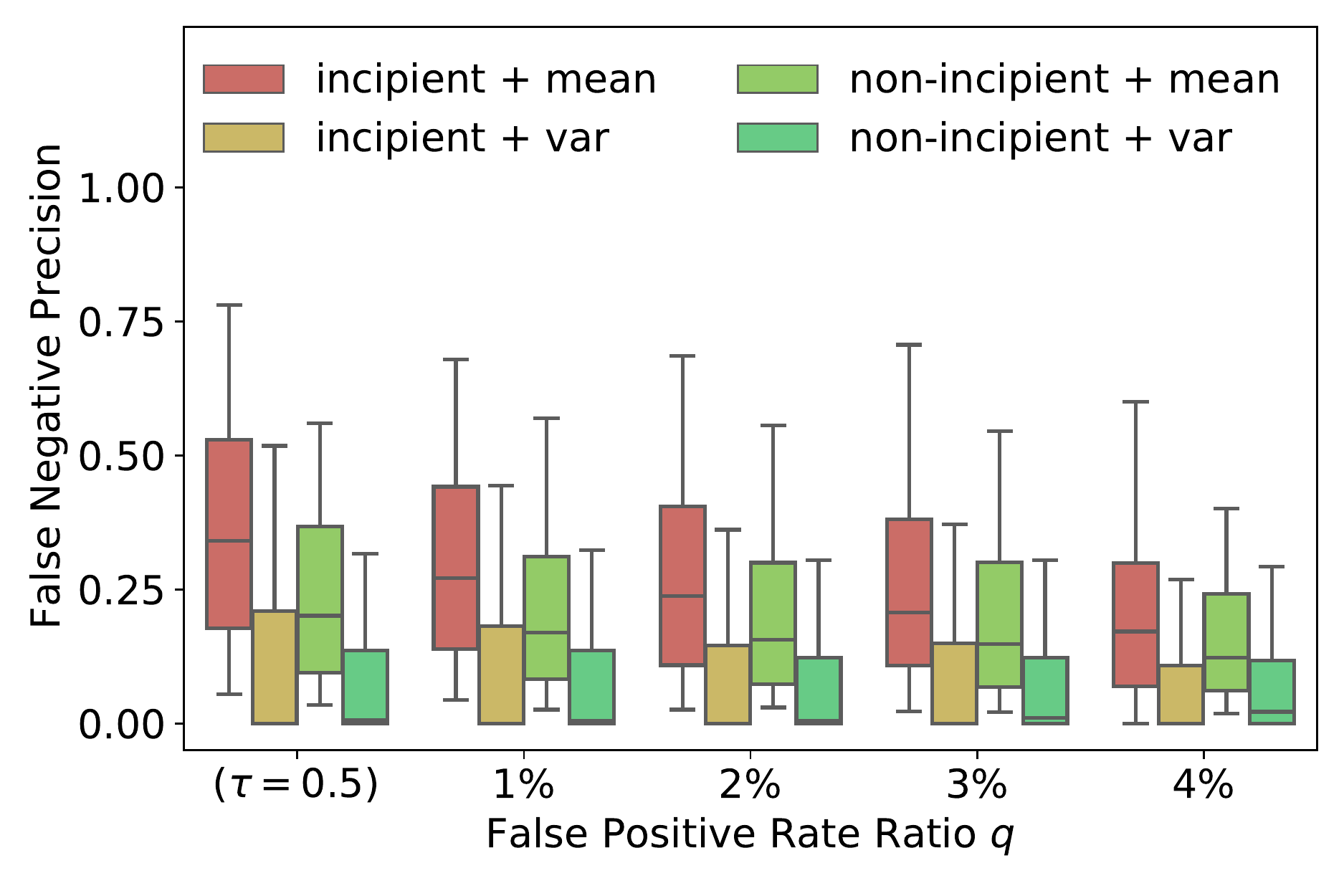}
    \caption{\acs{NN} + \acs{DR} ($\rho=1.0$)}
  \end{subfigure}
  \caption{Box plots showing the false negative precision metric for ensemble classifiers ($K=5,10,15,25$) under different settings of the \acs{FPR} percentile $q$ for the two datasets. Different colors indicate performance indices given by \textsc{mean} and \textsc{var} for the incipient and the non-incipient data.}
  \label{fig:FN-prec}
\end{figure}

Although the above analysis shows that \textsc{mean} compared to \textsc{var} can identify more false negatives among incipient anomalies, it is not sufficient to show that \textsc{mean} is more preferable to \textsc{var} because the increased number of corrected false negatives may simply be a consequence of more uncertain negatives being identified; in an extreme scenario, if all negative data points are marked as uncertain negatives, then all false negatives can be corrected. Therefore, we introduce the FN-precision metric to measure how precisely each model can identify the false negatives. As can be seen from Fig.~\ref{fig:FN-prec}, \textsc{mean} again outperforms \textsc{var} in terms of FN-precision.

\subsection{Detection Performance of One-Class Classifiers}
As a comparative study, we also experimented using \ac{OC-SVM}, a popular one-class model for semi-supervised and unsupervised learning, to learn a boundary of the normal data points (i.e., the inliers) that can be used to separate them from the outliers for the chiller dataset. We did not experimented applying \ac{OC-SVM} on the \ac{DR} dataset as it does apply to image data. Again, we conducted a grid search over various hyperparameter settings and picked out the best-performing models.

\begin{figure}[tb]
  \centering
  \begin{subfigure}[t]{0.32\linewidth}
    \centering
    \includegraphics[height=3cm]{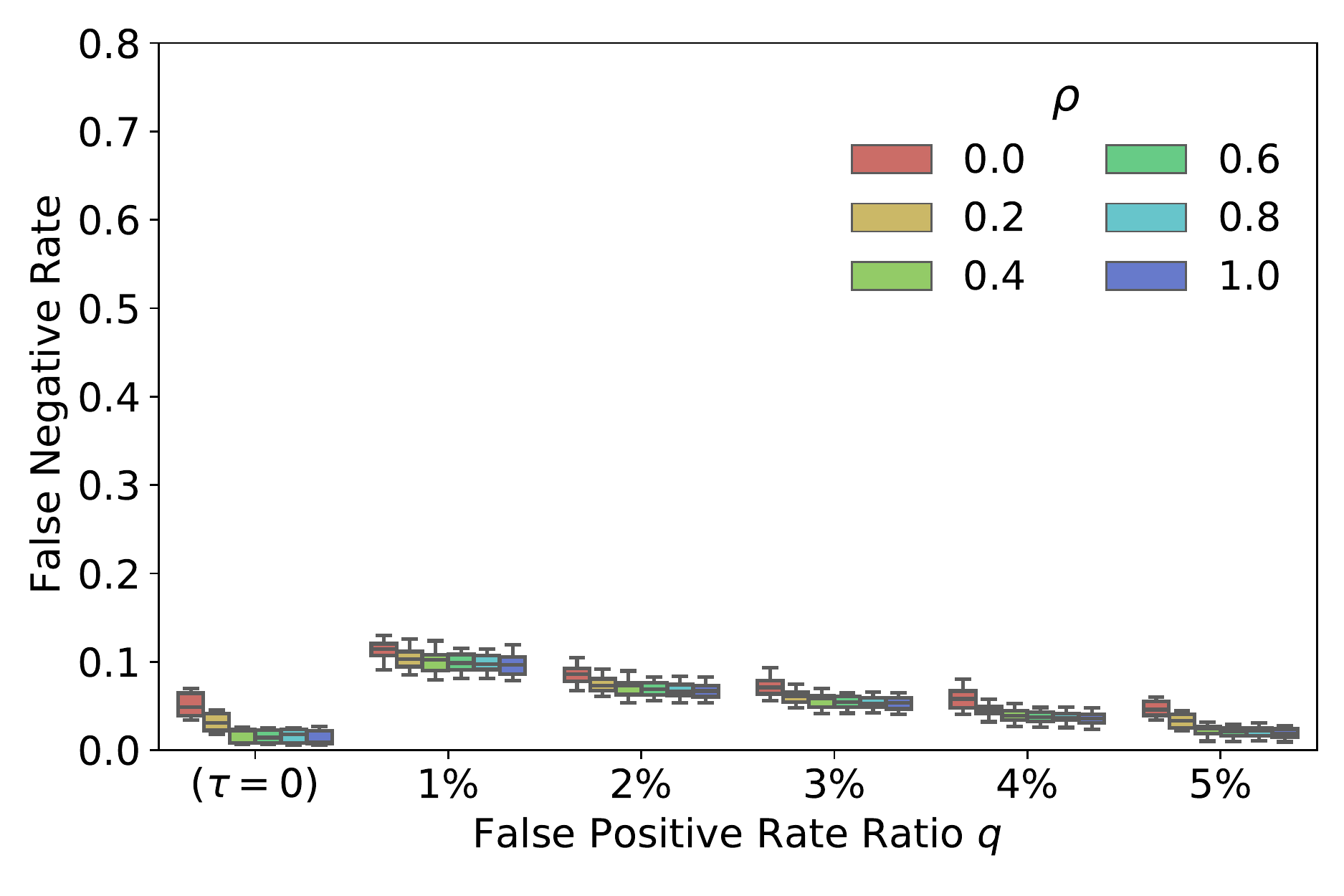}
    \caption{Non-incipient anomalies: $K=1$}
  \end{subfigure}
  \begin{subfigure}[t]{0.32\linewidth}
    \centering
    \includegraphics[height=3cm]{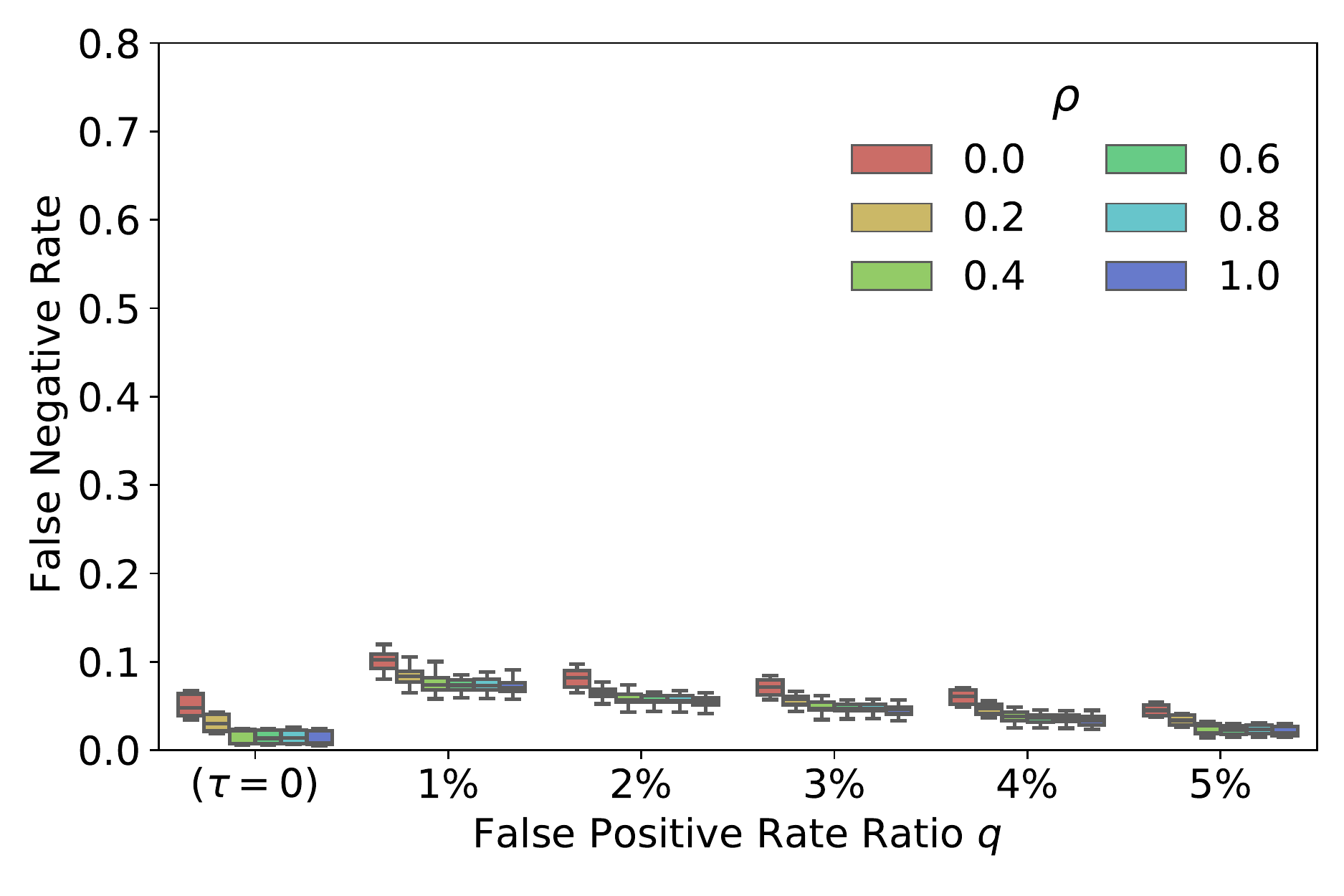}
    \caption{Non-incipient anomalies: $K=5$}
  \end{subfigure}
  \begin{subfigure}[t]{0.34\linewidth}
    \centering
    \includegraphics[height=3cm]{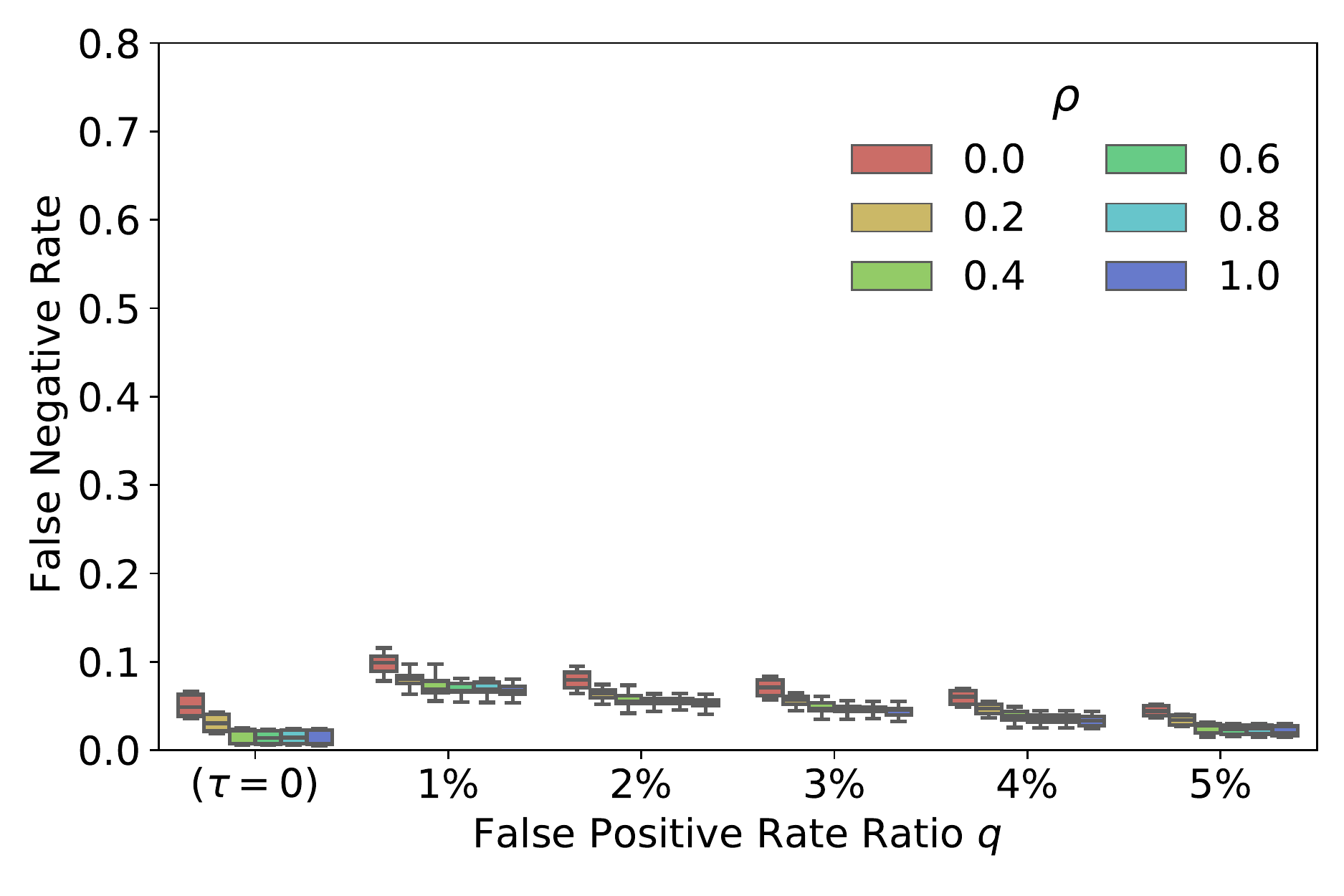}
    \caption{Non-incipient anomalies: $K=25$}
  \end{subfigure}
  
  \begin{subfigure}[t]{0.32\linewidth}
    \centering
    \includegraphics[height=3cm]{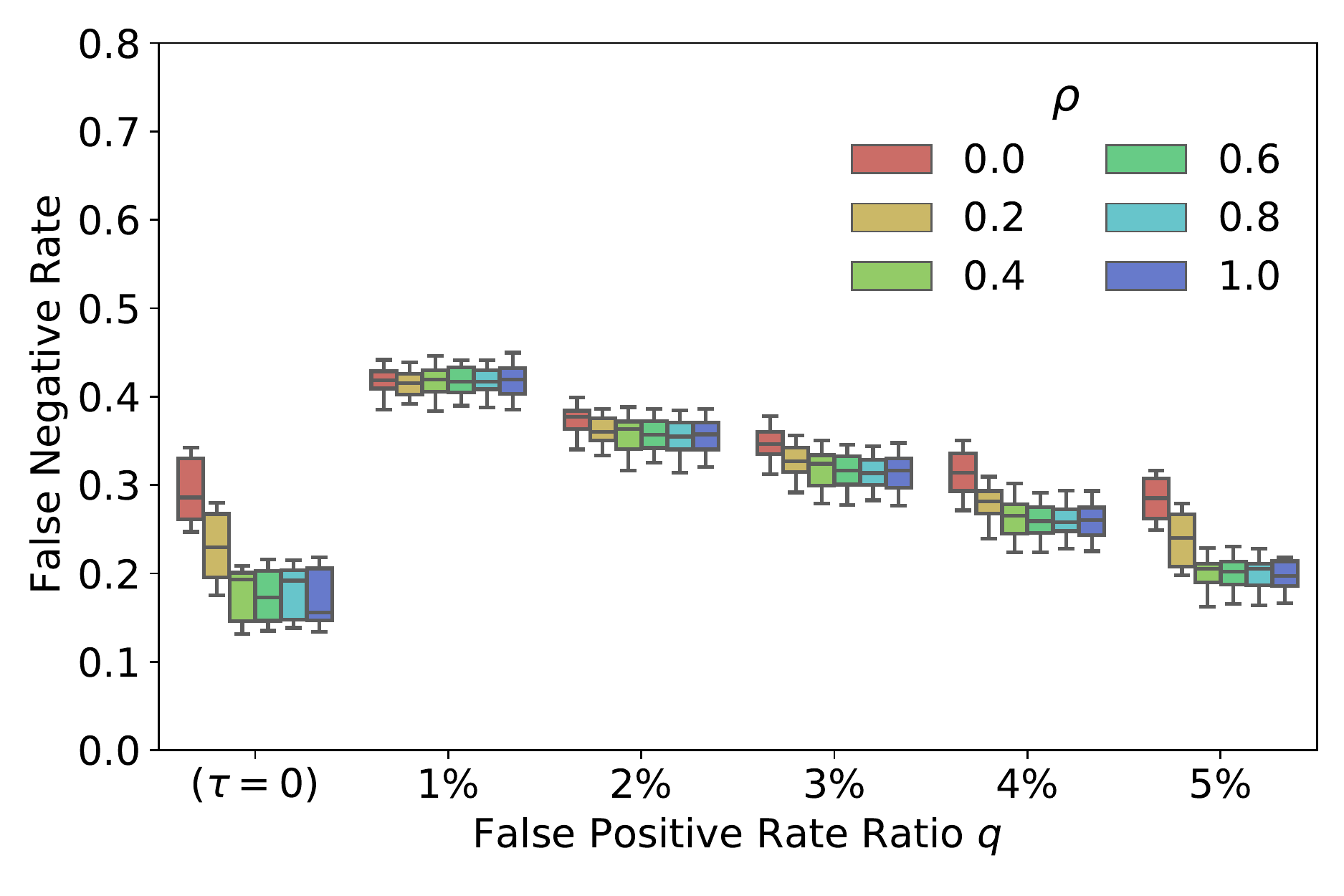}
    \caption{Incipient anomalies: $K=1$}
  \end{subfigure}
  \begin{subfigure}[t]{0.32\linewidth}
    \centering
    \includegraphics[height=3cm]{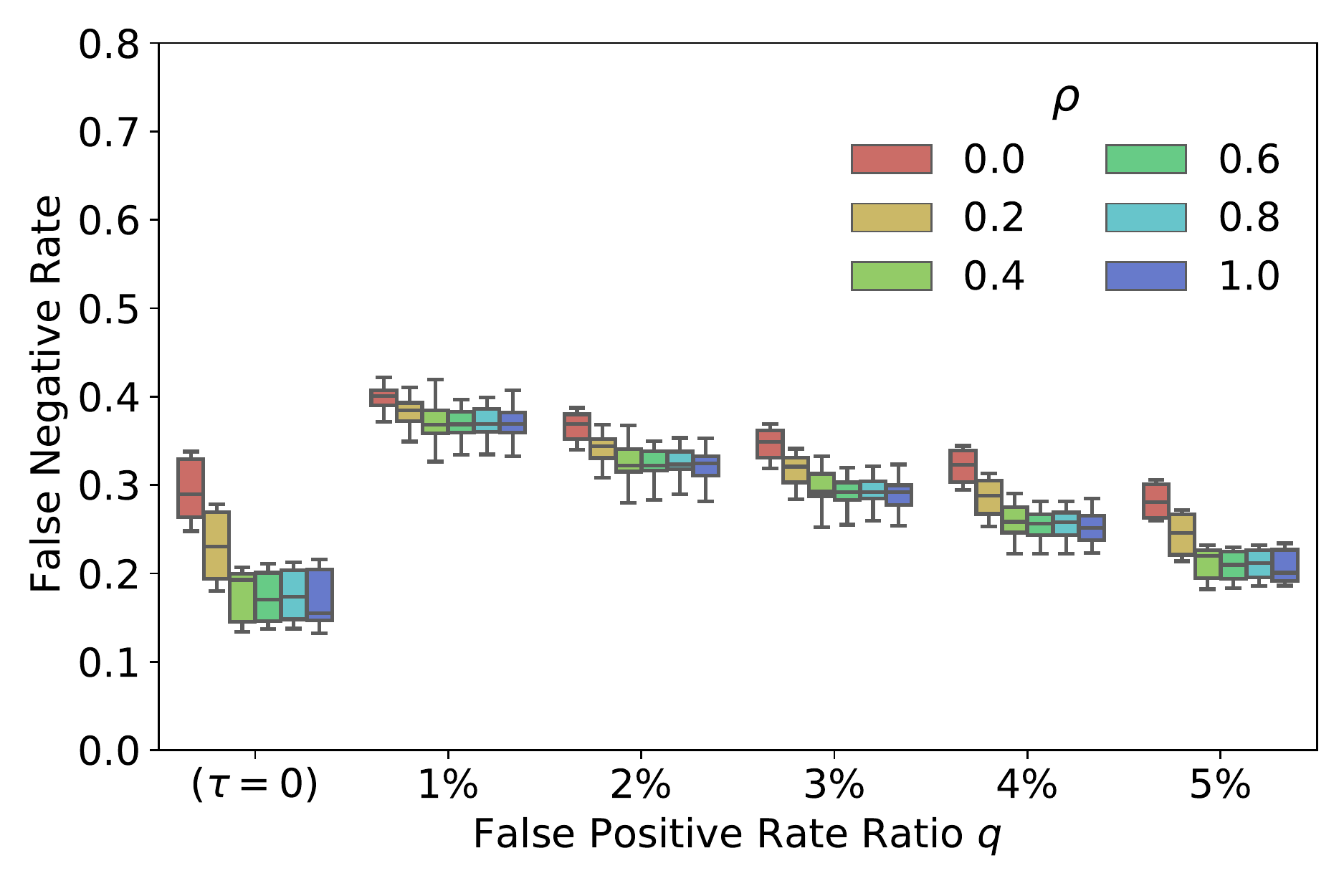}
    \caption{Incipient anomalies: $K=5$}
  \end{subfigure}
  \begin{subfigure}[t]{0.34\linewidth}
    \centering
    \includegraphics[height=3cm]{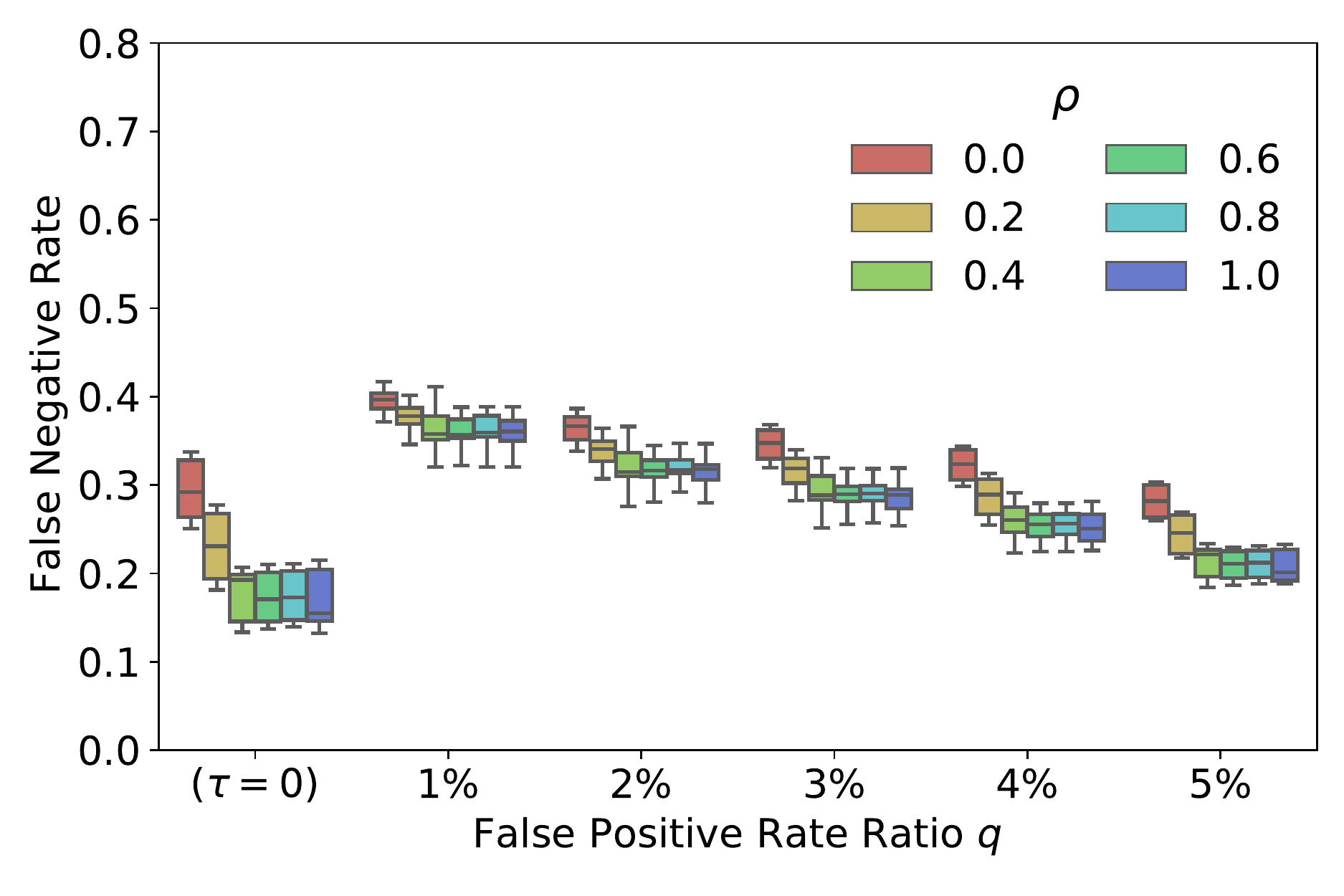}
    \caption{Incipient anomalies: $K=25$}
  \end{subfigure}
  
  \caption{The performance of \acp{OC-SVM} classifiers on non-incipient anomalies (top panel) and incipient anomalies (bottom panel). Box plots for ensembles of three different sizes $K=1,5,25$ are displayed.}
  \label{fig:ocsvm-FNR}
  \vspace{-3mm}
\end{figure}

In Fig.~\ref{fig:ocsvm-FNR}, we visualize the performance of \acp{OC-SVM} ensembles of three different sizes $K=1,5,25$, and show show how the how the detection performance in terms of \ac{FNR} varies with the \ac{FPR} percentile $q$. As with other learners for the chiller dataset, we used sample bagging to induce diversity among ensemble \ac{OC-SVM} learners. The experimental results for ensemble learners, however, did not demonstrate much improvement over the single learner cases. By comparing the results for \ac{OC-SVM} to those for \ac{DT} and \ac{NN} ensembles, we can see that \ac{OC-SVM} gives inferior detection performance for both incipient and non-incipient anomalies. A detailed discussion on \ac{OC-SVM} and other one-class methods (e.g., autoencoders) is beyond the scope of this paper. We believe there are also challenges ahead in applying one-class methods to anomaly detection, which will become promising directions for future research.

\subsection{Validation of the Beta Distribution Assumption}
In our theoretical analysis in Sec.~\ref{sec:theory}, we make an assumption that the individual predictions in an ensemble learner assume a Beta distribution $\mathcal{B}(\alpha, \beta)$ where $\alpha+\beta$ is held constant. We performed some observational studies to validate this assumption, as displayed by the five examples in Fig.~\ref{fig:beta-dist-five-SL}. 
Additional examples will be given in the appendix.   

\section{Related Work}\label{sec:related-work}

\paragraph{Out-of-distribution Input Detection and Uncertainty Estimation} 
In recent years, a number of research papers~\cite{lakshminarayanan2017simple,gal2016uncertainty} related to the detection of \ac{o.o.d.} data appeared in literature, especially in the deep learning community that has shown a strong and growing interest in utilizing ensemble methods in supervised learning to estimate the \textit{uncertainty} behind the decisions on data points. Lakshminarayanan~et~al.~\cite{lakshminarayanan2017simple} proposed using \textit{random initialization} and \textit{random shuffling} of training examples to diversify base learners of the same network architecture. Gal and Ghahramani proposed using MC-dropout~\cite{gal2016uncertainty} to estimate a network's prediction uncertainty by using dropout not only at training time but also at test time. By sampling a dropout model $\mathcal{M}$ using the same input for $T$ times, we can obtain an ensemble of prediction results with $T$ individual probability vectors. The dropout technique provides an inexpensive approximation to training and evaluating an ensemble of exponentially many similar yet different neural networks.

Although promising results from these ensemble approaches have been demonstrated on certain types of \ac{o.o.d.} data such as dataset shift and unseen/unknown classes~\cite{lakshminarayanan2017simple}, it is difficult to evaluate their effectiveness in general, because the \ac{o.o.d.} part of the world is obviously much ``larger'' than its in-distribution counterpart and is presumably much harder to analyze. In contrast, our work, although using similar algorithms to those on \ac{o.o.d.} detection, still embraces a closed-world assumption and restricts the focus to incipient anomalies---a special type of data distribution that has a close connection to the training distribution. We speculate that some knowledge necessary for detecting incipient anomalies is already entailed in the training data, thus making the detection of incipient anomalies possible with supervised methods.

\paragraph{Model calibration}
Another relevant line of work aims to produce good probability outputs using \textit{model calibration} techniques~\cite{niculescu2005predicting, guo2017calibration}, e.g. temperature scaling~\cite{guo2017calibration}. The calibration techniques are typically applied in a post-processing manner; in other words, a calibration method learns a transformation that is applied to a model's uncalibrated output probabilities, without affecting the parameters (weights) of the original model itself. Although good confidence measures are important in many other fields, we are skeptical about the role of model calibration in the context of anomaly detection. By design, calibration methods should only adjust the probability values without affecting the ranking among data points. Therefore, if a calibration method results in an isotonic transformation (as are popular methods), the rankings will not change. Nor will the detection decisions.

\section{Conclusions and Future Work}\label{sec:conclusion}
We show in this paper that, incipient anomalies can pose critical challenges to supervised anomaly detection systems built upon ML techniques, especially under situations where incipient anomalies are absent from the training data.
The resulting ML models can easily mistake incipient anomalies for normal ones, which can lead to costly consequences if the ideal time for intervention or treatment is missed. To address this challenge, we study how to exploit the uncertainty information from ensemble learners to identify incipient anomalies that are potentially wrongly classified. The three main takeaways from this study are summarized as follows:
\begin{itemize}
    \item Without sacrificing the detection performance on non-incipient anomalies, we can improve the classifier's performance on incipient anomalies by using models of higher sensitivity; this can be done by tuning down the a classifier's detection threshold $\tau$.
    \item The detection performance on incipient anomalies can be greatly improved by incorporating some incipient anomaly data, even in small amount, into the training distribution (i.e. the development set).
    \item \textsc{mean} is a more preferable uncertainty metric to \textsc{var}, as proved by our theoretical analysis and shown by our empirical results.
\end{itemize}
The three recommendations above are complementary and can lead to better results when applied together. It is worthy to note that in this paper we mainly focus on supervised \ac{ML} models and their ensembles. One-class methods such as \acp{OC-SVM} and autoencoders are also promising and interesting directions to explore, which will constitute our future work.

% \begin{acks}
% This work is supported by the National Research Foundation of Singapore through a grant
% to the Berkeley Education Alliance for Research in Singapore (BEARS) for the Singapore-Berkeley Building Efficiency and Sustainability in the Tropics (SinBerBEST) program, and by the Defence Science \& Technology Agency (DSTA) of Singapore. 
% \end{acks}

\bibliographystyle{ACM-Reference-Format}
\bibliography{refs.bib}

\appendix
\clearpage
\acresetall

\section{Implementation Details}

\subsection{Experiments on the Chiller Dataset}
% The RP-1043 chiller dataset~\cite{comstock1999development} is not publicly available but can be purchased from ASHRAE. The same sixteen features and six types of faults as used in our previous work~\cite{tan2019encoder} were used to train our models in our empirical study. We also tried to use other sets of selected features than the sixteen features in our case study, e.g., the features identified in Li~et~al.'s previous work~\cite{li2016fault}, and obtained similar results. Detailed descriptions of the sixteen features and the six fault types used in this study can be found in Table~\ref{tbl:features} and Table~\ref{tbl:faults}.

\begin{figure}[b]
    \centering
    \includegraphics[width=0.7\linewidth]{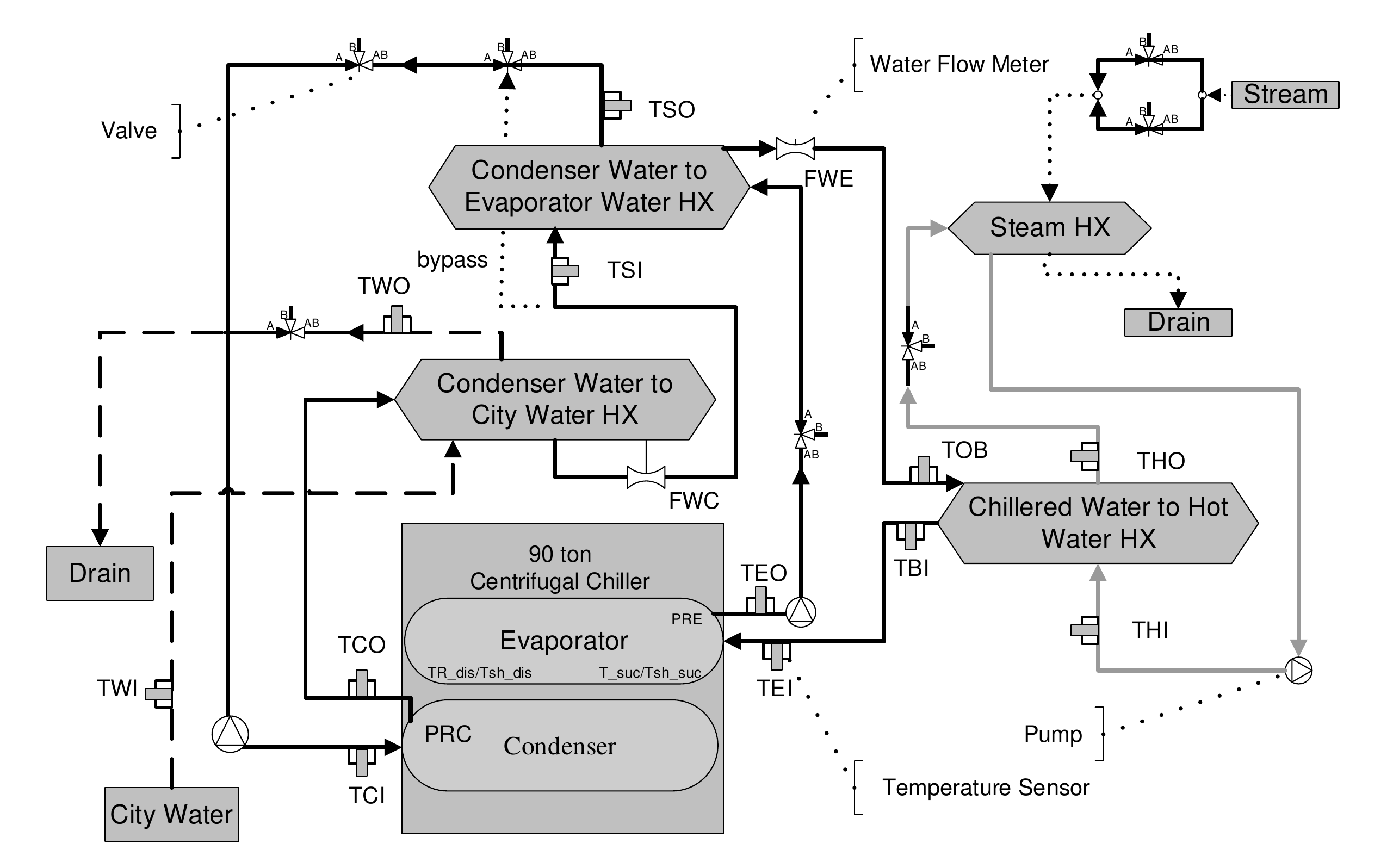}
    \caption{A schematic of the cooling system test facility and sensors mounted in the related water circuits~\cite{li2017fault}.}
    \label{fig:chiller-sensor}
\end{figure}

The RP-1043 chiller dataset~\cite{comstock1999development} is not public but is available for purchase from ASHRAE. The 90-ton chiller studied in the RP-1043 chiller dataset is representative of chillers used in larger installations~\cite{Comstock2002fault}, and consisted of the following parts: evaporator, compressor, condenser, economizer, motor, pumps, fans, and distribution pipes etc. with multiple sensor mounted in the system. Fig.~\ref{fig:chiller-sensor} depicts the cooling system with sensors mounted in both evaporation and condensing circuits.

The same sixteen features and six fault types as used in our previous work~\cite{tan2019encoder} were selected to train our models in the case study. We also attempted to use other sets of selected features than the aforementioned sixteen features, e.g., the features identified in Li~et~al.'s work~\cite{li2016fault}, and we obtained similar results. Detailed descriptions of the sixteen selected features and the six fault types are given in Table~\ref{tbl:features} and Table~\ref{tab:chiller-faults}. Each fault was introduced at four \acp{SL}, and we put fault data of all four \acp{SL} into the fault class.

For the chiller dataset, we used the \texttt{sklearn} package~\cite{scikit-learn} for implementing the \ac{ML} models used in our experiments. A few outlier points were first removed, and the data were first standardized before they were used for training. We experimented using \ac{DT}, \ac{NN}, \ac{OC-SVM} as base learners for constructing ensembles. The base learners were all implemented by using existing modules in \texttt{sklearn}; see our released code\footnote{The code will be released upon paper acceptance.} for further details.

\begin{table}[b]
\caption{Descriptions of the selected variables for the chiller dataset}
\label{tbl:features}
\begin{tabular}{lll}
\hline
\textbf{Sensor} & \textbf{Description} & \textbf{Unit} \\ \hline
TEI & Temperature of entering evaporator water & \degree F \\
TEO & \begin{tabular}[c]{@{}l@{}}Temperature of leaving evaporator water\end{tabular} & \degree F \\
TCI & \begin{tabular}[c]{@{}l@{}}Temperature of entering condenser water\end{tabular} & \degree F \\
TCO & \begin{tabular}[c]{@{}l@{}}Temperature of leaving condenser water\end{tabular} & \degree F \\
Cond Tons & \begin{tabular}[c]{@{}l@{}}Calculated condenser heat rejection rate\end{tabular} & Tons \\
Cooling Tons & Calculated city water cooling rate & Tons \\
kW & Compressor motor power consumption & kW \\
FWC & Flow rate of condenser water & gpm \\
FWE & Flow rate of evaporator water & gpm \\
PRE & \begin{tabular}[c]{@{}l@{}}Pressure of refrigerant in evaporator\end{tabular} & psig \\
PRC & \begin{tabular}[c]{@{}l@{}}Pressure of refrigerant in condenser\end{tabular} & psig \\
TRC & Subcooling temperature & \degree F \\
T\_suc & Refrigerant suction temperature & \degree F \\
Tsh\_suc & \begin{tabular}[c]{@{}l@{}}Refrigerant suction superheat temperature\end{tabular} & \degree F \\
TR\_dis & Refrigerant discharge temperature & \degree F \\
Tsh\_dis & \begin{tabular}[c]{@{}l@{}}Refrigerant discharge superheat temperature\end{tabular} & \degree F \\ \hline
\end{tabular}
\end{table}

% \usepackage{graphicx}
\begin{table}[tb]
\centering
\caption{The six chiller faults in the chiller dataset.}
\label{tab:chiller-faults}
\resizebox{\textwidth}{!}{%
\begin{tabular}{lll}
\hline
\textbf{Fault Type}              & \textbf{Normal Operation} & \textbf{Emulation Method}         \\ \hline
Reduced Condenser Water Flow (FWC)  & 270 gpm     & Reducing water flow rate in condenser  \\
Reduced Evaporator Water Flow (FWE) & 216 gpm     & Reducing water flow rate in evaporator \\
Refrigerant Leak (RL)       & 300 lb                    & Reducing the refrigerant charge   \\
Refrigerant Overcharge (RO) & 300 lb                    & Increasing the refrigerant charge \\
Condenser Fouling (CF)      & 164 tubes                 & Plugging tubes into condenser     \\
Non-condensables in System (NC)     & No nitrogen & Adding Nitrogen to the refrigerant     \\ \hline
\end{tabular}%
}
\end{table}

To carry out the hyperparameter search, we utilized the \texttt{GridSearchCV} module in \texttt{sklearn} to sweep over the prescribed hyperparameter space. For \ac{DT} models, we swept the \texttt{max\_depth} parameter over the range $\{8,10,12,15,20\}$ and attempted various parameters configurations such as \texttt{criterion} (measuring the quality of split) and \texttt{splitter} (strategy used to choose the split at each node). For \ac{NN} models (multilayer perceptrons), we tried several different network topologies with depth ranging from $2$ to $5$, various batches sizes ($32$, $64$, $128$) and optimizer settings (\texttt{sgd} or \texttt{adam}). For \ac{OC-SVM} models with \ac{RBF} kernels, we conducted a grid search over parameters $\nu$ and $\gamma$ \cite{jin2019one}. We refer interested readers to our released code base for further implementation details.

After the grid search, the top $R$ sets of hyperparameters for each model type were picked out and used for constructing the base learners for bagging ensembles. The bagging ensembles in this study were implemented using the \texttt{Bagging} module from \texttt{sklearn}, which enabled us to create bagging models with different types of base learners. The sizes of random sample subsets for training each base model can be specified through the \texttt{max\_samples} argument.

\subsection{Experiments on the \ac{DR} Dataset}

We used \ac{CNN} models for classifying image models in the \ac{DR} dataset. The \ac{CNN} models were implemented using \texttt{pytorch}~\cite{pytorch}. The deep learning models used to construct our ensembles vary in their architecture, image data resolution, training set selection, number of training epochs and data augmentation strengths. Two different \ac{CNN} architectures, \textit{ResNet34}~\cite{he2016deep} and \textit{VGG16}~\cite{simonyan2014very}, were used in our experiments. We used the binary-crossentropy loss function and the Adam~\cite{Kingma2015AdamAM} optimizer during training. All network parameters were initialized with the weights from pretrained models provided by the \texttt{torchvision}~\cite{torchvision} package that were created for classifying objects from the ImageNet database.

Since our experiments involved scanning various $\rho$ values, to reduce the total training effort, we first trained our models with non-incipient disease data (only SL0\,\&\,SL3\,\&\,SL4) for $130$ epochs, and then continued to train the resulting networks with all training data (SL0 to SL4) till convergence. Most trained models reached an \ac{AUC} of above $0.98$ on both the training and the validation sets. We discarded the bad performing models and put the rest into a pool. The retained models in the pool were then used as base learners for constructing ensembles. To create an ensemble model instance, we randomly picked $K$ single learners from the pool. In our experiment, we evaluated $K=5$ and $K=10$, two ensemble sizes used in previous works~\cite{lakshminarayanan2017simple,gulshan2016development}. Their individual predictions were then combined and grouped for later analysis.

\begin{figure}[htb]
  \centering
  \begin{subfigure}[t]{0.48\linewidth}
    \centering
    \includegraphics[height=2cm]{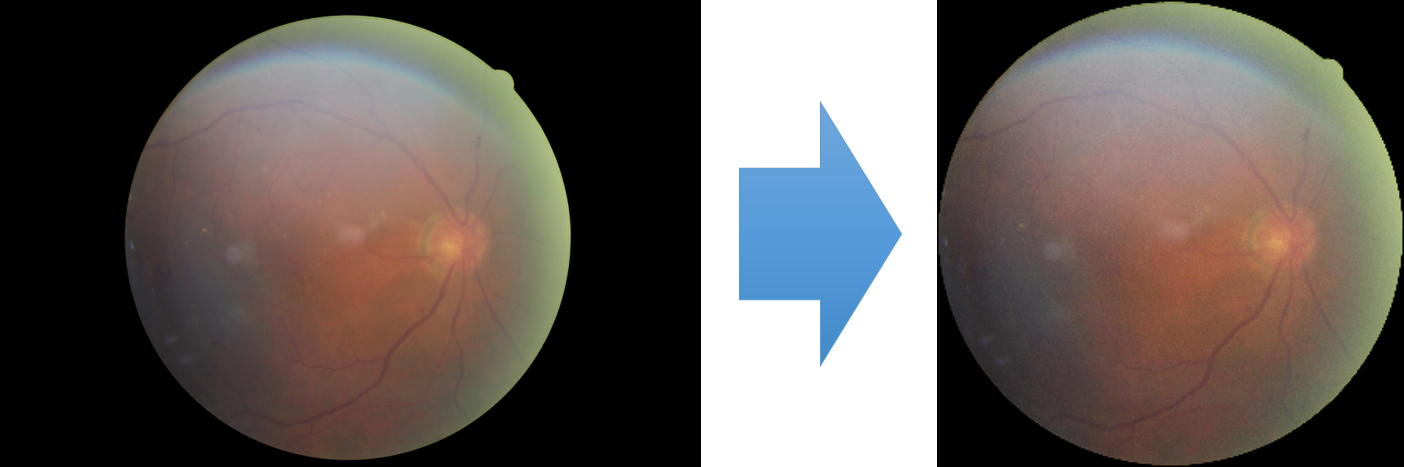}
    \caption{Cropping}
    \label{fig:fundus-format-1}
  \end{subfigure}
  \begin{subfigure}[t]{0.48\linewidth}
    \centering
    \includegraphics[height=2cm]{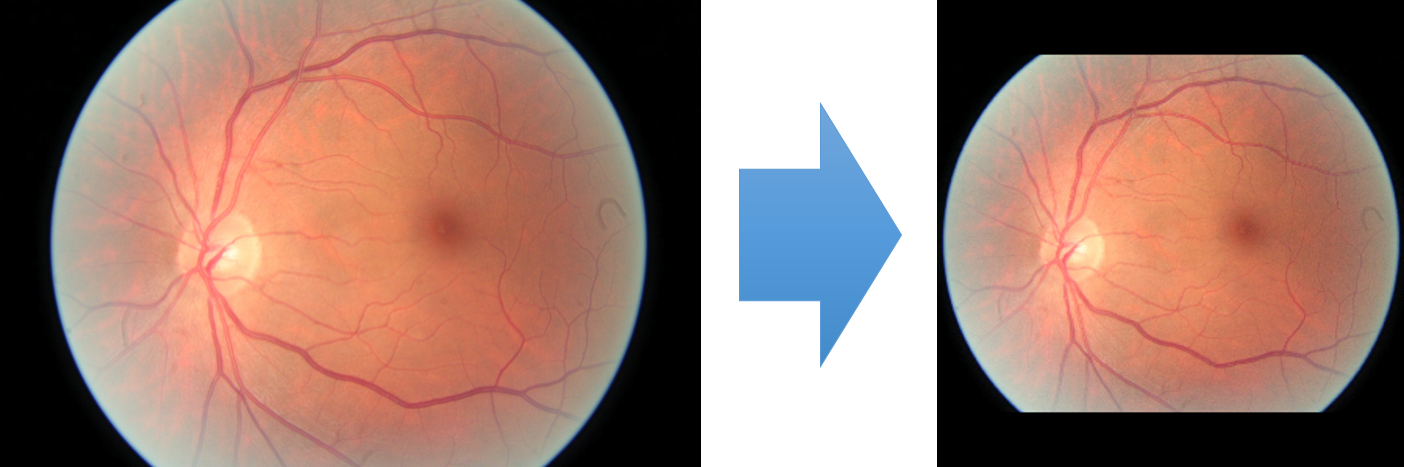}
    \caption{Blank padding}
    \label{fig:fundus-format-2}
  \end{subfigure}
  \caption{Preprocessing the fundus image data from the Kaggle-DR dataset~\cite{cuadros2009eyepacs}.}
  \label{fig:preprocessing}
\end{figure}

\paragraph{Data preprocessing} 
The image data used in our experiments were all unified into square-shaped images with resolutions $224\times 224$ or $384\times 384$ in our preprocessing procedure. For training each neural network model, only images of the same resolution were used. The original images came with either of the two formats as exemplified in Fig.~\ref{fig:preprocessing}. In the first format as shown in Fig.~\ref{fig:fundus-format-1}, the entire fundus was visible in the image. We cropped the original image such that the fundus would tightly fitted inside the square. In the second format of input images shown in Fig.~\ref{fig:fundus-format-2}, part of the fundus was not visible. We padded blank strips to make the image square-shaped and in a unified resolution; see our released code for further details.

\paragraph{Data augmentation}
Data augmentation~\cite{Perez2017TheEO} has proved to be an important technique for training deep learning models that can prevent overfitting and can enhance model's generalization ability. We utilized several different types of data augmentation operations at training time that are available from the \texttt{torchvision} package~\cite{torchvision}. These operations included \texttt{RandomResizedCrop}, \texttt{adjust\_brightness}, \texttt{adjust\_saturation} and \texttt{adjust\_contrast} that could randomly adjust the aspect ratio, the brightness, the saturation and the contrast respectively. The degree (strength) of data augmentation in our experiments was controlled by a multiplier $\gamma\in\{0.1,0.3,0.5,0.7\}$; see the provided code for further details.

\section{Modeling the Distributions of Ensemble Outputs}

\begin{figure}[p]
\centering
\begin{subfigure}[t]{0.19\textwidth}
    \centering
    \includegraphics[height=2cm]{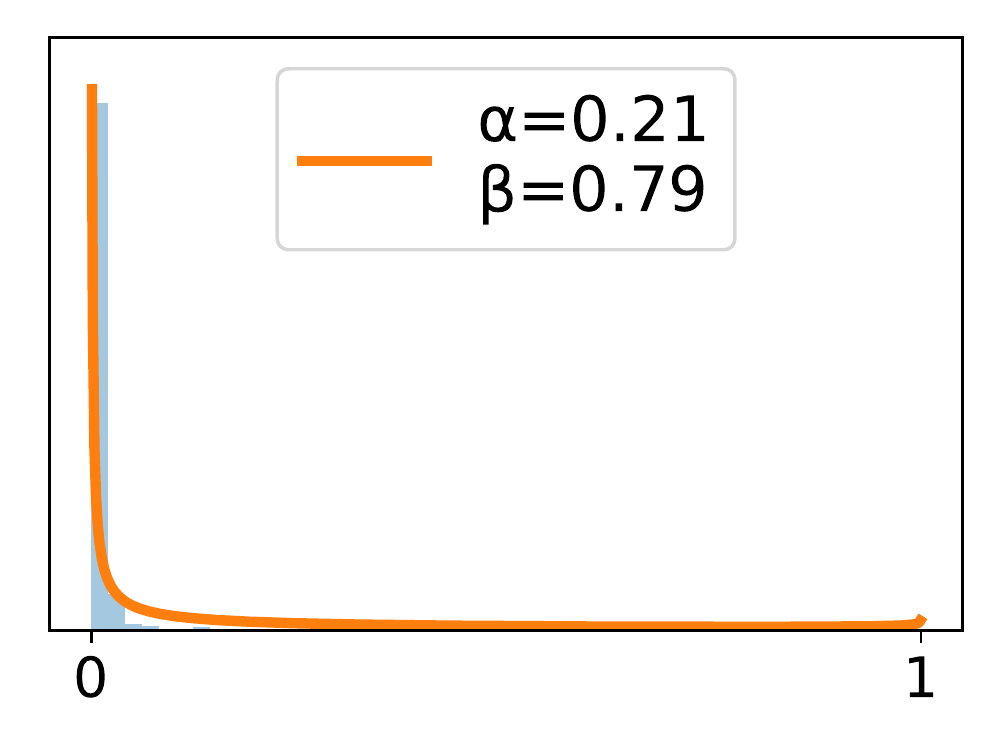}
    \includegraphics[height=2cm]{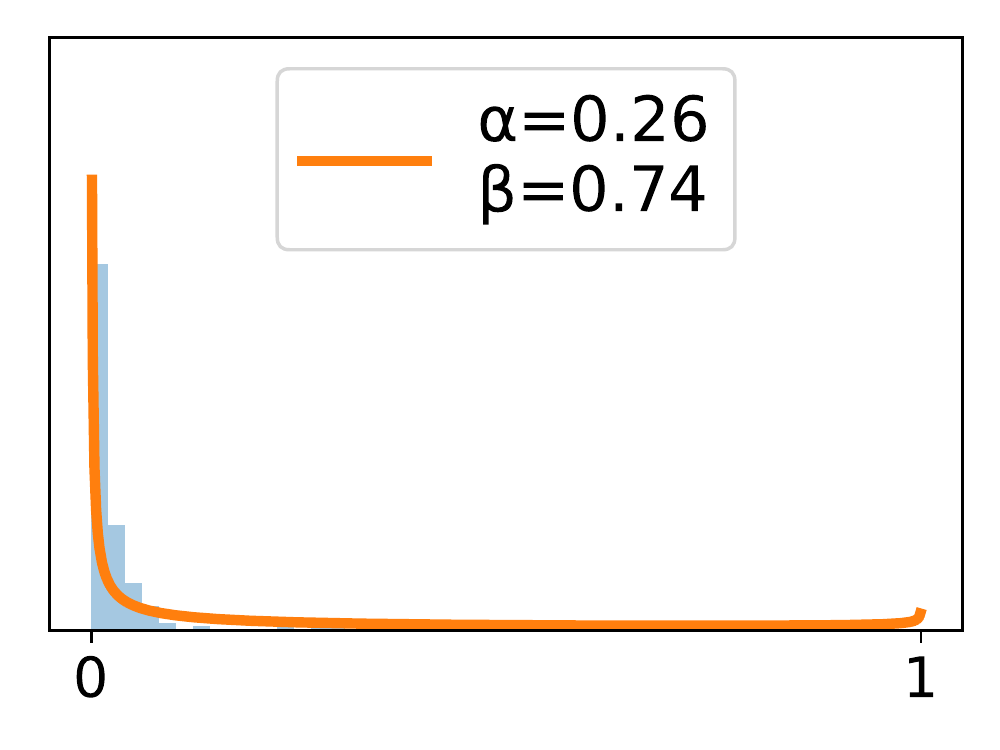}
    \includegraphics[height=2cm]{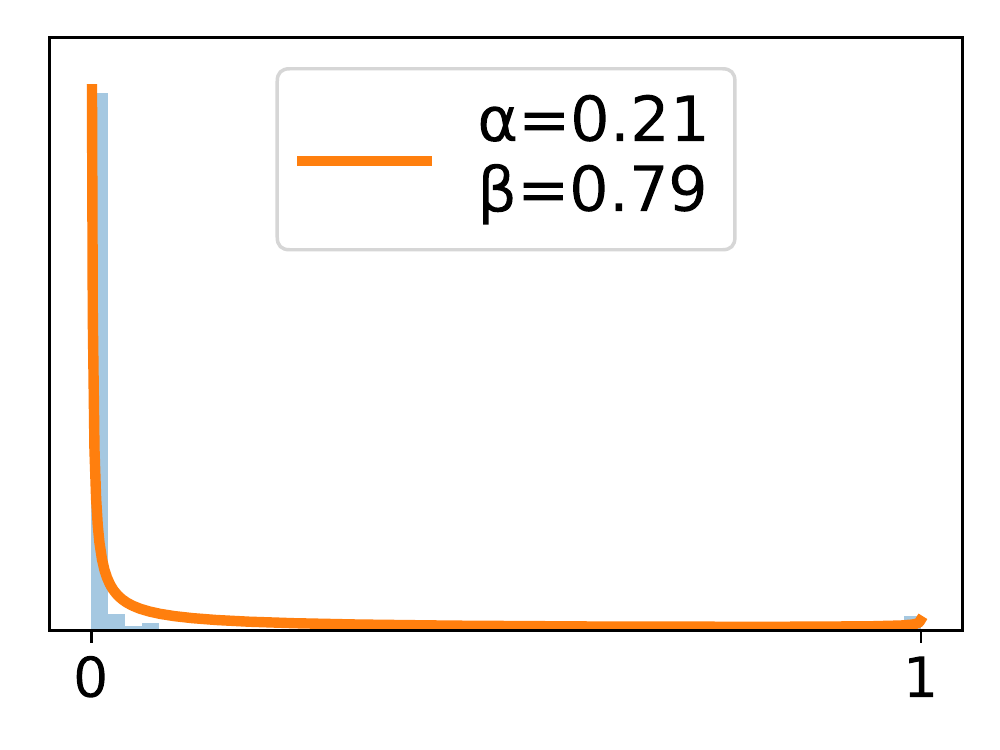}
    \includegraphics[height=2cm]{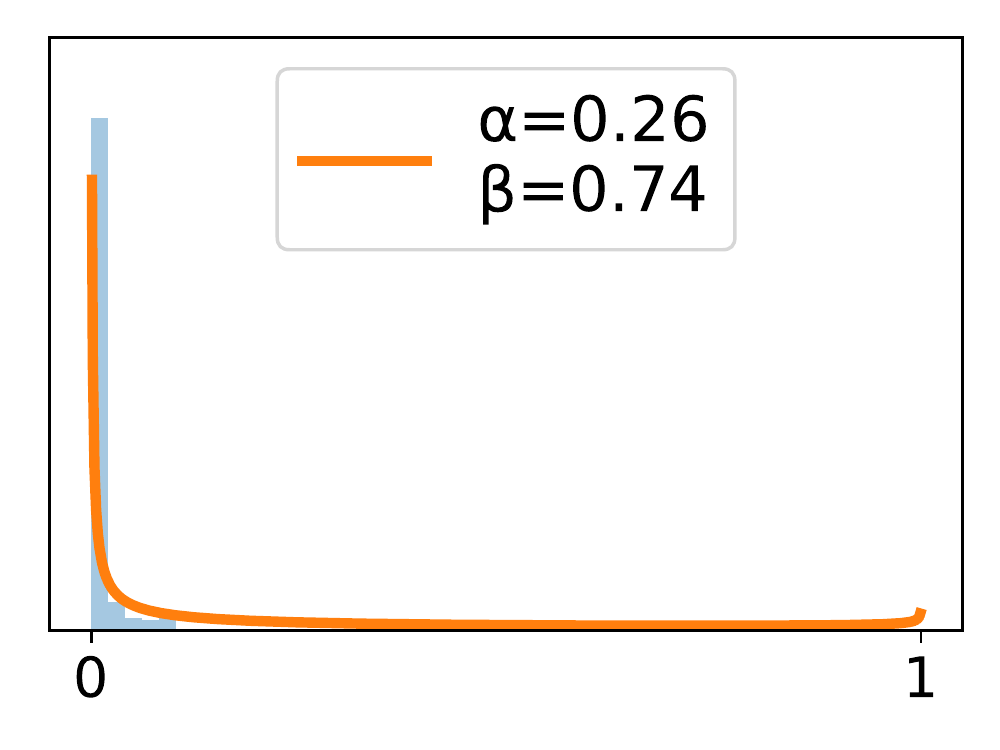}
    \includegraphics[height=2cm]{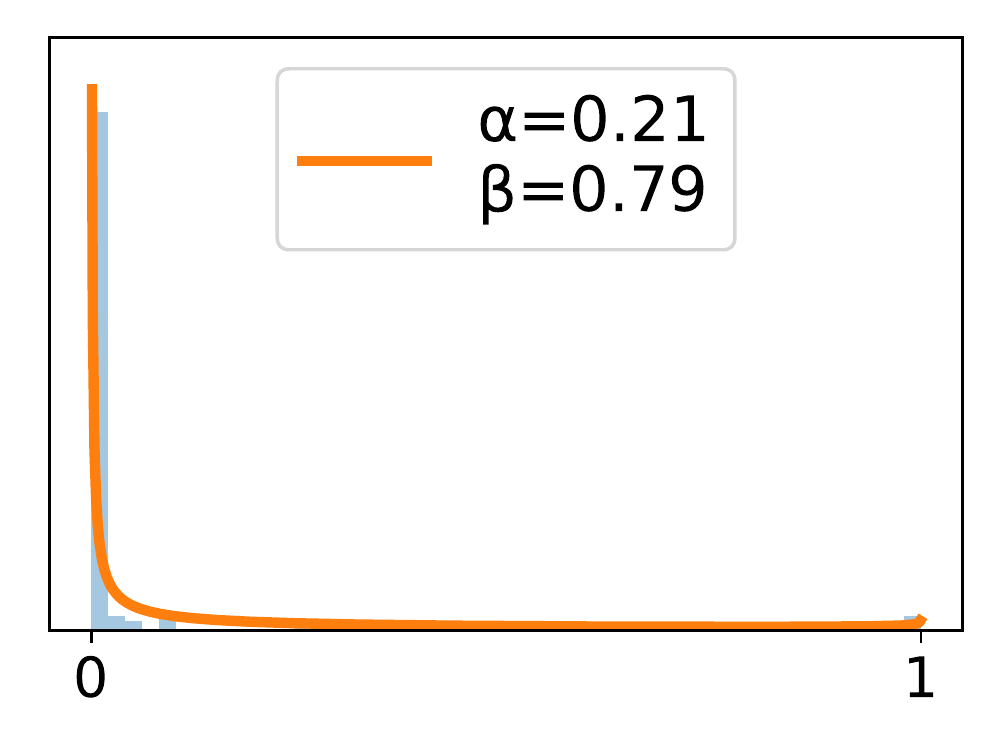}
    \includegraphics[height=2cm]{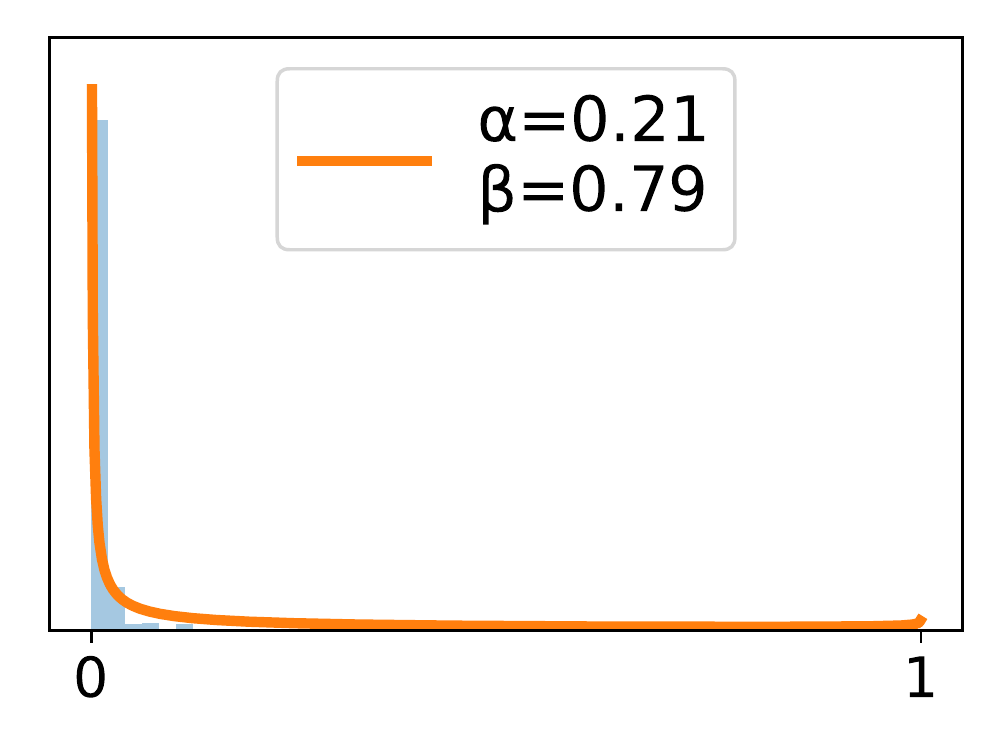}
    \includegraphics[height=2cm]{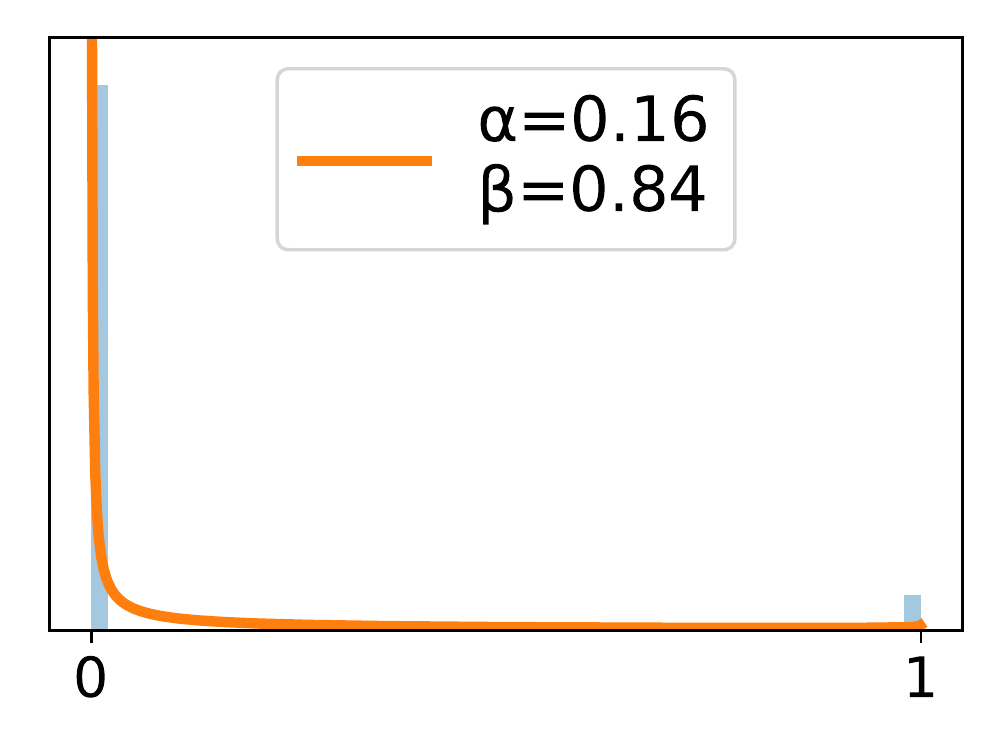}
    \includegraphics[height=2cm]{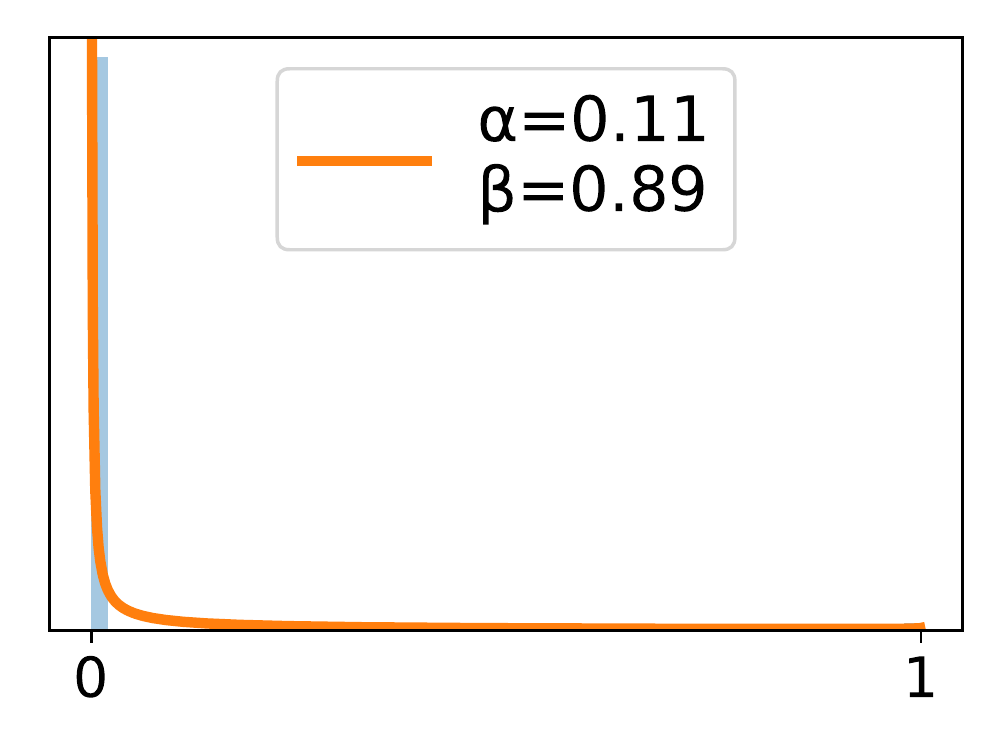}
    \includegraphics[height=2cm]{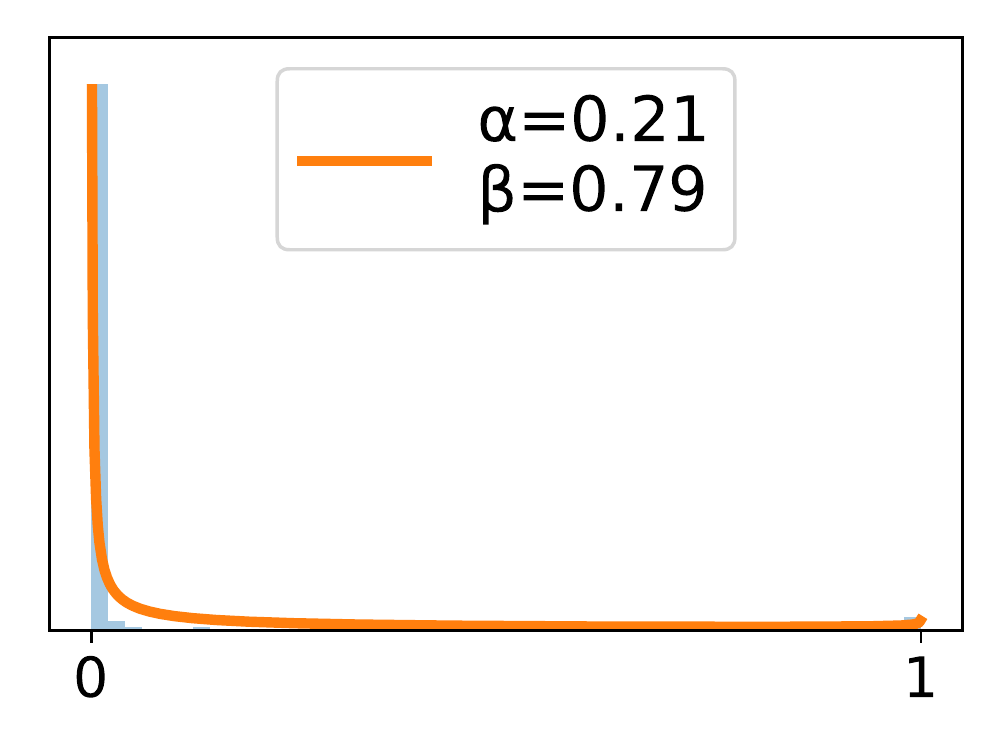}
    \caption{SL0}
\end{subfigure}\hfill
\begin{subfigure}[t]{0.19\textwidth}
    \centering
    \includegraphics[height=2cm]{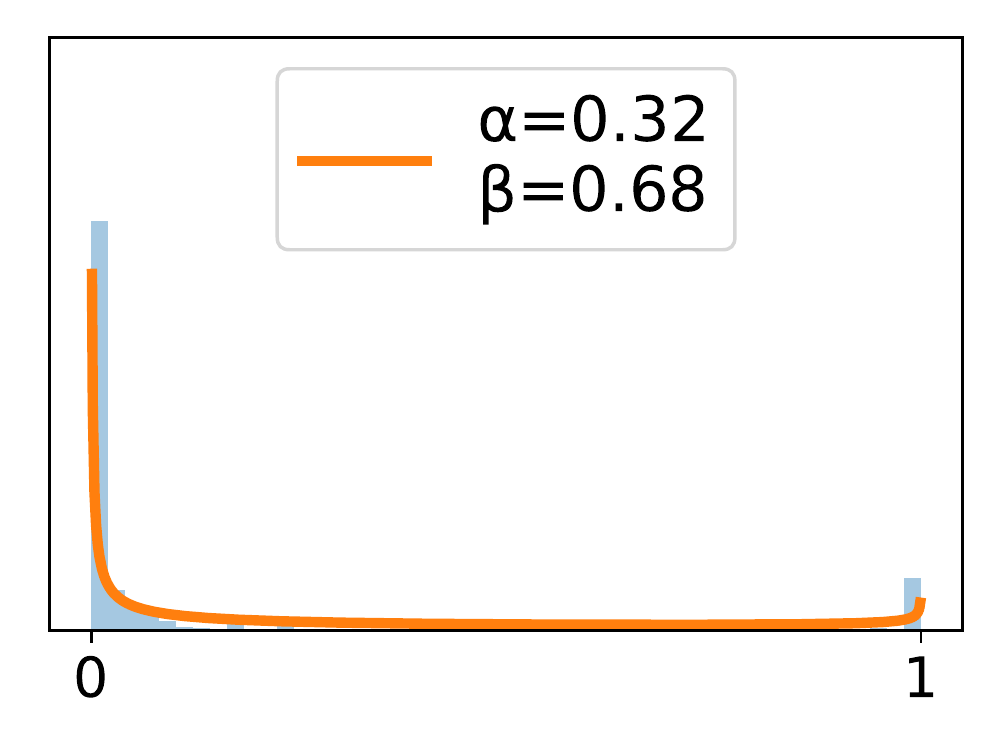}
    \includegraphics[height=2cm]{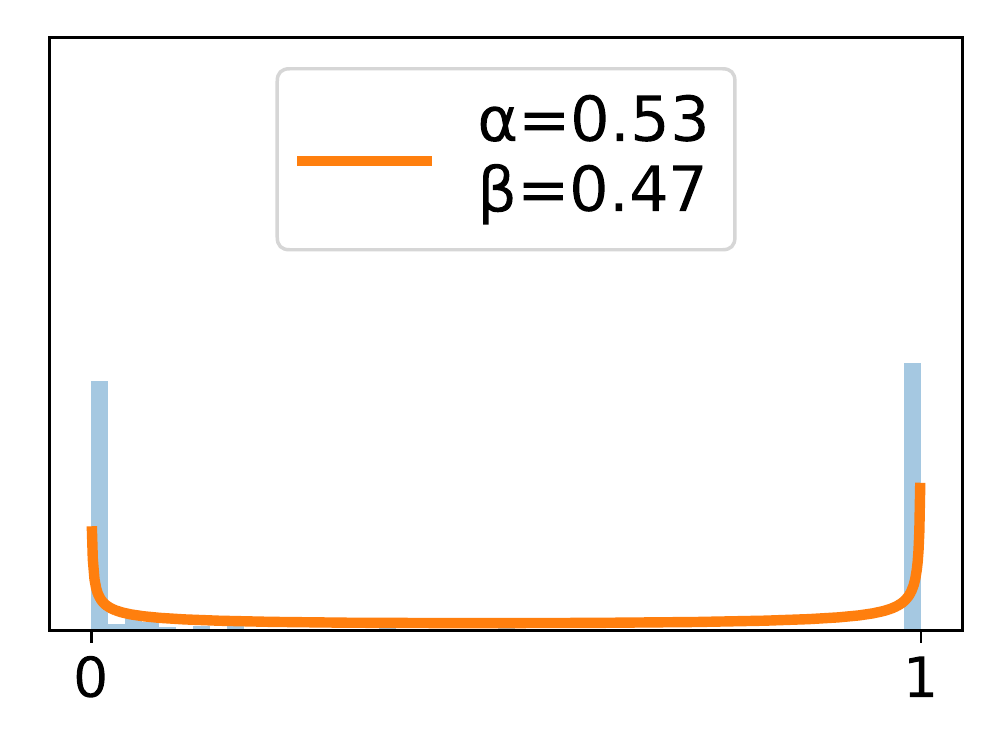}
    \includegraphics[height=2cm]{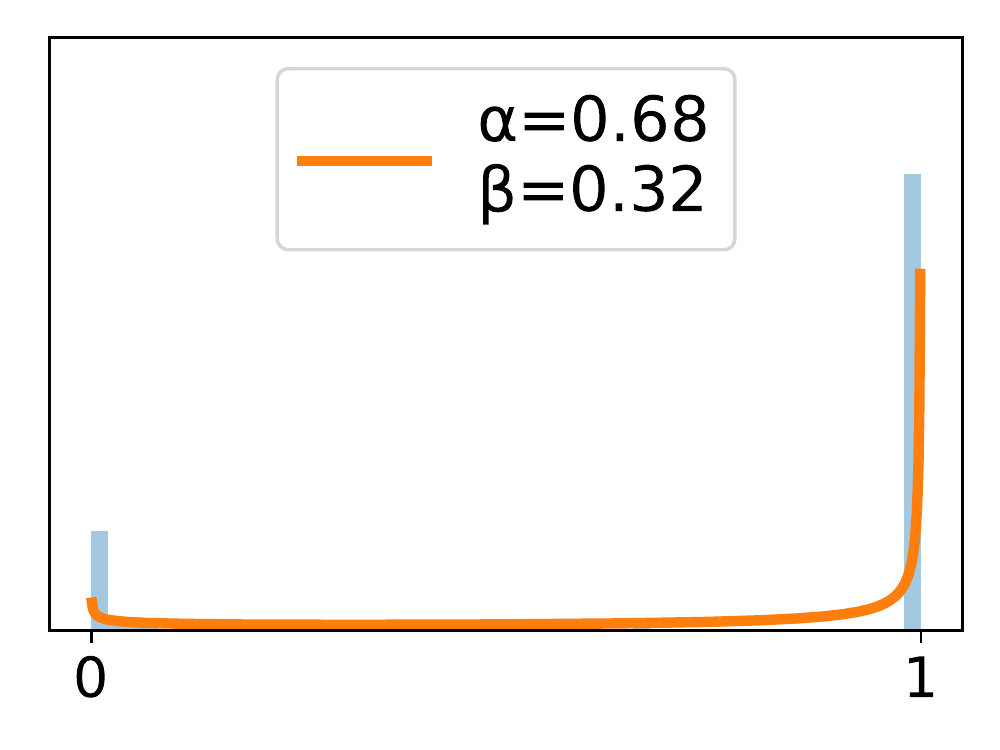}
    \includegraphics[height=2cm]{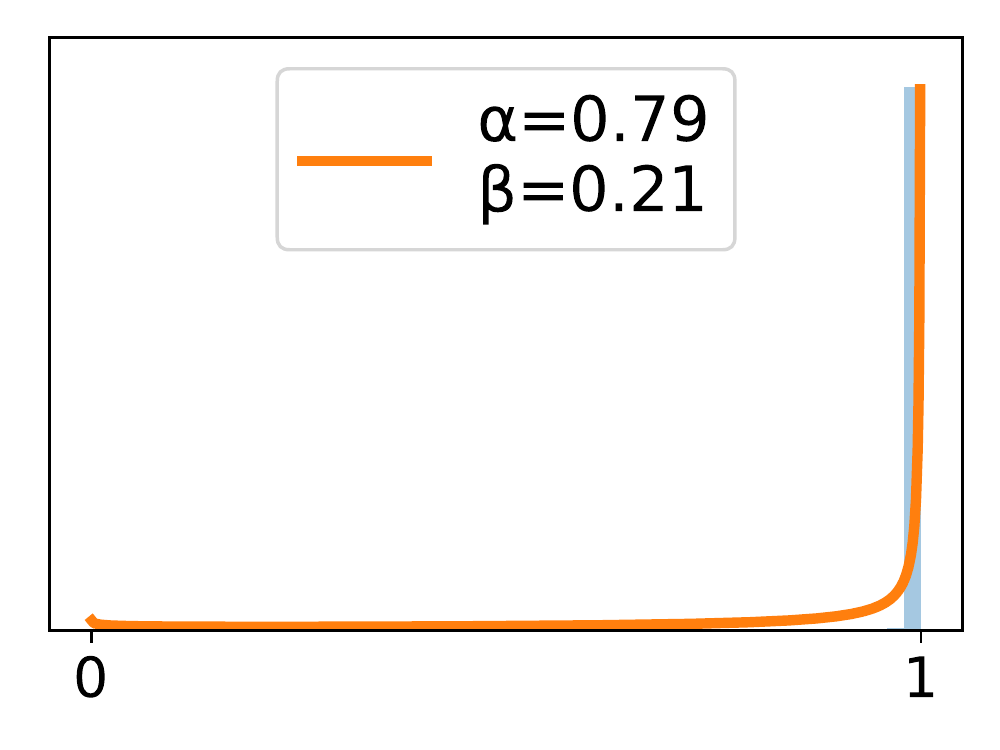}
    \includegraphics[height=2cm]{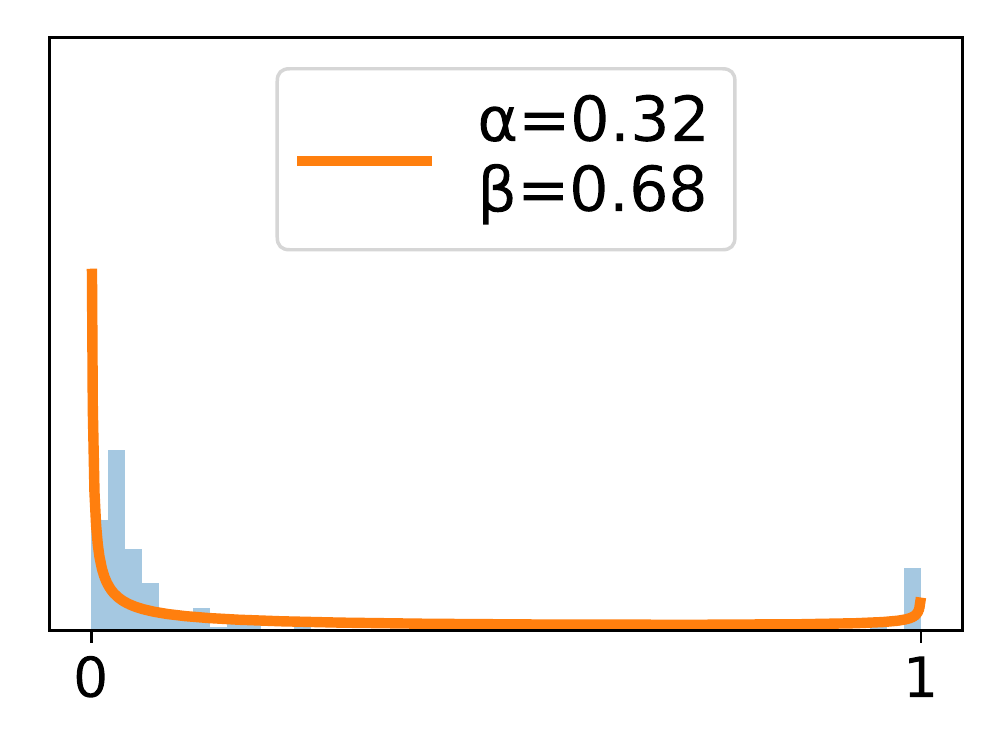}
    \includegraphics[height=2cm]{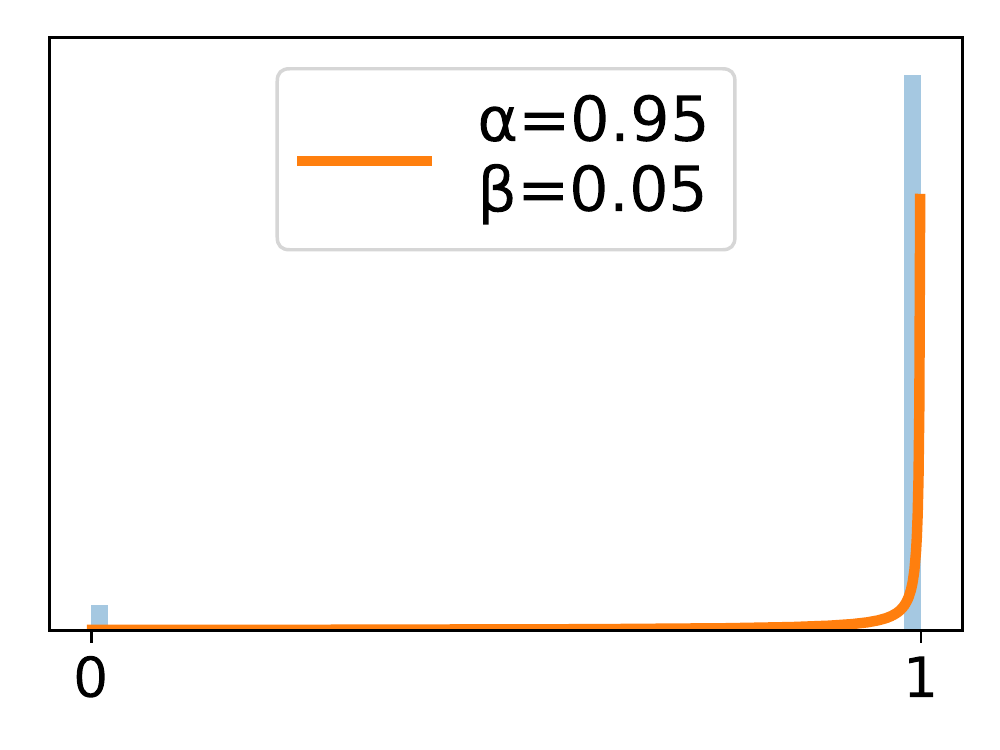}
    \includegraphics[height=2cm]{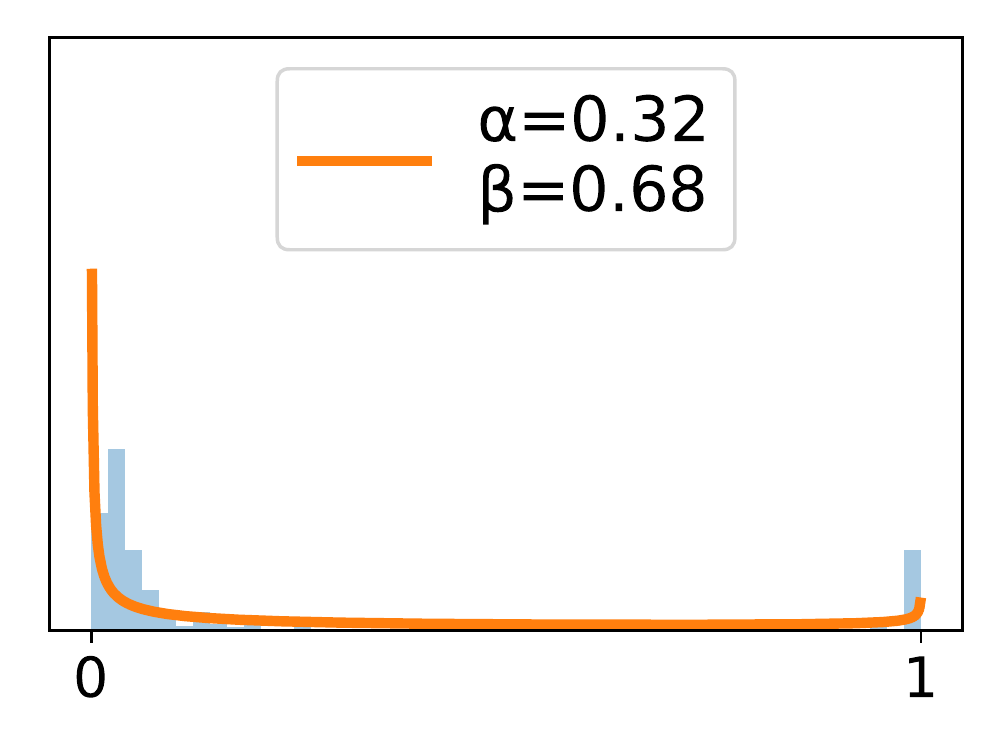}
    \includegraphics[height=2cm]{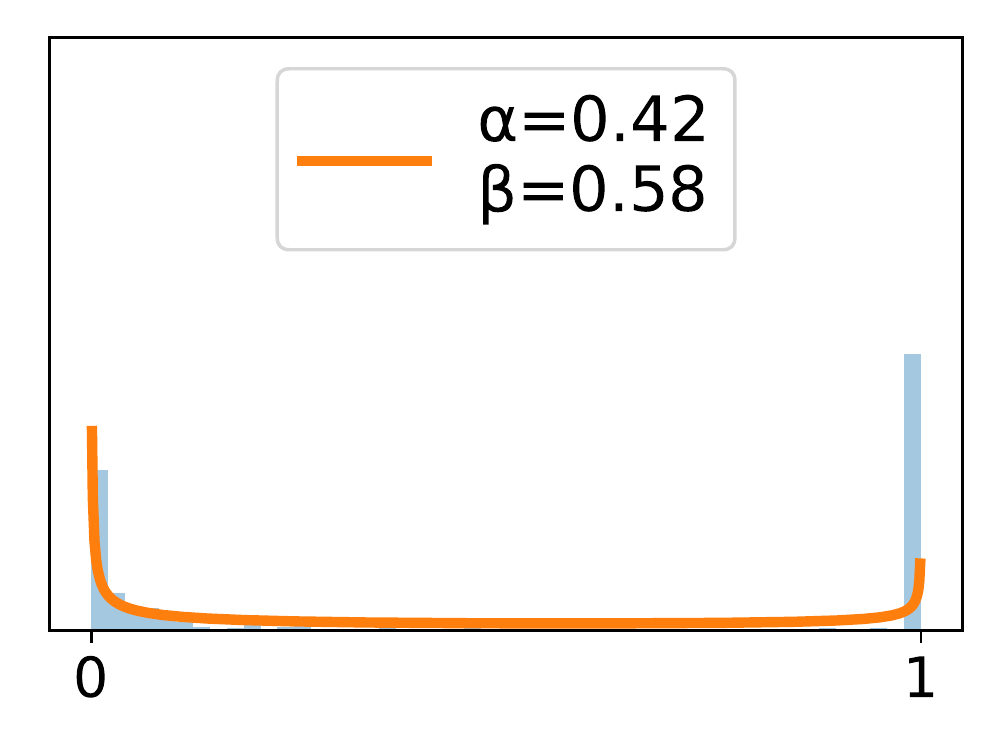}
    \includegraphics[height=2cm]{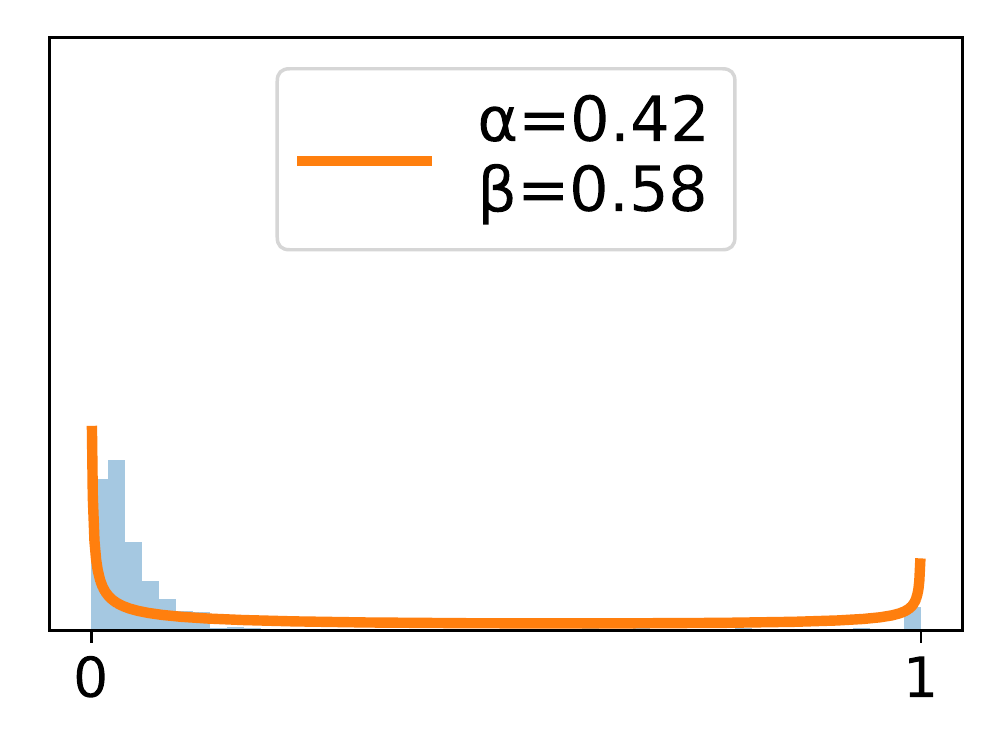}
    \caption{SL1}
\end{subfigure}\hfill
\begin{subfigure}[t]{0.19\textwidth}
    \centering
    \includegraphics[height=2cm]{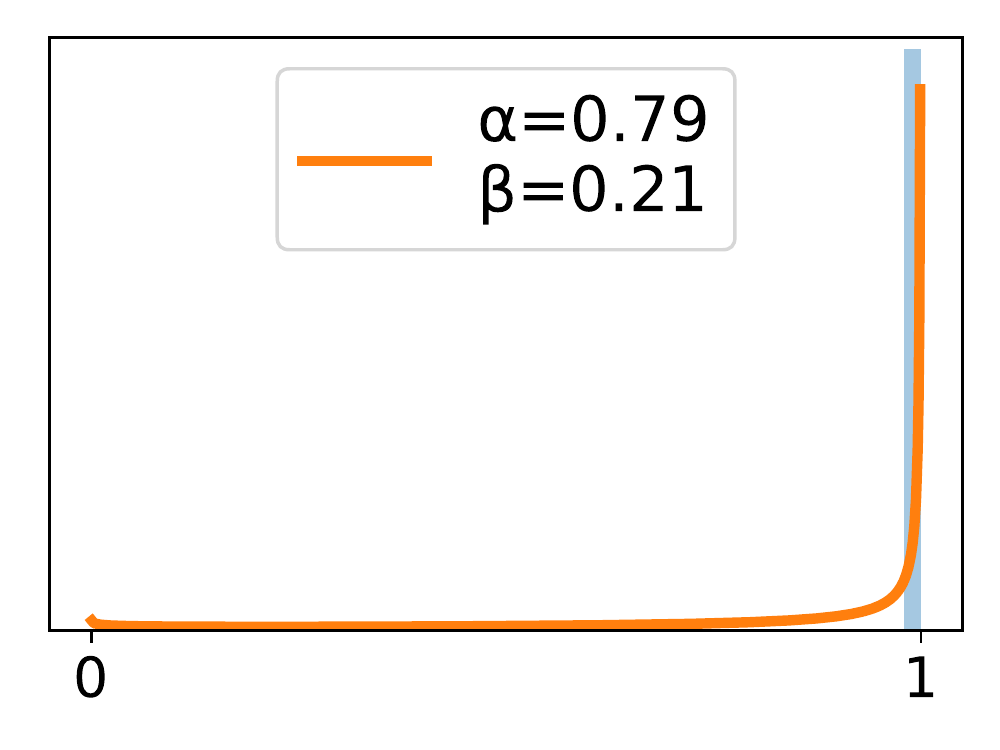}
    \includegraphics[height=2cm]{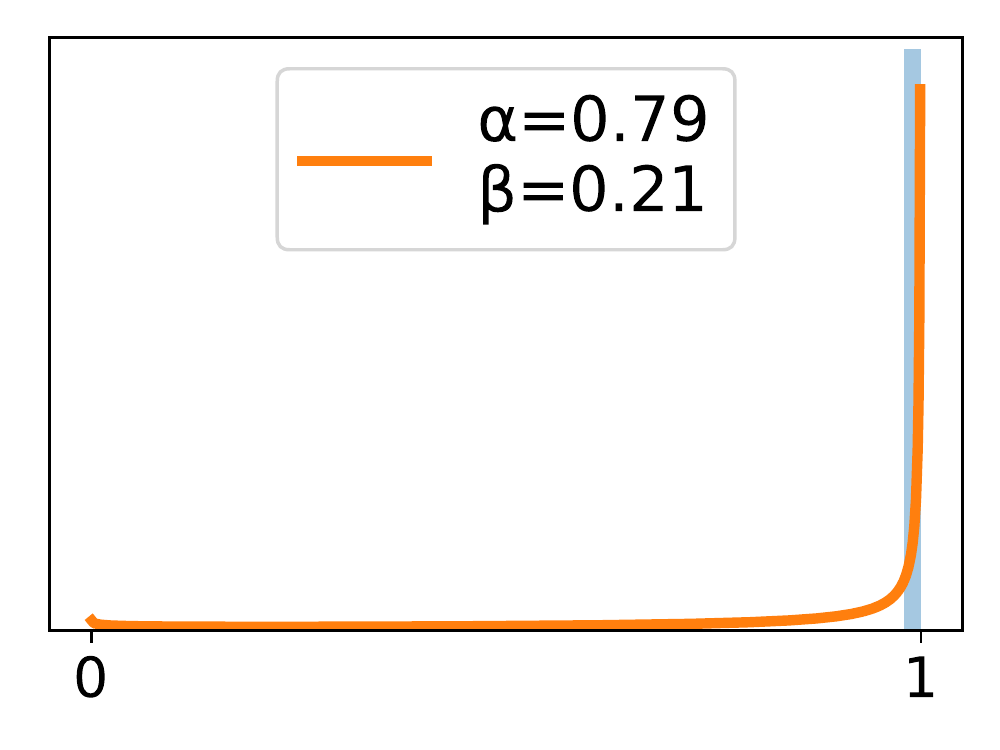}
    \includegraphics[height=2cm]{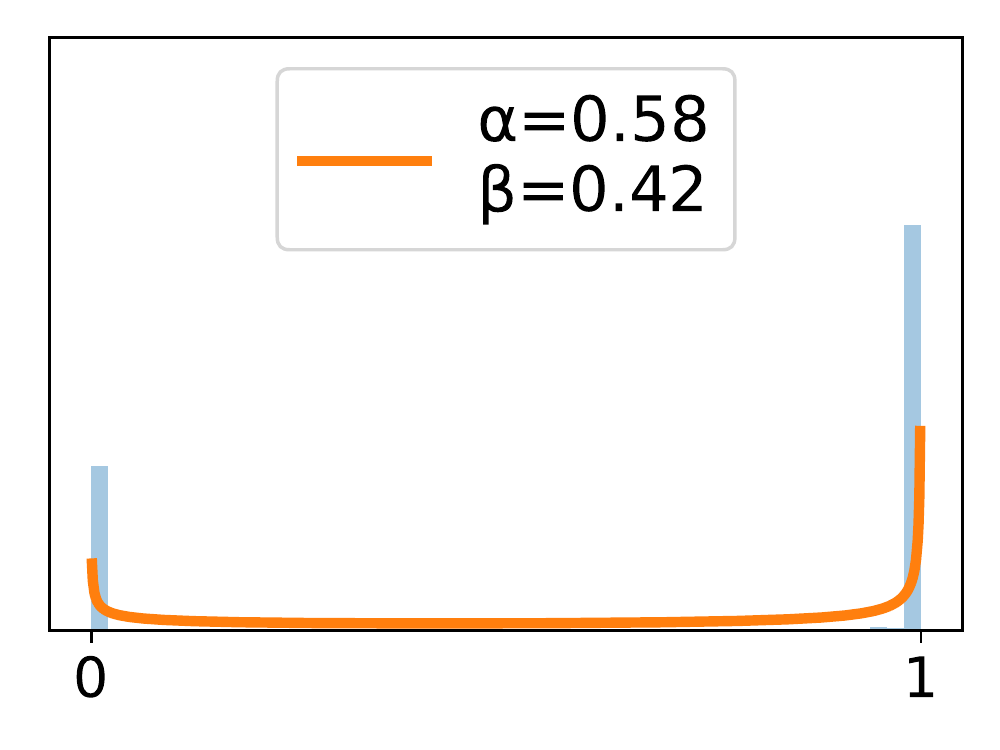}
    \includegraphics[height=2cm]{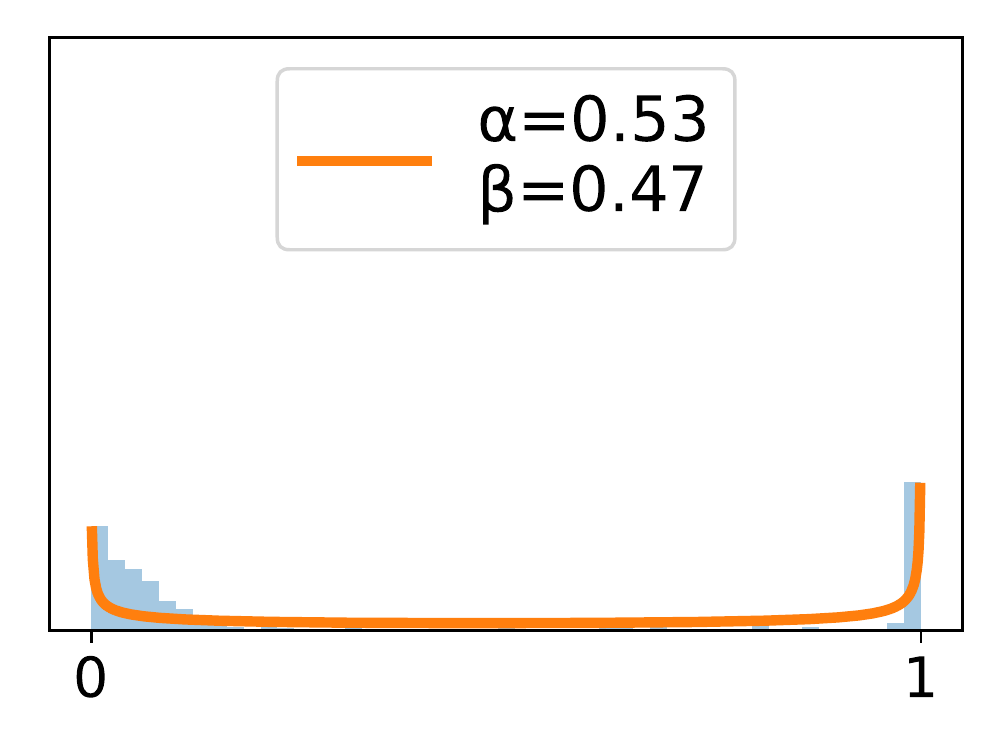}
    \includegraphics[height=2cm]{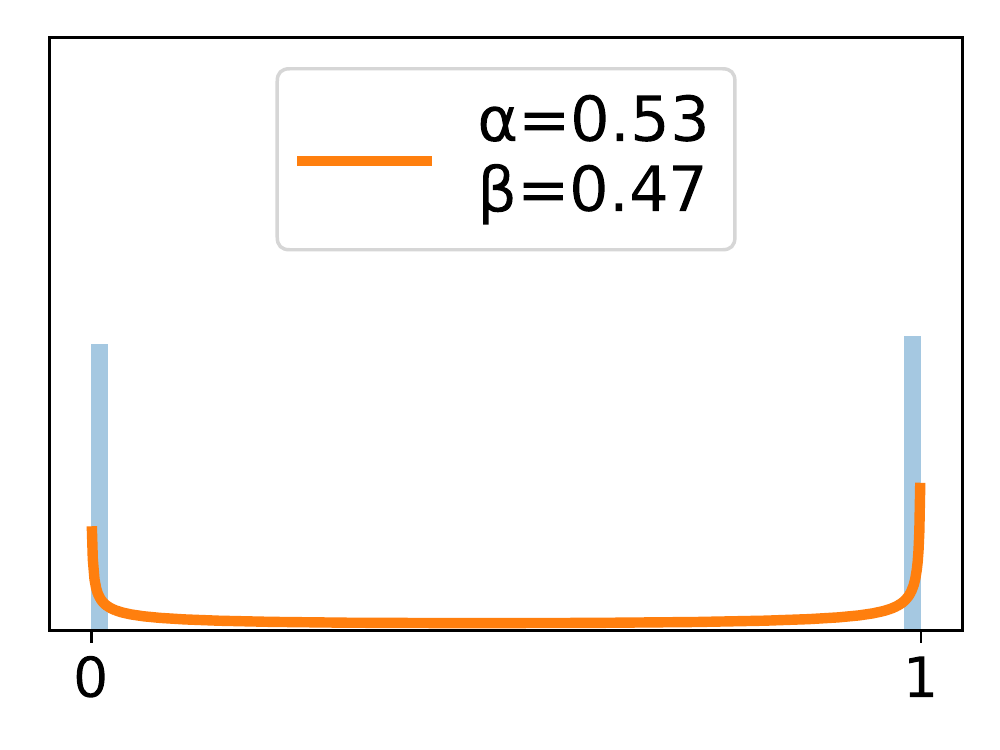}
    \includegraphics[height=2cm]{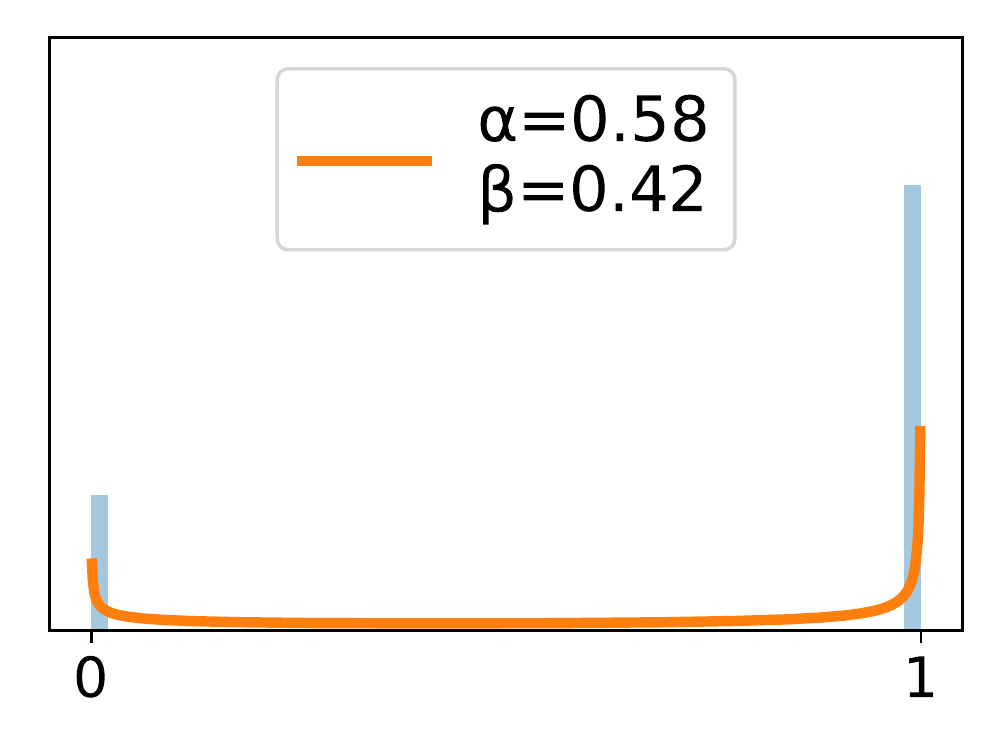}
    \includegraphics[height=2cm]{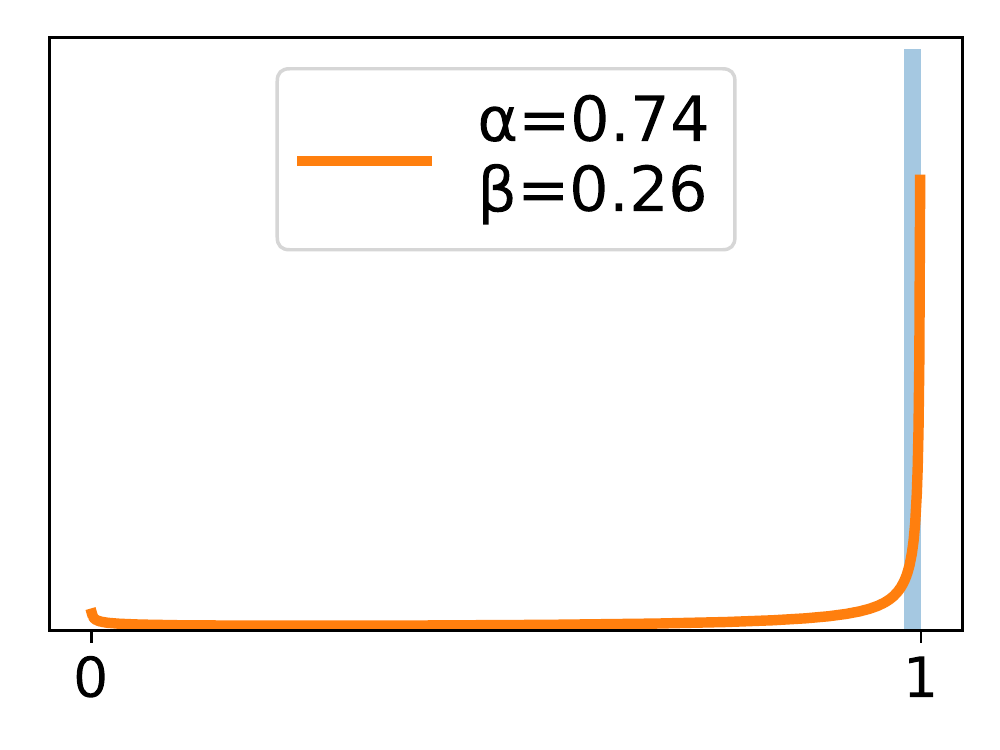}
    \includegraphics[height=2cm]{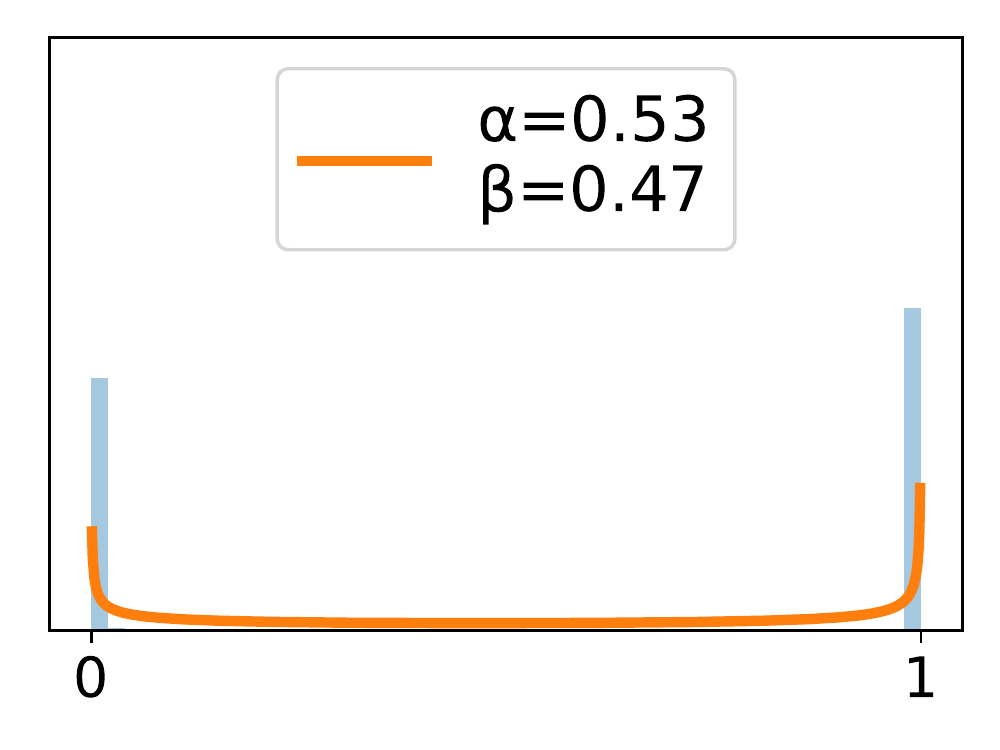}
    \includegraphics[height=2cm]{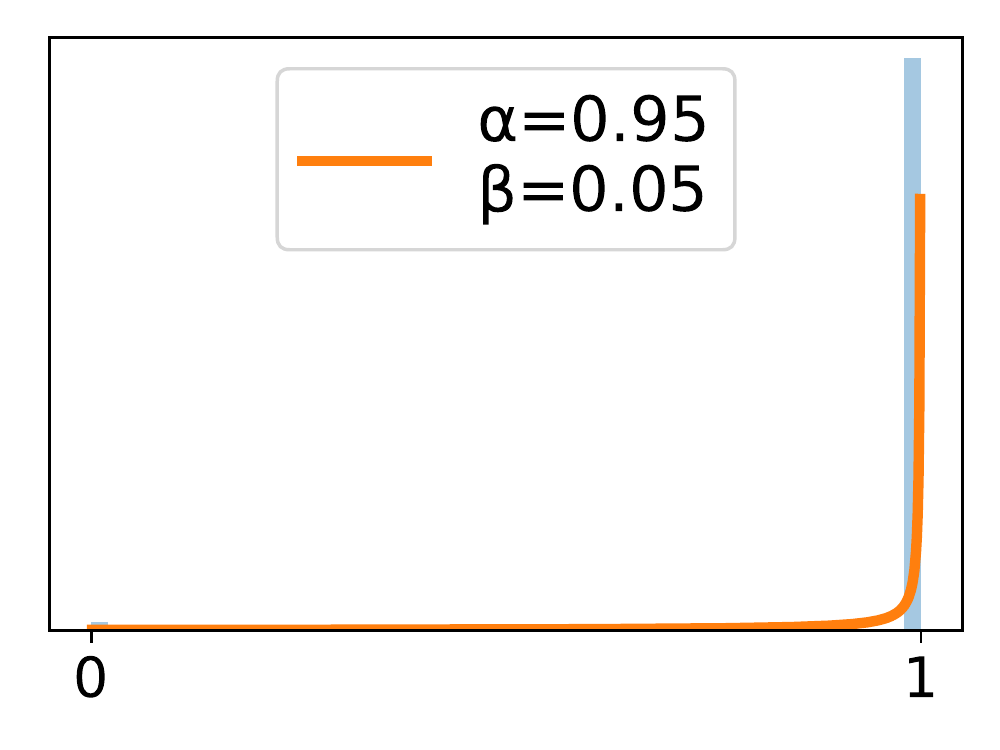}
    \caption{SL2}
\end{subfigure}\hfill
\begin{subfigure}[t]{0.19\textwidth}
    \centering
    \includegraphics[height=2cm]{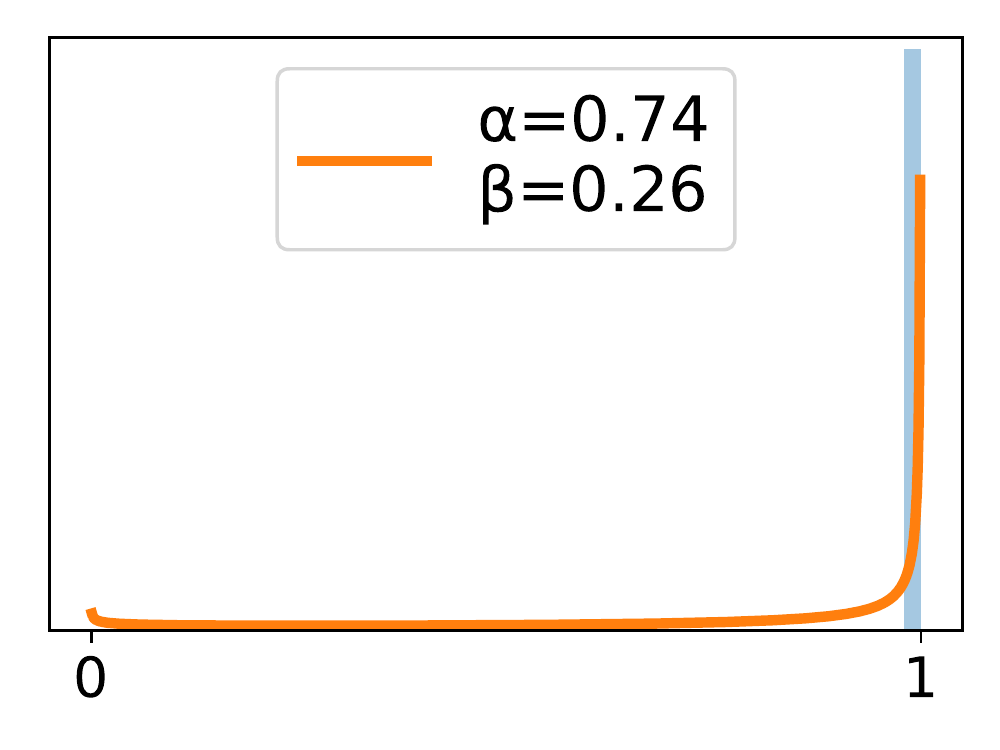}
    \includegraphics[height=2cm]{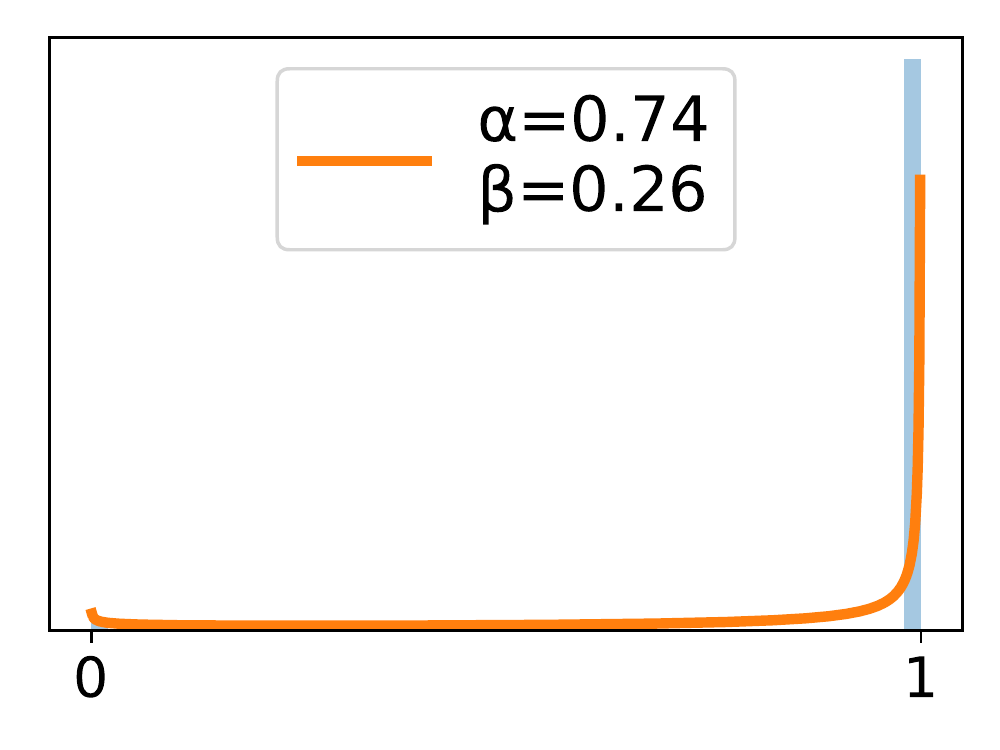}
    \includegraphics[height=2cm]{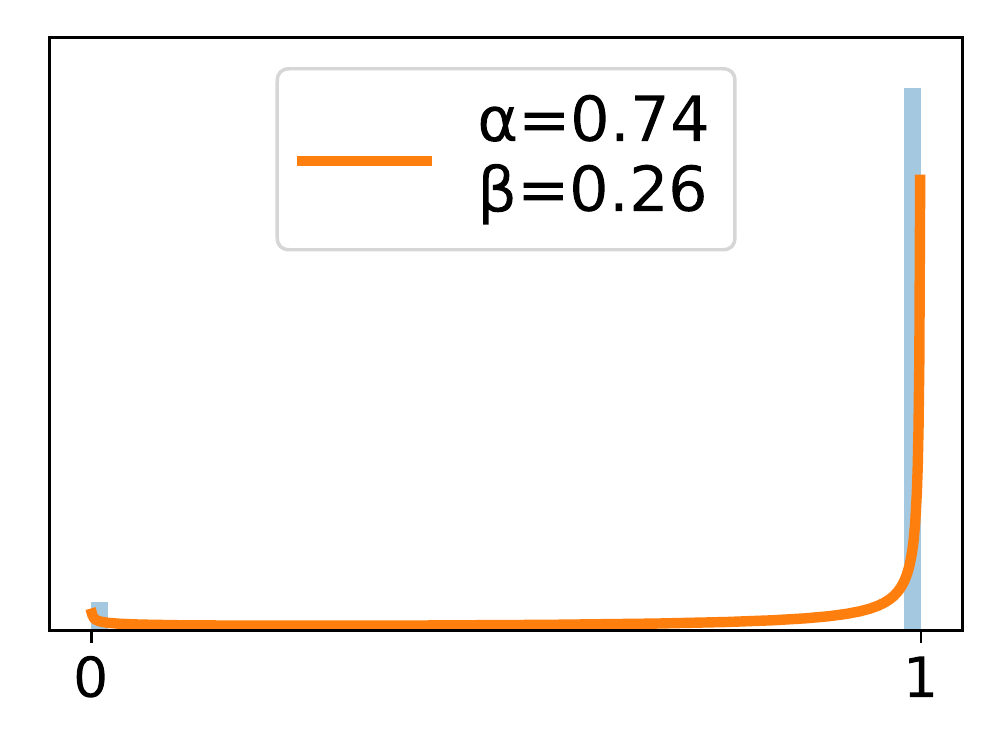}
    \includegraphics[height=2cm]{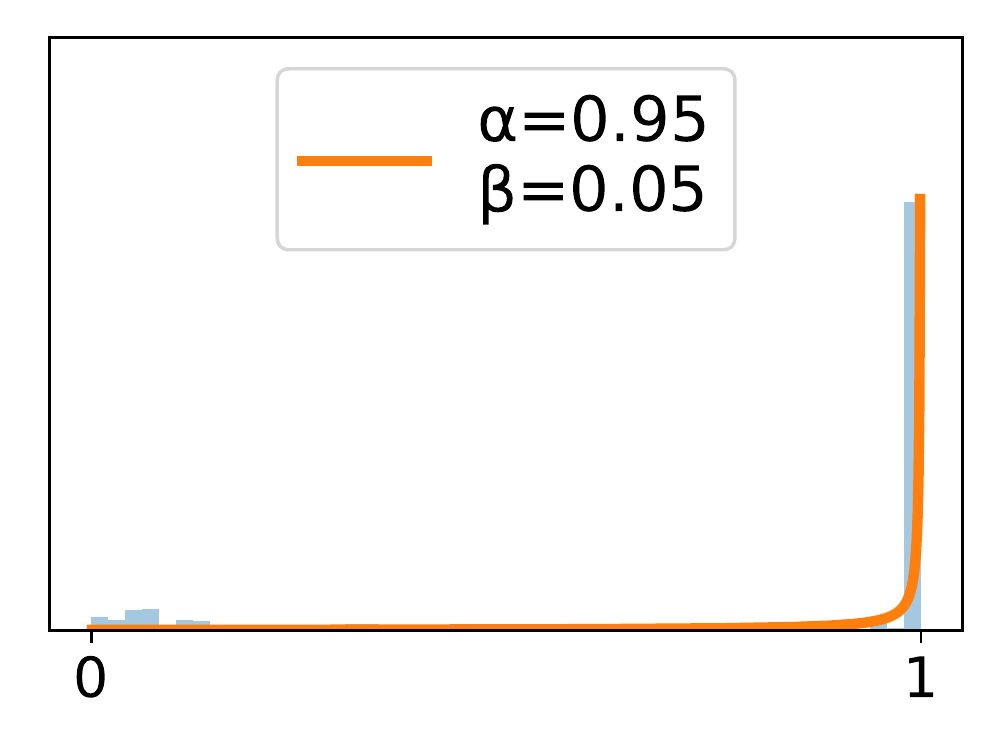}
    \includegraphics[height=2cm]{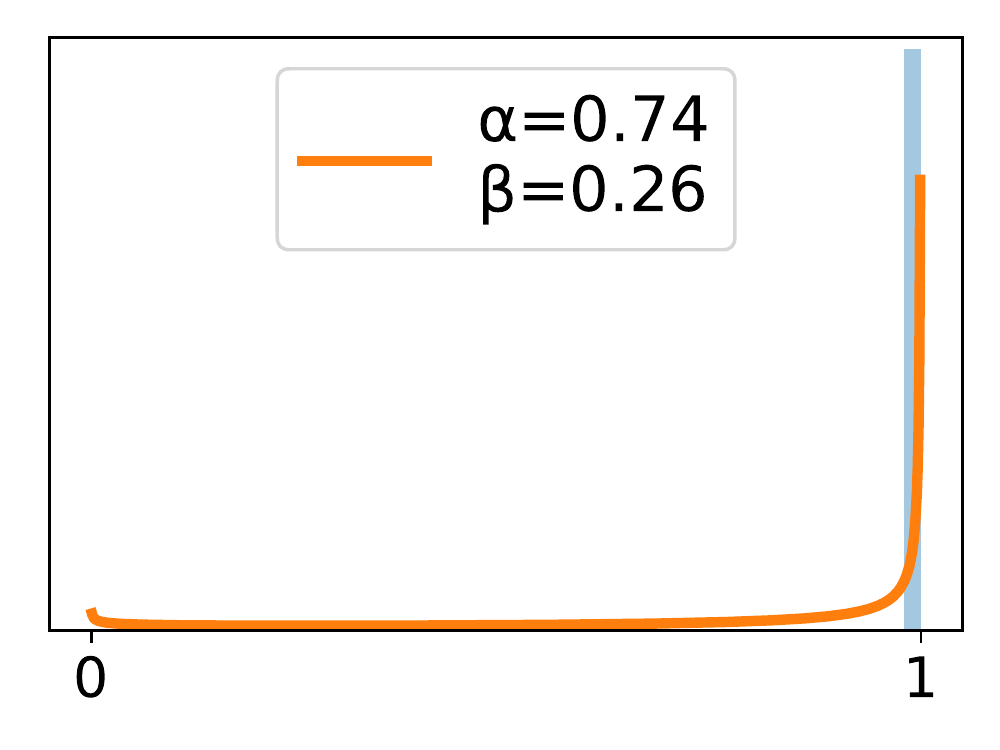}
    \includegraphics[height=2cm]{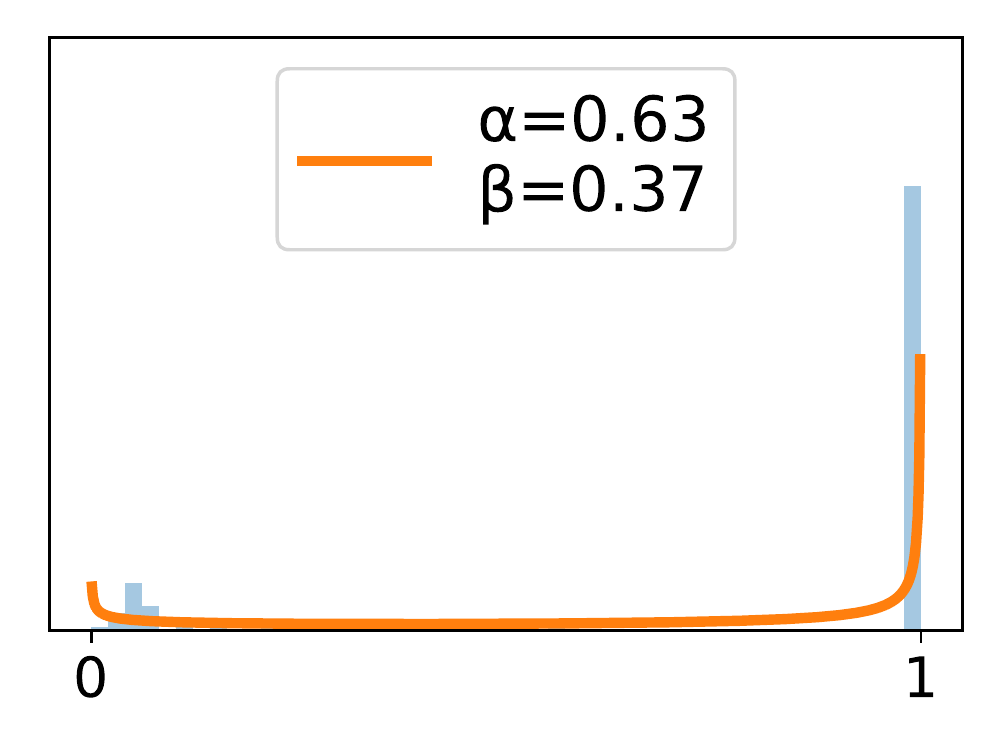}
    \includegraphics[height=2cm]{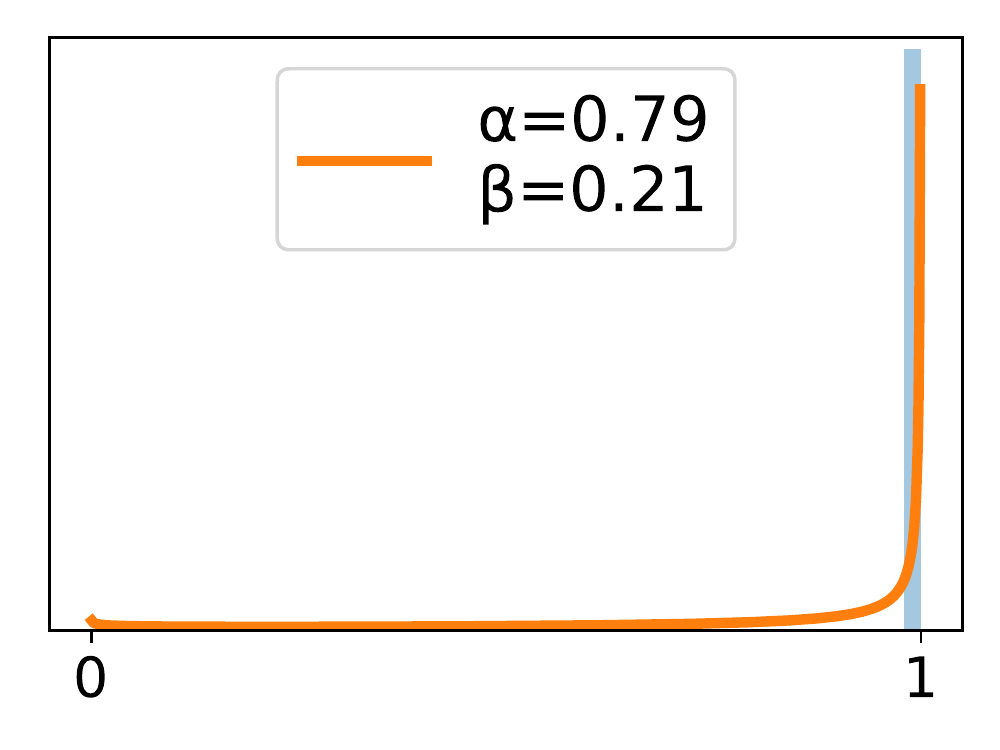}
    \includegraphics[height=2cm]{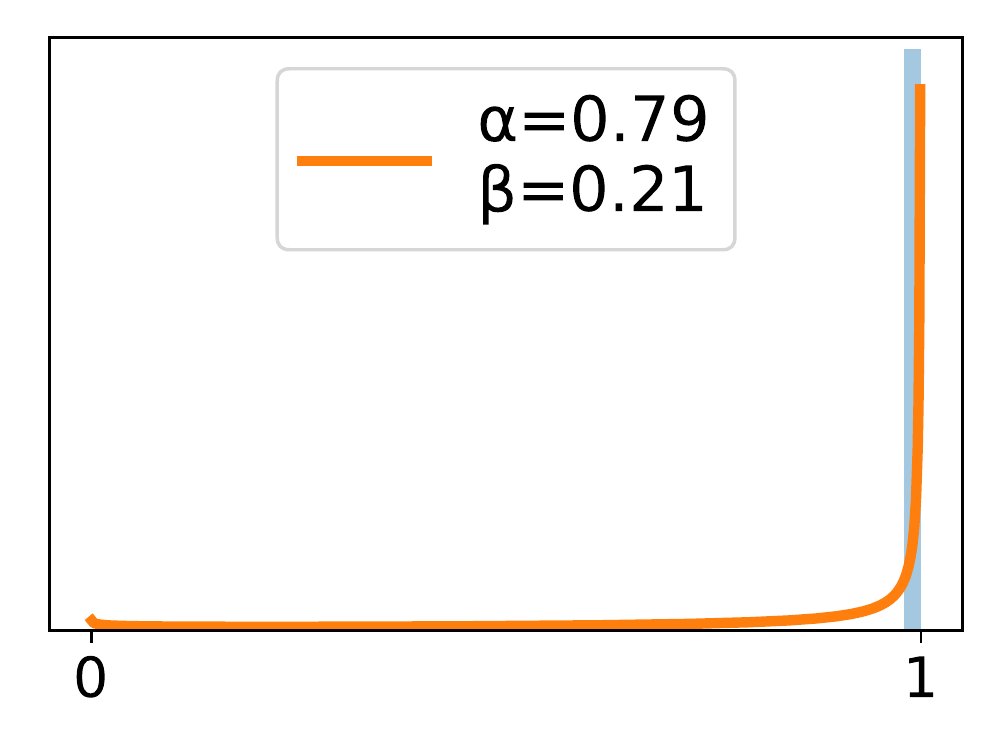}
    \includegraphics[height=2cm]{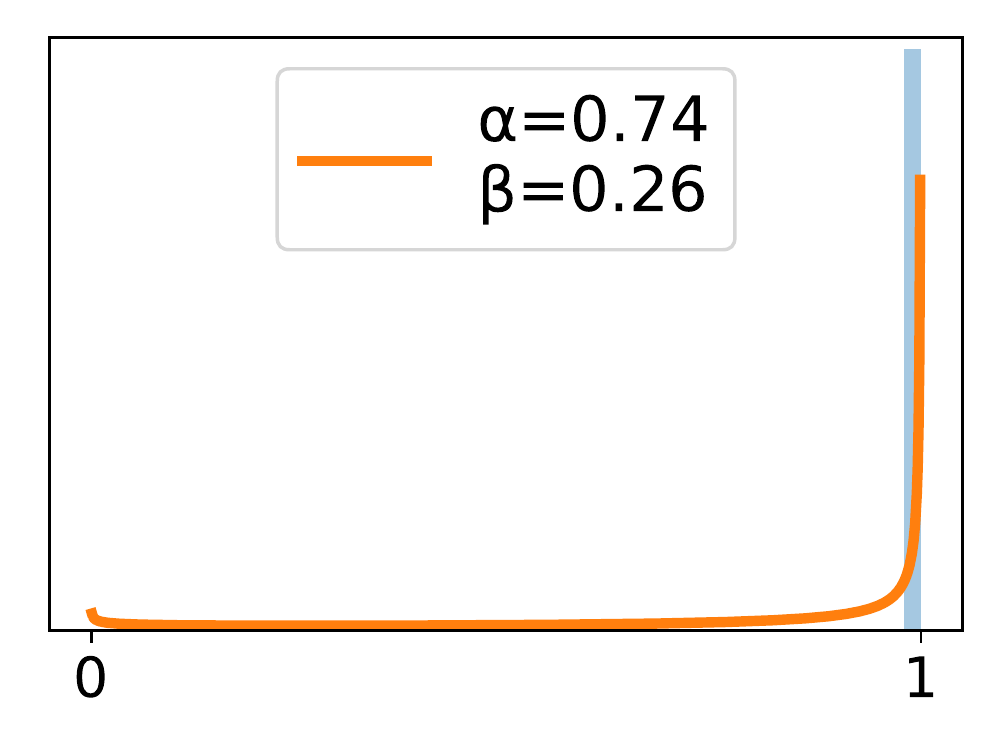}
    \caption{SL3}
\end{subfigure}\hfill
\begin{subfigure}[t]{0.19\textwidth}
    \centering
    \includegraphics[height=2cm]{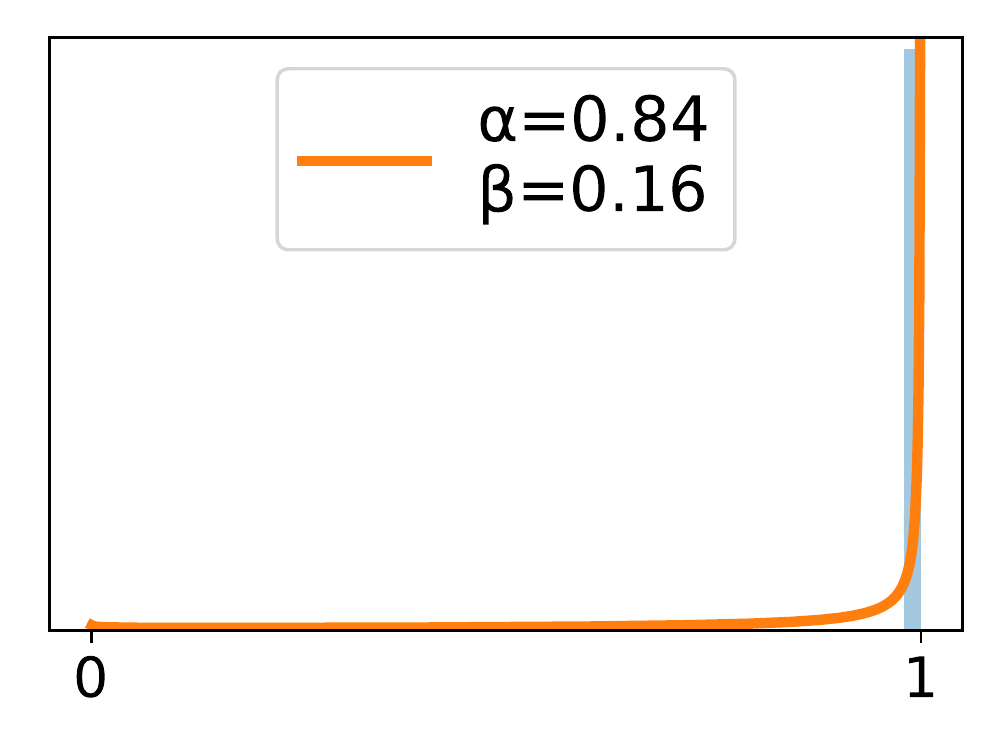}
    \includegraphics[height=2cm]{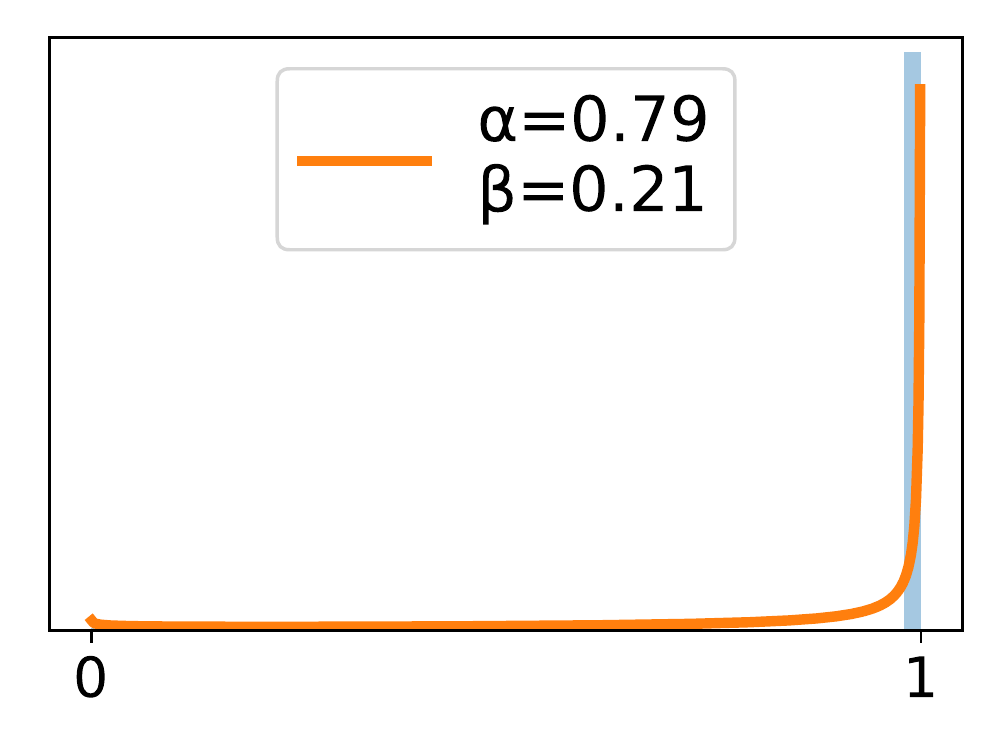}
    \includegraphics[height=2cm]{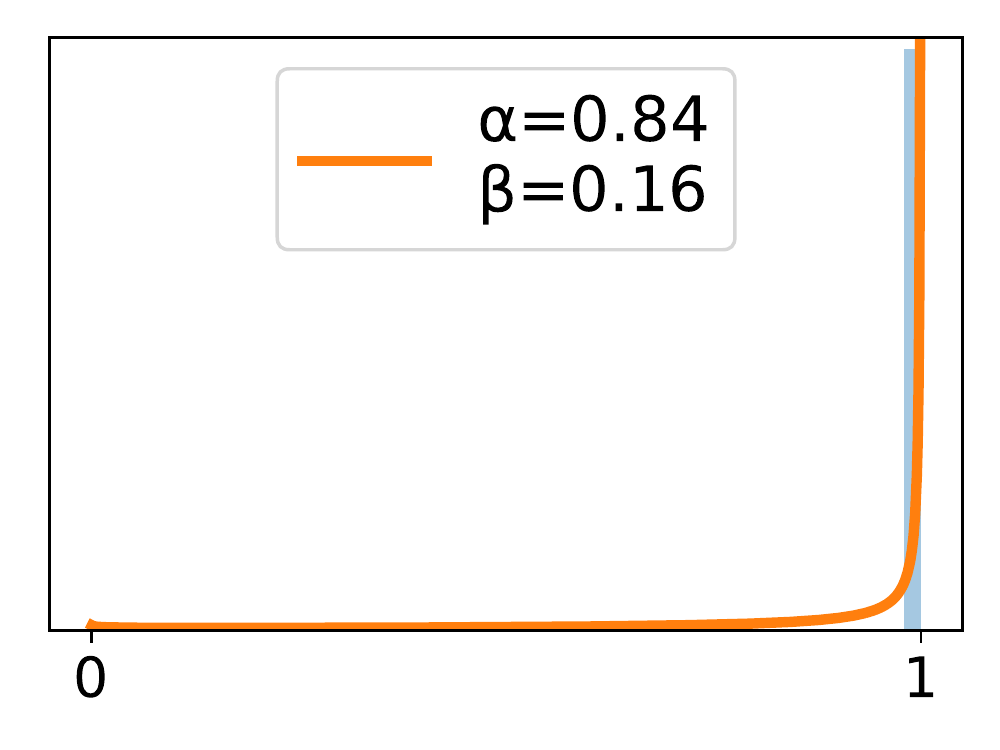}
    \includegraphics[height=2cm]{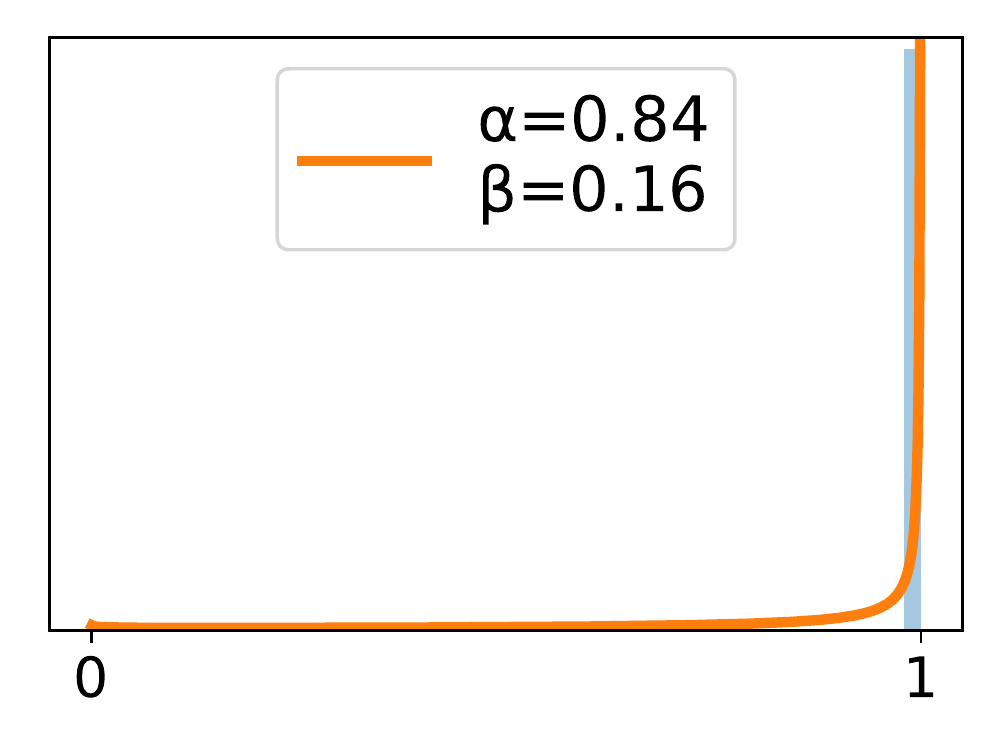}
    \includegraphics[height=2cm]{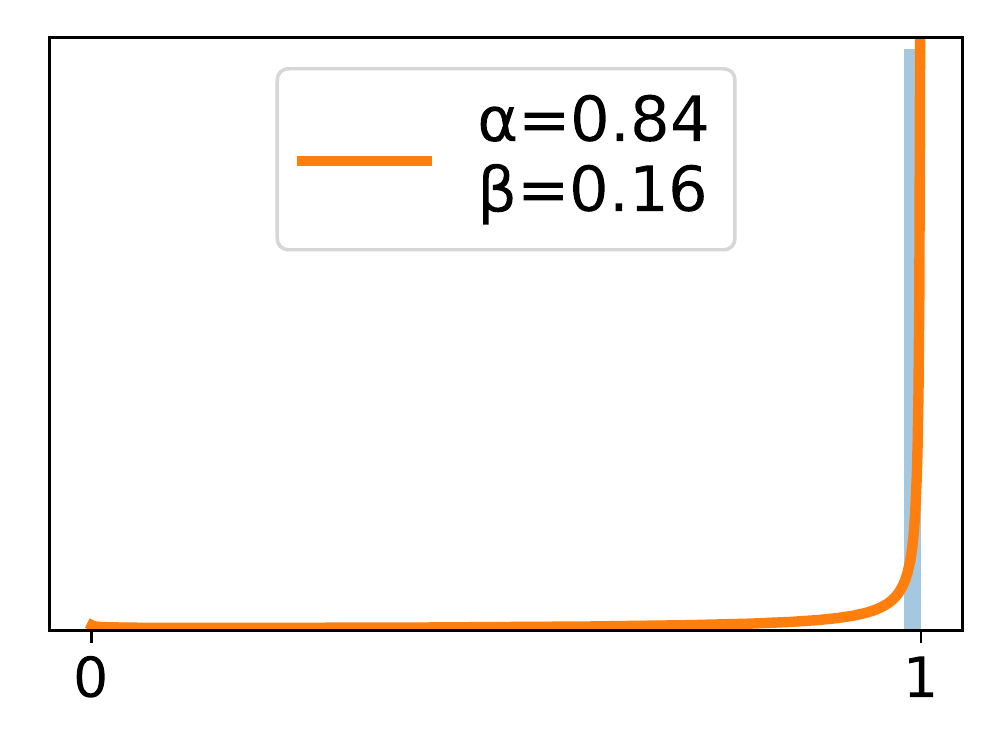}
    \includegraphics[height=2cm]{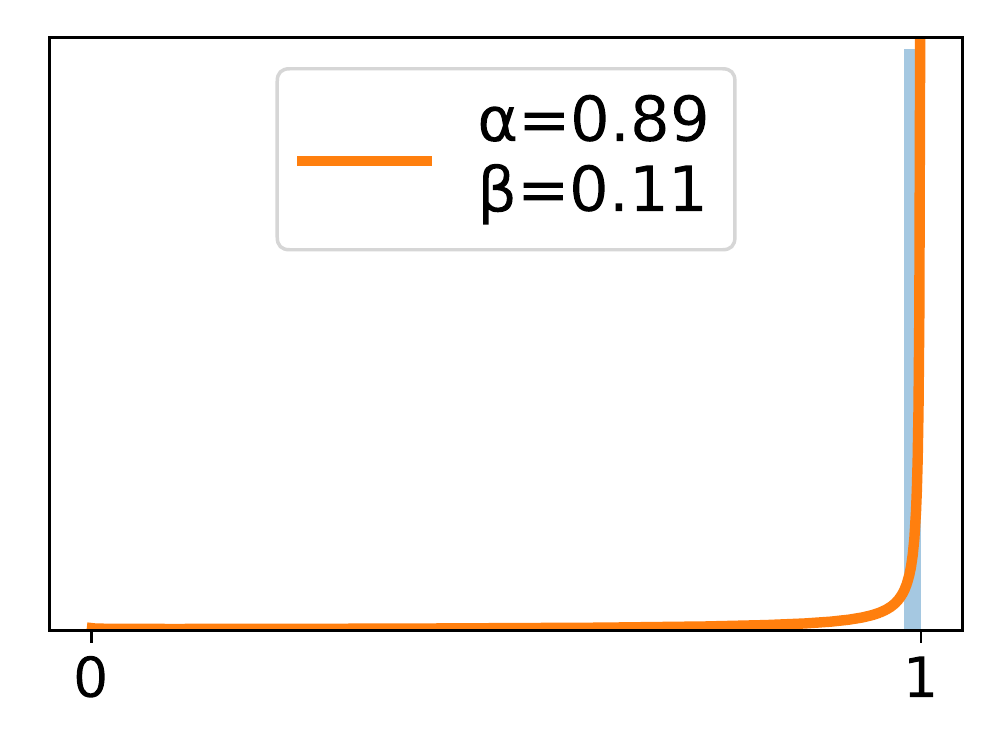}
    \includegraphics[height=2cm]{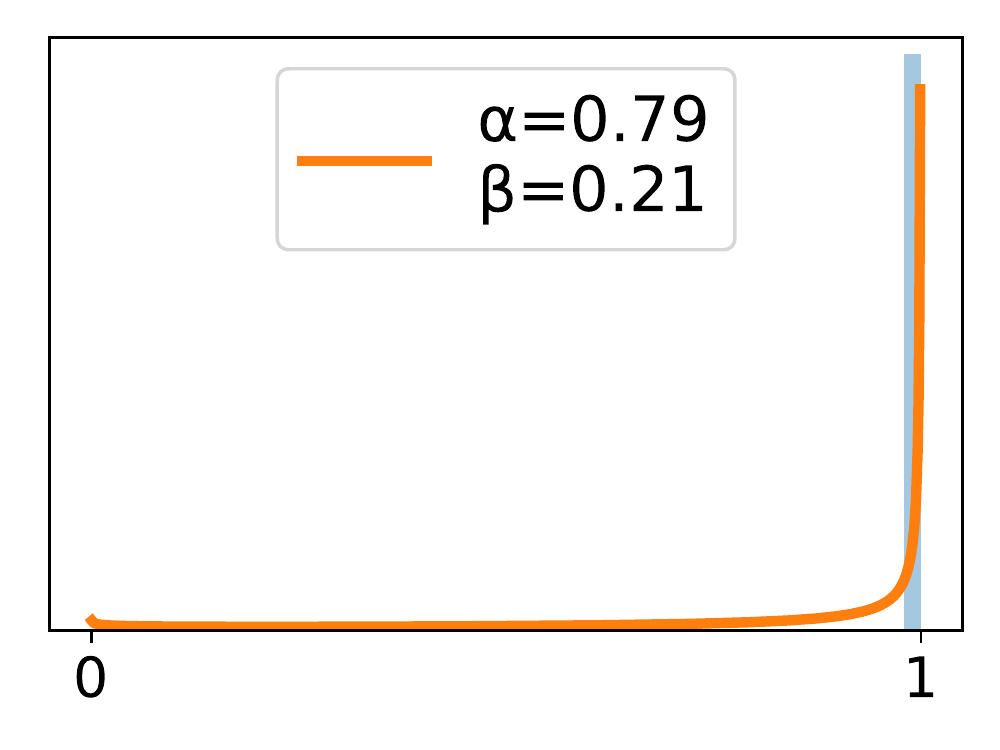}
    \includegraphics[height=2cm]{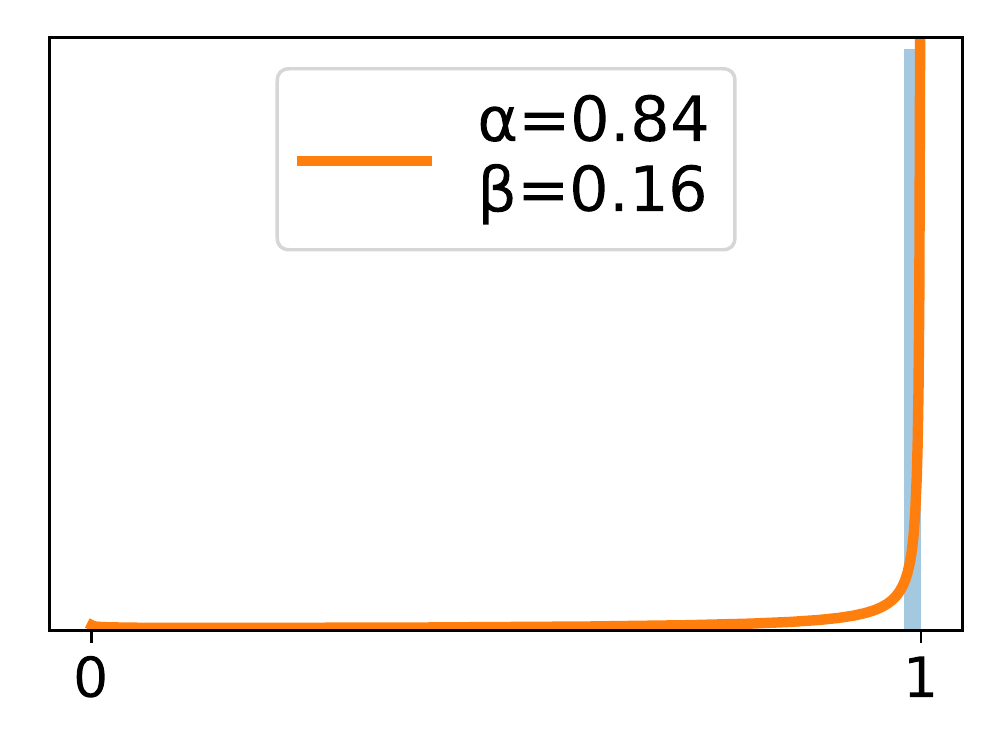}
    \includegraphics[height=2cm]{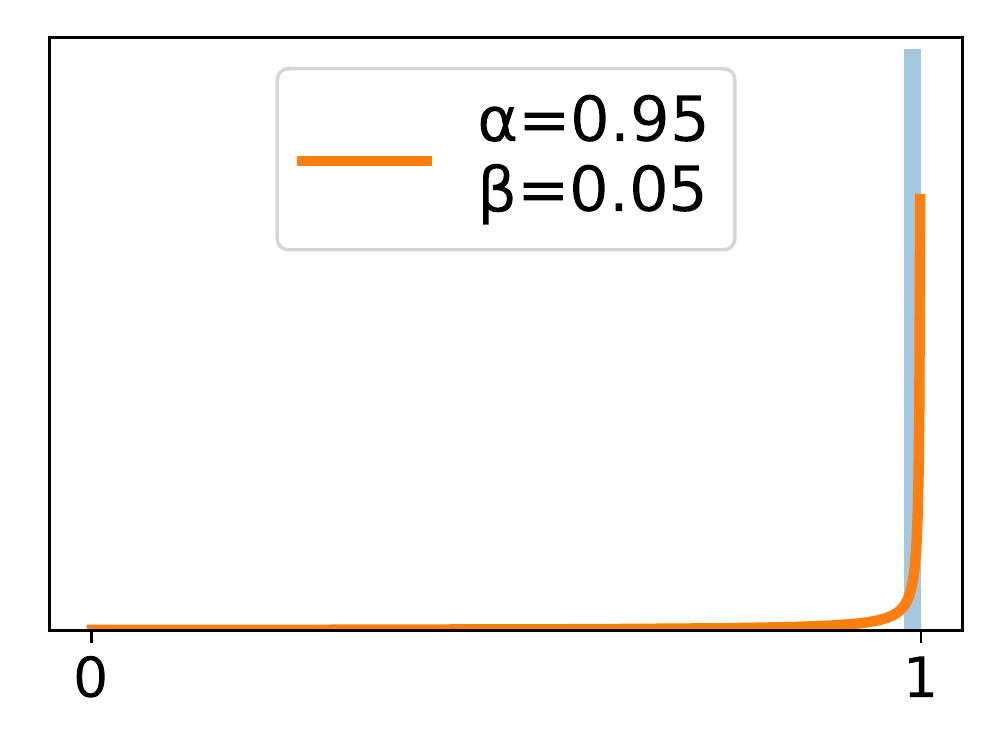}
    \caption{SL4}
\end{subfigure}
\caption{Histograms showing the spread of trained (with $\rho=0.2$) \acs{DT} models' predictions on a selected numbers of data examples from the chiller dataset, and a fitted Beta distribution $\mathcal{B}(\alpha,\beta)$ for each example.}
\label{fig:DT-chiller-distribution}
\end{figure}

\begin{figure}[p]
\centering
\begin{subfigure}[t]{0.19\textwidth}
    \centering
    \includegraphics[height=2cm]{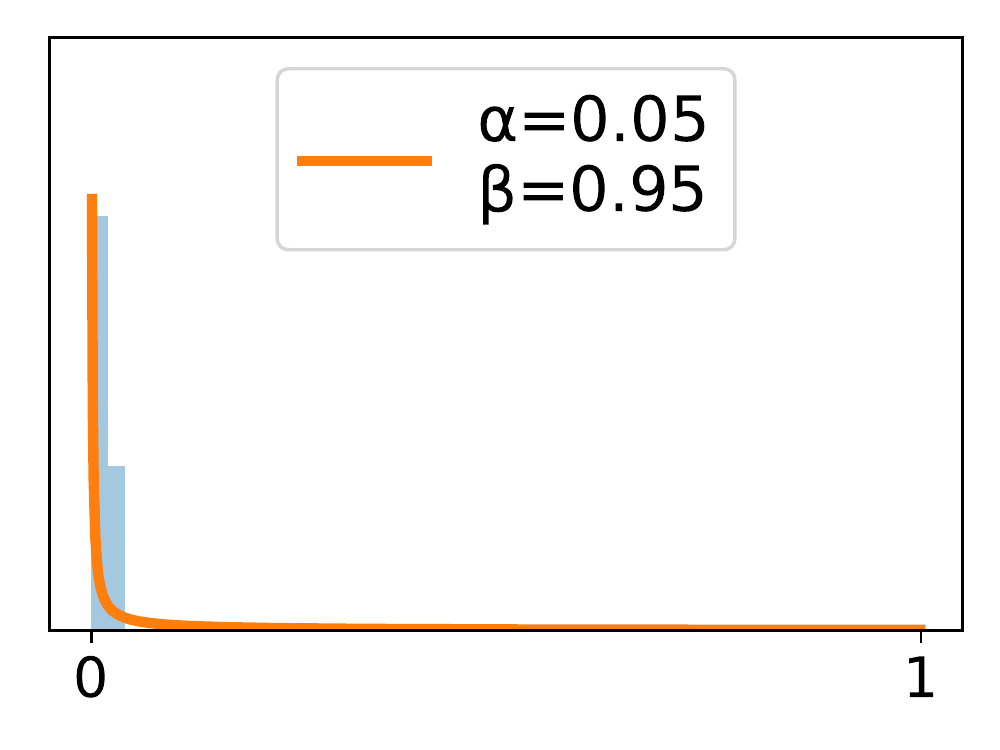}
    \includegraphics[height=2cm]{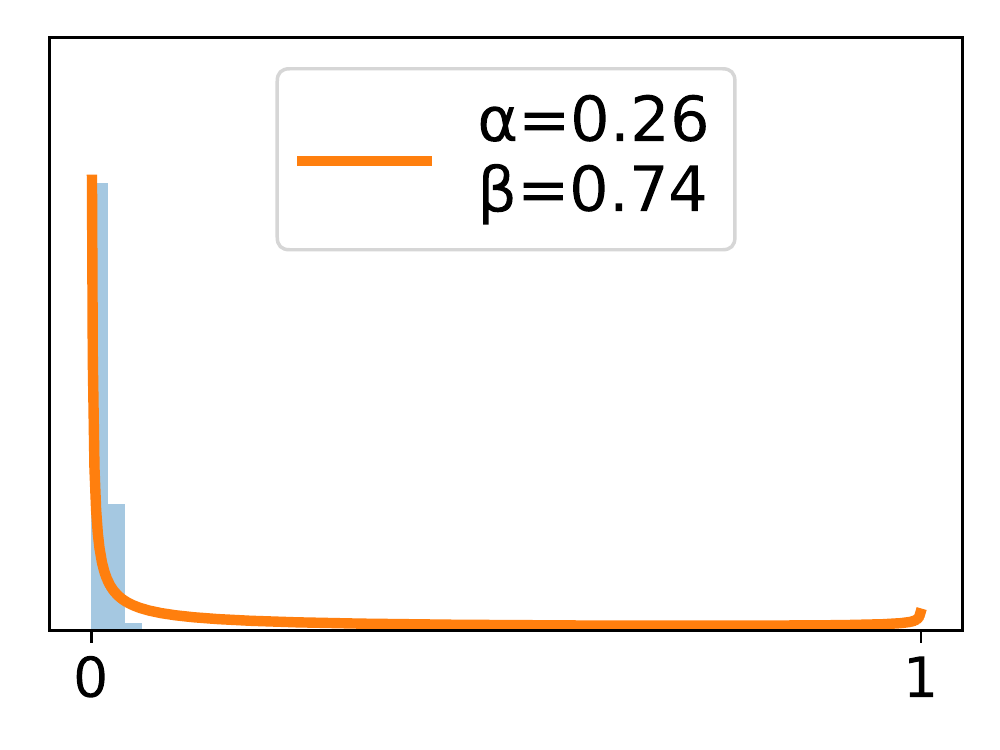}
    \includegraphics[height=2cm]{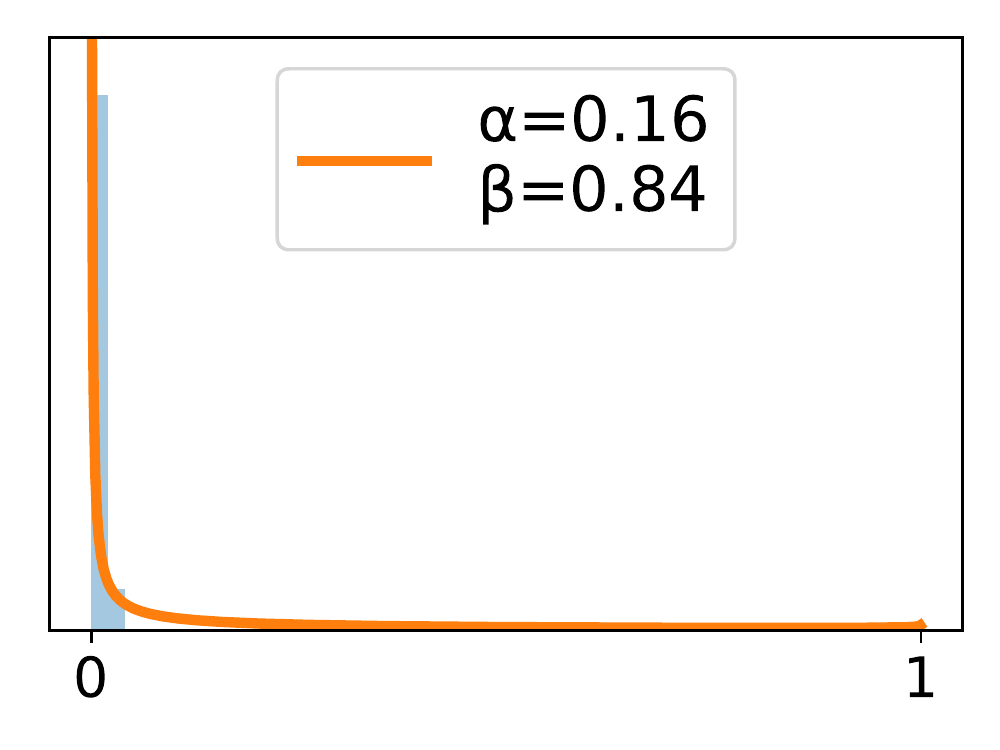}
    \includegraphics[height=2cm]{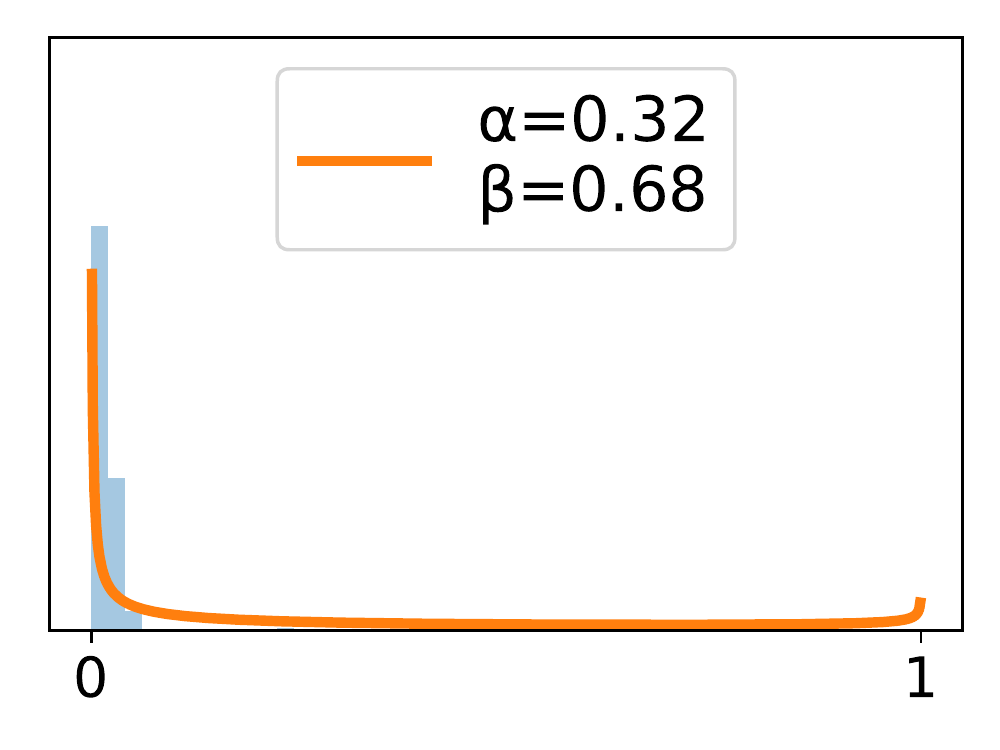}
    \includegraphics[height=2cm]{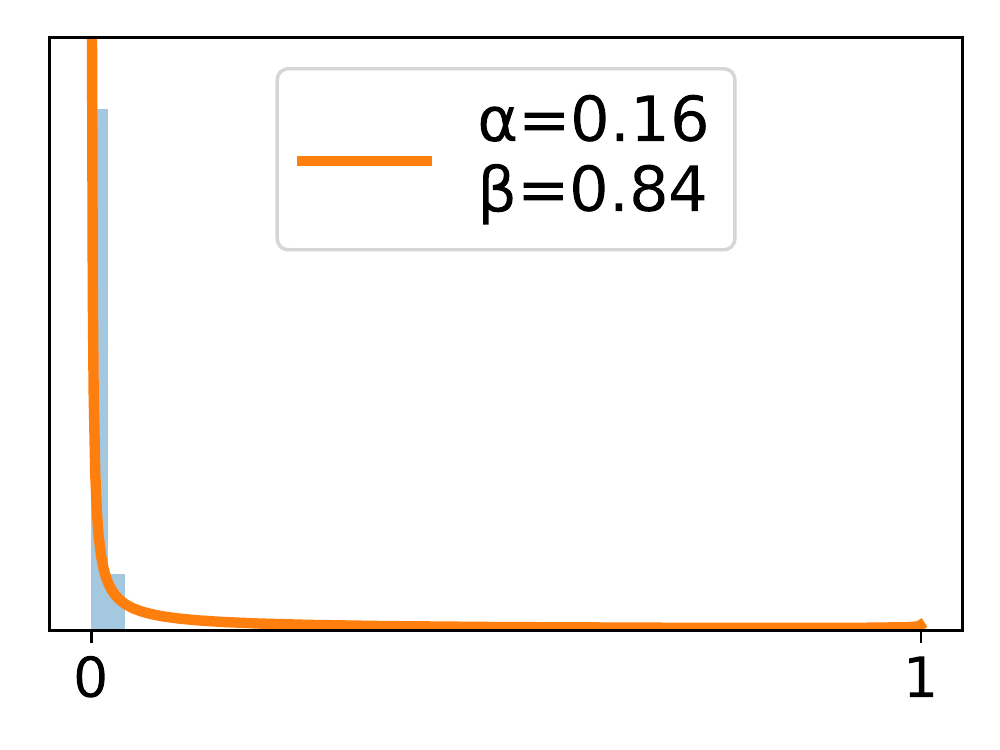}
    \includegraphics[height=2cm]{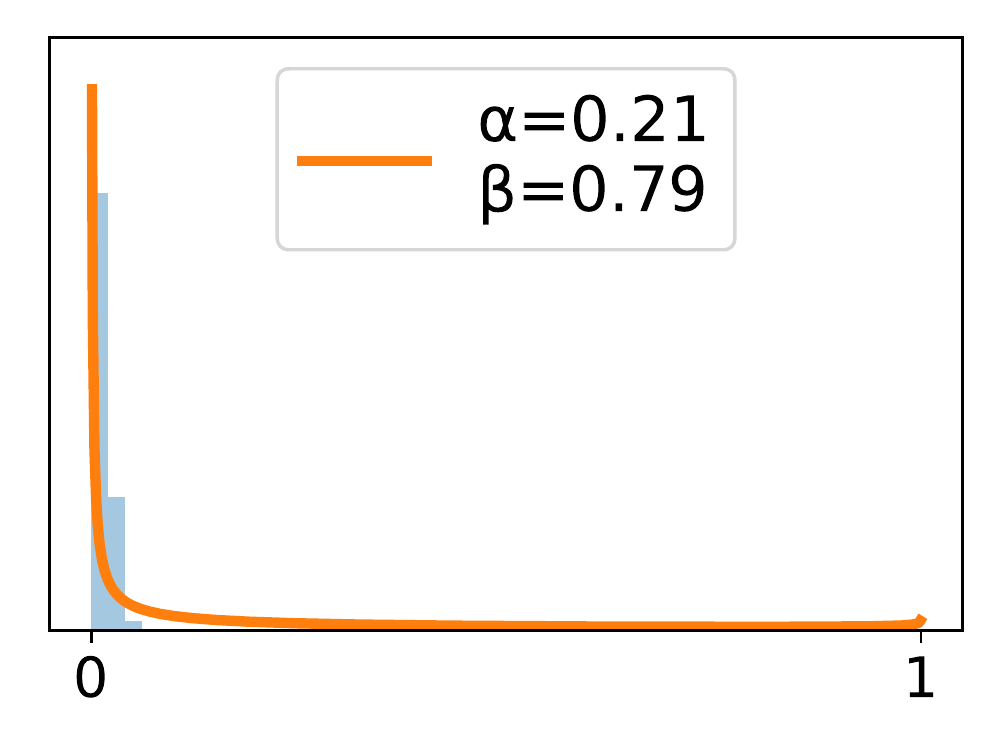}
    \includegraphics[height=2cm]{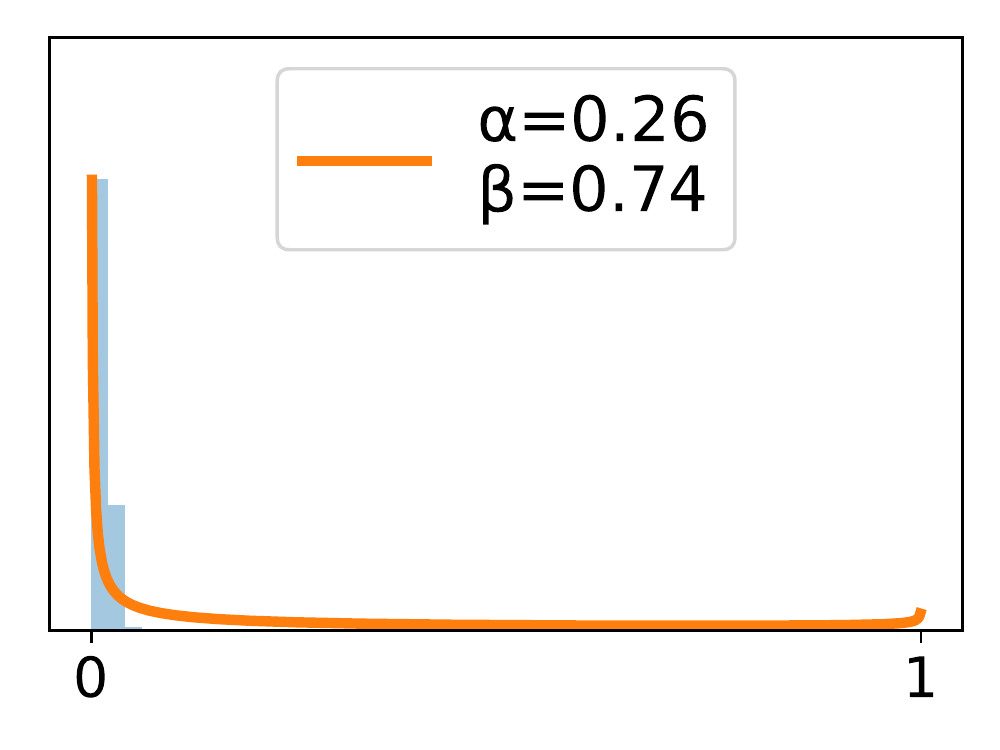}
    \includegraphics[height=2cm]{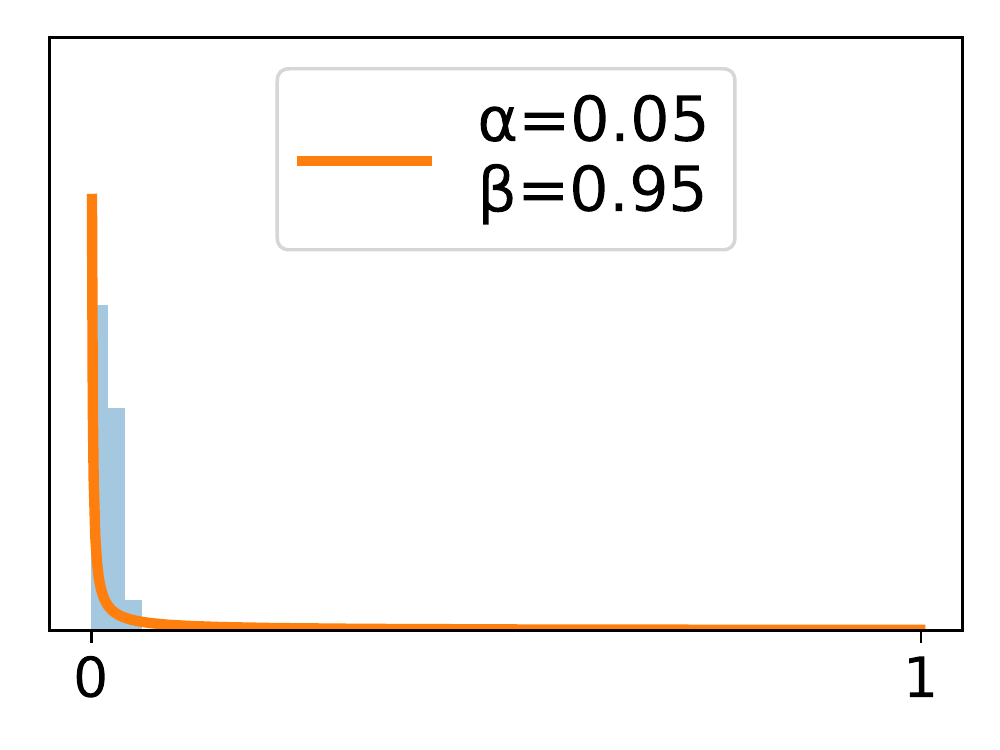}
    \includegraphics[height=2cm]{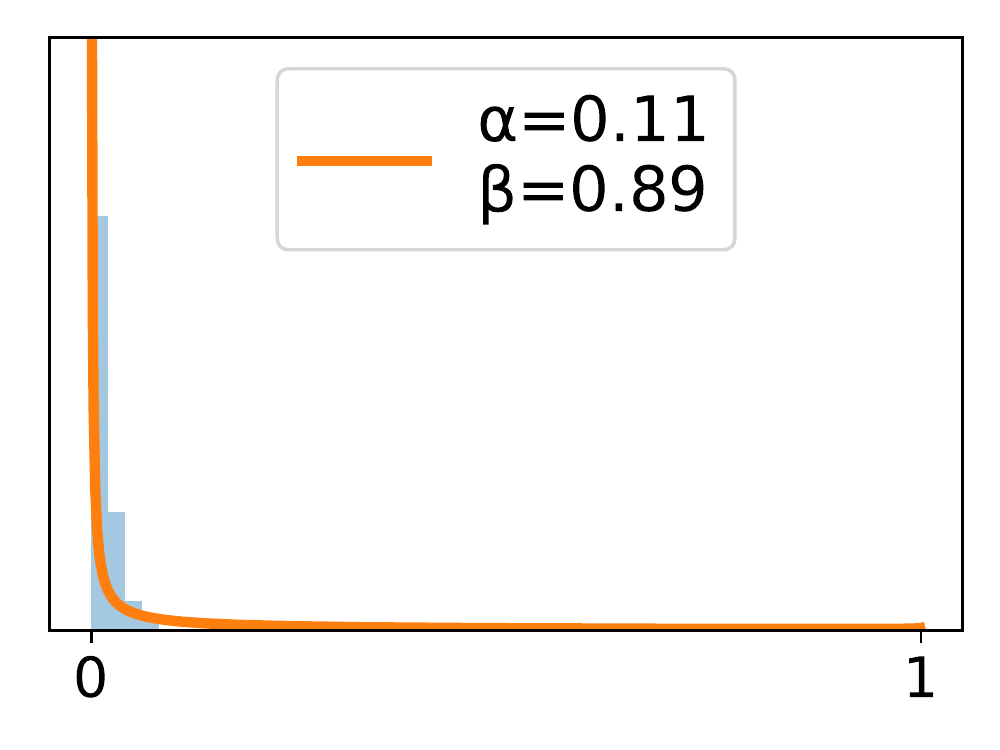}
    \caption{SL0}
\end{subfigure}\hfill
\begin{subfigure}[t]{0.19\textwidth}
    \centering
    \includegraphics[height=2cm]{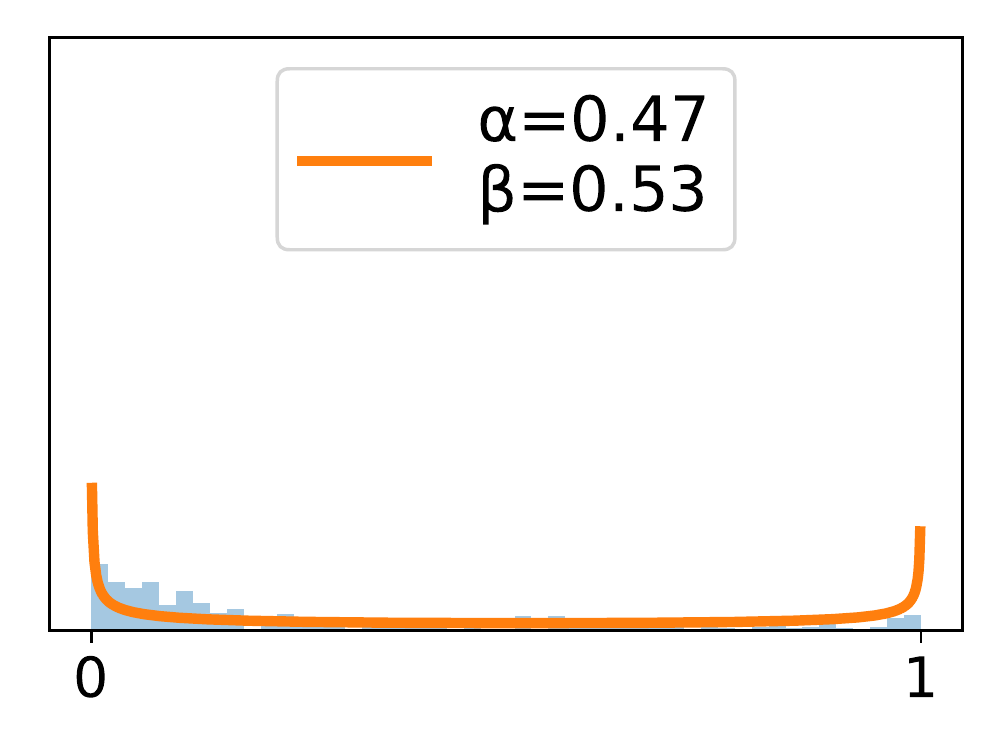}
    \includegraphics[height=2cm]{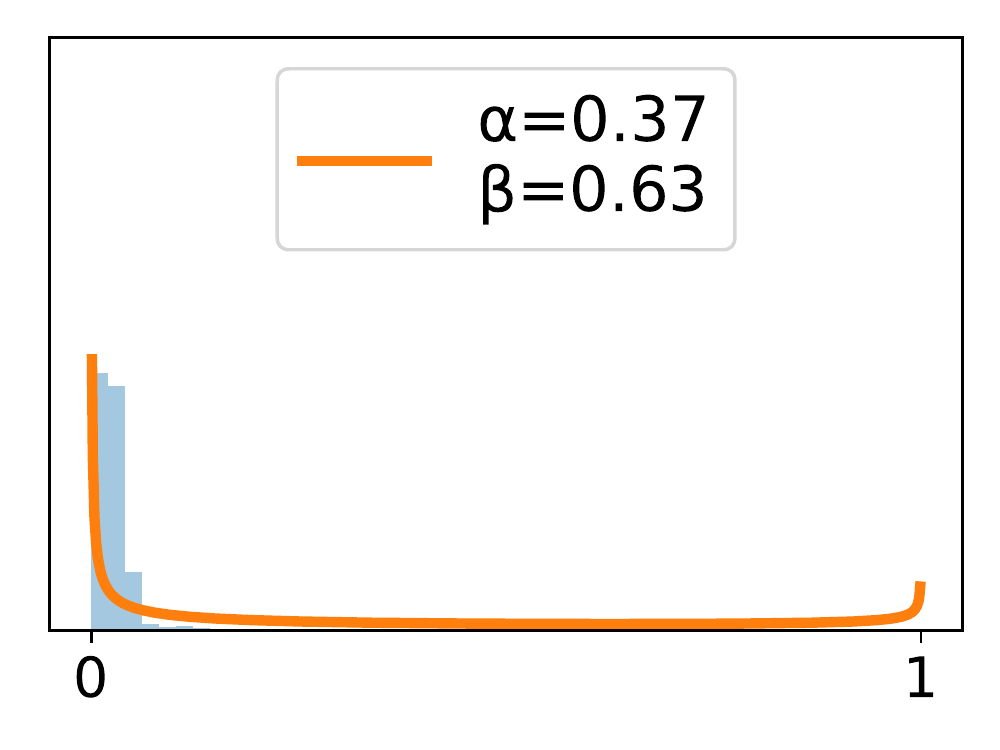}
    \includegraphics[height=2cm]{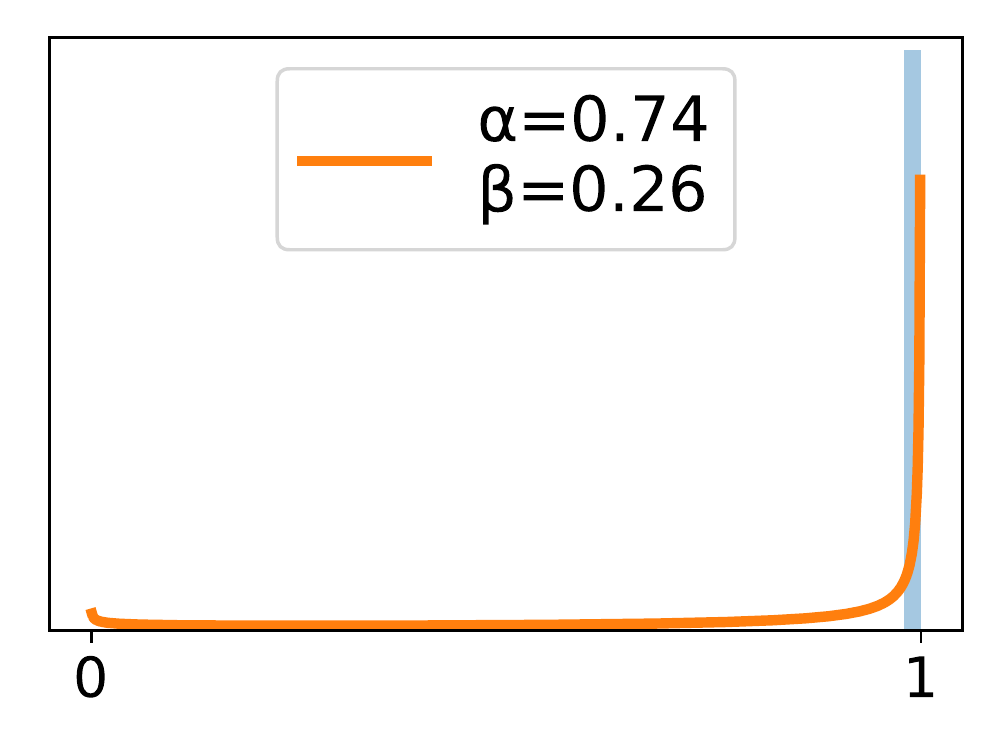}
    \includegraphics[height=2cm]{img/beta_detail/chiller_DT/mr10_sl1_idx_66654_a_79_b_21.pdf}
    \includegraphics[height=2cm]{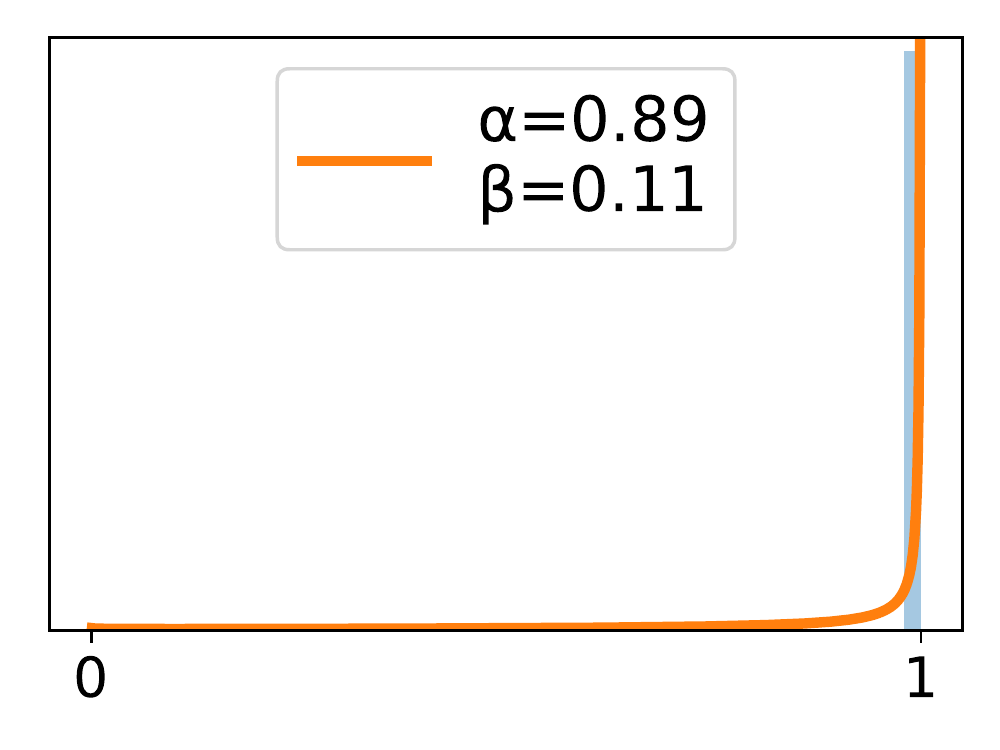}
    \includegraphics[height=2cm]{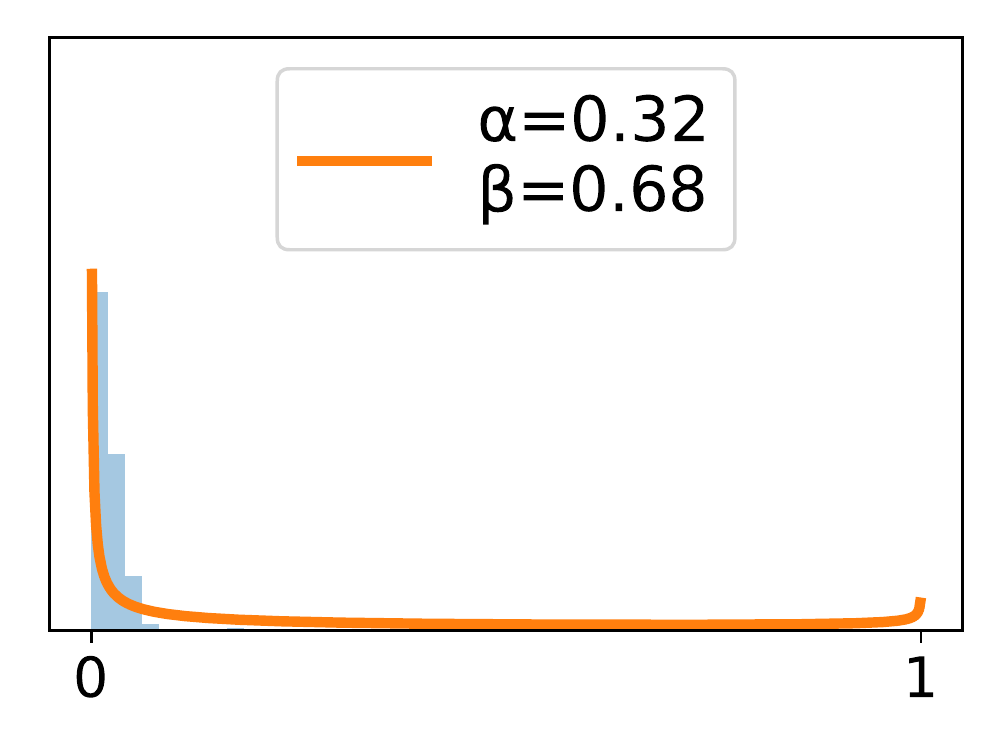}
    \includegraphics[height=2cm]{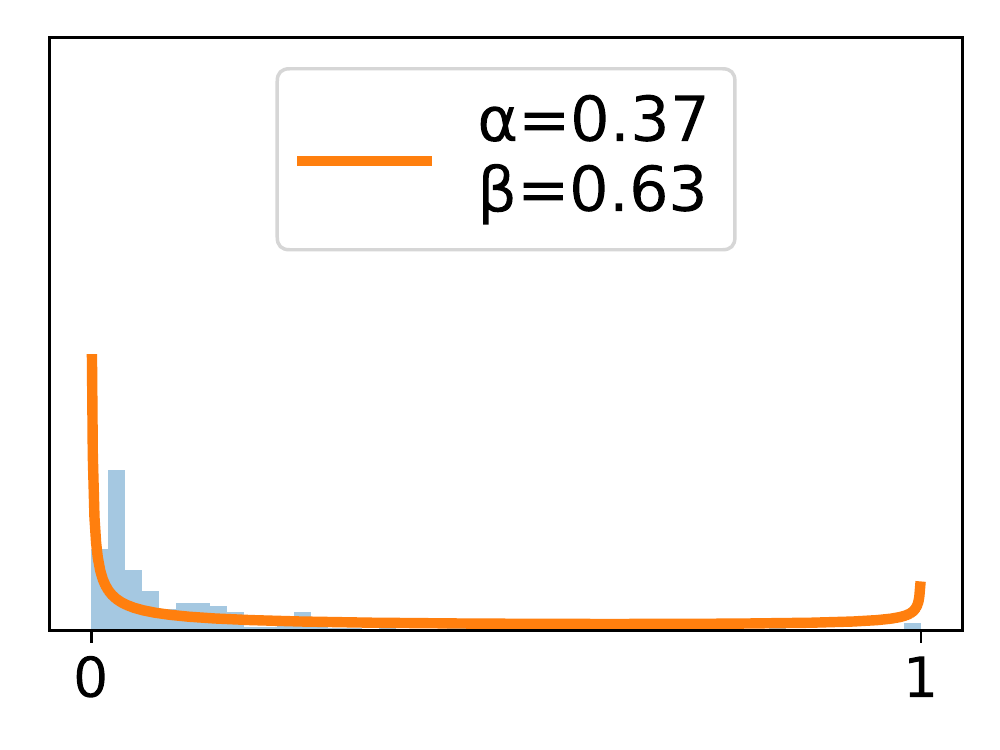}
    \includegraphics[height=2cm]{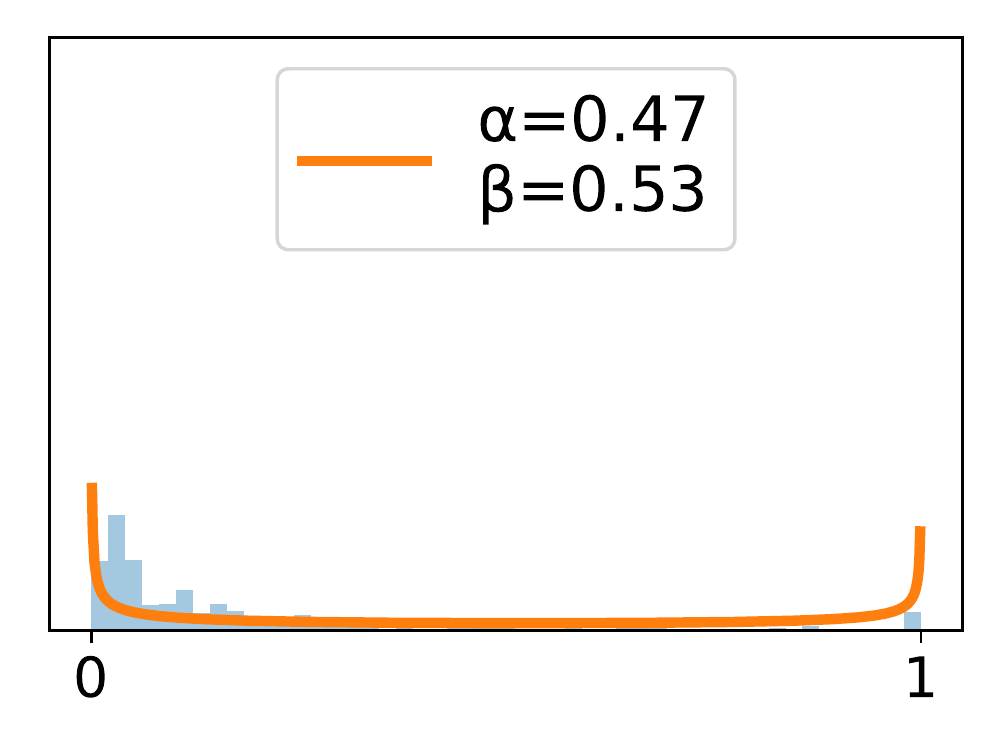}
    \includegraphics[height=2cm]{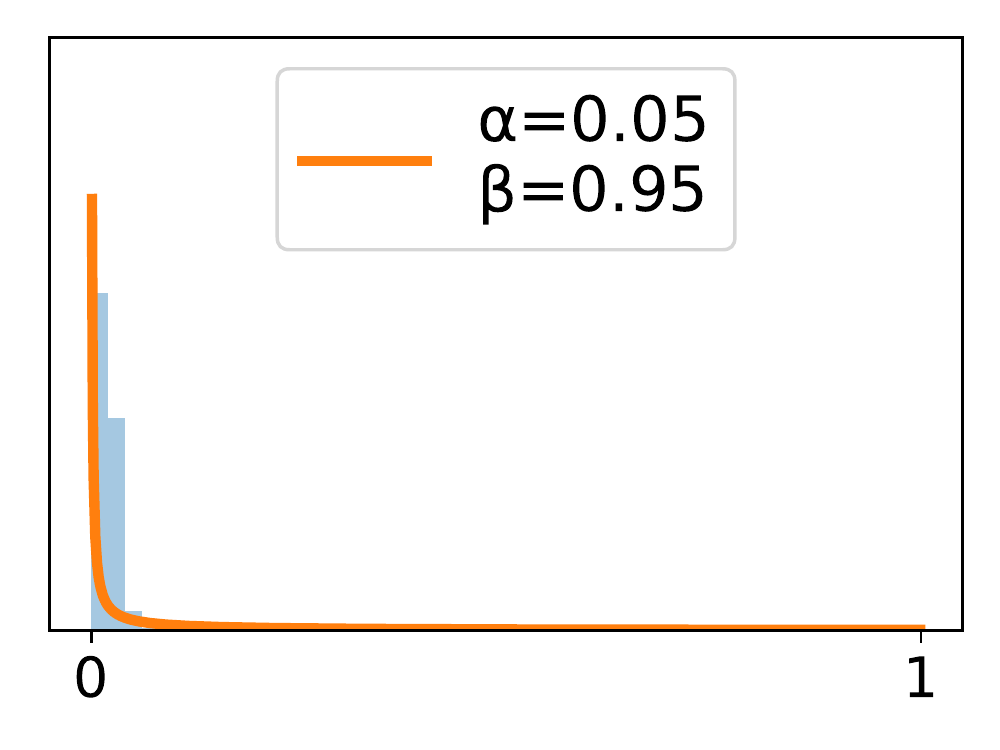}
    \caption{SL1}
\end{subfigure}\hfill
\begin{subfigure}[t]{0.19\textwidth}
    \centering
    \includegraphics[height=2cm]{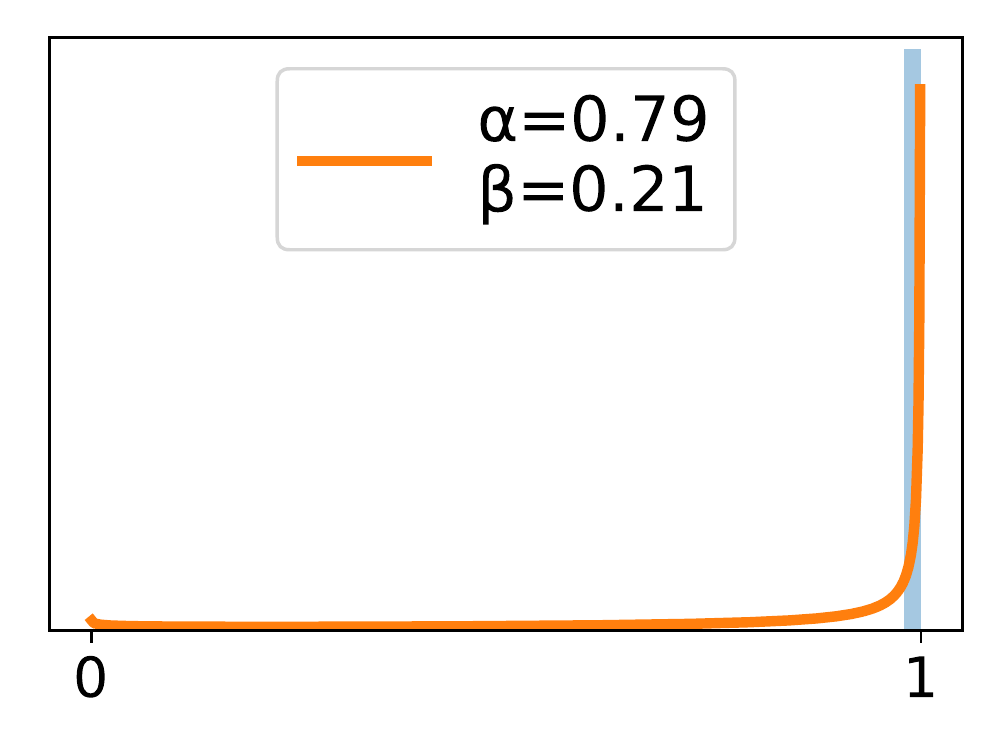}
    \includegraphics[height=2cm]{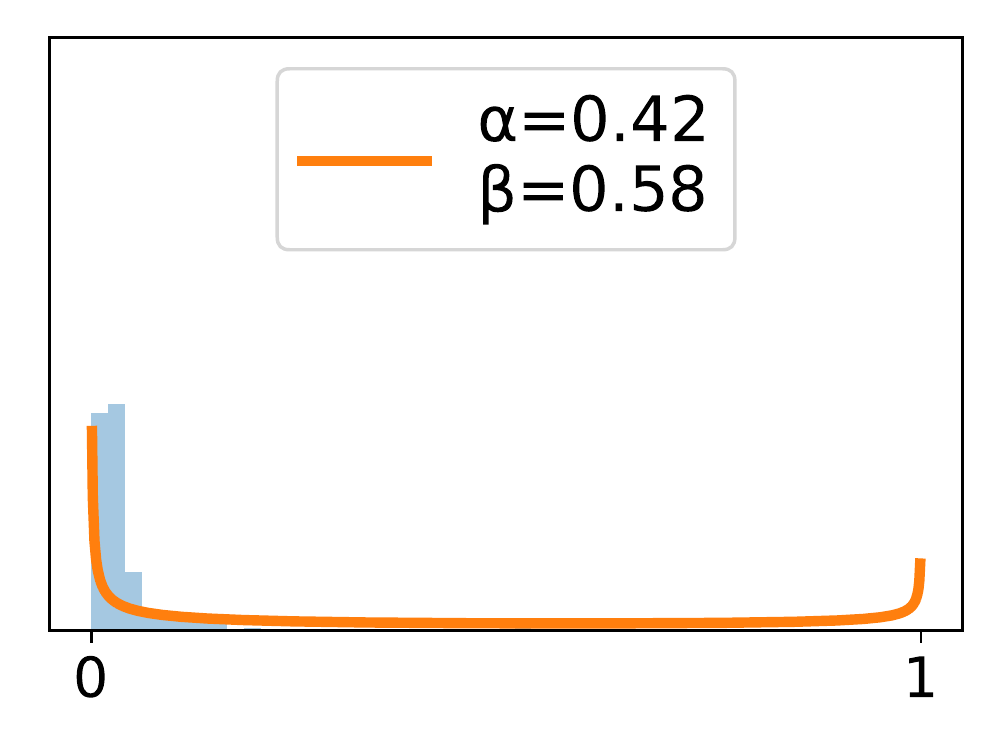}
    \includegraphics[height=2cm]{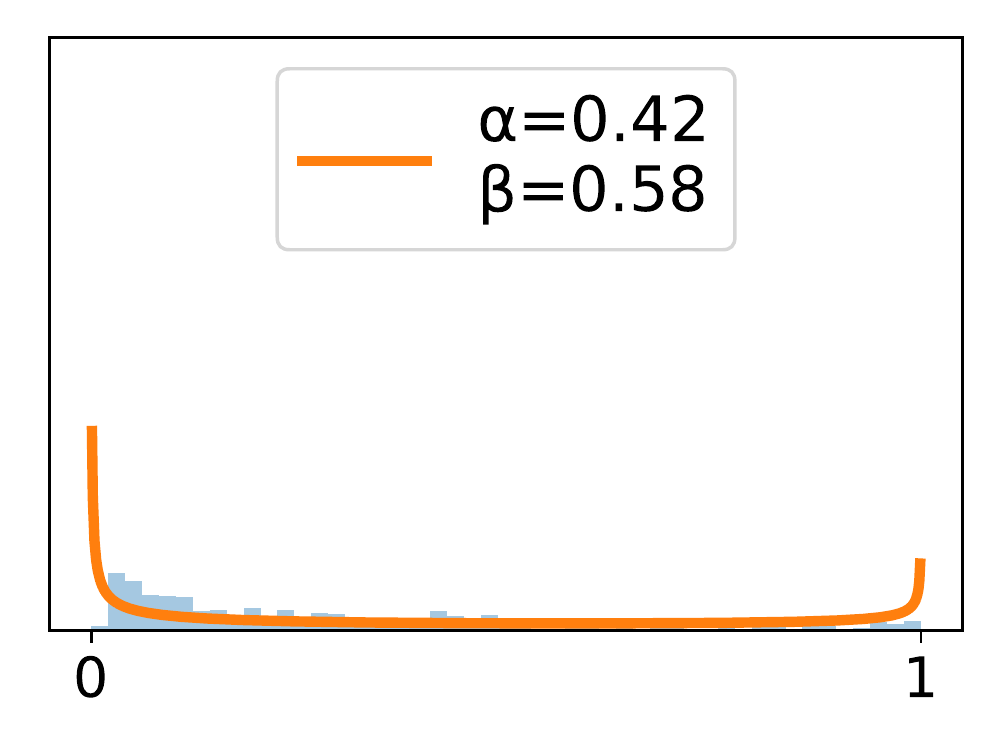}
    \includegraphics[height=2cm]{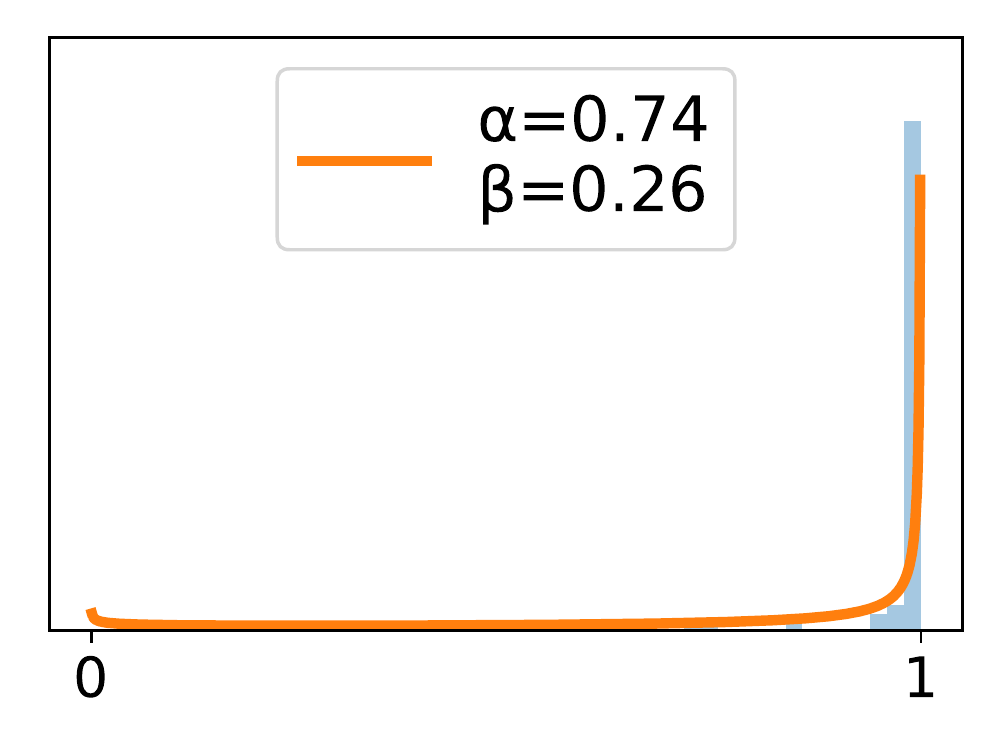}
    \includegraphics[height=2cm]{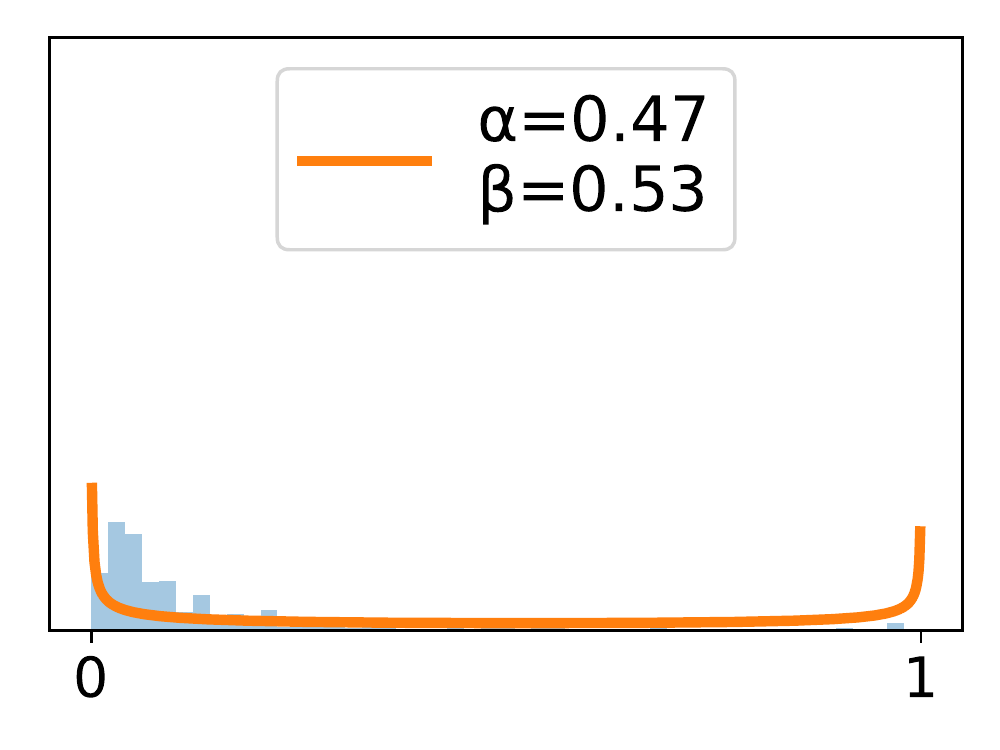}
    \includegraphics[height=2cm]{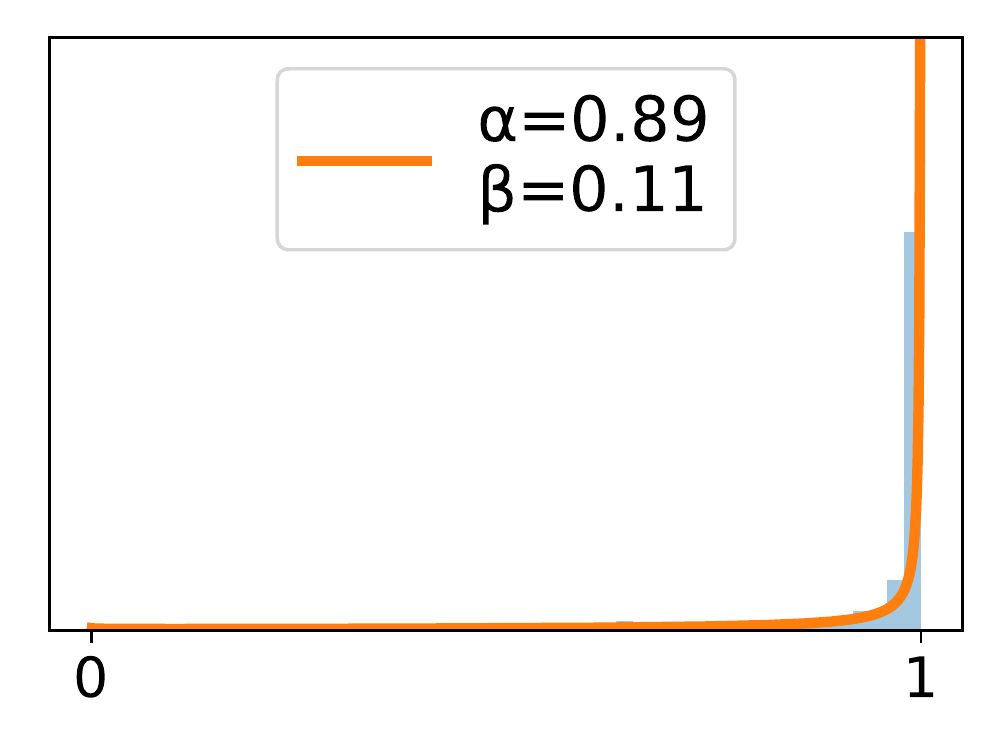}
    \includegraphics[height=2cm]{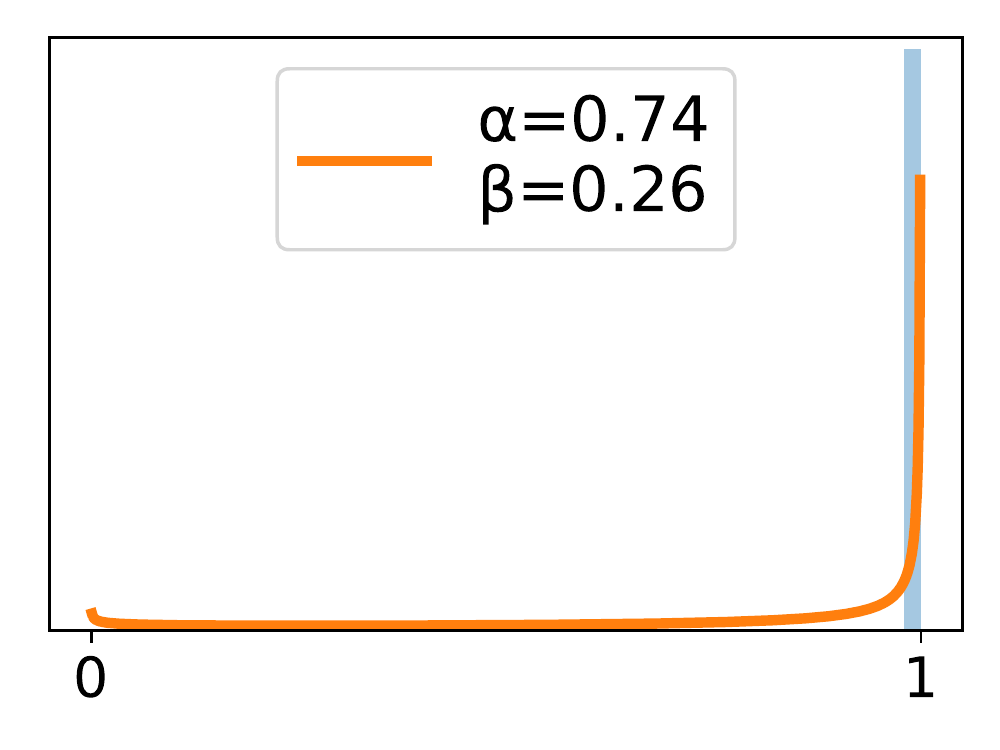}
    \includegraphics[height=2cm]{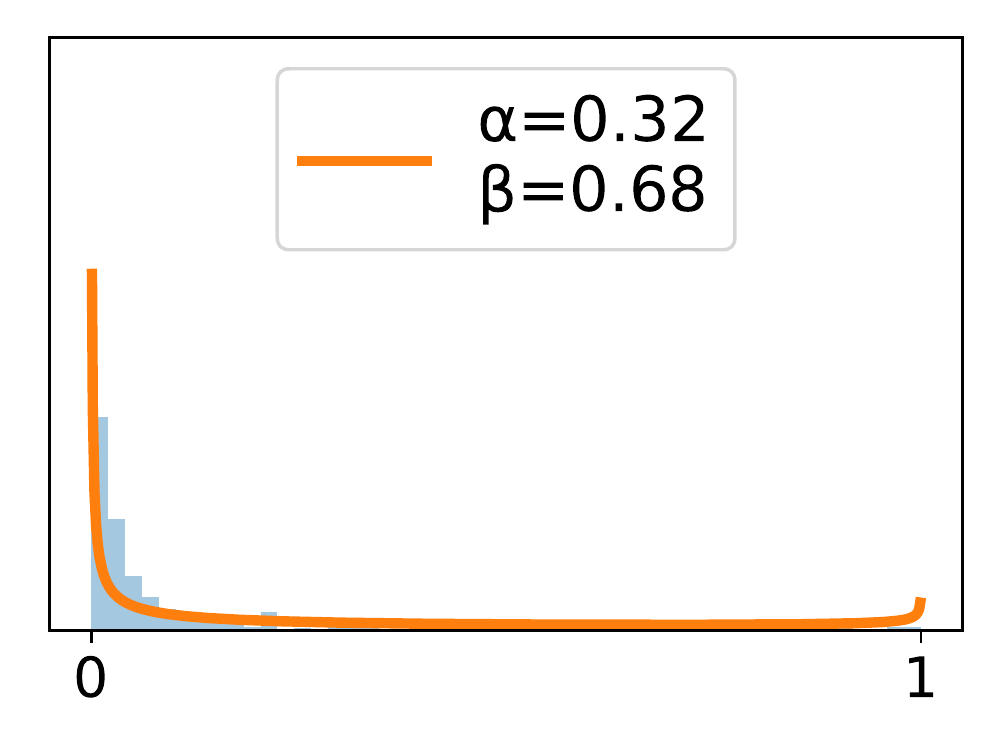}
    \includegraphics[height=2cm]{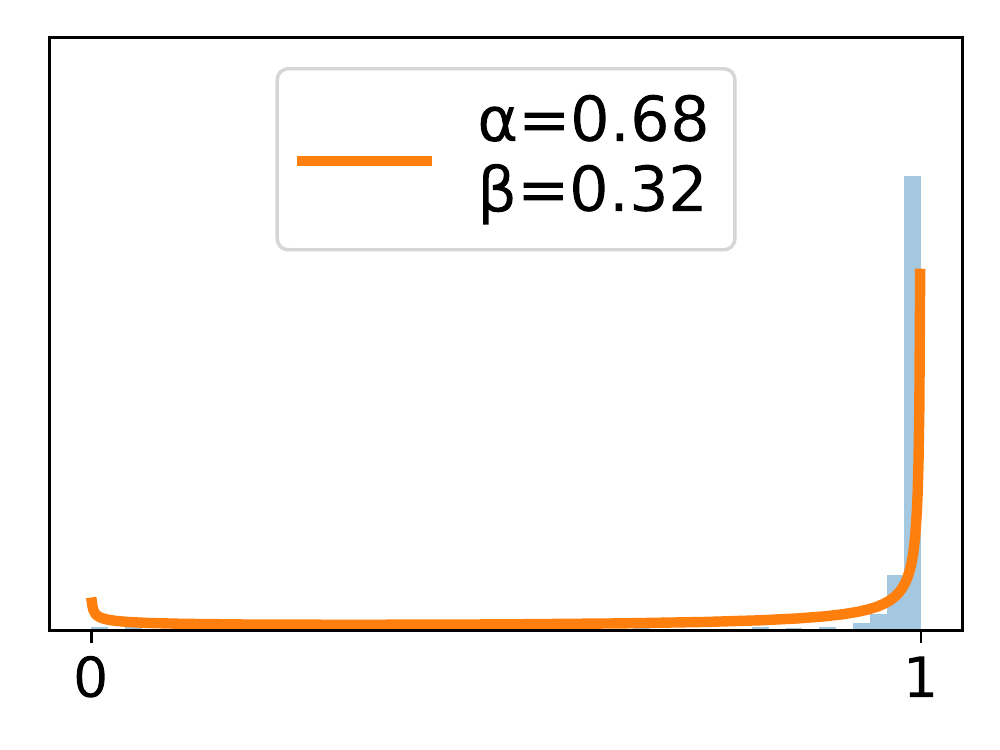}
    \caption{SL2}
\end{subfigure}\hfill
\begin{subfigure}[t]{0.19\textwidth}
    \centering
    \includegraphics[height=2cm]{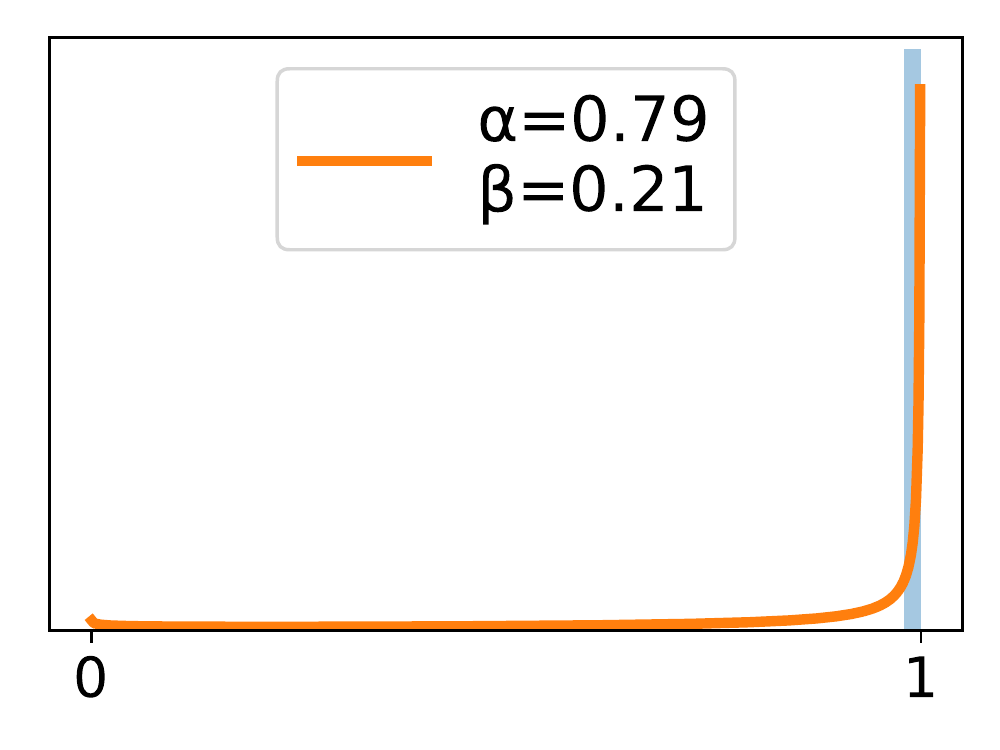}
    \includegraphics[height=2cm]{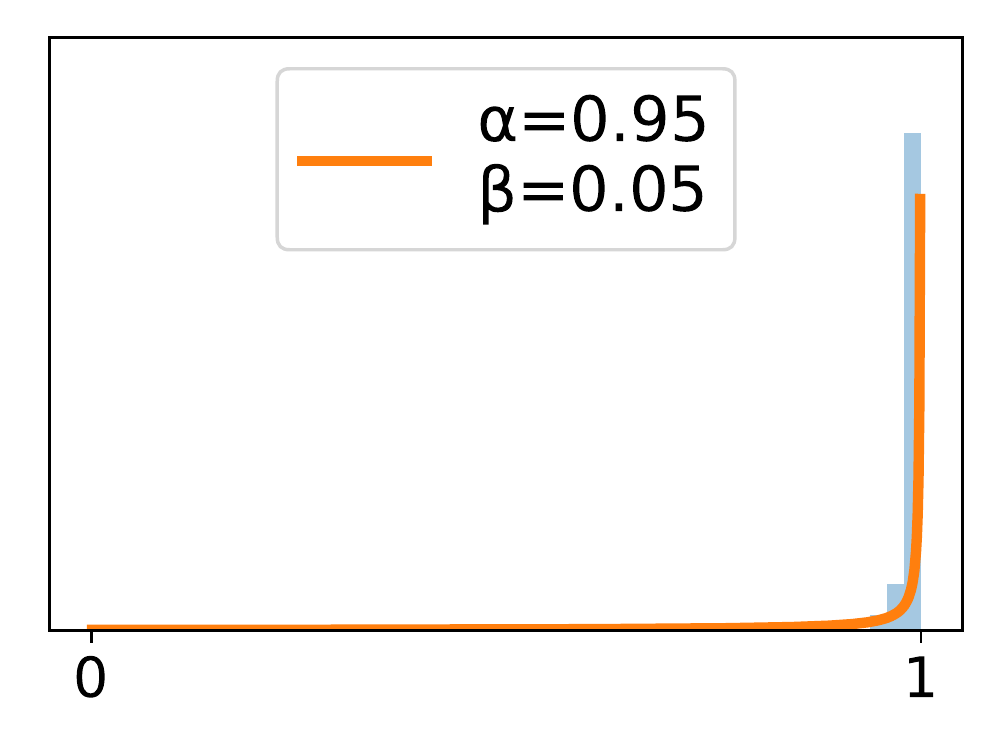}
    \includegraphics[height=2cm]{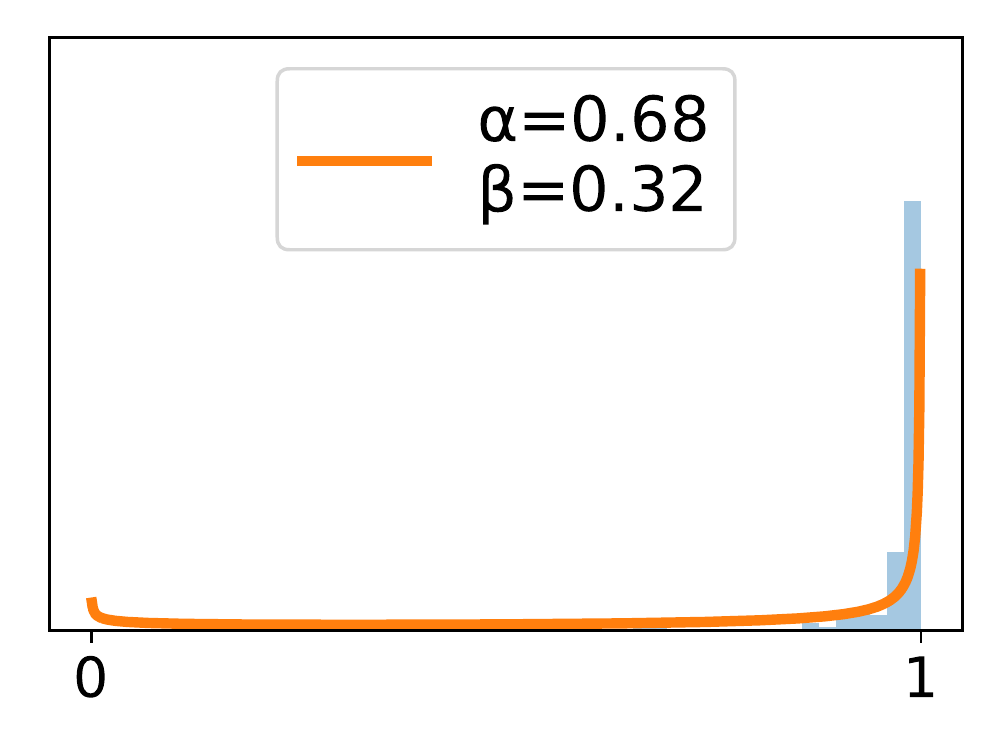}
    \includegraphics[height=2cm]{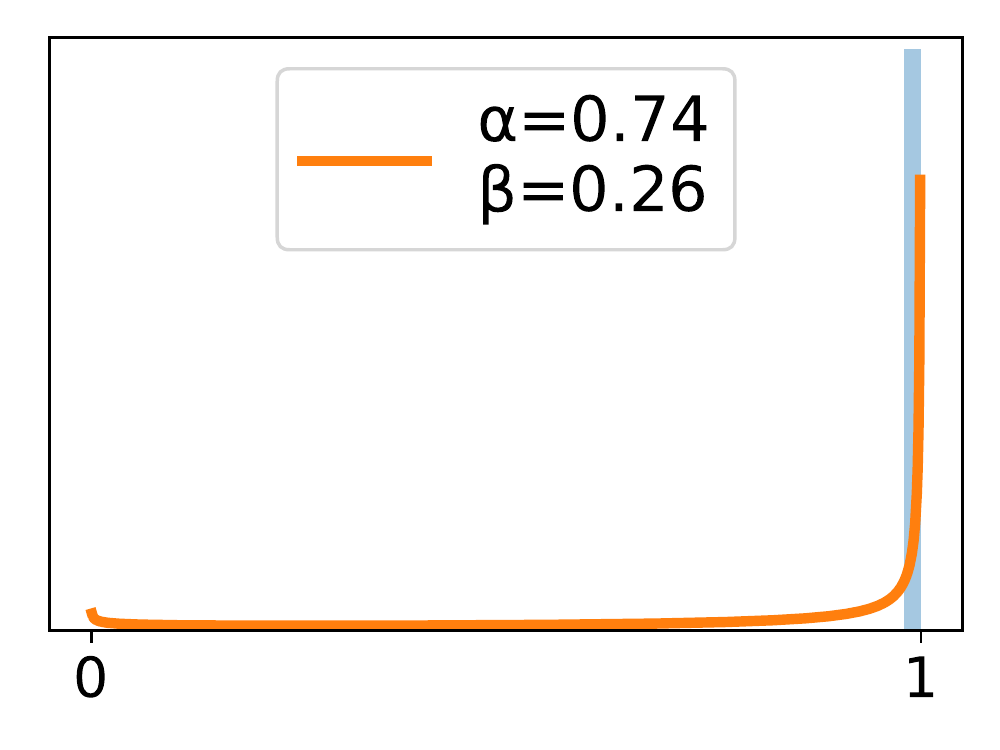}
    \includegraphics[height=2cm]{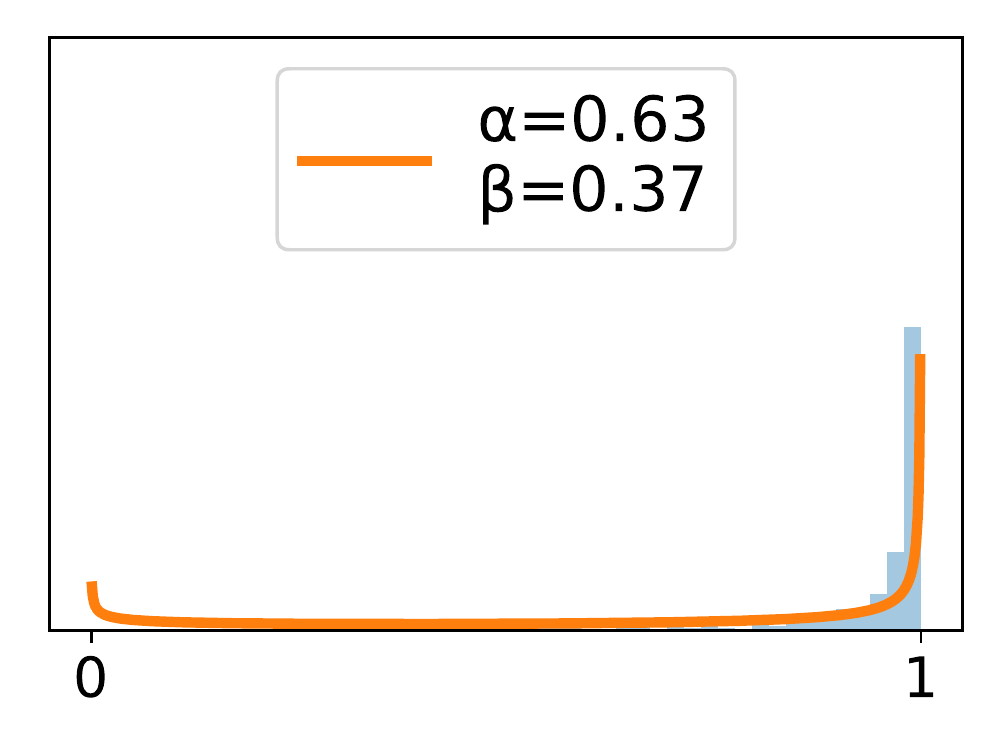}
    \includegraphics[height=2cm]{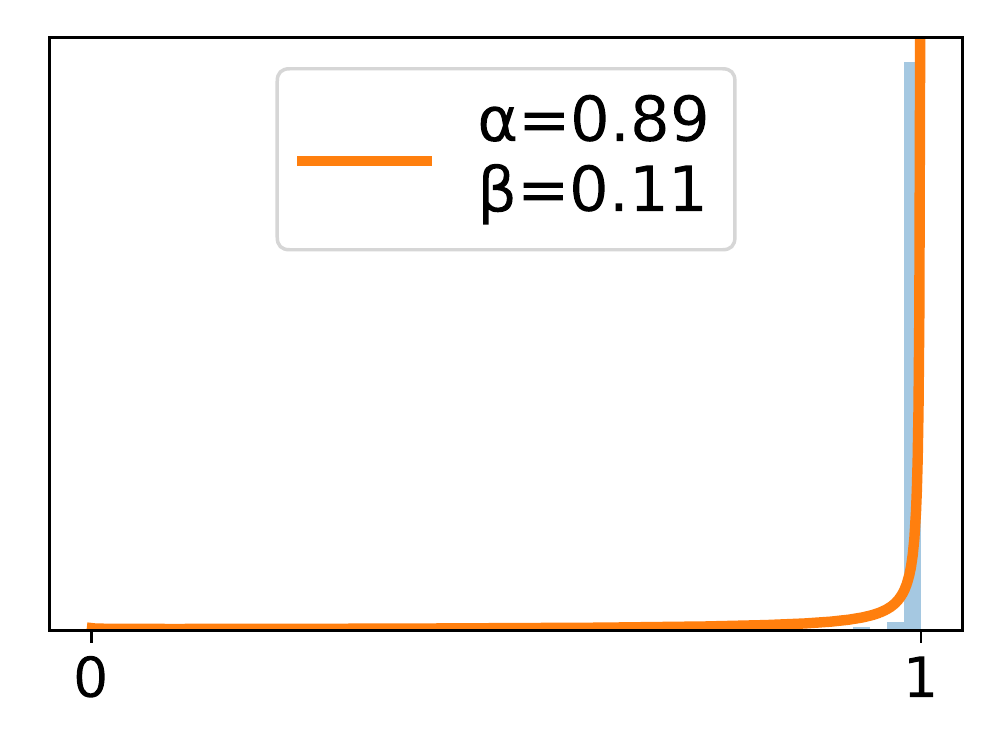}
    \includegraphics[height=2cm]{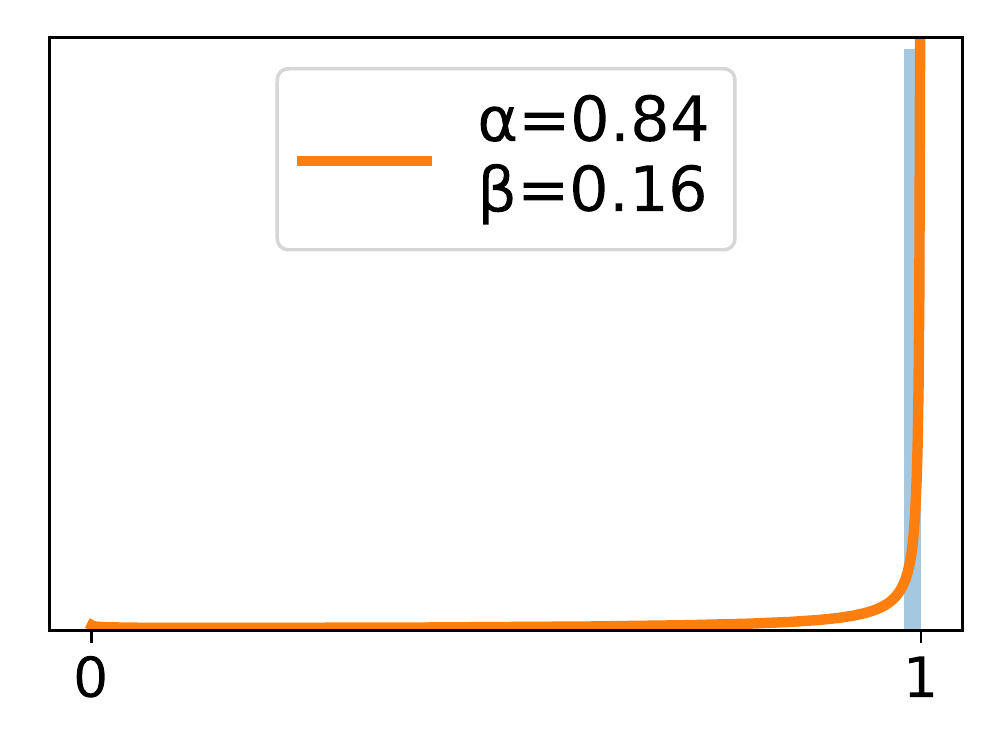}
    \includegraphics[height=2cm]{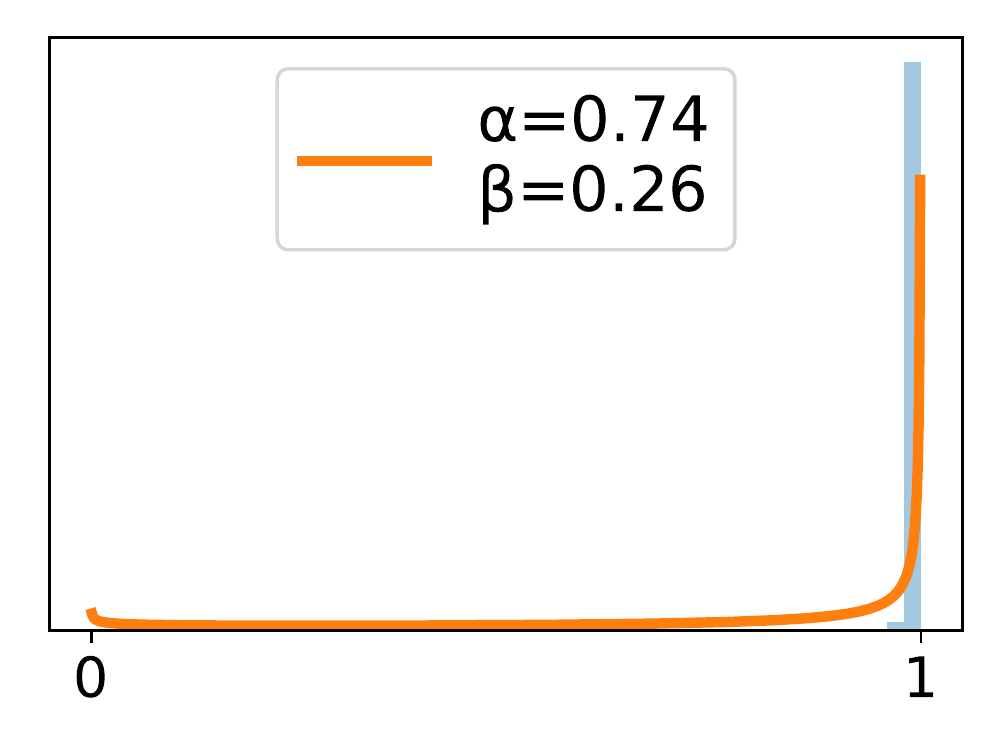}
    \includegraphics[height=2cm]{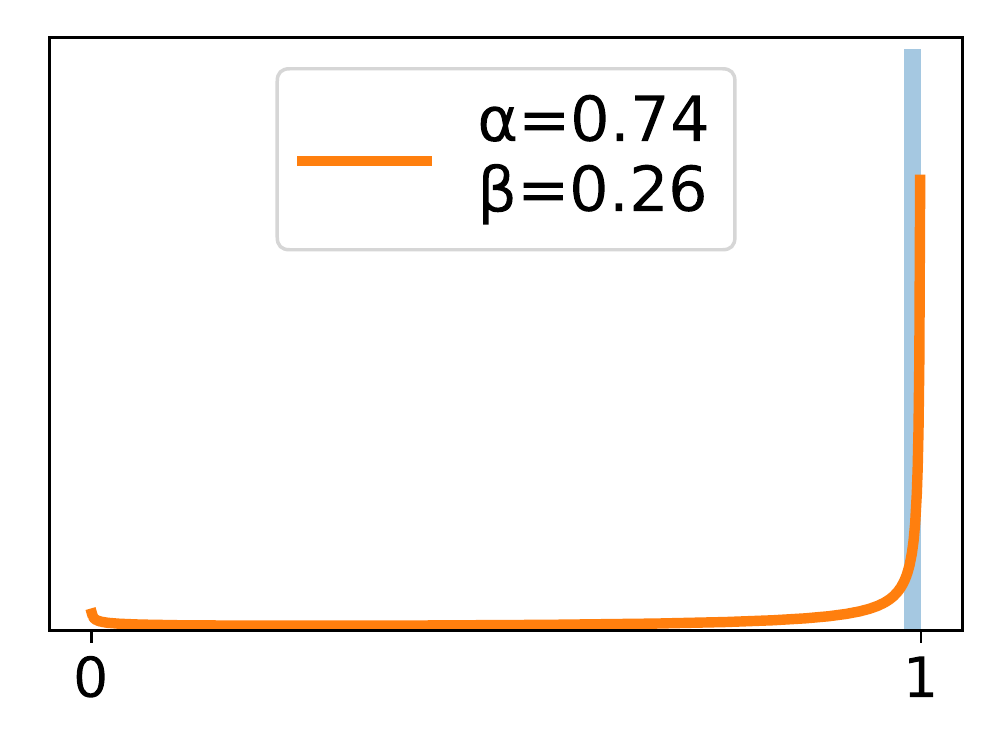}
    \caption{SL3}
\end{subfigure}\hfill
\begin{subfigure}[t]{0.19\textwidth}
    \centering
    \includegraphics[height=2cm]{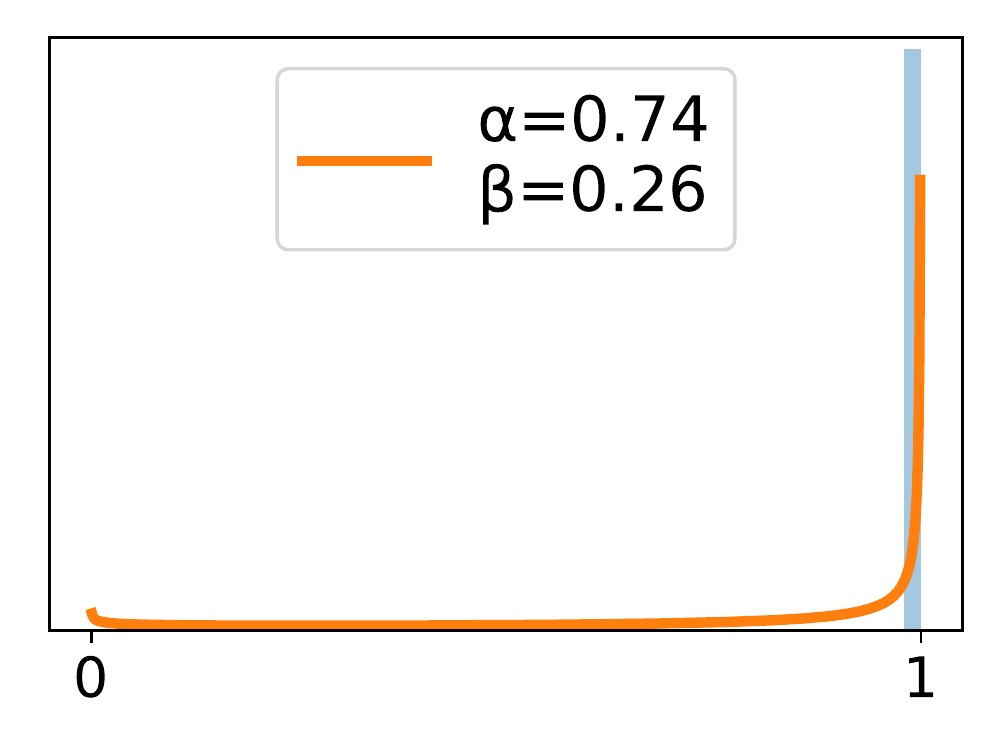}
    \includegraphics[height=2cm]{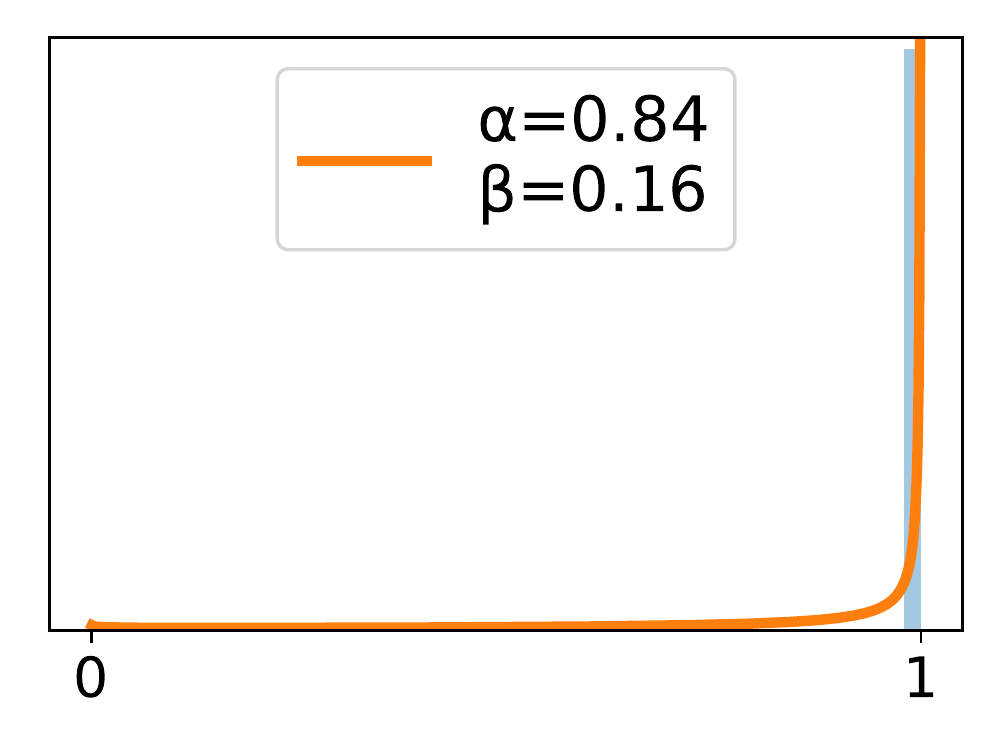}
    \includegraphics[height=2cm]{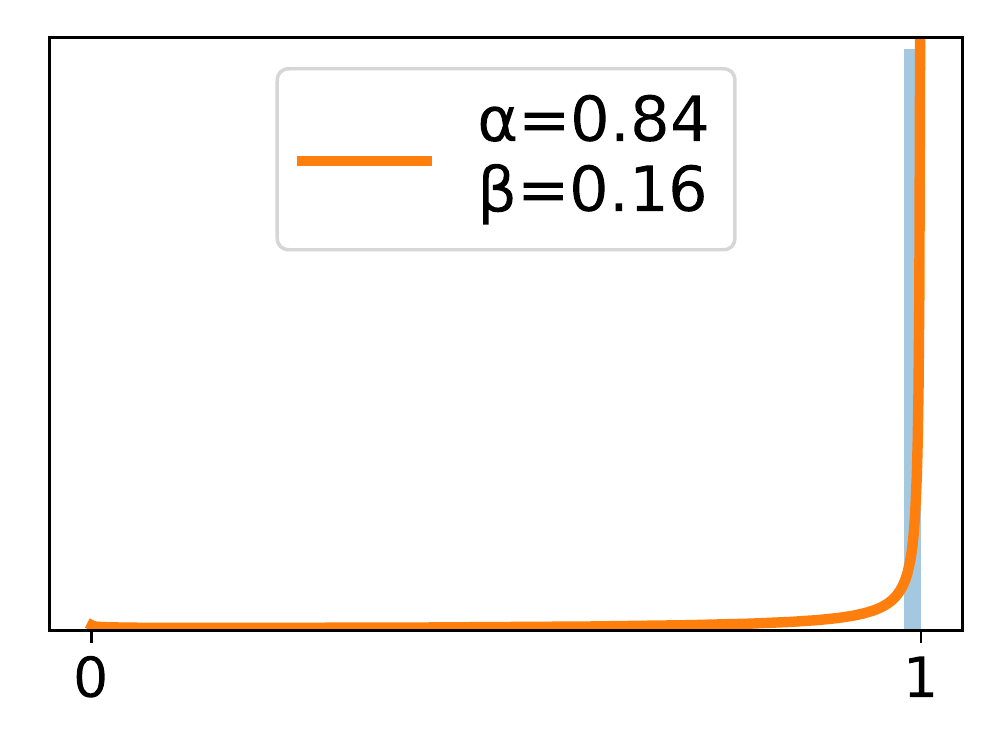}
    \includegraphics[height=2cm]{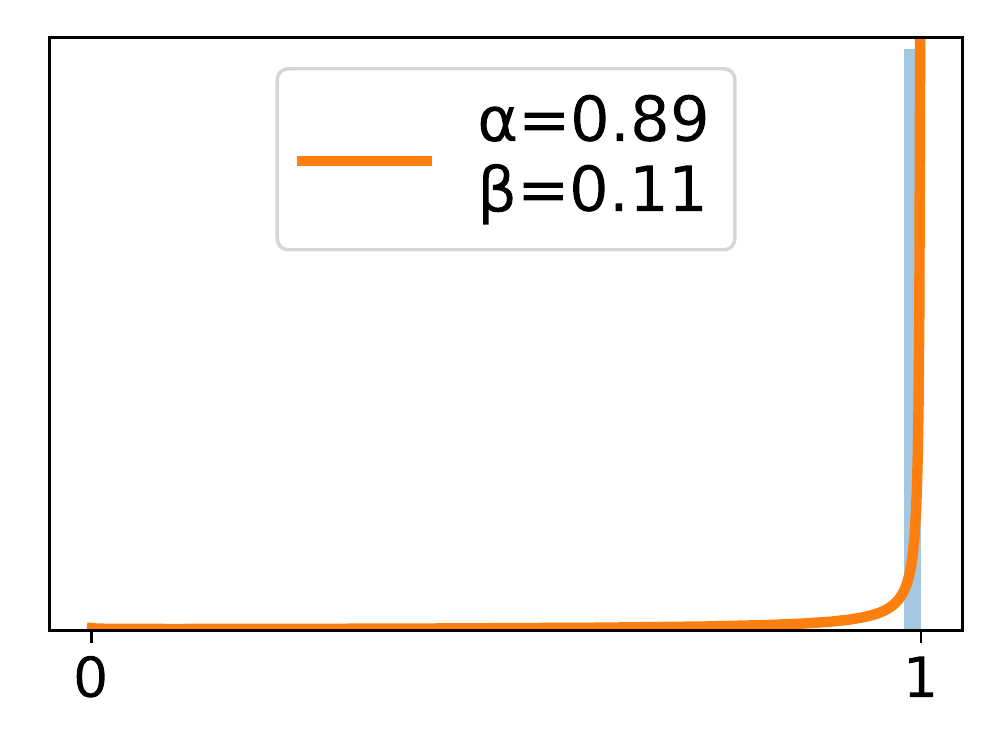}
    \includegraphics[height=2cm]{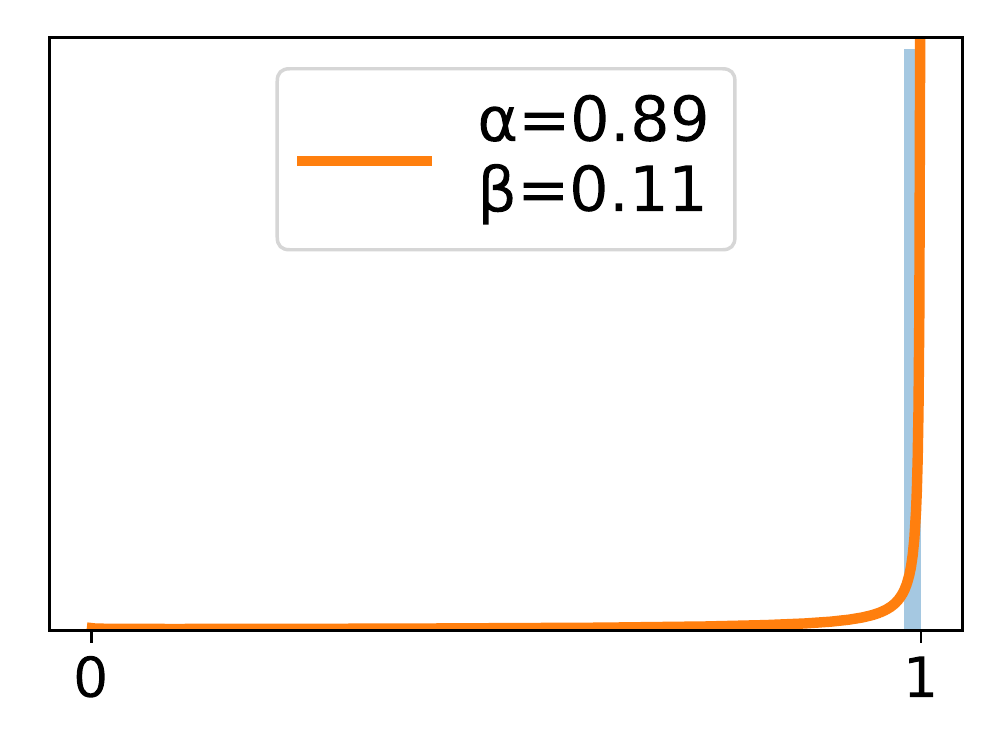}
    \includegraphics[height=2cm]{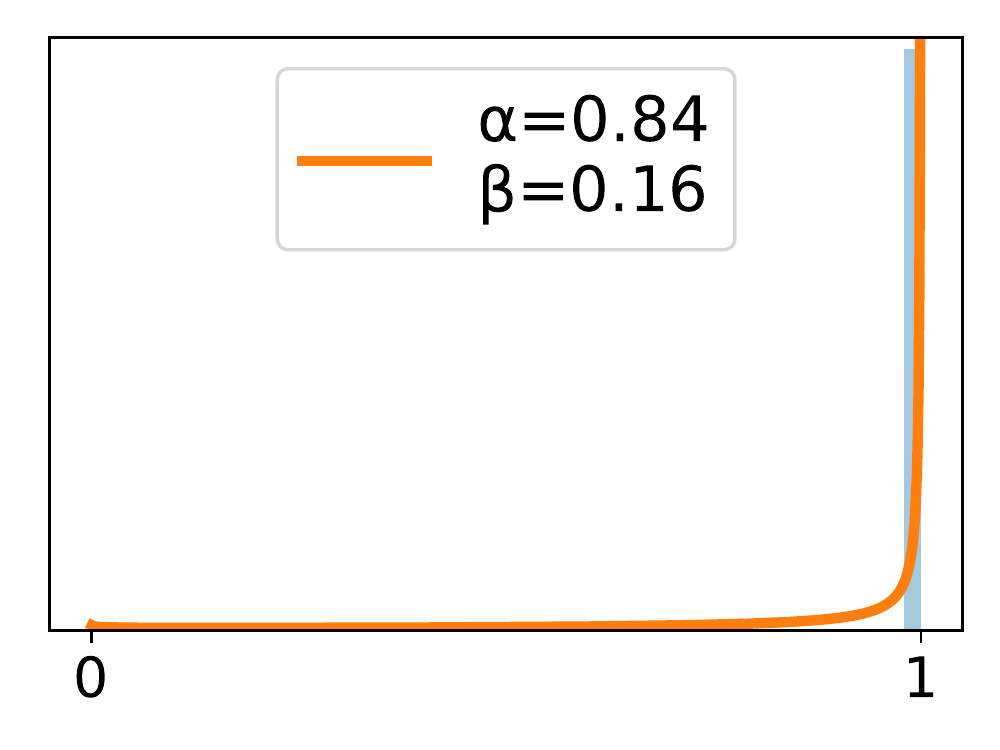}
    \includegraphics[height=2cm]{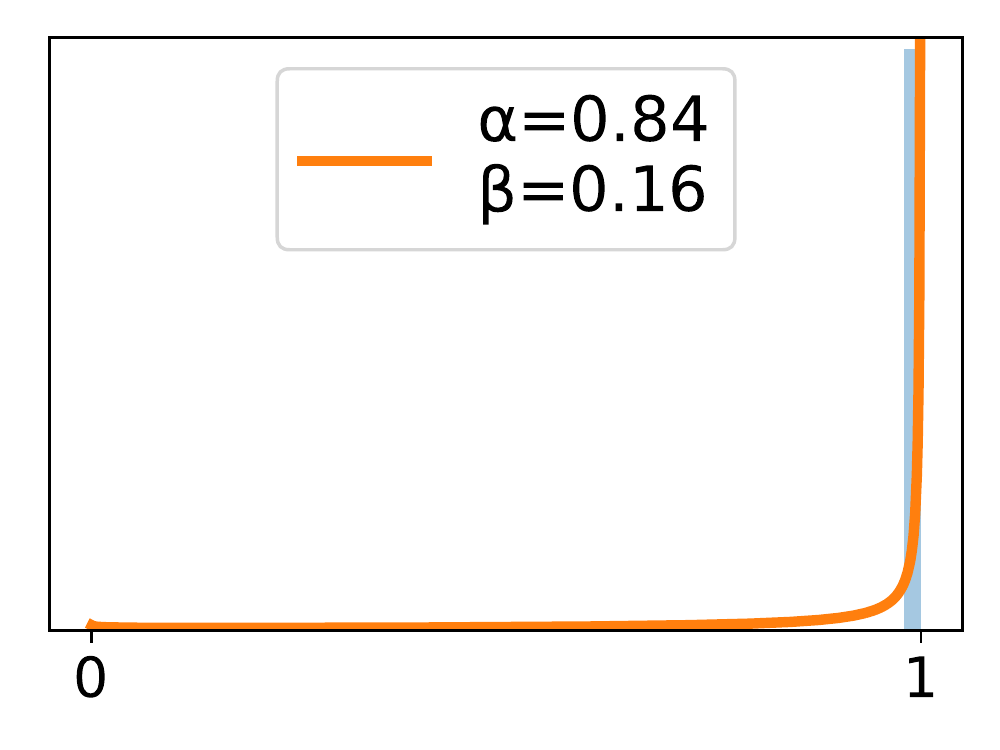}
    \includegraphics[height=2cm]{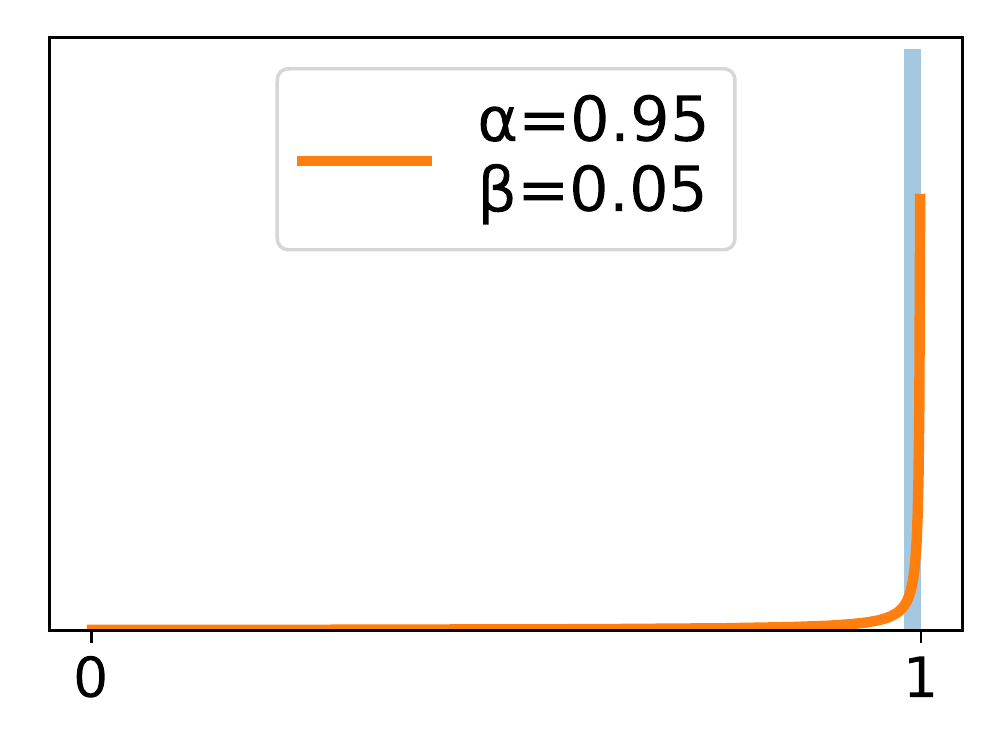}
    \includegraphics[height=2cm]{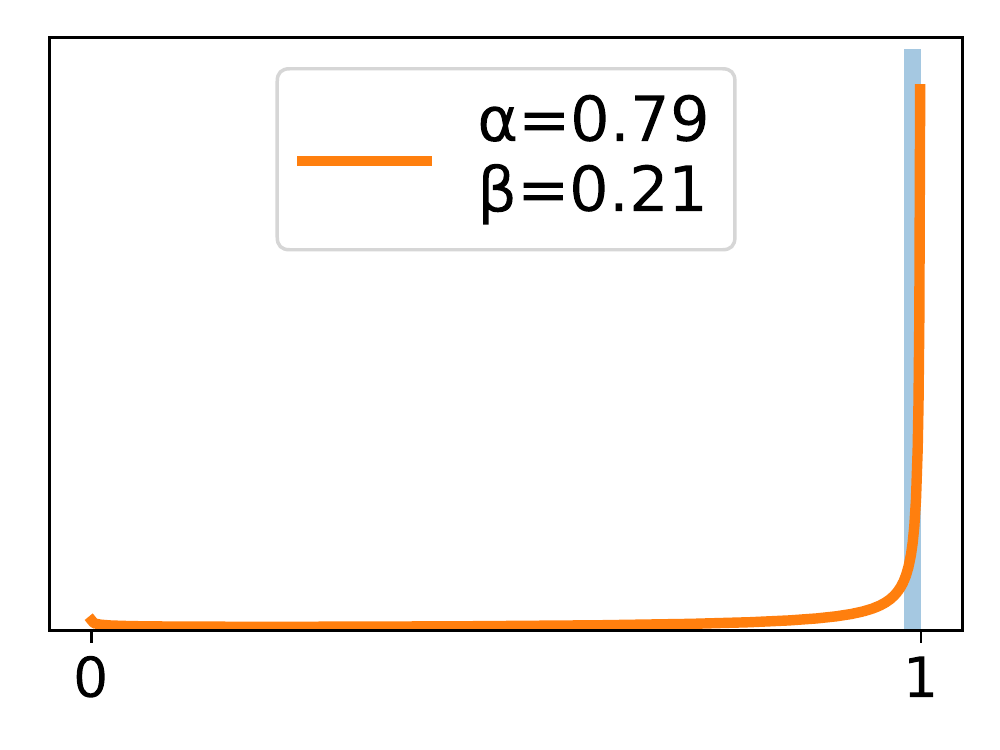}
    \caption{SL4}
\end{subfigure}
\caption{Histograms showing the spread of trained (with $\rho=0.2$) \acs{NN} models' predictions on a selected numbers of data examples from the chiller dataset, and a fitted Beta distribution $\mathcal{B}(\alpha,\beta)$ for each example.}
\label{fig:NN-chiller-distribution}
\end{figure}

\begin{figure}[p]
\centering
\begin{subfigure}[t]{0.19\textwidth}
    \centering
    \includegraphics[height=2cm]{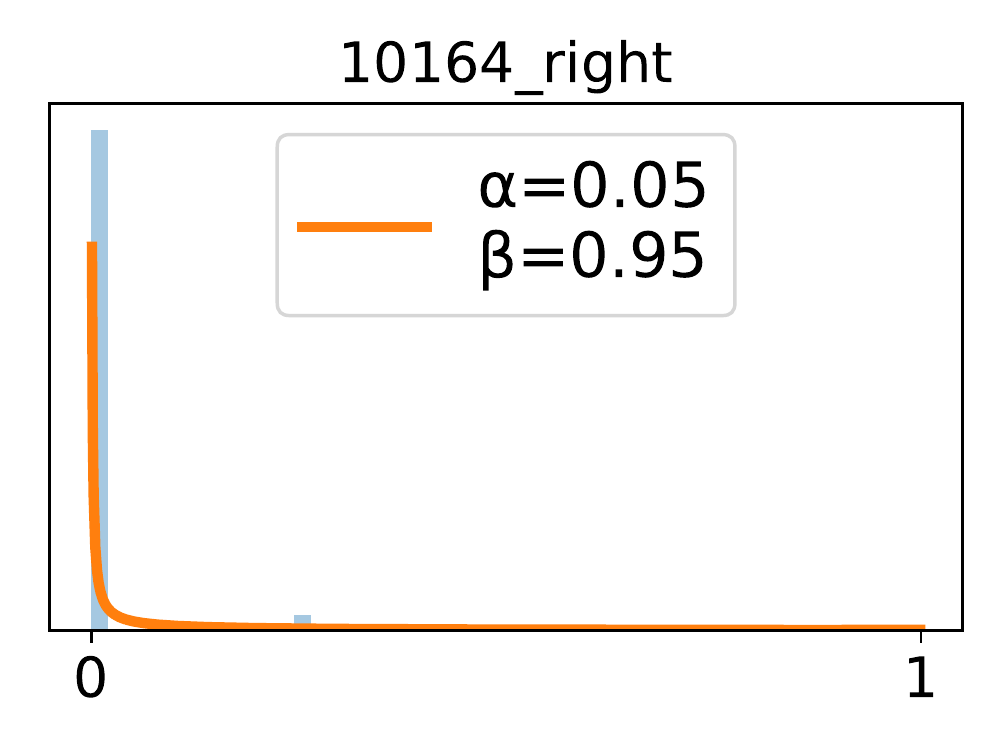}
    \includegraphics[height=2cm]{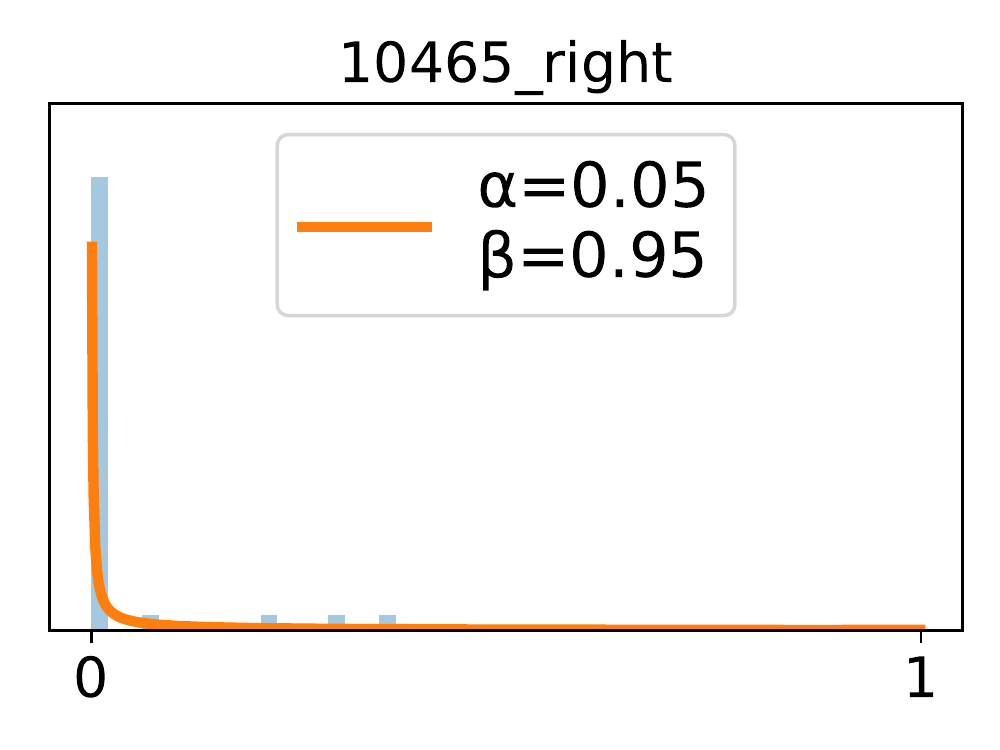}
    \includegraphics[height=2cm]{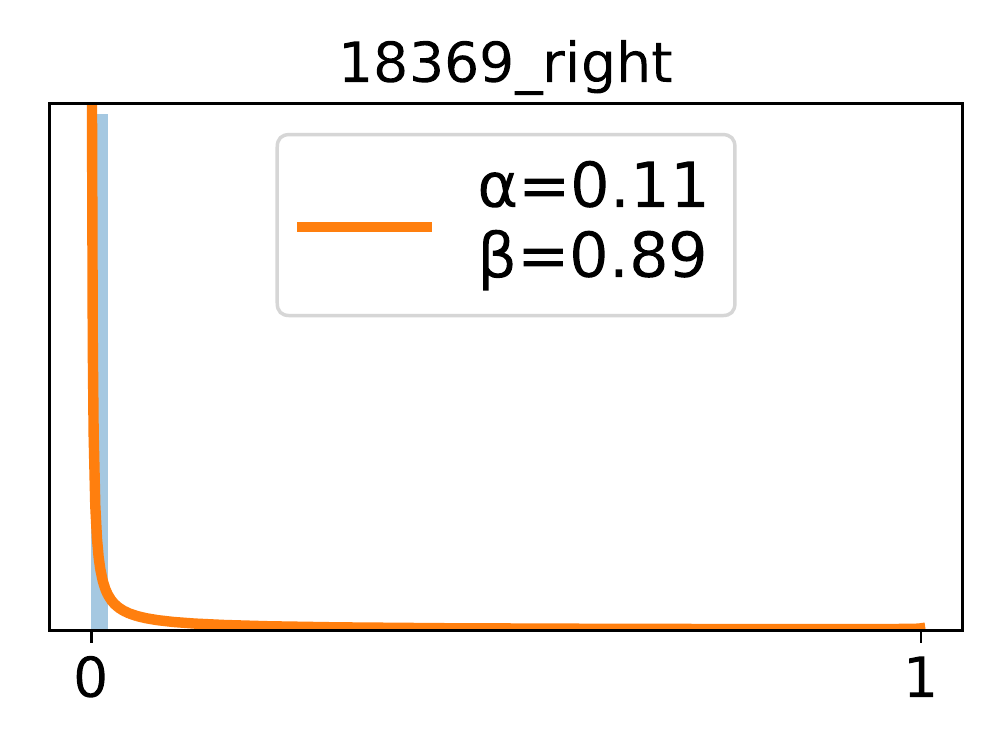}
    \includegraphics[height=2cm]{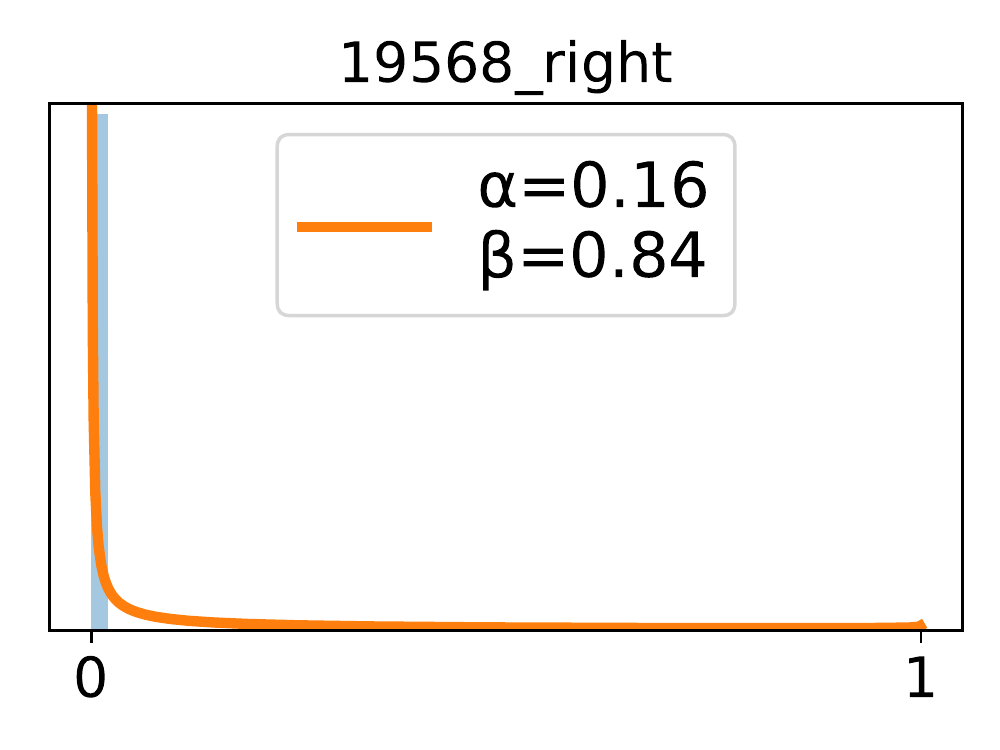}
    \includegraphics[height=2cm]{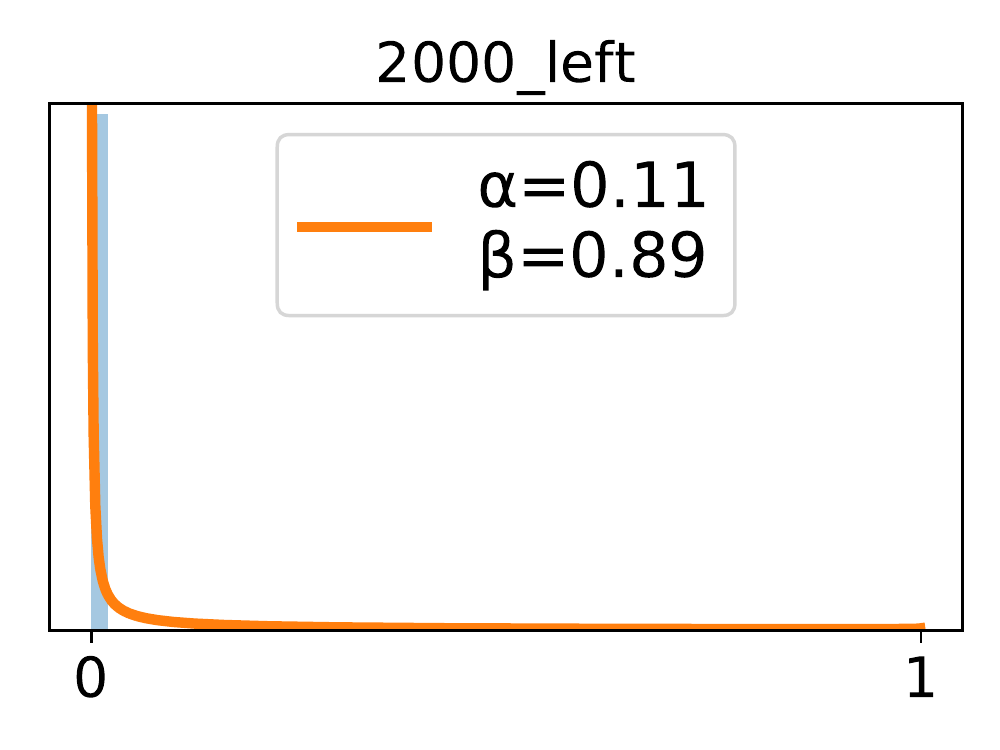}
    \includegraphics[height=2cm]{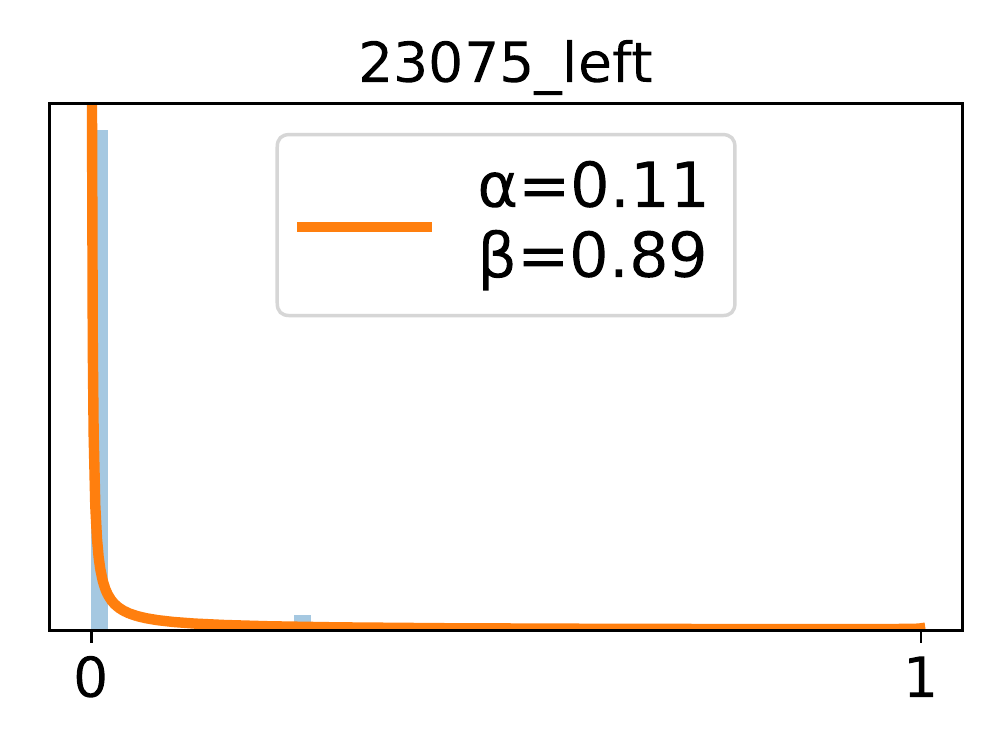}
    \includegraphics[height=2cm]{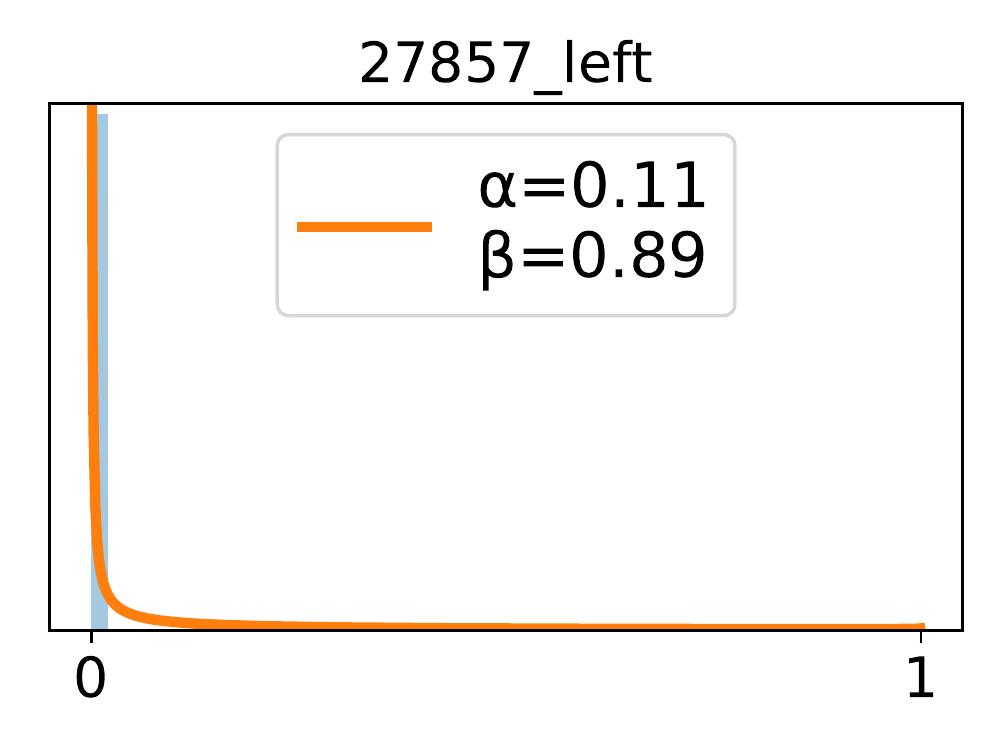}
    \includegraphics[height=2cm]{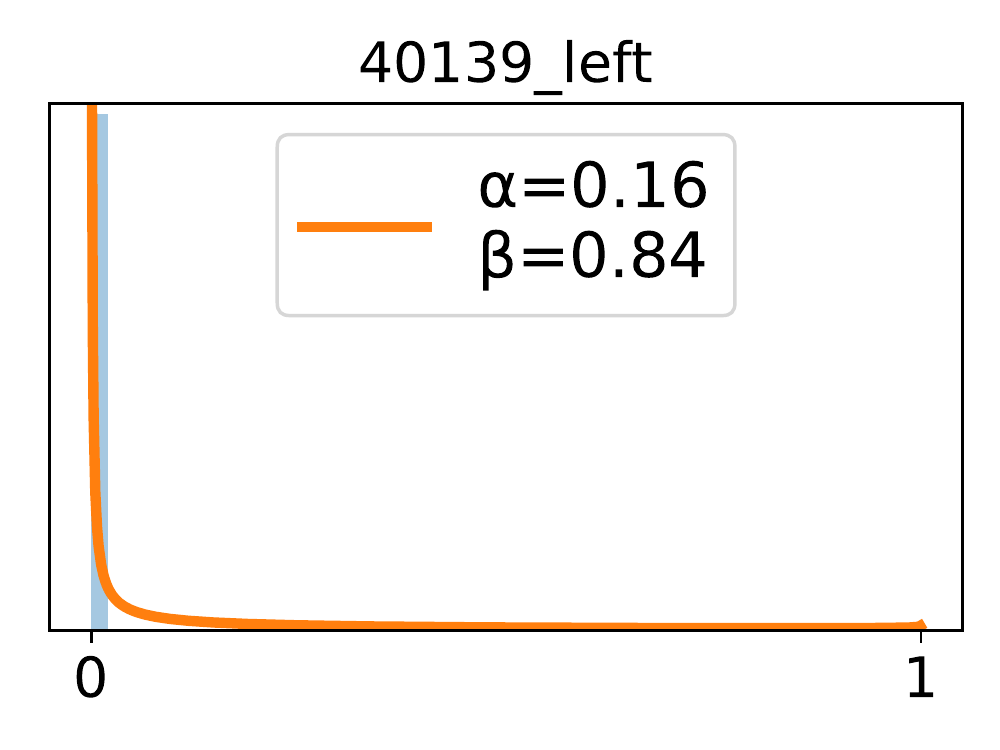}
    \includegraphics[height=2cm]{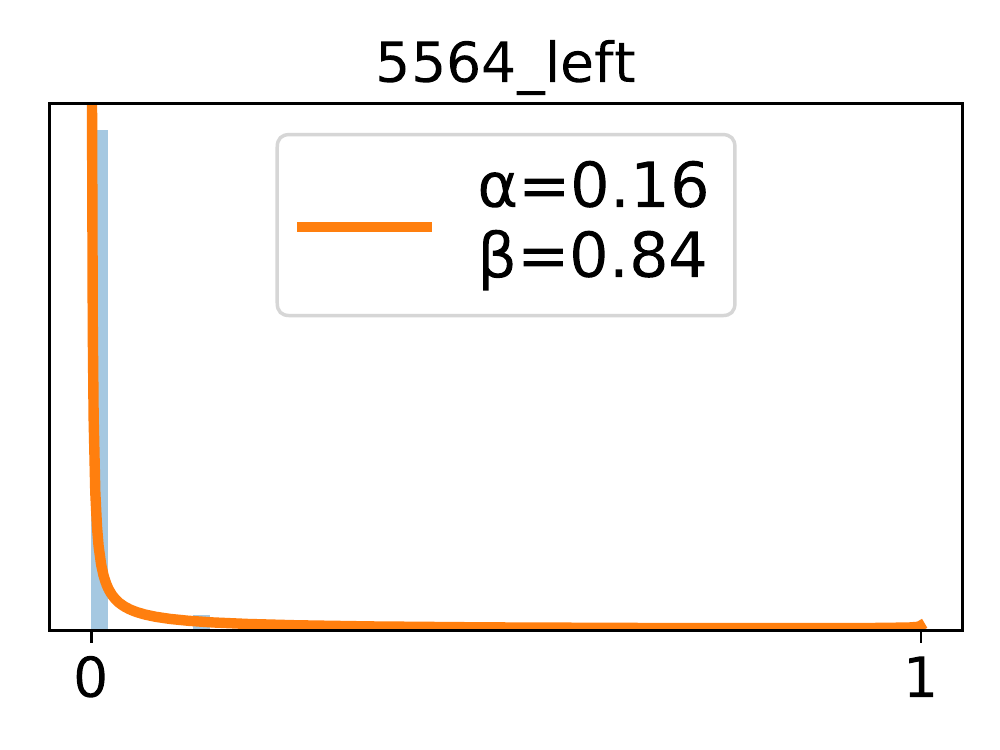}
    \caption{SL0}
\end{subfigure}\hfill
\begin{subfigure}[t]{0.19\textwidth}
    \centering
    \includegraphics[height=2cm]{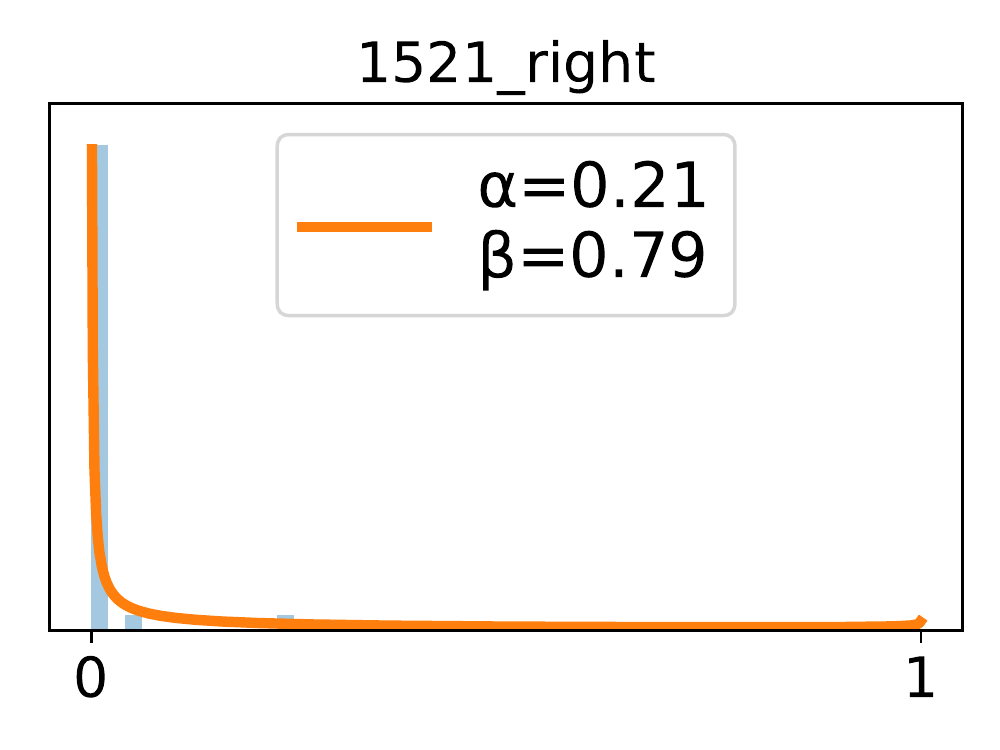}
    \includegraphics[height=2cm]{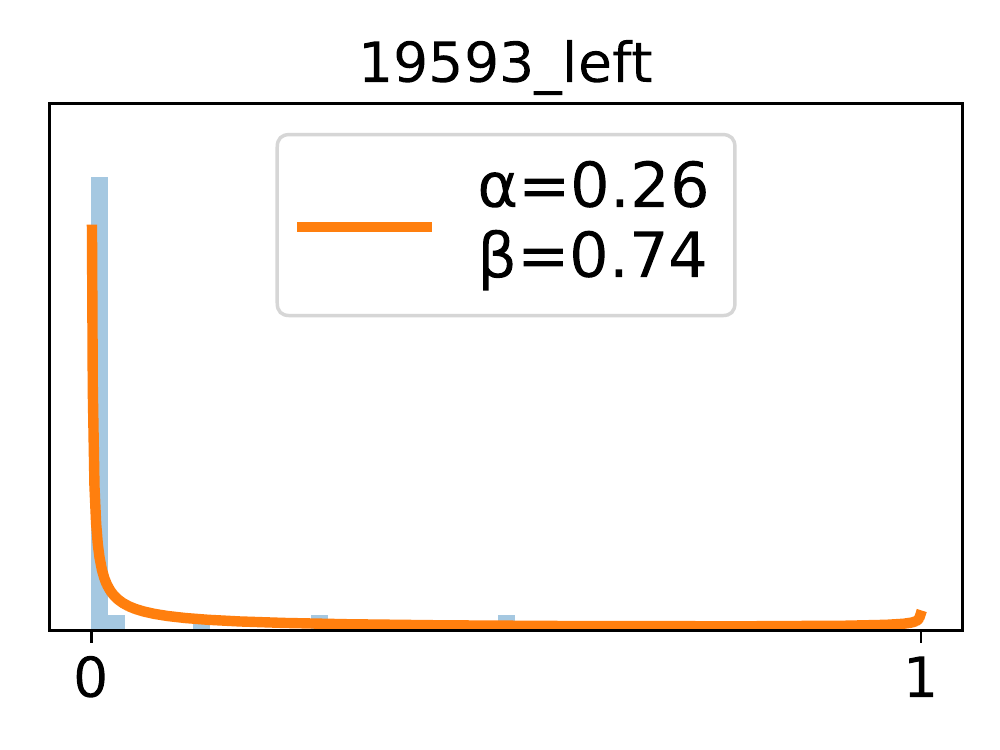}
    \includegraphics[height=2cm]{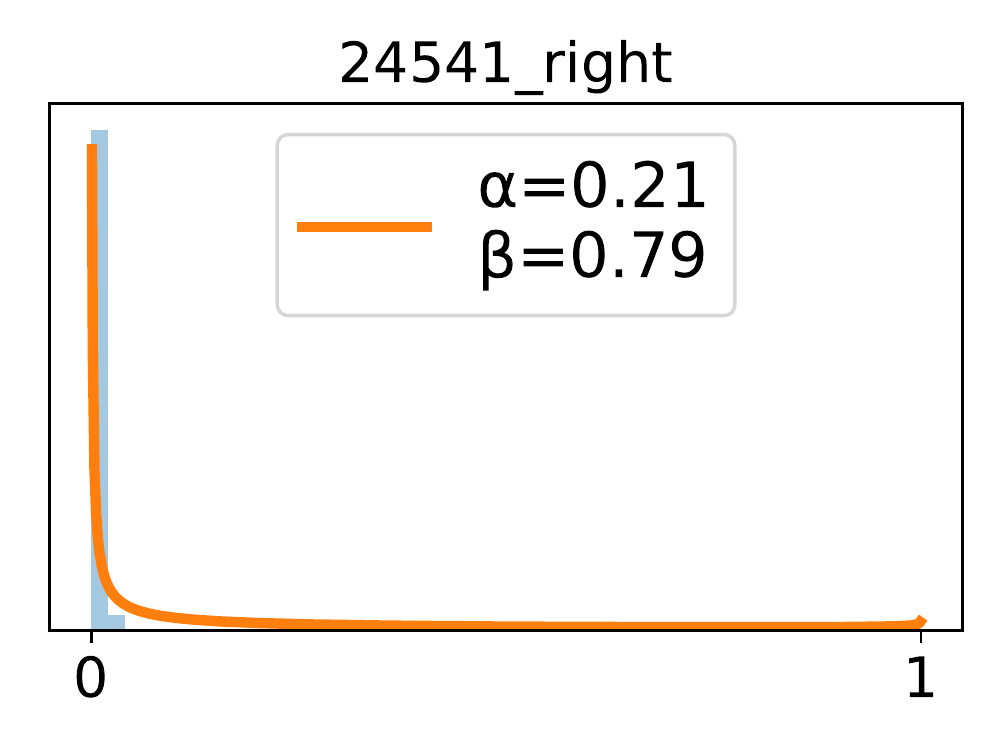}
    \includegraphics[height=2cm]{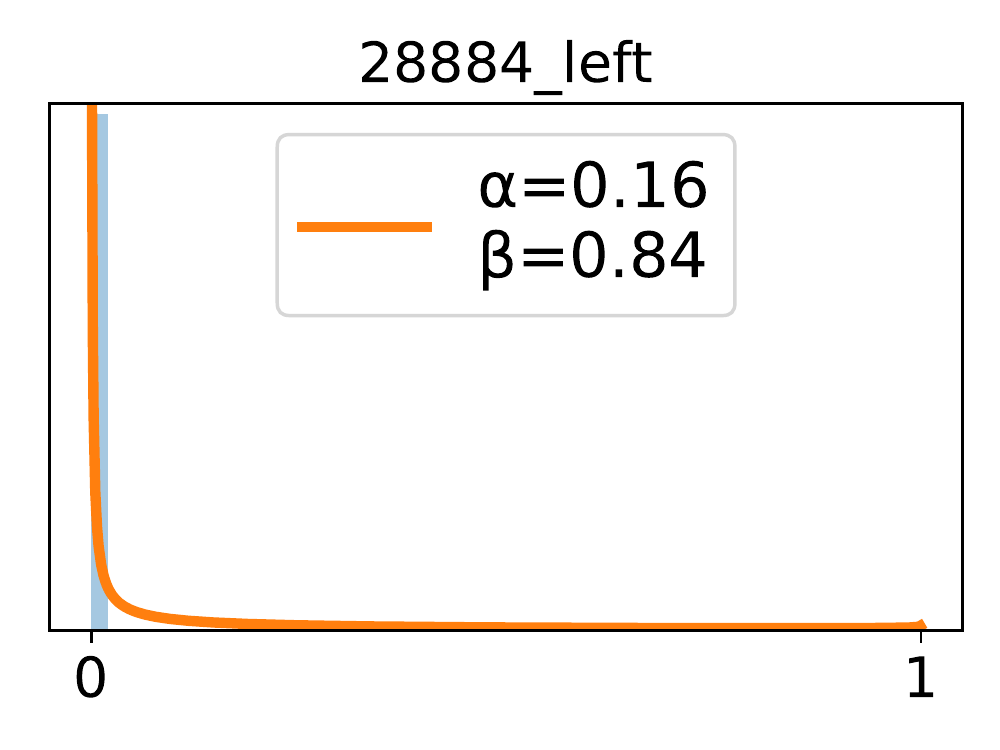}
    \includegraphics[height=2cm]{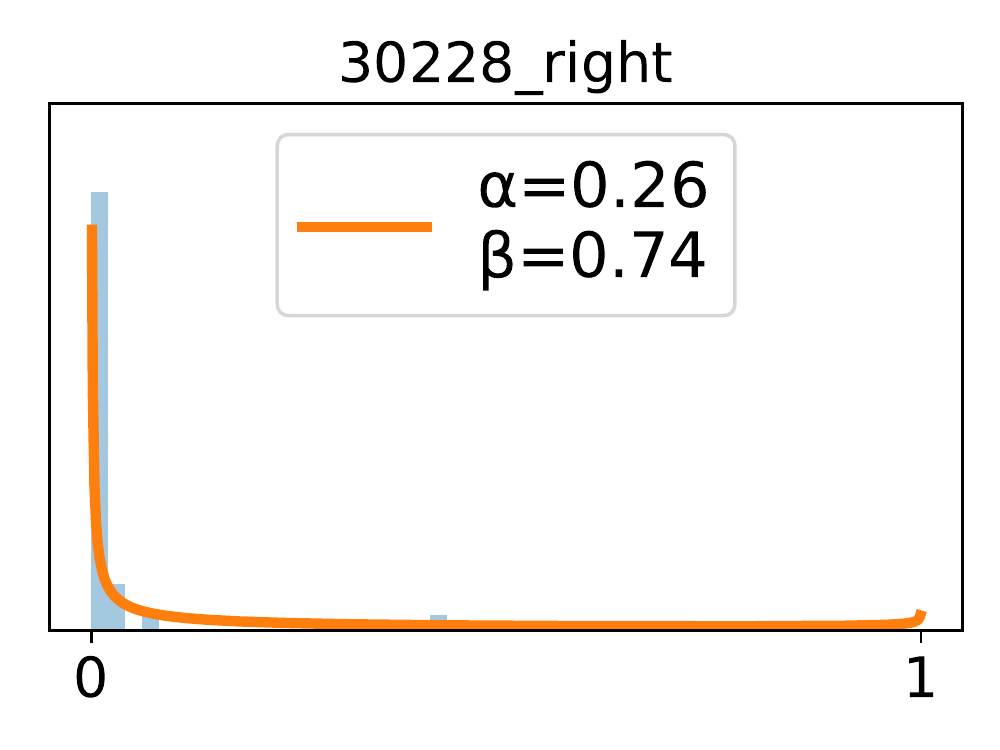}
    \includegraphics[height=2cm]{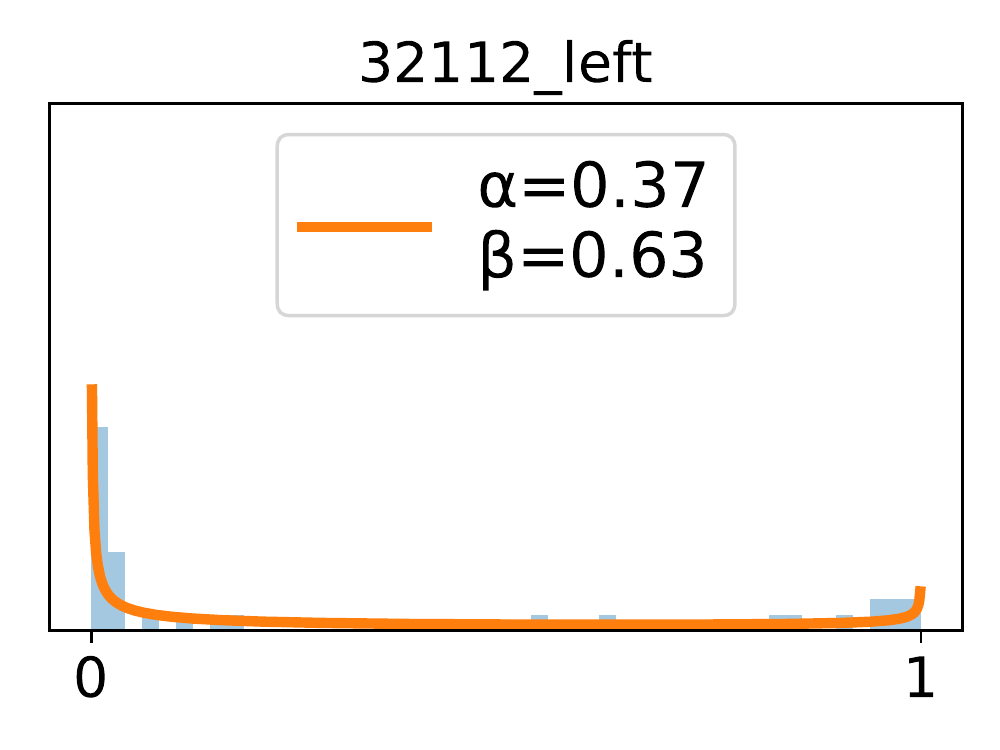}
    \includegraphics[height=2cm]{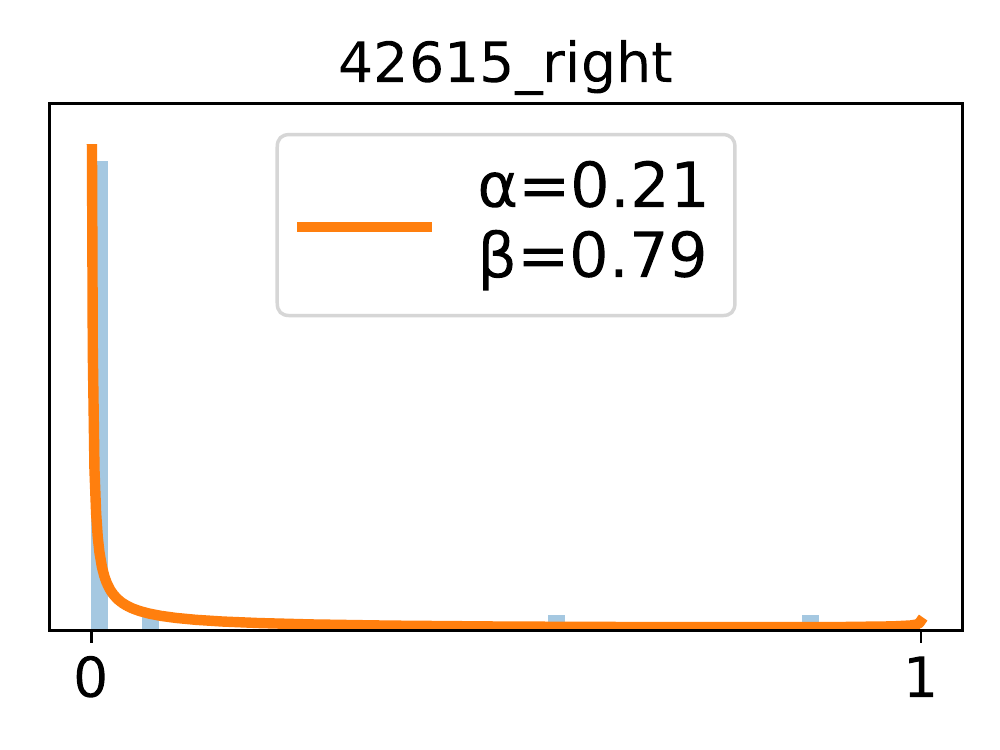}
    \includegraphics[height=2cm]{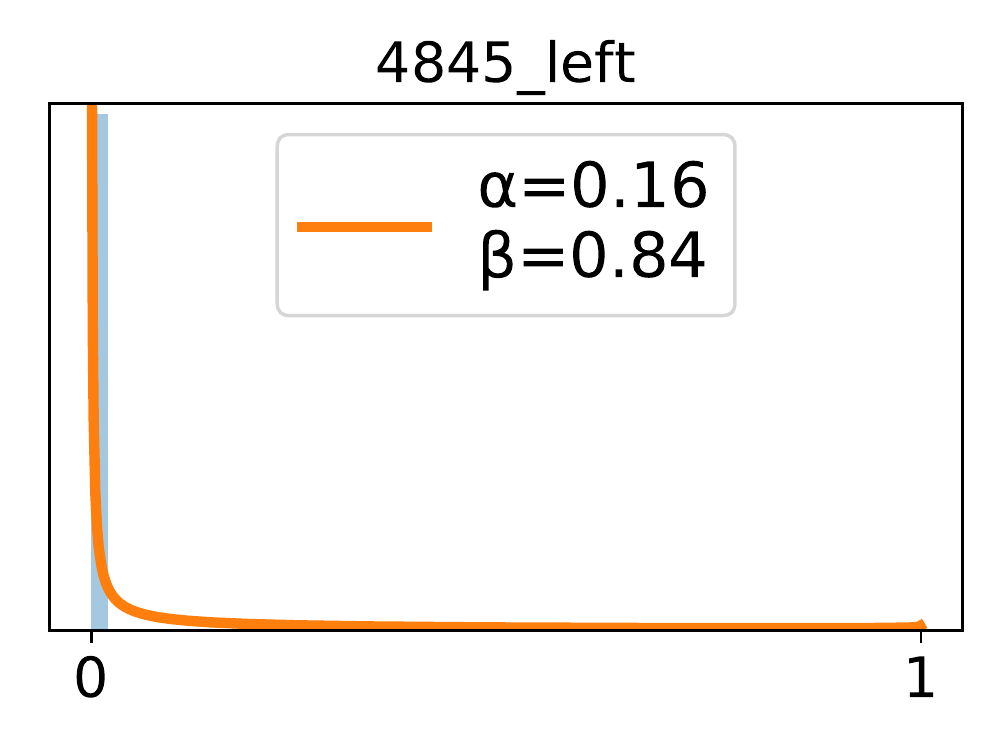}
    \includegraphics[height=2cm]{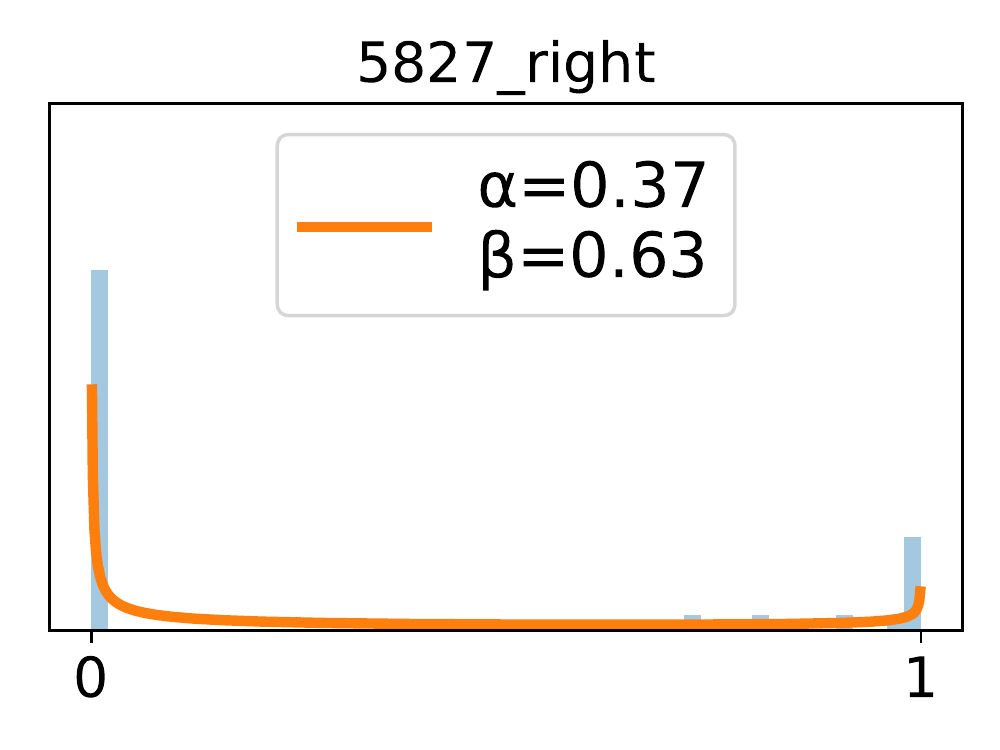}
    \caption{SL1}
\end{subfigure}\hfill
\begin{subfigure}[t]{0.19\textwidth}
    \centering
    \includegraphics[height=2cm]{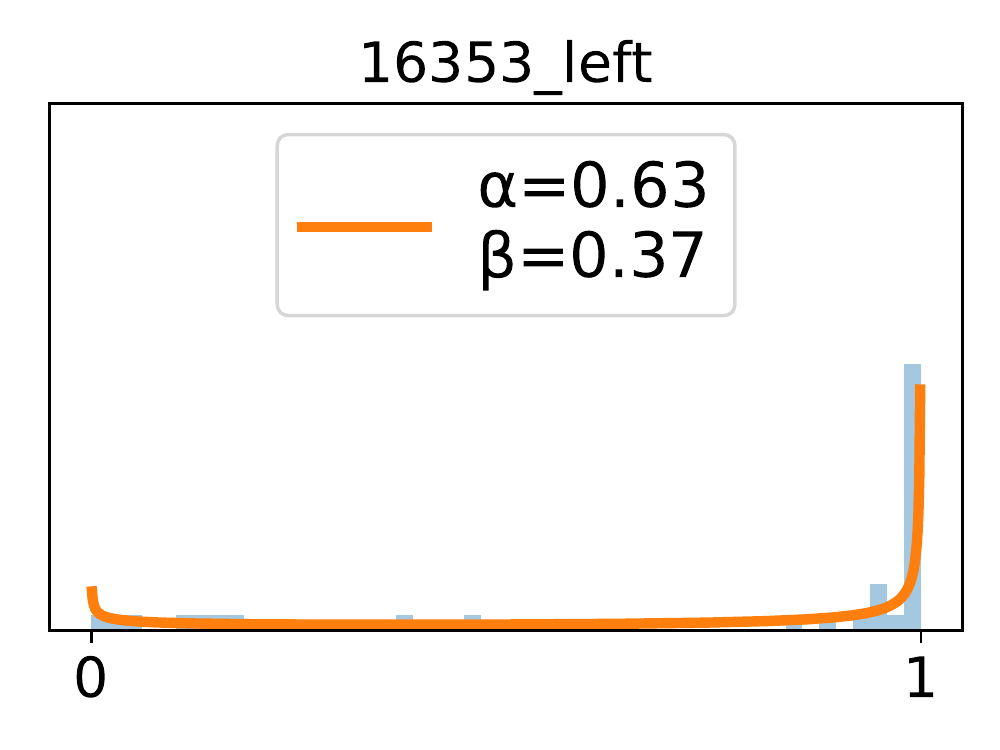}
    \includegraphics[height=2cm]{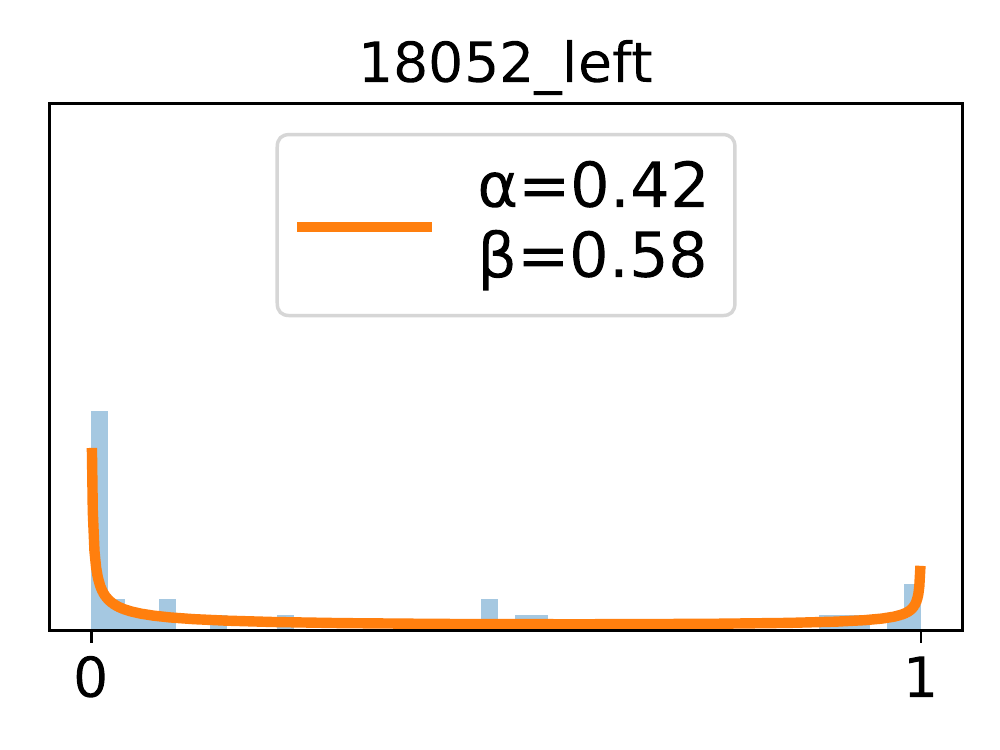}
    \includegraphics[height=2cm]{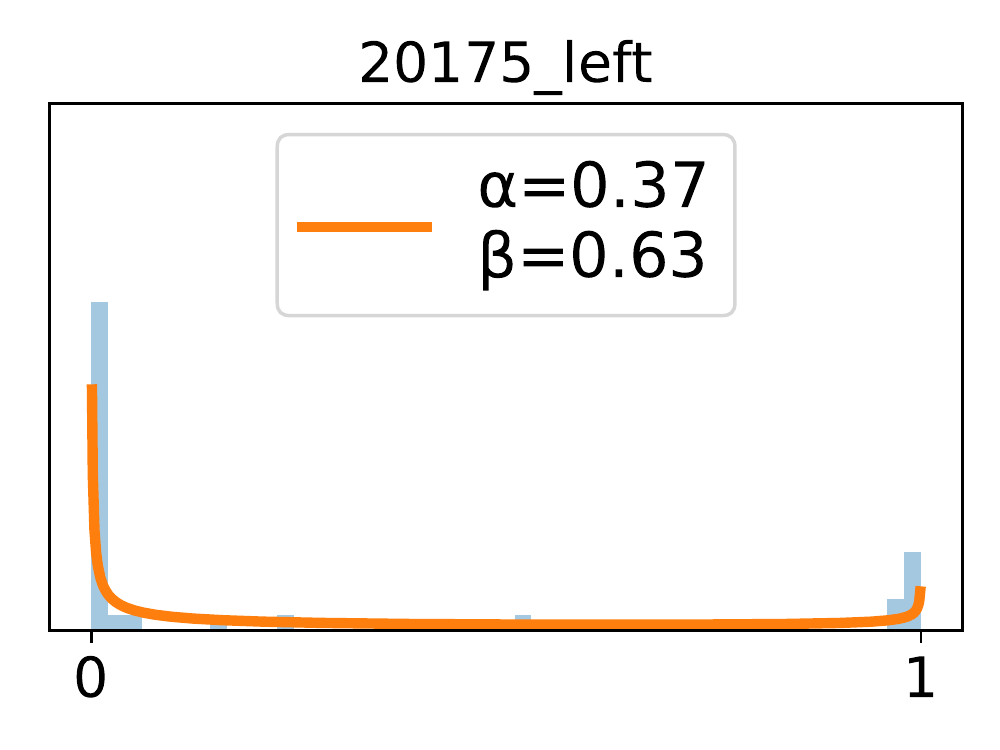}
    \includegraphics[height=2cm]{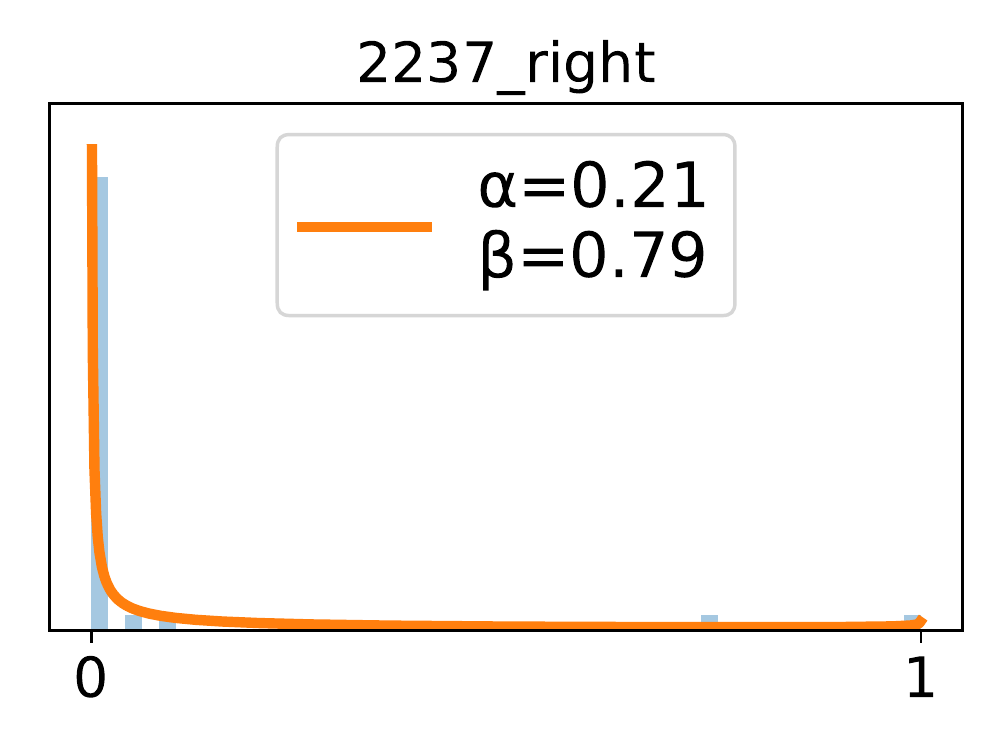}
    \includegraphics[height=2cm]{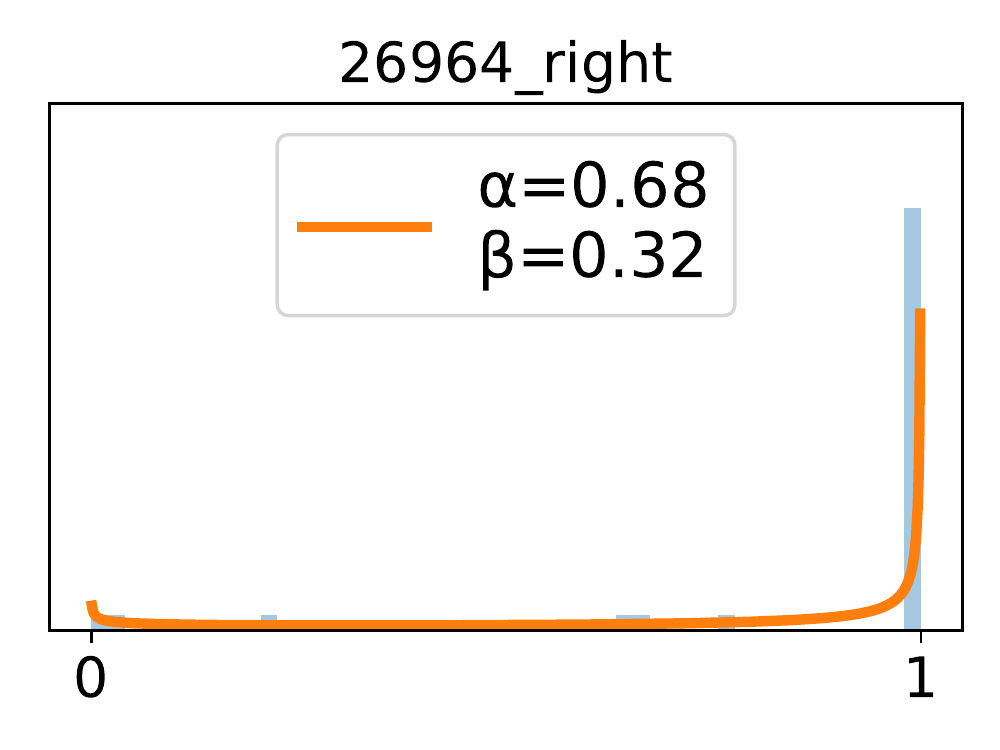}
    \includegraphics[height=2cm]{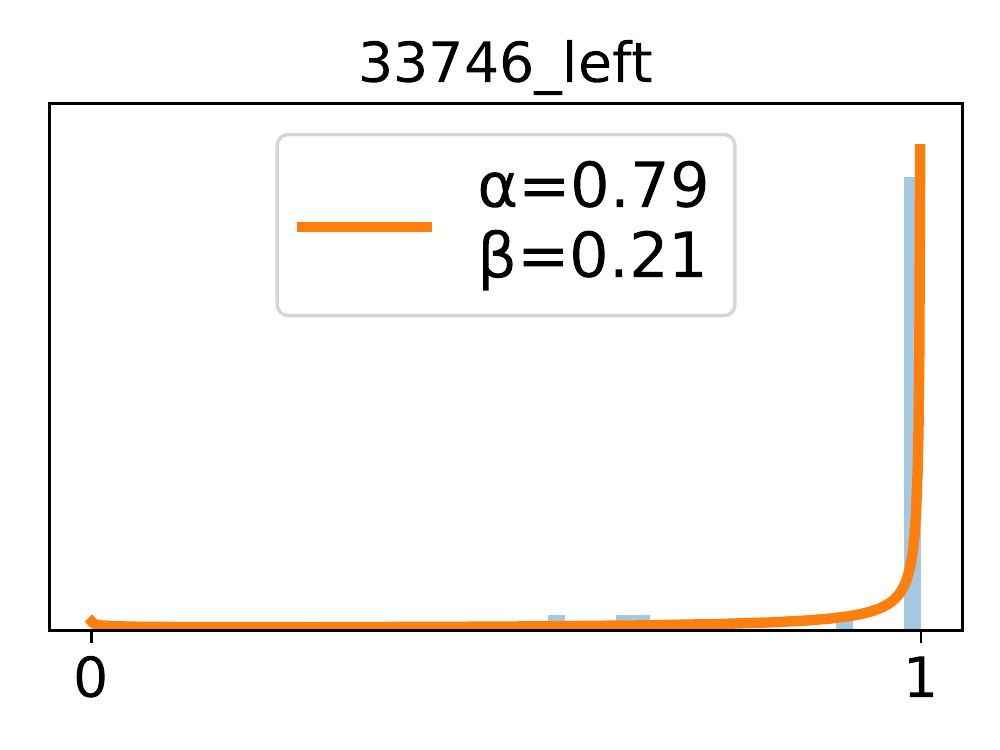}
    \includegraphics[height=2cm]{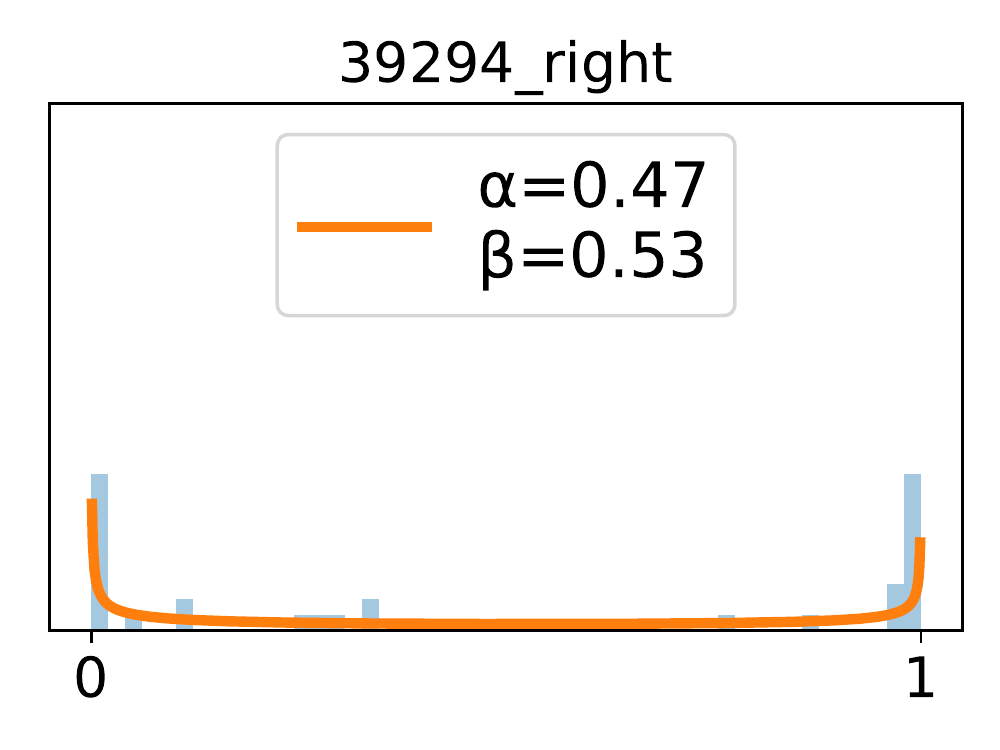}
    \includegraphics[height=2cm]{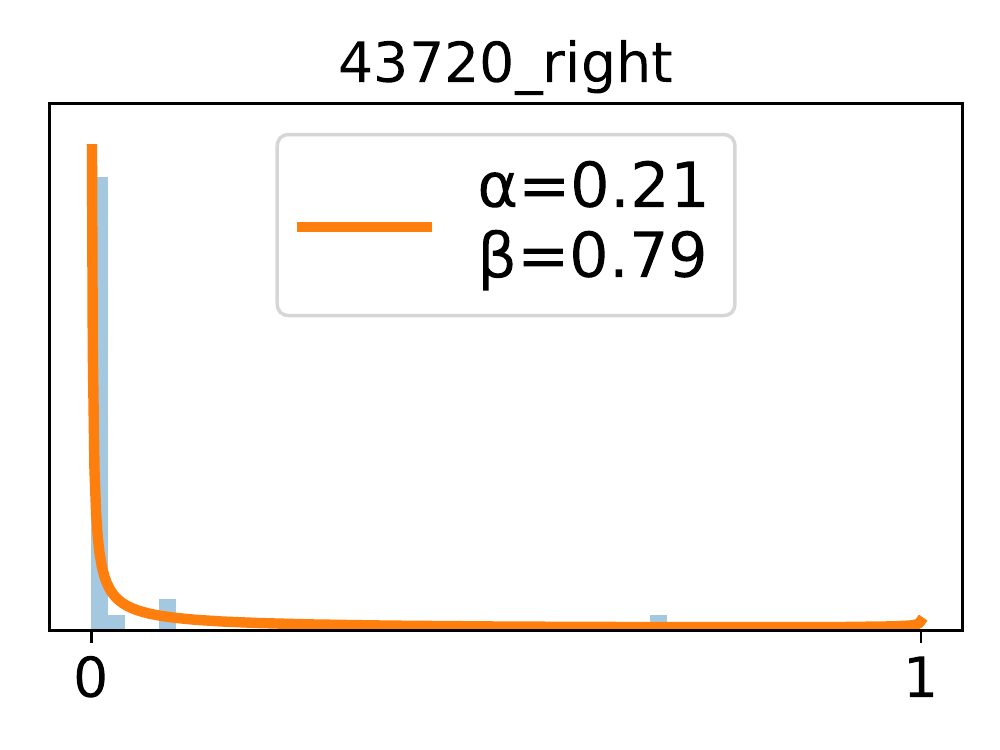}
    \includegraphics[height=2cm]{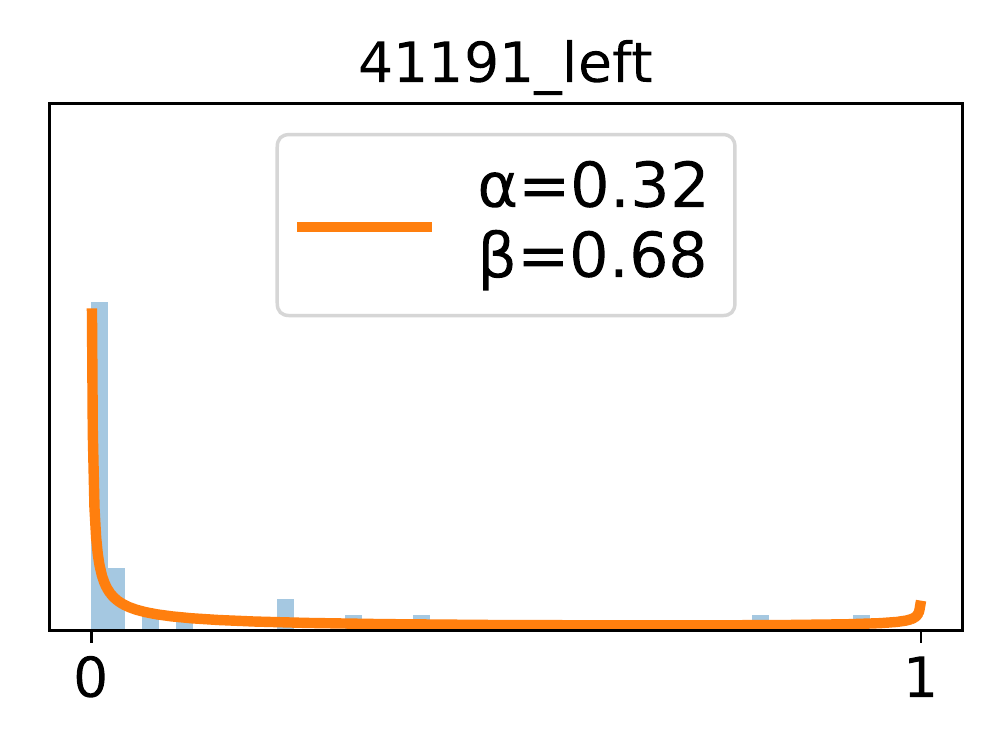}
    \caption{SL2}
\end{subfigure}\hfill
\begin{subfigure}[t]{0.19\textwidth}
    \centering
    \includegraphics[height=2cm]{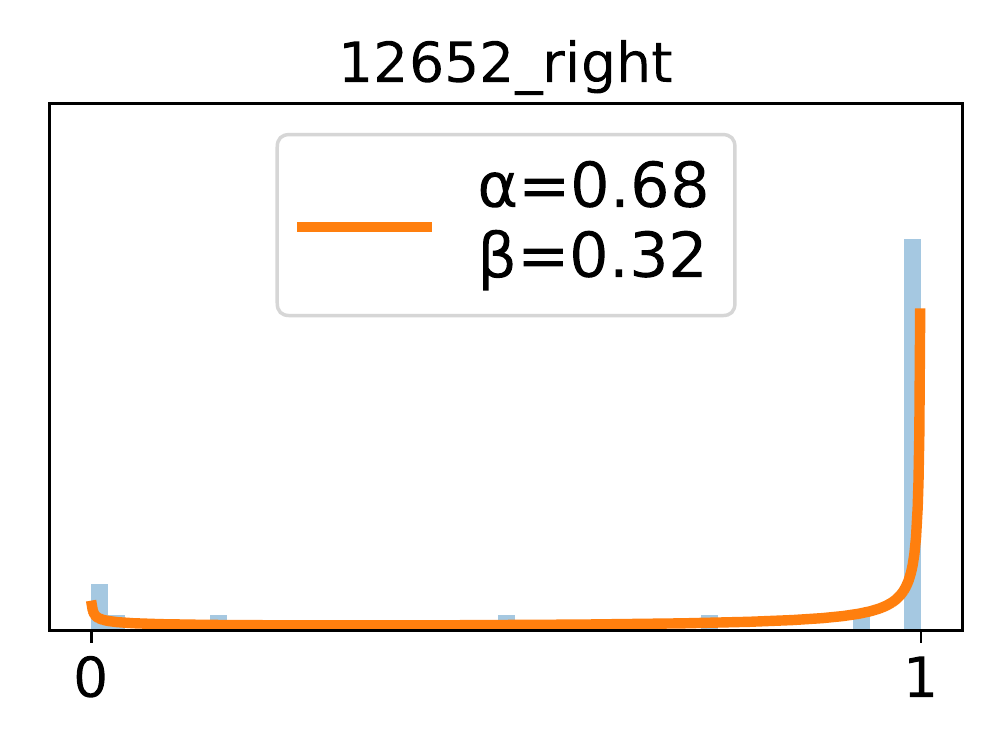}
    \includegraphics[height=2cm]{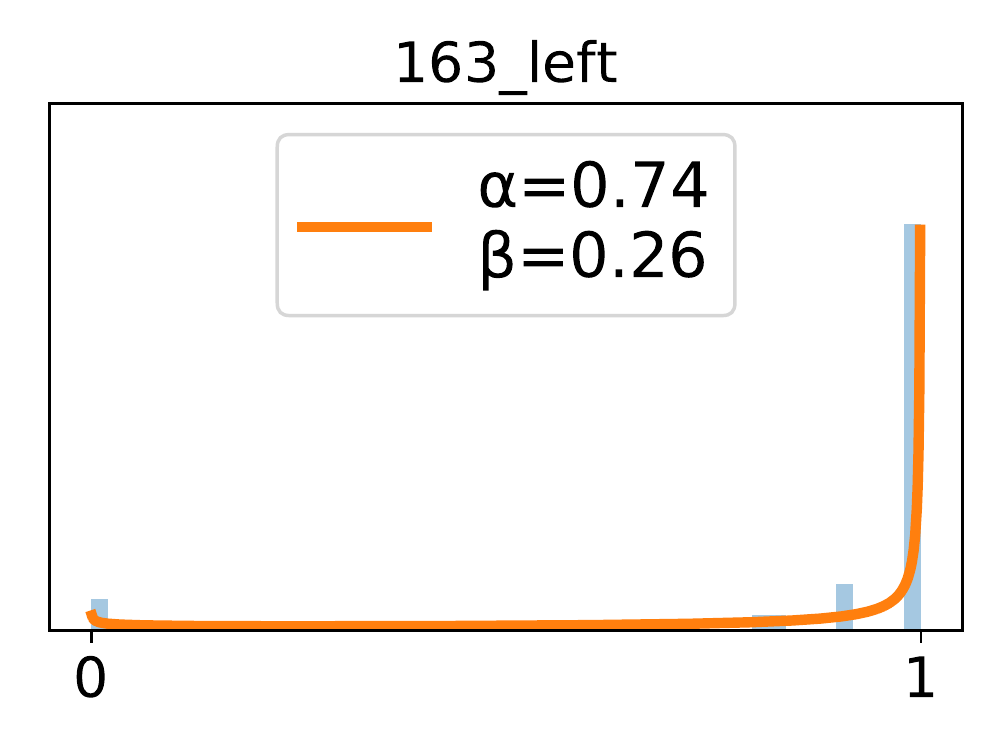}
    \includegraphics[height=2cm]{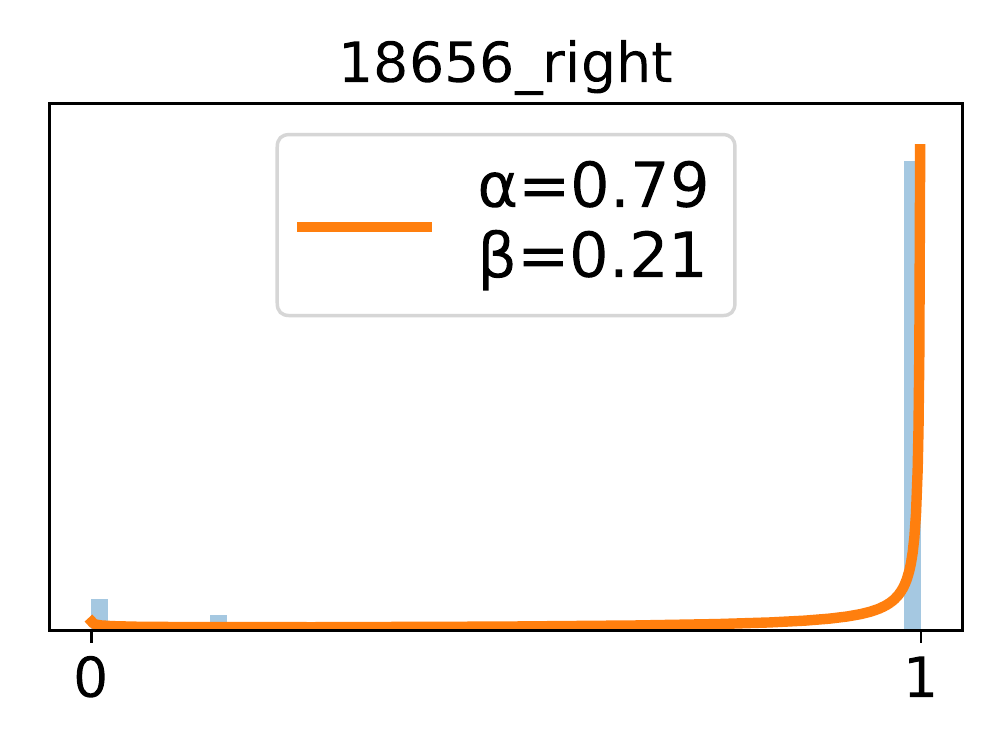}
    \includegraphics[height=2cm]{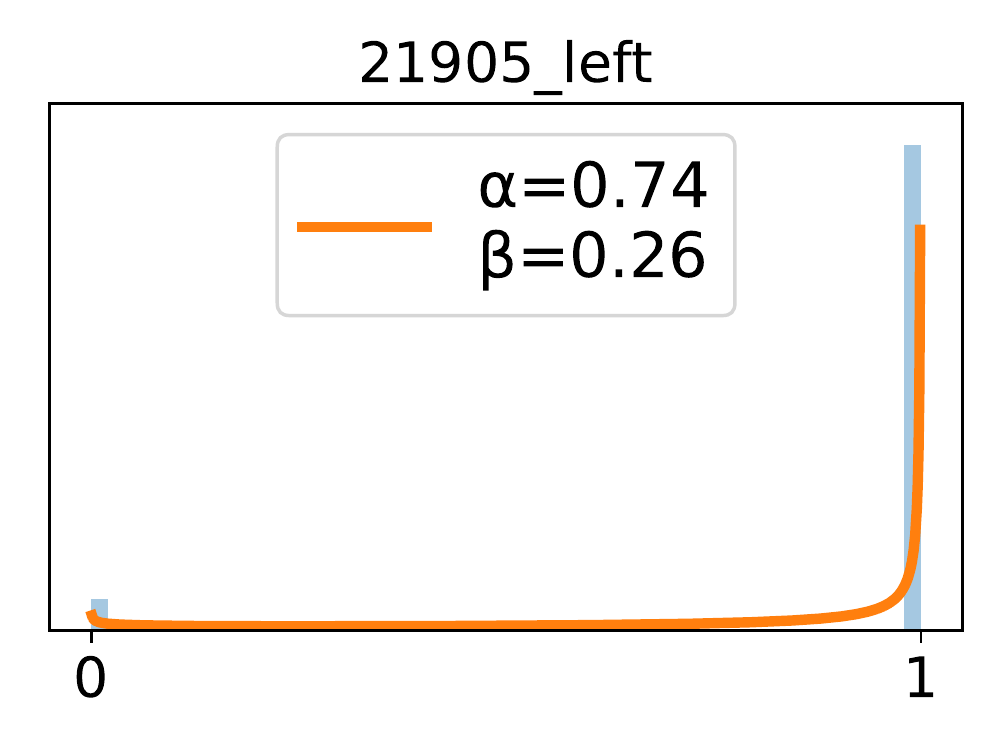}
    \includegraphics[height=2cm]{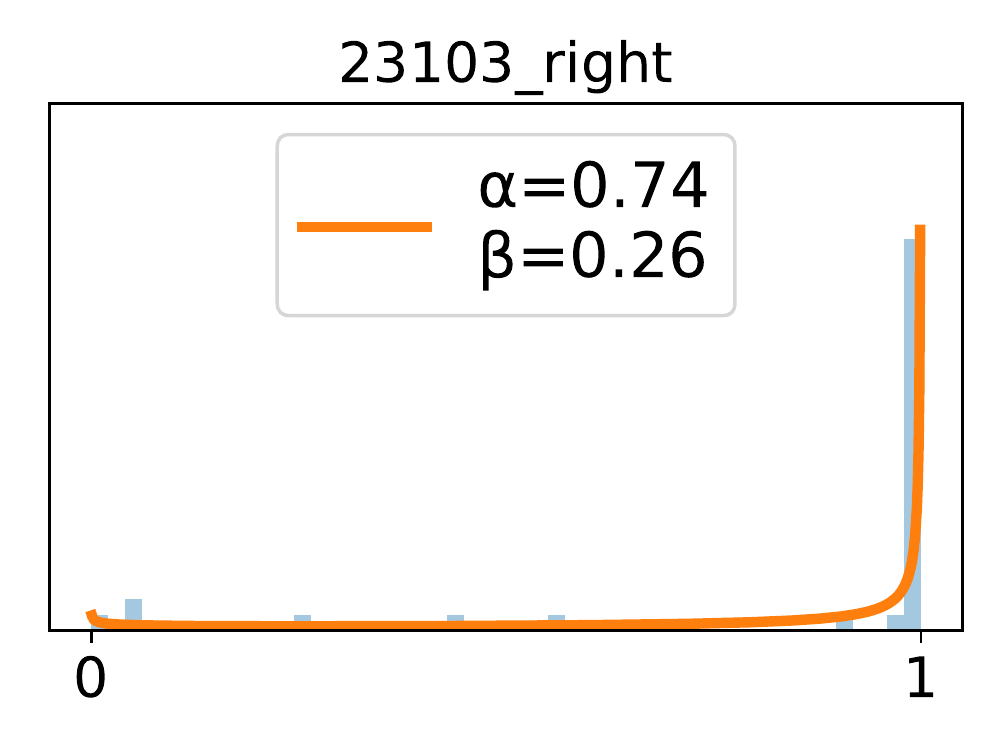}
    \includegraphics[height=2cm]{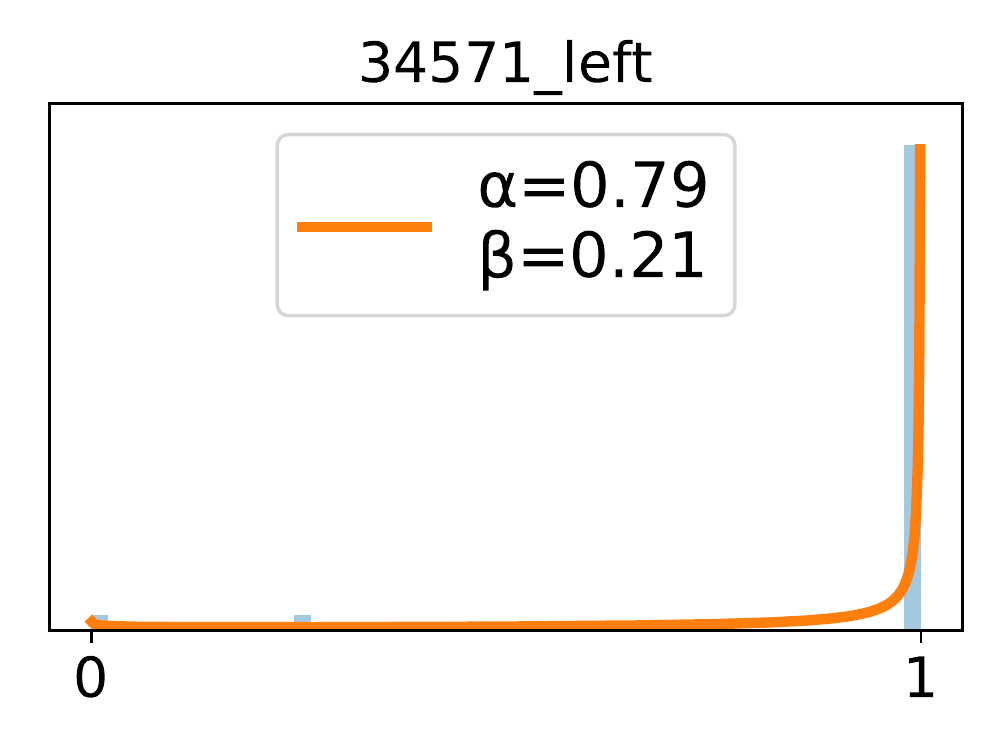}
    \includegraphics[height=2cm]{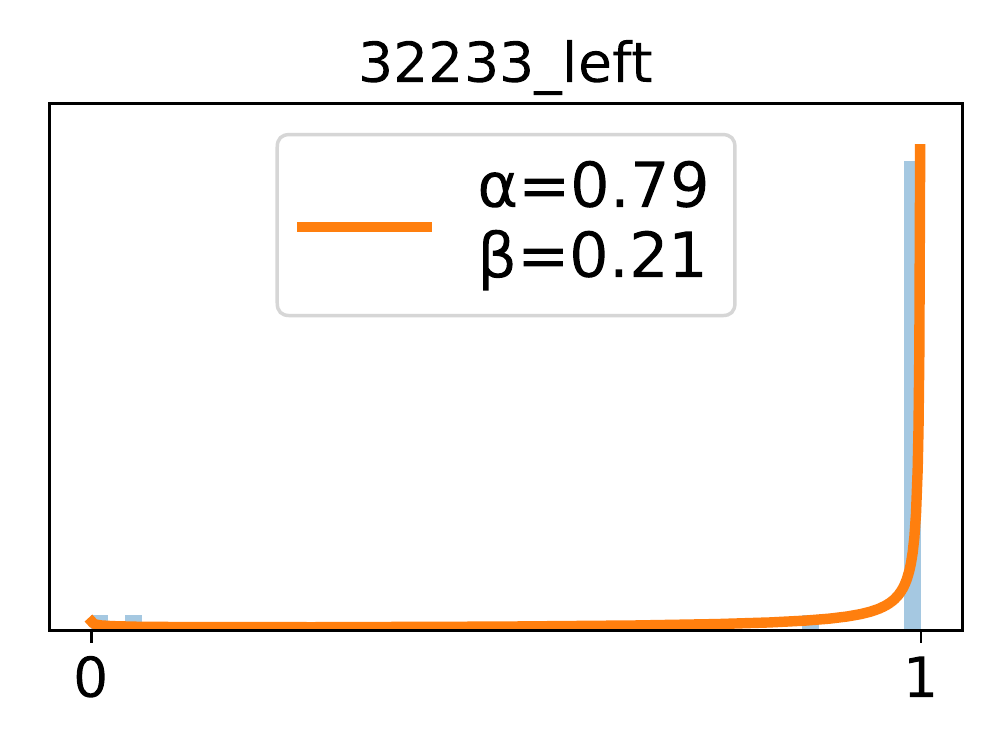}
    \includegraphics[height=2cm]{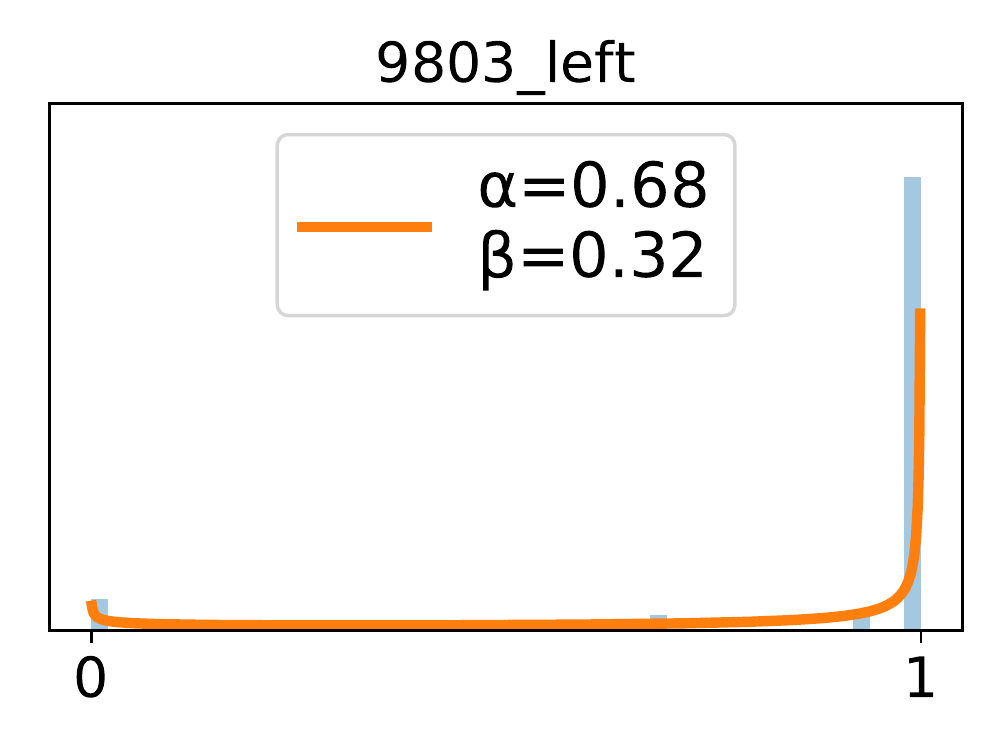}
    \includegraphics[height=2cm]{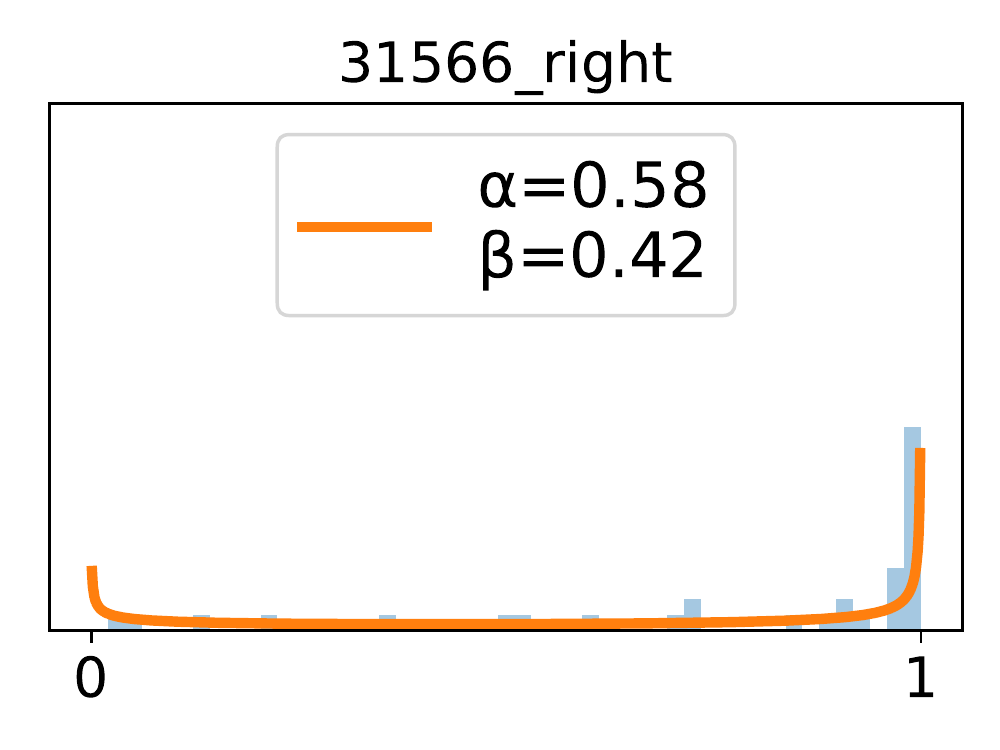}
    \caption{SL3}
\end{subfigure}\hfill
\begin{subfigure}[t]{0.19\textwidth}
    \centering
    \includegraphics[height=2cm]{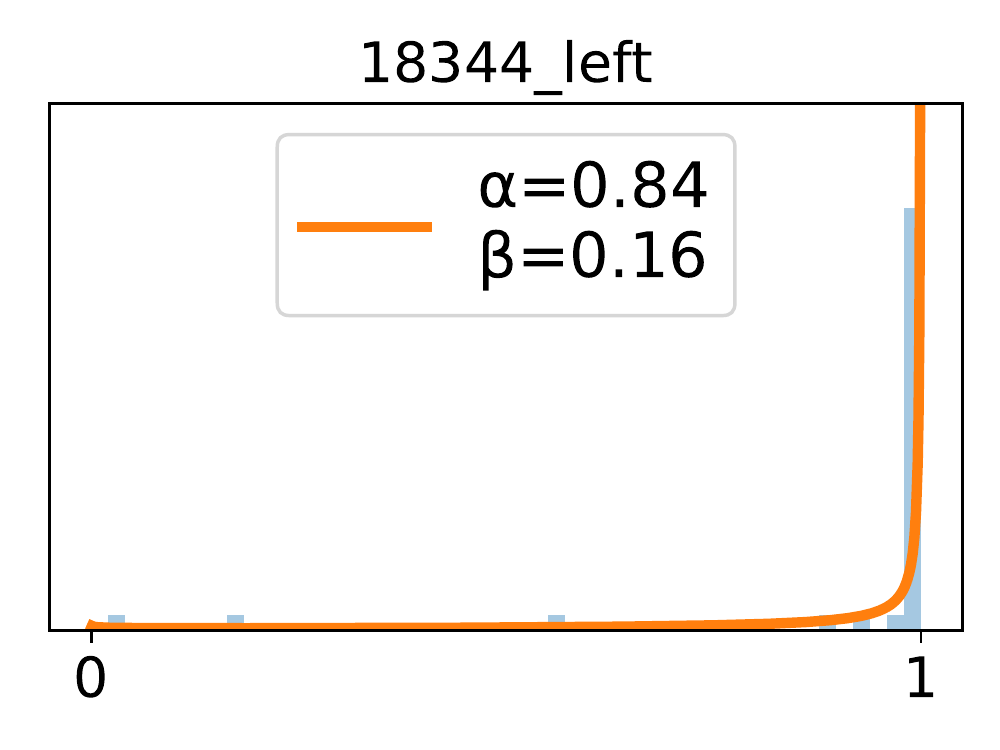}
    \includegraphics[height=2cm]{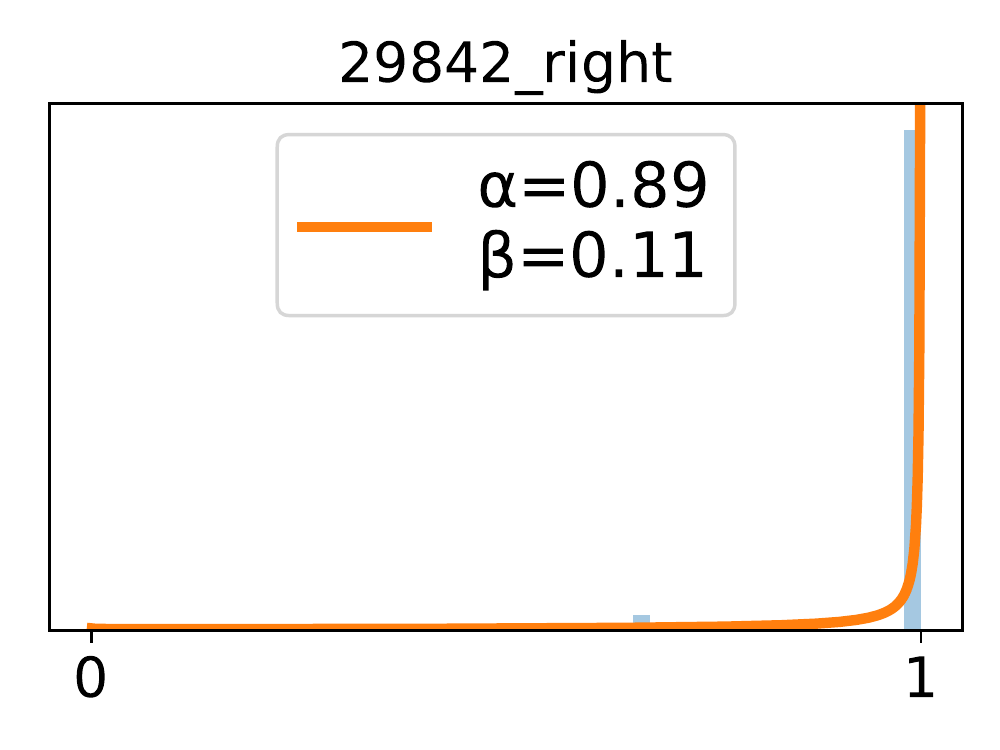}
    \includegraphics[height=2cm]{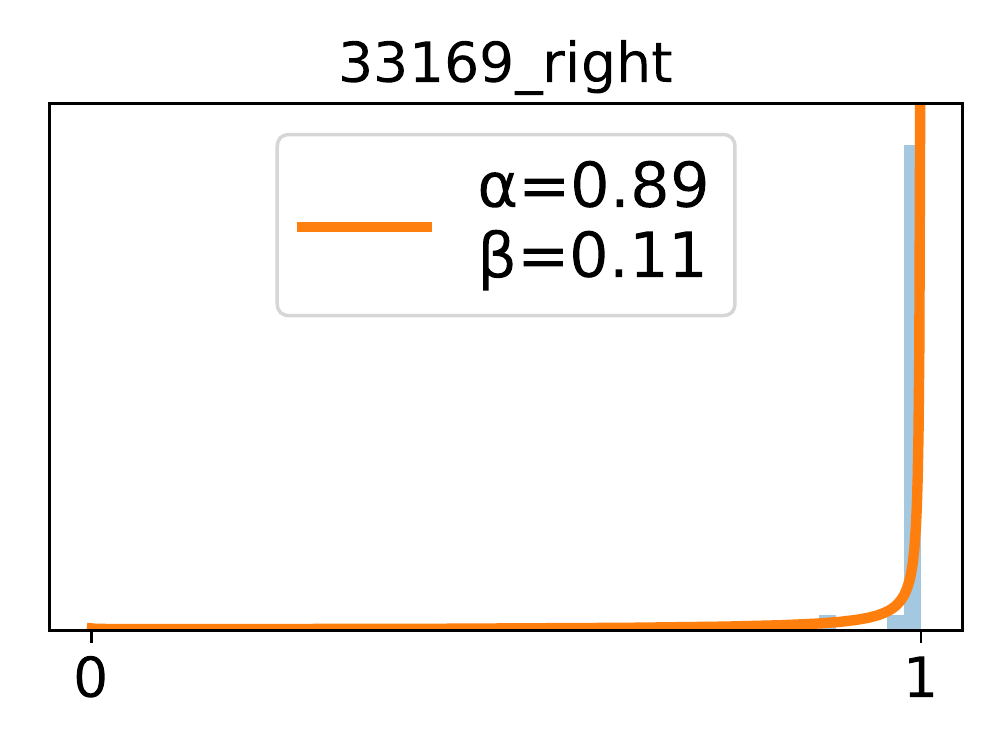}
    \includegraphics[height=2cm]{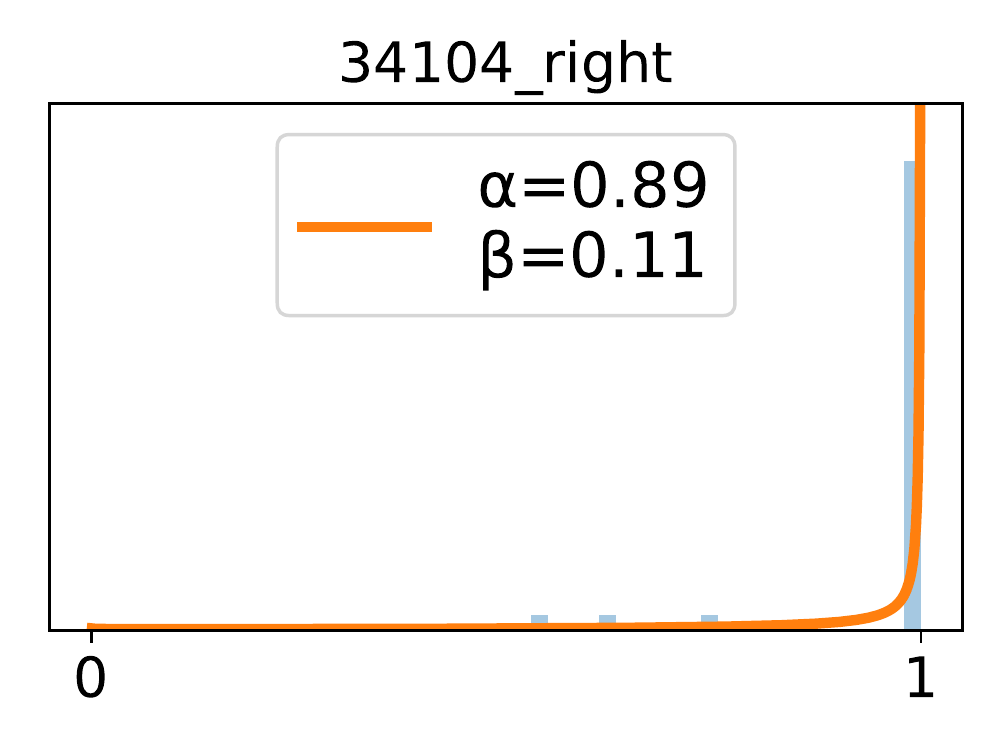}
    \includegraphics[height=2cm]{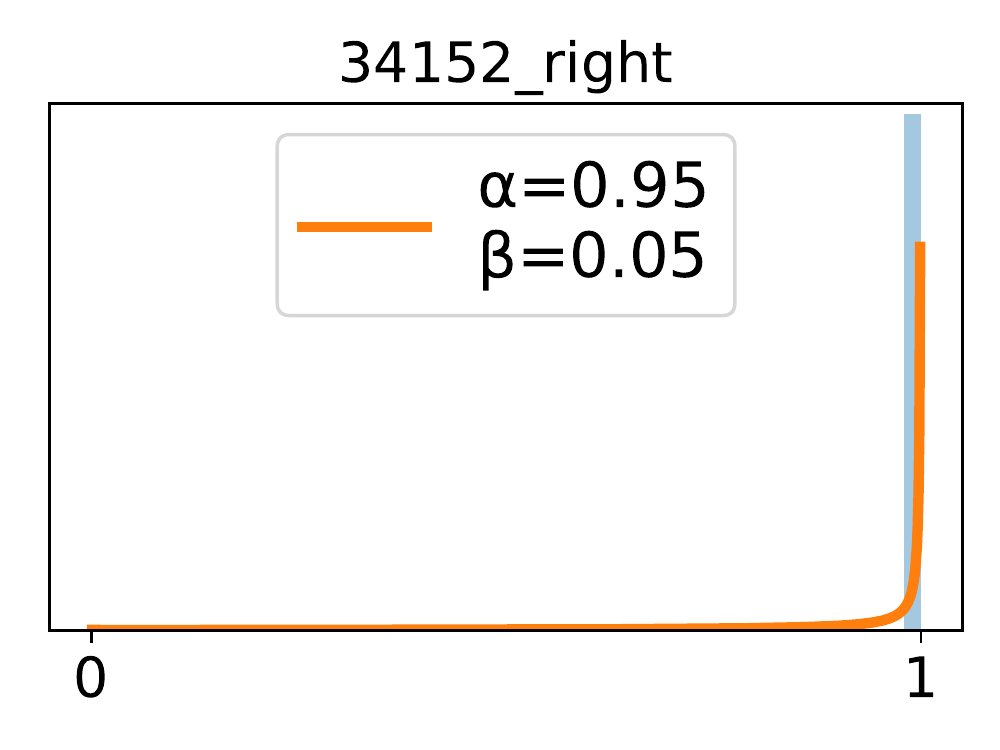}
    \includegraphics[height=2cm]{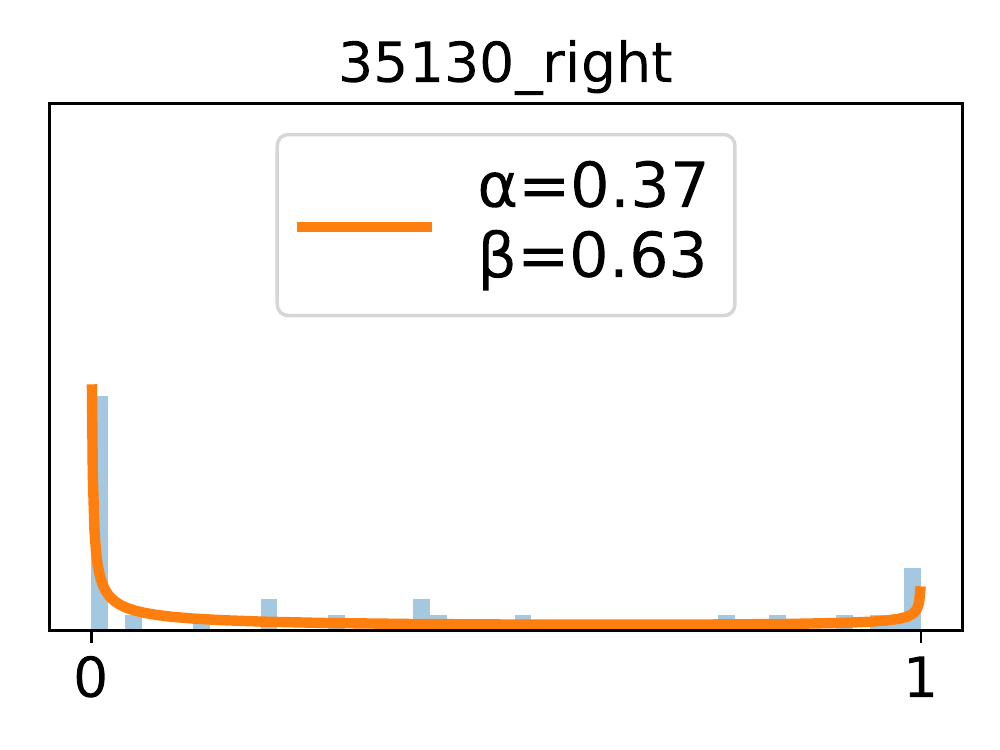}
    \includegraphics[height=2cm]{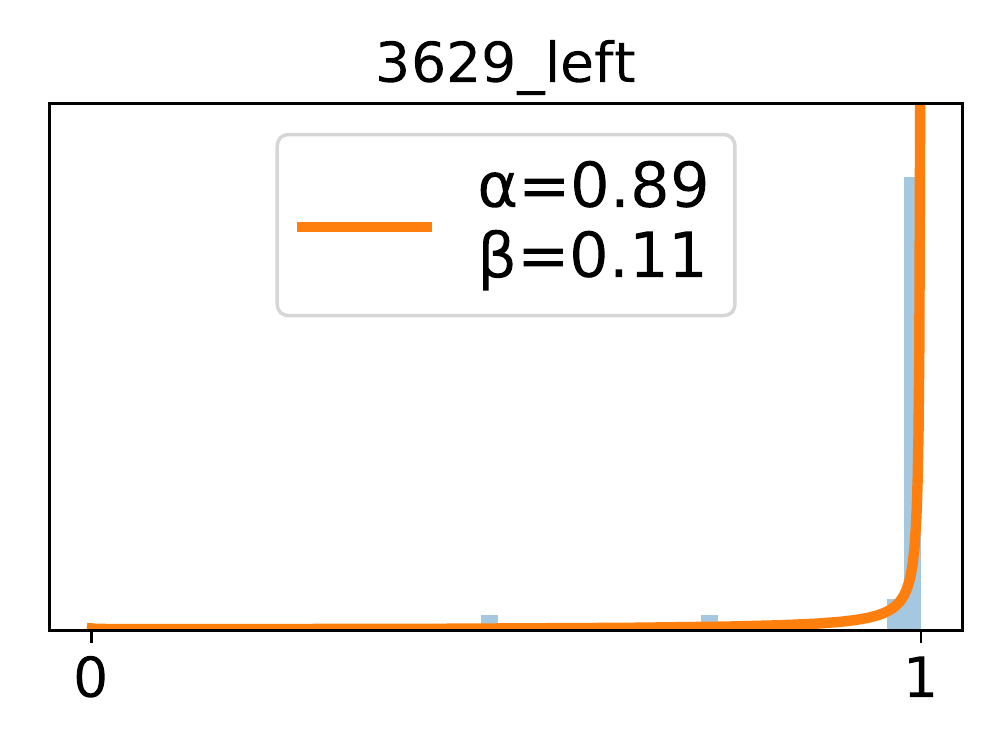}
    \includegraphics[height=2cm]{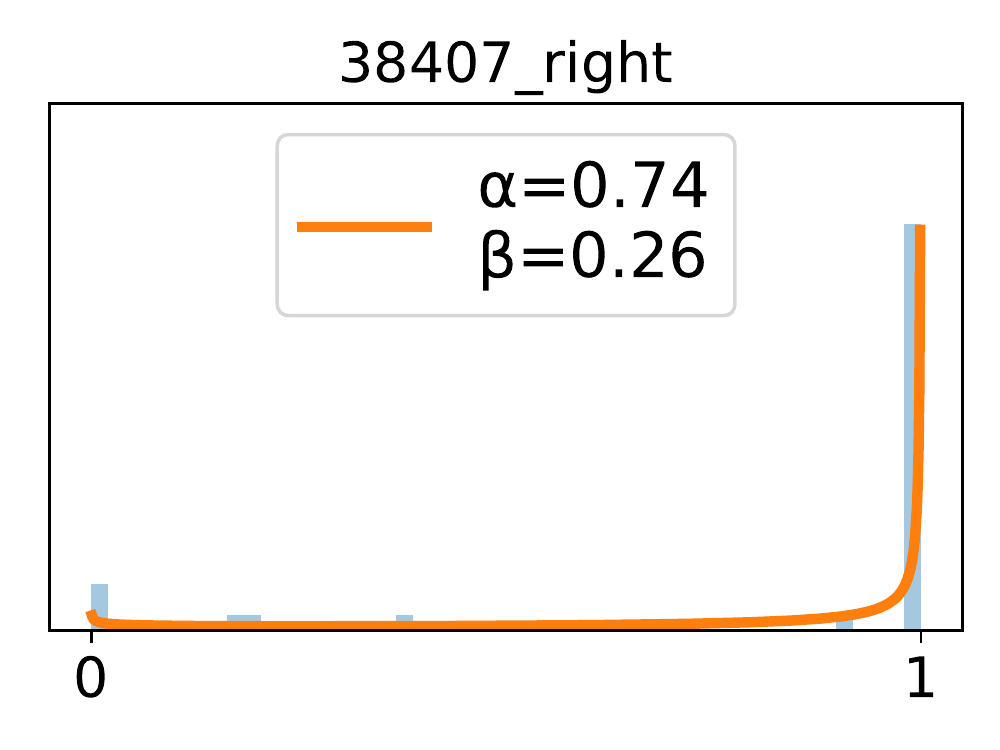}
    \includegraphics[height=2cm]{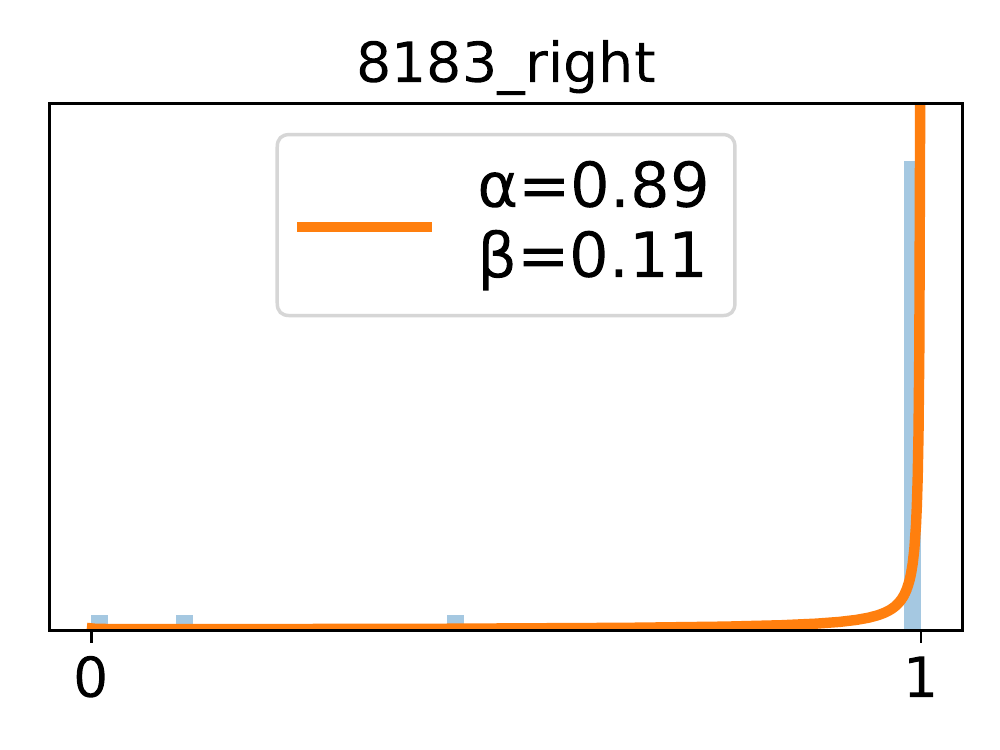}
    \caption{SL4}
\end{subfigure}
\caption{Histograms showing the spread of trained (with $\rho=0.2$) \acs{NN} models' predictions  on a selected numbers of data examples from the diabetic dataset, and a fitted Beta distribution $\mathcal{B}(\alpha,\beta)$ for each example.}
\label{fig:NN-diabetic-distribution}
\end{figure}

In this paper, we assume in our theoretical analysis that the responses of each member classifier in an ensemble follow a Beta distribution $\mathcal{B}(\alpha,\beta)$. We visualized the distributions of individual learner outputs for a select number of data points from both datasets in Figs.~\ref{fig:DT-chiller-distribution}\,\&\,\ref{fig:NN-chiller-distribution}\,\&\,\ref{fig:NN-diabetic-distribution}. For each model type and dataset, we randomly selected nine data points for each \ac{SL}, and showed as histograms the variations of the trained models' predictions on each data point.  

\end{document}